\documentclass[12pt,journal,onecolumn]{IEEEtran}
\usepackage{graphicx}
\usepackage{amsmath}
\usepackage{array}
\usepackage{subfig}
\usepackage{threeparttable}
\usepackage{algorithm}
\usepackage{enumitem}
\usepackage{algpseudocodex}
\usepackage{color}
\usepackage{siunitx}
\usepackage{multirow}
\usepackage{amssymb}
\usepackage{amsfonts}
\usepackage{bbm}
\usepackage[bbgreekl]{mathbbol}
\usepackage{makecell}
\usepackage{cite}
\usepackage{rotating}
\usepackage{cuted}
\usepackage{colortbl}

\newtheorem{assumption}{Assumption}
\newtheorem{definition}{Definition}
\newtheorem{proposition}{Proposition}

\newtheorem{proof}{Proof}

\makeatletter
\newcommand*{\addFileDependency}[1]{% argument=file name and 
\typeout{(#1)}
\@addtofilelist{#1}
\IfFileExists{#1}{}{\typeout{No file #1.}}
}\makeatother
\newcommand*{\myexternaldocument}[1]{%
\externaldocument{#1}%
\addFileDependency{#1.tex}%
\addFileDependency{#1.aux}%
}
\usepackage{xr}
\myexternaldocument{Supp0528}

\begin{document}

\title{Harmonized Gradient Descent for \\Class Imbalanced Data Stream Online Learning}

\author{Han Zhou, 
        Hongpeng Yin,
        Xuanhong Deng,
        Yuyu Huang,
        Hao Ren% <-this % stops a space
\thanks{Han Zhou is with the Peng Cheng Laboratory, and School of Automation, Chongqing University, China. Email: zhouhan1515@foxmail.com.}
\thanks{Hongpeng Yin, Xuanhong Deng and Yuyu huang are with the School of Automation, Chongqing University, China.  \textit{(Hongpeng Yin is the Corresponding Author. Email:yinhongpeng@cqu.edu.cn.)}}
\thanks{Hao Ren is with School of Information Science and Engineering, East China University of Science and Technology, Shanghai, China. Email: renhaocheng@vip.163.com.}}


\IEEEtitleabstractindextext{%
\begin{abstract}
Many real-world data are sequentially collected over time and often exhibit skewed class distributions, resulting in imbalanced data streams. 
While existing approaches have explored several strategies, such as resampling and reweighting, for imbalanced data stream learning, our work distinguishes itself by addressing the imbalance problem through training modification, particularly focusing on gradient descent techniques. 
We introduce the harmonized gradient descent (HGD) algorithm, which aims to equalize the norms of gradients across different classes. By ensuring the gradient norm balance, HGD mitigates under-fitting for minor classes and achieves balanced online learning. 
Notably, HGD operates in a streamlined implementation process, requiring no data-buffer, extra parameters, or prior knowledge, making it applicable to any learning models utilizing gradient descent for optimization.
Theoretical analysis, based on a few common and mild assumptions, shows that HGD achieves a satisfied sub-linear regret bound. 
The proposed algorithm are compared with the commonly used online imbalance learning methods under several imbalanced data stream scenarios. Extensive experimental evaluations demonstrate the efficiency and effectiveness of HGD in learning imbalanced data streams.
\end{abstract}

\begin{IEEEkeywords}
Data Streams, Imbalance, Online Learning, Online Convex Optimization.
\end{IEEEkeywords}}

\maketitle

\IEEEdisplaynontitleabstractindextext
\IEEEpeerreviewmaketitle

\section{Introduction}
\label{sec:Intro}

\begin{figure*}[t]
    \centering
    \subfloat[Biased Problem]{
        \includegraphics[width=3.3cm]{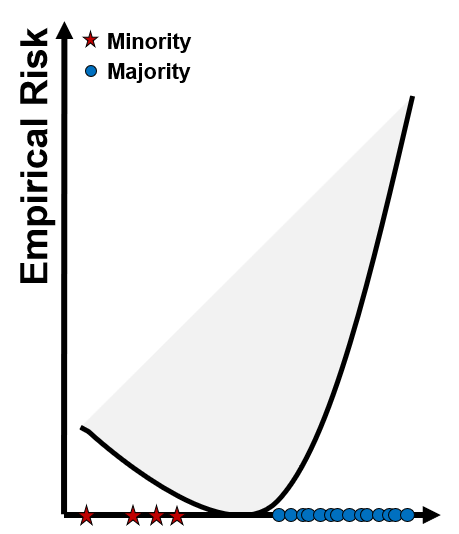}
    }
    \subfloat[Biased Learning]{
        \includegraphics[width=3.3cm]{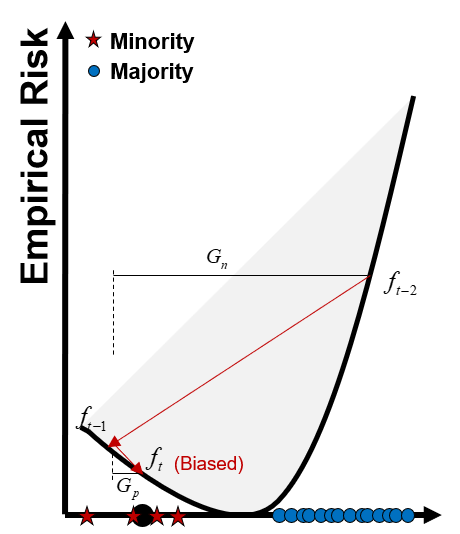}
    }  
    \subfloat[Resampling Learning]{
        \includegraphics[width=3.3cm]{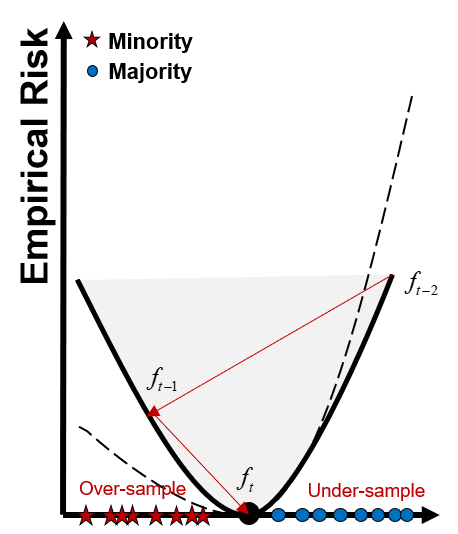}
    }  
    \subfloat[Reweighting Learning]{
        \includegraphics[width=3.3cm]{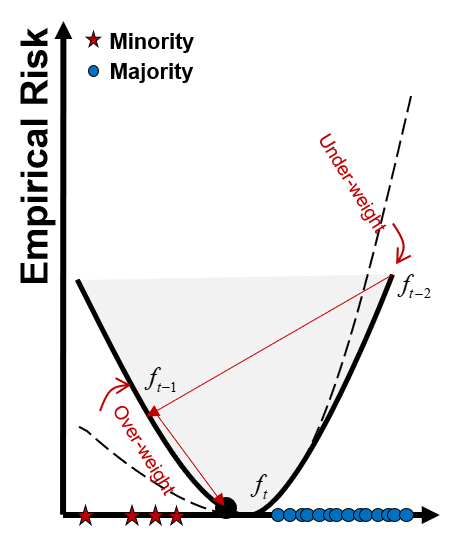}
    }  
    \subfloat[Ours]{
        \includegraphics[width=3.3cm]{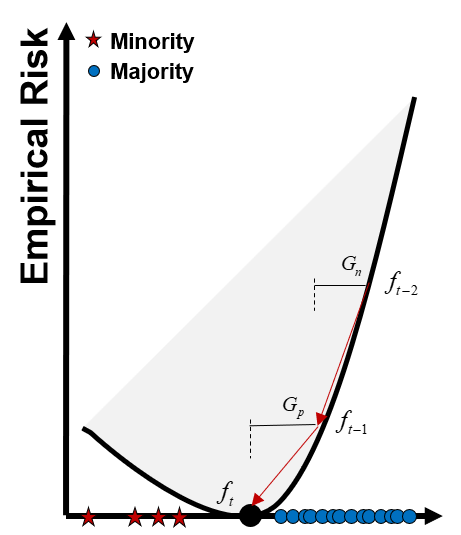}
    }  
    \caption{The imbalanced learning problem and strategies. (a) Imbalanced data streams result biased optimization problem. (b) Traditional online learning methods becomes biased towards the majority. (c) Resamplinging adjusts the number of instances. (d) Reweighting adjusts the importance of classes. (e) Our method ensures equal contribution from each class in gradient descent.}
    \label{fig:ImbalanceLearning}
\end{figure*}

In many real-world scenarios, data are frequently obtained sequentially over time, often exhibiting skewed class distributions, commonly referred to as imbalanced data streams. Learning from imbalanced data streams introduces a biased optimization problem due to the unequal representation of classes, as depicted in Figure~\ref{fig:ImbalanceLearning}(a), and thus many online learning algorithms \cite{hazan2016introduction} become biased towards the majority class. This bias emerges due to their tendency to prioritize minimizing overall risk, leading to suboptimal performance, particularly concerning the minority class, as illustrated in Figure~\ref{fig:ImbalanceLearning}(b).
Such biased performance is problematic, especially in contexts where misclassifying minority class samples could have severe consequences, such as in industrial fault diagnosis \cite{zhou2025FD, Zhou2022FaultDiagnosis}, disease diagnosis \cite{ahsan2022machine}, anomaly detection \cite{Abdelmoumin2022Intrusion, Liu2023Traffic} and among others. Consequently, the combination of online learning and class imbalance learning presents new challenges and research avenues.

Recently, various strategies have been commonly employed to mitigate the effects of biased optimization in imbalanced data stream learning. We divide them into three primary groups: data level resampling \cite{Wang2013resampling, wang2014resampling, wang2016online}, cost sensitive reweighting \cite{Wang2014CSOGD, Chen2023CSAKL, You2023OLI2DS, Chen2023CSRDA}, and others \cite{cieslak2008learning, krawczyk2018dynamic}.
\begin{itemize}
    \item In data level resampling, it involves modifying the dataset by either reducing the number of instances from the majority class (undersampling) or increasing the number of instances from the minority class (oversampling), thereby balancing the class distribution. Accordingly, the optimization problem becomes more balanced, with each class contributing proportionally to the overall objective function, as shown in Figure~\ref{fig:ImbalanceLearning}(c).
    \item In cost sensitive reweighting, the loss function used for training the model is modified to assign higher weights to instances from the minority class and lower weights to instances from the majority class. As shown in Figure~\ref{fig:ImbalanceLearning}(d), by reweighting, the optimization algorithm is encouraged to prioritize minimizing errors on the minority class, allowing the loss function to be more accurately estimated for all classes.
    \item Additional methods for addressing imbalance in data streams include the integration of resampling and reweighting techniques, ensamble learning, one-class learning, among others. 
\end{itemize}
A comprehensive examination of these approaches is presented in Section~\ref{sec:RelatedWork}.

 %Although these methods are intuitive and working reasonably well, an important thing is that these methods inevitably involve heuristic sampling rate selection or empirical cost determination, limiting their easy implementation to practical applications.

%============================================================
%=============================================================

The important feature that makes our work depart from prior work is
that we try to solve the imbalance problem from the perspective of the training modification, particularly in gradient descent. 
We starts the investigation from the following assumption: In the learning procedure, the gradient steps are dominated by the majority class, driving the classifier weights towards configurations that fitting the majority class, while under fitting the minority class.
We first analyze this claim through a didactic experiment and underscore its association with under-fitting for minor classes.
Then, we introduce the proposed harmonized gradient descent (HGD) algorithm, which aims to equalize the norms of gradients across different classes, as shown in Figure~\ref{fig:ImbalanceLearning}(e). %A detailed description of Figure~\ref{fig:ImbalanceLearning}(e) is presented in Section~\ref{sec:Methodology}.
By ensuring gradient balance, HGD mitigates under-fitting for minor classes and thus achieves balanced online learning. Notably, HGD requires no data-buffer, extra parameters, or prior knowledge, and treats imbalanced data streams the same as balanced ones, enabling its straightforward implementation.

%Unlike previous approaches that primarily focused on addressing instance number inequality or cost asymmetry, this paper

Our contributions are listed below.
\begin{itemize}
    \item[$\bullet$] \textit{Motivation.} We identify that gradient imbalances lead to biased learning performance in the context of imbalanced data streams and validate this assumption experimentally.
    \item[$\bullet$] \textit{Method.} We introduce the harmonized gradient descent (HGD) method for imbalanced data stream learning, emphasizing its ease of implementation and versatility.
    \item[$\bullet$] \textit{Theoretical guarantee.} We establish a satisfying sub-linear regret bound for HGD under common and mild assumptions.
    \item[$\bullet$] \textit{Experimental results.} Extensive experiments illustrate the efficiency and effectiveness of HGD, showing its practical utility.
\end{itemize}

The subsequent sections of this paper are structured as follows. Section~\ref{sec:RelatedWork} provides an overview of related research. In Section~\ref{sec:Methodology}, we present the problem formulation and detail the proposed method. Theoretical analysis is presented in Section~\ref{sec:Theory}. Comprehensive experimental results and discussions are presented in Section~\ref{sec:Exp}. Lastly, Section~\ref{sec:Conclusion} offers conclusions of this study.

\section{Related Work}
\label{sec:RelatedWork}

\subsection{Online Learning}

Online learning is a machine learning paradigm where models are trained incrementally by processing data instances as they arrive sequentially over time. Let us denote a sequence of data instances $\mathcal{D} = \{(\mathbf{x}_1,y_1), (\mathbf{x}_2,y_2), \cdots, (\mathbf{x}_T,y_T) \}$, where $\mathbf{x}_t \in \mathbb{R}^d$ represents the features of the $t$-th instance and $y_t \in \{+1, -1\}$ represents its corresponding label. Based on a single data instance and its corresponding label, the learning objective at each time step $t$ is to learn a predictive model $f \in \mathcal{F}$ that minimizes the prediction error or loss $\mathcal{L}(y_t, \hat{y}_t)$:
\begin{align}\label{Eq:EmRisk}
    \min_f \mathcal{L}(f; \mathbf{x}_t, y_t).
\end{align}
Here, $f: \mathbb{R}^d\to\mathbb{R}$ is a function from a convex function space $\mathcal{F}$. $\mathcal{L}:\mathcal{F}\to\mathbb{R}$ is a loss function. $\hat{y}_t = f(\mathbf{x}_t)$ is the predicted label. The performance of an online learning model is commonly measured by the regret:
\begin{align}\label{Eq:regret}
    Regret = \sum_{t=1}^T\mathcal{L}(f_t;\mathbf{x}_t,y_t) - \sum_{t=1}^T\mathcal{L}(f^*;\mathbf{x}_t,y_t),
\end{align}
where $f^* \in \mathcal{F}$ is the best learner that minimizes the loss function. This measurement shows the difference between the cumulative loss incurred by the online learning model $f$ and the cumulative loss of the best fixed  model $f^*$ that could have been chosen in retrospect.

Popular solutions for online learning include online gradient descent (OGD) \cite{Zinkevich2003OGD}, passive-aggressive algorithms \cite{crammer2006PA}, and others \cite{hazan2007logarithmic, McMahan2011FTRL, lu2016large, zhou2025OLEF, zhang2025OLEF}. Among these, OGD has received significant attention due to its simplicity and effectiveness. It optimizes the problem in Eq.\eqref{Eq:EmRisk} by iteratively adjusting its parameters in the direction of the steepest descent of the loss function. This gives the updating rule as $f_{t+1} = f_t-\eta\nabla\mathcal{L}(f_t, \mathbf{x}_t, y_t)$, where $\eta$ is the learning rate. Notably, OGD achieves sublinear regret $\mathcal{O}(\sqrt{T})$ on any convex loss functions with bounded gradients. The sublinearity implies that the time average regret $\frac{Regret (T)}{T}$ converges to zero as the number of instances goes to infinity. Namely, the performance difference between the online learning model $f_t$ and the optimal model $f^*$ ultimately approaching zero.

\subsection{Learning Imbalanced Data}

The issue of imbalanced learning has garnered significant attention, initially emerging within the batch learning settings. One of the popular strategies is data level resampling. It aims to balance the class distribution by adjusting the number of instances. This can involve reducing the number of instances from majority classes (undersampling), generating new instances for minority classes (oversampling), or a combination of both strategies. For example, one very straightforward method can randomly add the minority class instances or eliminate the available majority class instances. However, randomly resampling may lead to over-fitting or loss of important information. Thus, several informed sampling techniques have been developed including: Synthetic Minority Oversampling Technique (SMOTE) \cite{chawla2002smote, fernandez2018smote}, which uses the nearest neighbors technique to create synthetic instances; Adaptive Synthetic Algorithm (ADASYN) \cite{He2008ADASYN}, which resamples the minority class instances that are more difficult to learn; Majority Weighted Minority Oversampling Technique (MWMOTE) \cite{Barua2014MWMOTE}, which generates informative minority class instances; and others \cite{Ng2015Diversified, Abdi2016MDO, Xie2022GDO, Dablain2023DeepSMOTE}.

Another popular strategy for imbalanced data is cost-sensitive reweighting. It adjusts the importance of classes by assigning higher costs to minority classes and lower costs to majority classes \cite{aurelio2019learning, Jing2021Multiset,Zhou2024Interpretable, Cui2019EffetiveNumber}. By reweighting, the optimization algorithm is encouraged to evenly minimizing errors on all classes, allowing the loss function to be more accurately estimated for all classes. Moreover, one can also modify the optimization problem by optimizing a more feasible metric that can avoid biased performance measurement, instead of minimizing empirical risk. For example, one can maximize AUC \cite{wang2020AUC, kim2023AUC}, F score \cite{Wang2023FMetric}, Matthews correlation coefficient \cite{chicco2020MCC}.

Ensemble-based approaches have also been extensively studied for imbalanced data learning. Two variants ensemble methods are widely utilized: Bagging and Boosting. In bagging, a set of base learners are trained with subsets of dataset. In boosting, multiple base learners are sequentially trained whose input is instances incorrectly classified by the previous learner. 
The idea of combining multiple learners can reduce the probability of overfitting \cite{Wang2013Diversity}. In addition, by combining either resample strategies (SMOTEBagging \cite{Wang2009SMOTEBag}, SMOTEBoosting \cite{Chawla2003SMOTEBoost}, UnderOverBagging \cite{Wang2009SMOTEBag}, RUBoosting \cite{Seiffert2010RUSBoost}, Imputation Ensemble \cite{Roozbeh2021Imputation} etc.) or reweighting strategies (CostBagging \cite{Zhang2008CostBagging}, CostAdapBoosting \cite{sun2007CostBoost, tao2019CostBoost}, etc.), they can further boost the performance on imbalanced data learning.

The concept of modifying training techniques to address imbalanced data is also explored in \cite{He2024CILGra, Li2019GraH, Tan2020Equalization, Tan2021Equalization, Huang2024GraHarUDA}. However, unlike most of these studies, which primarily focus on batch training, the present research investigates the challenges posed by data streams, an online learning setting. Furthermore, this study provides theoretical guarantees for learning in such dynamic environments, an aspect that has been overlooked in previous work.

\subsection{Learning Imbalanced Data Streams}

Existing methods for mining imbalanced data streams operate in two distinctive modes: chunk-by-chunk and one-by-one. In the chunk-by-chunk approach, data is processed in batches, allowing the model to be updated periodically with each chunk. This approach facilitates the extension of familiar techniques from batch learning settings to chunk-by-chunk online learning. Techniques such as resampling \cite{anupama2019novel, Bernardo2020CSMOTE, zyblewski2021preprocessed} and reweighting \cite{Hu2018AUC, Vong2018GMEANS, liu2021comprehensive} can be naturally adapted to this context.

However, it becomes more challenging in scenarios where learning paradigms must promptly respond to one instance and then discard it, as in one-pass learning. In such stricter settings, efficient and effective strategies are needed to address class imbalance and ensure timely learning simultaneously. One possibility \cite{Wang2013resampling, wang2014resampling, wang2016online} involves sequentially resampling instances using a Poisson distribution, considering the probability basis of event occurrences over a fixed interval of time. Additionally, reweighting-based solutions \cite{Wang2014CSOGD, Zhao2019ACSOGD, cano2020kappa, Du2021OEB, Chen2023CSAKL, You2023OLI2DS, Chen2023CSRDA} also can be easily implemented to adjust the cost weights in a timely manner, without requiring data buffers. Unfortunately, many existing approaches often necessitate heuristic weighting parameter design, prior cost determination, or sampling rate selection. These complexities in implementation constrain the applicability of these methods in practical scenarios.

There are also some methods using algorithm-level training modification to mitigate class imbalance. This can be done: by focusing solely on modeling one class while mitigating the influence of other classes \cite{krawczyk2018dynamic, Huang2024OneClass}; by learning a box-constrained minimax classifier \cite{Gilet2022BoXConstraint}; by utilizing Hellinger distance for the decision tree splitting metric \cite{cieslak2008learning}; by acting on the initial conditions, given empirical evidence that pretraining can be beneficial \cite{hendrycks2019using}; or by adapting the direction of the gradient steps in order to boost/suppress the domination of the minority/majority class \cite{anand1993improved}. Importantly, most methods mainly focus on batch learning or do not have well-understood performance guarantees in online learning.

Imbalanced data streams are often connected with concept drift \cite{Wang2018CD, liu2021comprehensive, zyblewski2021preprocessed}. For instance, in fraud detection, fraudulent transactions may be rare compared to legitimate ones, leading to class imbalance. Further, fraudsters constantly adapt their tactics, causing concept drift. Henceforth, the combined problem of addressing imbalance issue and concept issue also has enjoyed relatively much research.  Although simultaneously considering imbalance issue and concept issue is quite interesting and challenging, this study, however, mainly focuses on imbalance issue, and concept drift is out of our scope.

We would like to state the difference from our previous work in \cite{zhou2023OHGD}. In this study, we have improved the previous work in several key areas. Firstly, we expanded the idea of harmonized gradient to encompass any learners utilizing gradient descent optimization techniques, whereas previous work focused solely on linear classifiers.  Additionally, we provide a comprehensive theoretical analysis to support this broader application. Then, we extended our method to address dynamic imbalance environments and multi-class learning scenarios. Finally, we provided more extensive experimental results to demonstrate the competitive performance of the proposed method.

\section{Methodology}
\label{sec:Methodology}

This section presents the core idea of this work for imbalanced online learning. We start with an inspiring example to show the the rationale behind our study, and then give the details of the proposed HGD algorithm.

\subsection{A Didactic Example}

%============================================================
\begin{figure}[t]
    \centering
    \subfloat[Gradient Imbalance]{
        \includegraphics[width=6cm]{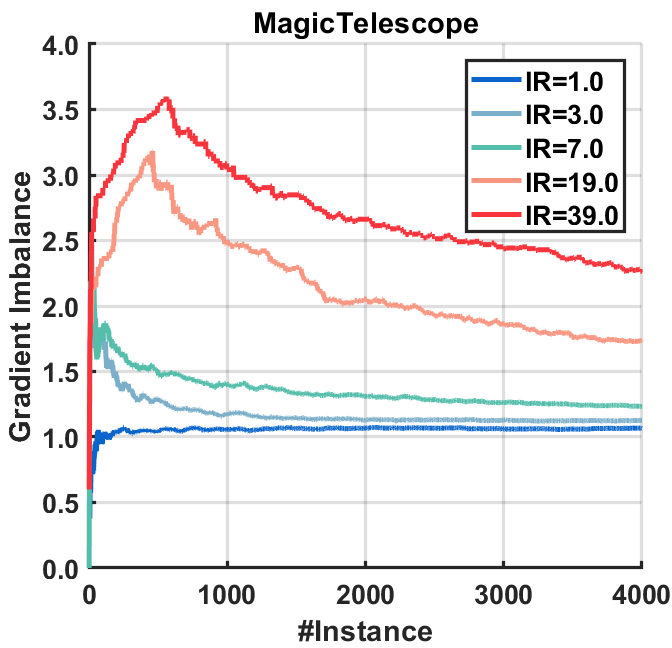}
    }
    \subfloat[Accuracy]{
        \includegraphics[width=6cm]{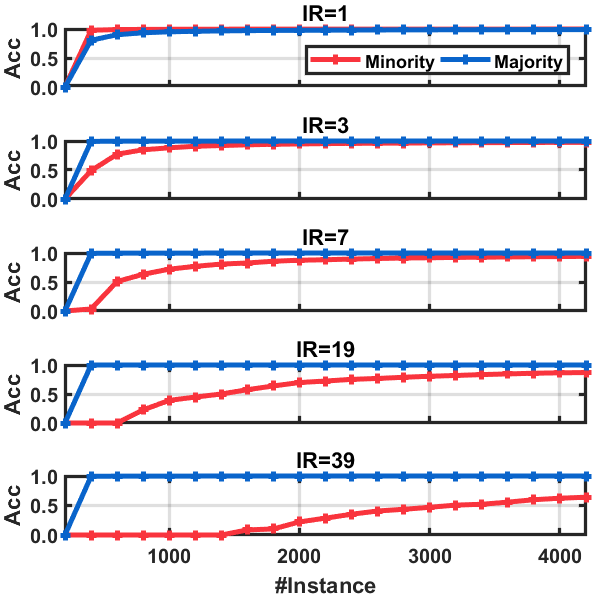}
    }  
    \caption{The gradient imbalance when applying OGD on the \textit{MagicTelescope} dataset with varied imbalance ratio.(a) The gradient imbalance curves; (b) Accuracy on each class.}
    \label{fig:GradientImbalance}
\end{figure}
%=============================================================

Eq.\eqref{Eq:EmRisk} suggests that every instance in the data stream $\mathcal{D}$ plays an equal role in updating the model. However, this symmetrical treatment may introduce bias towards samples from majority classes. While previous research has primarily focused on data-level or cost-level strategies to address this issue, our paper adopts an optimization-level approach to analysis, particularly focusing on gradient descent-based optimization methods. We define the gradient imbalance (GI) in the following.

\begin{definition}
Let the label of majority class be $-1$ while that of minority class be $+1$. Denote the gradient norm of the $t$-th sample as $G_t = \left\| \eta_t \nabla \mathcal{L} (f_t;\mathbf{x}_t,y_t) \right\|^2_2$. Then, the gradient imbalance (GI) is defined as 
\begin{align}
\label{GI}
GI_t =  \frac{\sum^t_{i\in\mathbb{I}_{(y_i=-1)}}G_i}{\sum^t_{i\in\mathbb{I}_{(y_i=+1)}}G_i}.
\end{align}
\end{definition}
where $\mathbb{I}_{(y_i=-1)}$ denotes the dataset where the label of the $i$-th sample is negative. In the following, we will use $\mathbb{I}_{(-)}$ for simplicity. 

Based on the above definition, we provided a didactic example using OGD and a linear classifier on the \textit{MagicTelescope} dataset. Specifically, we varied the imbalance ratio of the dataset from 1.0 to 39.0 and recorded the gradient imbalance and accuracy for both the minority and majority classes. The experimental results are depicted in Figure~\ref{fig:GradientImbalance}. It is evident in Figure~\ref{fig:GradientImbalance}(a) that when the dataset is balanced (the blue line), the gradient imbalance remains around one. However, as the imbalance ratio increases, the positive gradient norms are increasingly overwhelmed by the negative gradient norms at the very beginning ($GI>1$). This implies that the OGD performs a smaller number of gradient descent actions guided by positive samples, resulting in under-fitting to the minority class. The under-fitting of the minority class is further evident in the performance curves shown in Figure~\ref{fig:GradientImbalance}(b). As the imbalance ratio increases, the accuracy curves of the minority class demonstrates slow improvement over time, resulting imbalanced performance.

These observations inspired us to harmonize the gradient norms influenced by different classes throughout the learning process. This needs ensure that the gradient imbalance ($GI$) remains close to one at all times, thus ensuring equal contribution from each class in the optimization process ($G_p=G_n$ in Figure~\ref{fig:ImbalanceLearning}(e)). Ultimately, this attempt aims to achieve balanced performance in both the minority and majority classes.

\subsection{Harmonized Gradient Descent}

To harmonized gradient norms on each class, we design a weight factor $\alpha_t$ for updating step at $t$-th round as follows:
\begin{align}\label{Eq:Alpha}
    \alpha_t = 2\frac{\rho_t \left [\sum^{t-1}_{i\in\mathbb{I}_{-}} \alpha_i G_{i}\right ]\mathbb{I}_{+} + \left[\sum^{t-1}_{i\in\mathbb{I}_{+}} \alpha_i G_{i}\right]\mathbb{I}_{-}}{\sum^{t-1}_{i\in\mathbb{I}_{-}} \alpha_i G_{i} + \sum^{t-1}_{i\in\mathbb{I}_{+}} \alpha_i G_{i}},
\end{align}
where the parameter $\rho_t$ is the ratio of the negative to positive instance number before $t$-th iterations. It can be obtained in a timely manner during the learning procedure, and ensures that the GI converges to one faster. 

Applying the weights to gradient descent actions lead to the proposed HGD algorithm:
\begin{align}\label{Eq:HGD}
f_{t+1}&=f_{t} - \eta_t \alpha_t \nabla\mathcal{L}(f_t; \mathbf{x}_t,y_t) \\ \notag
&=\begin{cases}
f_t- \eta_t \frac{2 \rho_t \sum^{t-1}_{i\in\mathbb{I}_{(-)}}\alpha_i G_i}{\sum_{i=1}^{t-1} \alpha_i G_i }
\nabla\mathcal{L}(f_t; \mathbf{x}_t,y_t), & \text{ if } y_t=+1, \\
f_t- \eta_t \frac{2\sum^{t-1}_{i\in\mathbb{I}_{(+)}}\alpha_i G_i}{\sum_{i=1}^{t-1} \alpha_i G_i }
\nabla\mathcal{L}(f_t; \mathbf{x}_t,y_t), & \text{ if } y_t=-1.
\end{cases}
\end{align}
As we can see from Eq~\eqref{Eq:HGD}, the proposed algorithm adjusts the model in the direction of the steepest descent of the loss function. It employs a more aggressive step for instances from minority classes and a more conservative step for instances from majority classes. This approach encourages that each class contributes equally to the optimization process.

We present the pseudo-code of the proposed HGD in Algorithm~\ref{alg:alg1}. The time complexity of HGD is $\mathcal{O}(Td)$, where $T$ represents the number of instances in the data stream $\mathcal{D}$ and $d$ denotes the dimensionality of instances. This computational efficiency scales linearly with both $T$ and $d$, rendering it comparable to that of traditional OGD.

%=================================================================================
\begin{algorithm}[t]
\caption{Harmonized Gradient Descent.}% with time-smoothed gradients
\label{alg:alg1}
\begin{algorithmic}[1]
    \Require initial learning model $f_1$; learning rate $\eta$;
    \State Initialize $\alpha_0=1$;
    \For{$t=1,\cdots,T$}
        \State Receive the instance $\mathbf{x}_t \in \mathbb{R}^d$ from the data stream $\mathcal{D}$;
        \State Predict the label $\hat{y}_t $ using the model $f_{t-1}$;
        \State Receive the true label $ y_t $;
        \State Incur the loss $\mathcal{L}(f_t; \mathbf{x}_t,y_t)$;
        \State Calculate the weight $\alpha_t$ using Eq.\eqref{Eq:Alpha};
        \If{$\mathcal{L}>0$}
            \State Update the model using Eq.\eqref{Eq:HGD}.
        \EndIf
    \EndFor
\end{algorithmic}
\end{algorithm}
%=================================================================================

\subsection{Several Extensions}

\subsubsection{Dynamic Imbalance} In certain real-world online scenarios, the imbalance ratio may vary over time. For example, fraudulent activities increase sporadically during the holiday seasons, resulting in variation in the imbalance between fraudulent and legitimate transactions over time. To dynamically capture the variation, we can employ exponential smoothing, given by $\rho_t = \lambda \rho_{t-1} + (1-\lambda) \mathbb{I}_{(-)}$, where $\lambda \in (0,1)$ is a predefined smoothing parameter. A higher value of $\lambda$ places more emphasis on recent observations, resulting in faster adaptation to changes in the imbalance ratio.

\subsubsection{Multiclass Situation} The idea of harmonized gradient can be easily extended to the multi-class learning setting. Let us assume $y_t\in\{1,\cdots,C\}$ where $C$ is the number of classes and denote the number of instances of the $c$-th class as $N^c_t$ before time $t$. Then, the weight factor at the $t$-th iteration can be defined as
\begin{align}\label{Eq:HGD_MC}
    \alpha_t =  \frac{\max \{N^{c}_t\}}{N_t^{y_t}} \cdot \frac{C \cdot \sum^{t-1}_{i \in \mathbb{I}_{y_t}} \alpha_i G_i}{\sum_{c=1}^C \sum^{t-1}_{i \in \mathbb{I}_{c}} \alpha_i G_i},
\end{align}
where $\max \{N^{c}_t\}$ represents the maximum number of instances among all classes before time $t$. $\mathbb{I}_{c}$ denote a dataset in which instances belong to the $c$-th class. The updating mechanism in Eq~\eqref{Eq:HGD_MC} is similar to that of Eq~\eqref{Eq:HGD}. The model undergoes more aggressive adjustments during training with minority classes, whereas it undergoes more conservative adjustments during training with majority classes. This approach ensures a balanced optimization contribution across all classes within the dataset.

\section{Theoretical Analysis}
\label{sec:Theory}

This section gives the theoretical performance of HGD, in terms of the regret defined in Eq.\eqref{Eq:regret}. Similarly to the previous work in OCO, we adopt the following standard assumptions \cite{Zinkevich2003OGD}.

\begin{assumption}
\label{asump:1D}
    The function set $\mathcal{F}$ is a closed convex set. It contains the origin $\mathbf{0}$, and belongs to an Euclidean ball $R\mathcal{B}$ with the diameter $D = 2R$, i.e.,
    \begin{align}
        \forall f_i,f_j \in \mathcal{F}, \left \| f_i - f_j \right \|_2^2 \leq D^2.
    \end{align}
\end{assumption}

\begin{assumption}
\label{asump:2C}
    The loss function $\mathcal{L}(\cdot)$ is convex, i.e., 
    \begin{align}
        \forall f_i,f_j \in \mathcal{F},
        \mathcal{L}(f_j) \geq \mathcal{L}(f_i) + \nabla \mathcal{L}(f_i)^{\top} (f_j - f_i).
    \end{align}
\end{assumption}

\begin{assumption}
\label{asump:3L}
    The loss function $\mathcal{L}(\cdot)$ is Lipschitz continuous with Lipschitz constant $L  > 0 $, i.e., 
    \begin{align}
        \forall f_i,f_j \in \mathcal{F},
        \left|\mathcal{L}(f_i)-\mathcal{L}(f_j) \right|\leq L\left\| f_i - f_j \right\|_2.
    \end{align}
\end{assumption}

\begin{assumption}
\label{asump:4Alpha}
    The value of the gradient weighting factor $\alpha_t$ is bounded in $[\frac{1}{\rho}, 2\rho]$.
\end{assumption}

Assumption~\ref{asump:1D} imposes a bounded requirement, serving as a technical simplification for analysis purposes, and its validity has been verified through extensive computer simulations. Assumptions~\ref{asump:2C} and \ref{asump:3L} establish conditions for convexity and stability within the problem domain. Within Assumption~\ref{asump:4Alpha}, the upper bound for $\alpha$ is a natural consequence of its definition, while the empirical lower bound prevents $\alpha_t$ from reaching 0, thereby avoiding potential tractability issues in analysis.

Then, we can give the following regret bounds for Algorithm~\ref{alg:alg1}.

\begin{proposition}
\label{prop:RegBound}
    Let $\{f_i\}$ be the sequence obtained by Algorithm~\ref{alg:alg1}. Let $\rho$ be the maximum of $\rho_t$. Under Assumption~\ref{asump:1D}-\ref{asump:4Alpha}, for any learning model $f^{*} \in \mathcal{F}$, we have the following regret bound for Algorithm~\ref{alg:alg1} if we set the learning rate as $\eta_t = \frac{1}{\sqrt{t}}$
    \begin{align}
        Regret(T) = &\sum_{t=1}^T\mathcal{L}(f_t;\mathbf{x}_t,y_t) - \sum_{t=1}^T\mathcal{L}(f^*;\mathbf{x}_t,y_t) \\ \notag
        \leq &\rho D^2 \sqrt{T} + \rho L^2 \sqrt{T}.
    \end{align}
\end{proposition}

The proofs of Proposition~\ref{prop:RegBound} follows quite readily from the work by \cite{Zinkevich2003OGD}, but need to be generalized to a different gradient reweighting setting.

\begin{proof}

Recalling the updating rule in Eq.\eqref{Eq:HGD} and Assumption~\ref{asump:3L}, we have
\begin{align}
    \left \| f_{t+1} - f^* \right \|^2_2  = &\left \| f_{t} - \alpha_t \eta_t \nabla \mathcal{L}(f_t;\mathbf{x}_t, y_t) - f^* \right \|^2_2 \\ \notag
    = & \left \| f_{t} - f^* \right \|^2_2 + \alpha_t^2 \eta^2_t \left\| \nabla \mathcal{L}(f_t;\mathbf{x}_t, y_t) \right\|_2^2 \\\notag
    & - 2\left \langle f_{t} - f^*,  \alpha_t \eta_t \nabla \mathcal{L}(f_t;\mathbf{x}_t, y_t) \right \rangle  \\\notag
    \leq & \left \| f_{t} - f^* \right \|^2_2 + \alpha_t^2 \eta^2_t L^2 \\\notag
    & - 2\alpha_t \eta_t \nabla \mathcal{L}(f_t;\mathbf{x}_t, y_t)^{\top} (f_{t} - f^*).
\end{align}

Using Assumption~\ref{asump:2C} and rearranging the above inequality yields
\begin{align}
    \mathcal{L}(f_t;\mathbf{x}_t,y_t) &- \mathcal{L}(f^*;\mathbf{x}_t,y_t) \\\notag 
    \leq & \nabla \mathcal{L}(f_t;\mathbf{x}_t, y_t)^{\top} (f_{t} - f^*) \\\notag
    \leq & \frac{\left \| f_{t} - f^* \right \|^2_2 - \left \| f_{t+1} - f^* \right \|^2_2}{2\alpha_t\eta_t } + \frac{\alpha_t \eta_t L^2 }{2}.
\end{align}

Summing over $T$ leads to the upper bound of regret
\begin{align}
    Regret(T) = &\sum_{t=1}^T\mathcal{L}(f_t;\mathbf{x}_t,y_t) - \sum_{t=1}^T\mathcal{L}(f^*;\mathbf{x}_t,y_t) \\ \notag
    \leq & \sum_{t=1}^T \left \{ \frac{\left \| f_{t} - f^* \right \|^2_2 - \left \| f_{t+1} - f^* \right \|^2_2}{2\alpha_t\eta_t} +  \frac{\alpha_t \eta_t L^2}{2} \right \} \\\notag
    \leq & \frac{\left \| f_{1} - f^* \right \|^2_2}{2 \alpha_1 \eta_1} - \frac{\left \| f_{T+1} - f^* \right \|^2_2}{2 \alpha_T \eta_T}  +\sum_{t=2}^T \left\{ \left( \frac{1}{2\alpha_{t} \eta_{t}} - \frac{1}{2\alpha_{t-1} \eta_{t-1}}\right) \left \| f_{t} - f^* \right \|^2_2 \right\} \\\notag
    & + \sum_{t=1}^T \frac{\alpha_t \eta_t L^2 }{2}.
\end{align}

Using $\alpha \in [\frac{1}{\rho}, 2\rho]$, Assumption~\ref{asump:1D}, the inequality becomes
\begin{align}
    &Regret(T) \\\notag
    \leq & \frac{ D^2}{2}  \left\{ \frac{1}{\alpha_1 \eta_1}+ \sum_{t=2}^T \left( \frac{1}{\alpha_t \eta_{t}} - \frac{1}{\alpha_{t-1}\eta_{t-1}}\right) \right\} + \sum_{t=1}^T \frac{\alpha_t \eta_t L^2 }{2} \\\notag
    \leq & \frac{\rho D^2}{\eta_T} + \rho L^2 \sum_{t=1}^T\eta_t.
\end{align}

If we define $\eta_t = \frac{1}{\sqrt{t}}$, then we have sublinear regret for HGD
\begin{align}
    Regret(T) \leq \rho D^2 \sqrt{T} + \rho L^2 \sqrt{T}.
\end{align}

This completes the proof. \hfill $\square$
    
\end{proof}

The sublinear regret shows that the average regret approaches zero as $T$ increases. This means that on average, the performance gap between the proposed HGD algorithm and the optimal strategy diminishes over time.

\section{Experimental Results}
\label{sec:Exp}

This section evaluates the performance of the HGD algorithm in imbalanced data stream learning tasks. We initially test HGD on data streams with a static imbalance ratio and then assess its performance on streams with dynamic imbalance ratios. Lastly, we extend the application of HGD to a multiclass setting and implement it in deep neural networks.

%========================================
\begin{table*}[]
\centering
\renewcommand\arraystretch{1}
\caption{The Selected Datasets for Experiments}
\resizebox{\linewidth}{!}{
\begin{tabular}{cccccc|cccccc|cccccc}
\hline
\textbf{No.} & \textbf{Dataset} & \textbf{\#Instances} & \textbf{\#Fea} & \textbf{IR} & \textbf{Group} & \textbf{No.} & \textbf{Dataset} & \textbf{\#Instances} & \textbf{\#Fea} & \textbf{IR} & \textbf{Group} & \textbf{No.} & \textbf{Dataset} & \textbf{\#Instances} & \textbf{\#Fea} & \textbf{IR} & \multicolumn{1}{c}{\textbf{Group}} \\ \hline
\textbf{D1} & bank & 10578 & 7 & 1 & E & \textbf{D25} & diabetes & 768 & 8 & 1.87 & L & \textbf{D49} & pc4 & 1458 & 37 & 7.19 & M \\
\textbf{D2} & credit & 16714 & 10 & 1 & E & \textbf{D26} & steel-plates-fault & 1941 & 33 & 1.88 & L & \textbf{D50} & yeast3 & 1484 & 8 & 8.1 & M \\
\textbf{D3} & default & 13272 & 20 & 1 & E & \textbf{D27} & breast-w & 699 & 9 & 1.9 & L & \textbf{D51} & pc3 & 1563 & 37 & 8.77 & M \\
\textbf{D4} & electricity & 38474 & 7 & 1 & E & \textbf{D28} & BNG & 39366 & 9 & 1.91 & L & \textbf{D52} & pageblocks0 & 5472 & 10 & 8.79 & M \\
\textbf{D5} & heloc & 10000 & 22 & 1 & E & \textbf{D29} & QSAR & 1055 & 41 & 1.96 & L & \textbf{D53} & ijcnn1 & 49990 & 22 & 9.3 & M \\
\textbf{D6} & house16H & 13488 & 16 & 1 & E & \textbf{D30} & pol & 15000 & 48 & 1.98 & L & \textbf{D54} & kc3 & 458 & 39 & 9.65 & M \\
\textbf{D7} & jannis & 57580 & 54 & 1 & E & \textbf{D31} & rna & 59535 & 8 & 2 & L & \textbf{D55} & HTRU2 & 17898 & 8 & 9.92 & M \\
\textbf{D8} & jannisSample & 2000 & 54 & 1 & E & \textbf{D32} & vertebra & 310 & 6 & 2.1 & M & \textbf{D56} & vowel0 & 988 & 13 & 9.98 & M \\
\textbf{D9} & twonorm & 7400 & 20 & 1 & E & \textbf{D33} & phoneme & 5404 & 5 & 2.41 & M & \textbf{D57} & modelUQ & 540 & 20 & 10.74 & H \\
\textbf{D10} & Friedman & 1000 & 25 & 1.01 & L & \textbf{D34} & yeast1 & 1484 & 8 & 2.46 & M & \textbf{D58} & led7digit & 443 & 7 & 10.97 & H \\
\textbf{D11} & equity & 96320 & 21 & 1.02 & L & \textbf{D35} & ILPD & 583 & 10 & 2.49 & M & \textbf{D59} & mw1 & 403 & 37 & 12 & H \\
\textbf{D12} & ringnorm & 7400 & 20 & 1.02 & L & \textbf{D36} & MiniBooNE & 130064 & 50 & 2.56 & M & \textbf{D60} & shuttle04 & 1829 & 9 & 13.87 & H \\
\textbf{D13} & stock & 950 & 9 & 1.06 & L & \textbf{D37} & SPECTF & 349 & 44 & 2.67 & M & \textbf{D61} & Ozone & 2534 & 72 & 14.84 & H \\
\textbf{D14} & mushroom & 8124 & 21 & 1.07 & L & \textbf{D38} & AutoUniv & 1000 & 20 & 2.86 & M & \textbf{D62} & Sick & 3772 & 29 & 15.33 & H \\
\textbf{D15} & splice & 3175 & 60 & 1.08 & L & \textbf{D39} & ada & 4147 & 48 & 3.03 & M & \textbf{D63} & ecoli4 & 336 & 7 & 15.8 & H \\
\textbf{D16} & higgs & 98050 & 28 & 1.12 & L & \textbf{D40} & a8a & 32561 & 123 & 3.15 & M & \textbf{D64} & spills & 937 & 49 & 21.85 & H \\
\textbf{D17} & eye & 14980 & 14 & 1.23 & L & \textbf{D41} & a9a & 48842 & 123 & 3.18 & M & \textbf{D65} & yeast4 & 1484 & 8 & 28.1 & H \\
\textbf{D18} & banknote & 1372 & 4 & 1.25 & L & \textbf{D42} & svmguide3 & 1243 & 21 & 3.2 & M & \textbf{D66} & w8a & 64700 & 300 & 32.47 & H \\
\textbf{D19} & svmguide1 & 7089 & 4 & 1.29 & L & \textbf{D43} & FRMTC & 748 & 4 & 3.2 & M & \textbf{D67} & yeast5 & 1484 & 8 & 32.73 & H \\
\textbf{D20} & bank8FM & 8192 & 8 & 1.48 & L & \textbf{D44} & vehicle0 & 846 & 18 & 3.25 & M & \textbf{D68} & yeast6 & 1484 & 8 & 41.4 & H \\
\textbf{D21} & spambase & 4601 & 57 & 1.54 & L & \textbf{D45} & Click & 39948 & 11 & 4.94 & M & \textbf{D69} & mammography & 11183 & 6 & 42.01 & H \\
\textbf{D22} & analcatdata & 841 & 70 & 1.65 & L & \textbf{D46} & kc1 & 2109 & 21 & 5.47 & M & \textbf{D70} & Satellite & 5100 & 36 & 67 & H \\
\textbf{D23} & magic04 & 19020 & 10 & 1.84 & L & \textbf{D47} & musk2 & 6598 & 166 & 5.49 & M & \textbf{D71} & mc1 & 9466 & 38 & 138.21 & H \\
\textbf{D24} & MagicTelescope & 19020 & 11 & 1.84 & L & \textbf{D48} & segment0 & 2308 & 19 & 6.02 & M & \textbf{D72} & pc2 & 5589 & 36 & 242 & H \\ \hline
\end{tabular}
}
\begin{tablenotes}
    \item{$\cdot$} We empirically divide datasets into four groups regarding their imbalance ratio. 
    \item{$\cdot$} Equal: $IR=1$; Low $1< IR\leq 2$; Medium $2<IR\leq 10$; High $IR>10$.
    \item{$\cdot$} Due to computational memory and time constraints, we utilized datasets with fewer than 20,000 instances for the kernel model.
\end{tablenotes}
\label{tab:datasets}
\end{table*}
%========================================

\subsection{General Settings}

\subsubsection{Datasets}

Seventy-two public datasets with different imbalance ratios were selected to evaluate the performance of the proposed method. 
Each instance in these datasets were provided sequentially, one at a time, in a randomly shuffled order, to stimulate one-pass data streams. Table~\ref{tab:datasets} summarizes the details of the datasets, including the number of instances, the number of features, and the imbalance ratio. In particular, we empirically divided the datasets into four groups regarding their imbalance ratio, encompassing Equal (E), Low (L), Medium (M) and High (H) imbalance groups.

\subsubsection{Competitors}

We conducted a comprehensive comparison between HGD and several popular strategies for learning from imbalanced data streams, encompassing the baseline method (OGD), data-level approaches, cost-sensitive approaches, and ensemble learning techniques. Details of these methods are summarized in Table~\ref{S:tab:methods} in the supplementary material. Given that our method aims to enhance the performance of gradient descent-based models in imbalanced data streams, we selected three widely-used gradient descent-based classifiers as base learners: perceptron, linear support vector machine, and kernel models. The learning rate $\eta_t$ was uniformly set to 0.3 across all methods. While many techniques may necessitate intricate hyper-parameter tuning to accommodate varying imbalance ratios, the specific settings utilized are provided in Table~\ref{S:tab:methods}.

\subsubsection{Metrics}
As the data distributions are highly skewed, we adopted area under an ROC curve (AUC), GMEANS and F1 Score to evaluate the performance of each algorithm, instead of accuracy. Specifically, the definition of AUC can be found in \cite{Huang2005AUC}. GMEANS and F1 are calculated as follows:
\begin{align}
    GMEANS = \sqrt{TPR \times TNR};\\\notag
    F1 =\frac{2\times precision \times recall}{precision + recal}.
\end{align}
Here, True Positive Rate (TPR) or Recall is $\frac{TP}{TP+FN}$. Precision is $\frac{TP}{TP+FP}$. True Negative Rate (TNR) is $\frac{TN}{FP+TN}$ \footnote{TP: True Positive; TN: True Negative; FP: False Positive; FN: False Negative.}.

Additionally, to verify the validity of our motivation, we introduce a gradient imbalance indicator. Recalling the defined gradient imbalance in Eq.\eqref{GI}, to further measure it over the learning procedure, we modify it as GI Indicator (GII):
\begin{equation}
GII = \sum_{t=1}^T(GI_t-1)^2
\end{equation}
A smaller GII clearly indicates a more equal gradient ratio over the online learning procedure.

Directly averaging metrics across different datasets may not yield commensurate results due to the varying complexity of the datasets. To address this issue, we scaled the metrics to a common scale, with the best performance set to 1. By normalizing the metrics, we eliminate dataset-related variations, allowing us to focus on the relative performance of different approaches. Unless otherwise stated, all metrics presented in this manuscript are normalized.

%===============================================
\begin{figure*}[th]
    \centering
    \subfloat[Perceptron]{
    \begin{minipage}[c]{0.3\textwidth}
        \includegraphics[width=5.5cm]{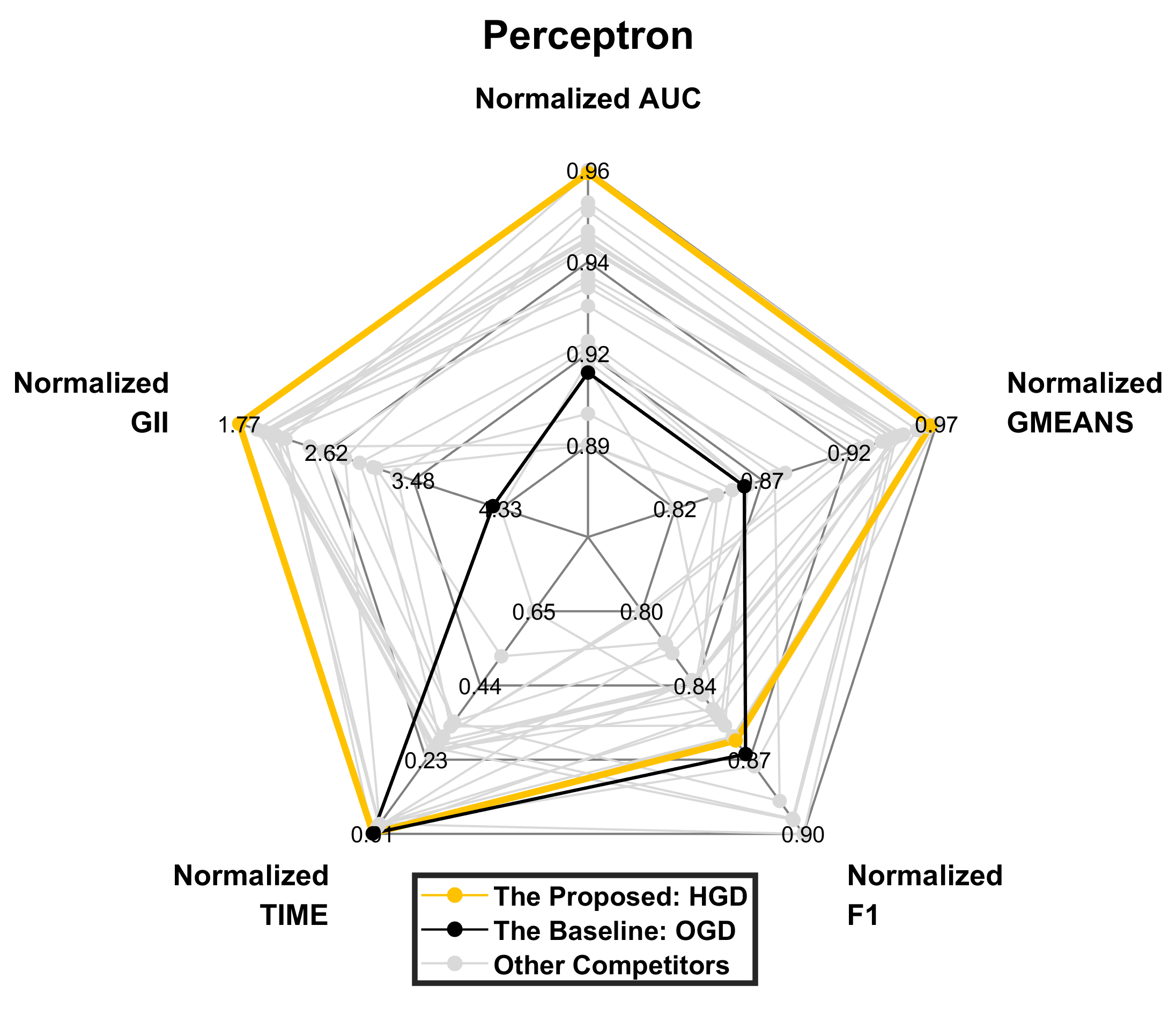}
    \end{minipage}}
    \subfloat[Linear SVM]{
    \begin{minipage}[c]{0.3\textwidth}
        \includegraphics[width=5.5cm]{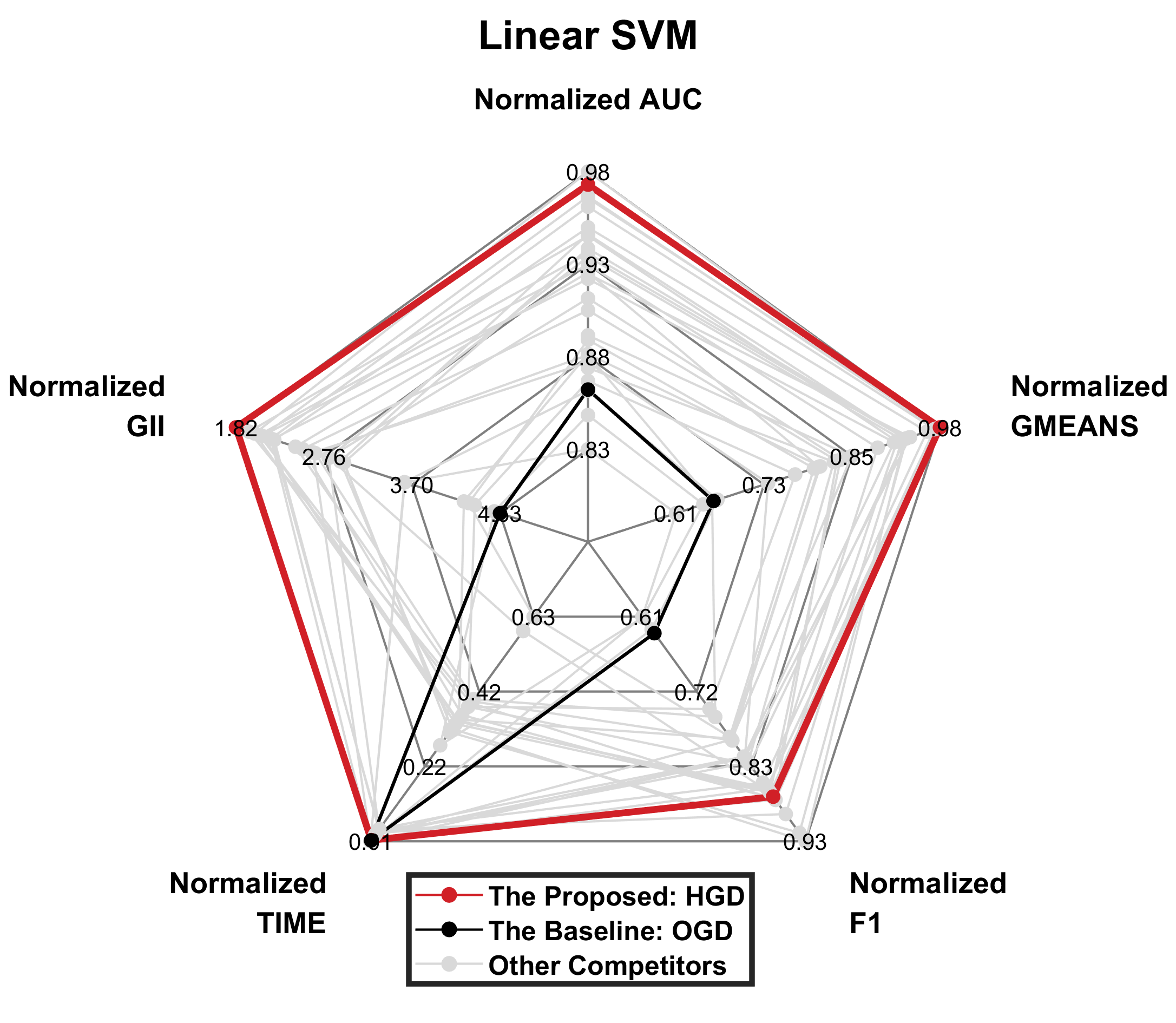}
    \end{minipage}}
        \subfloat[Kernel Model]{
    \begin{minipage}[c]{0.3\textwidth}
        \includegraphics[width=5.5cm]{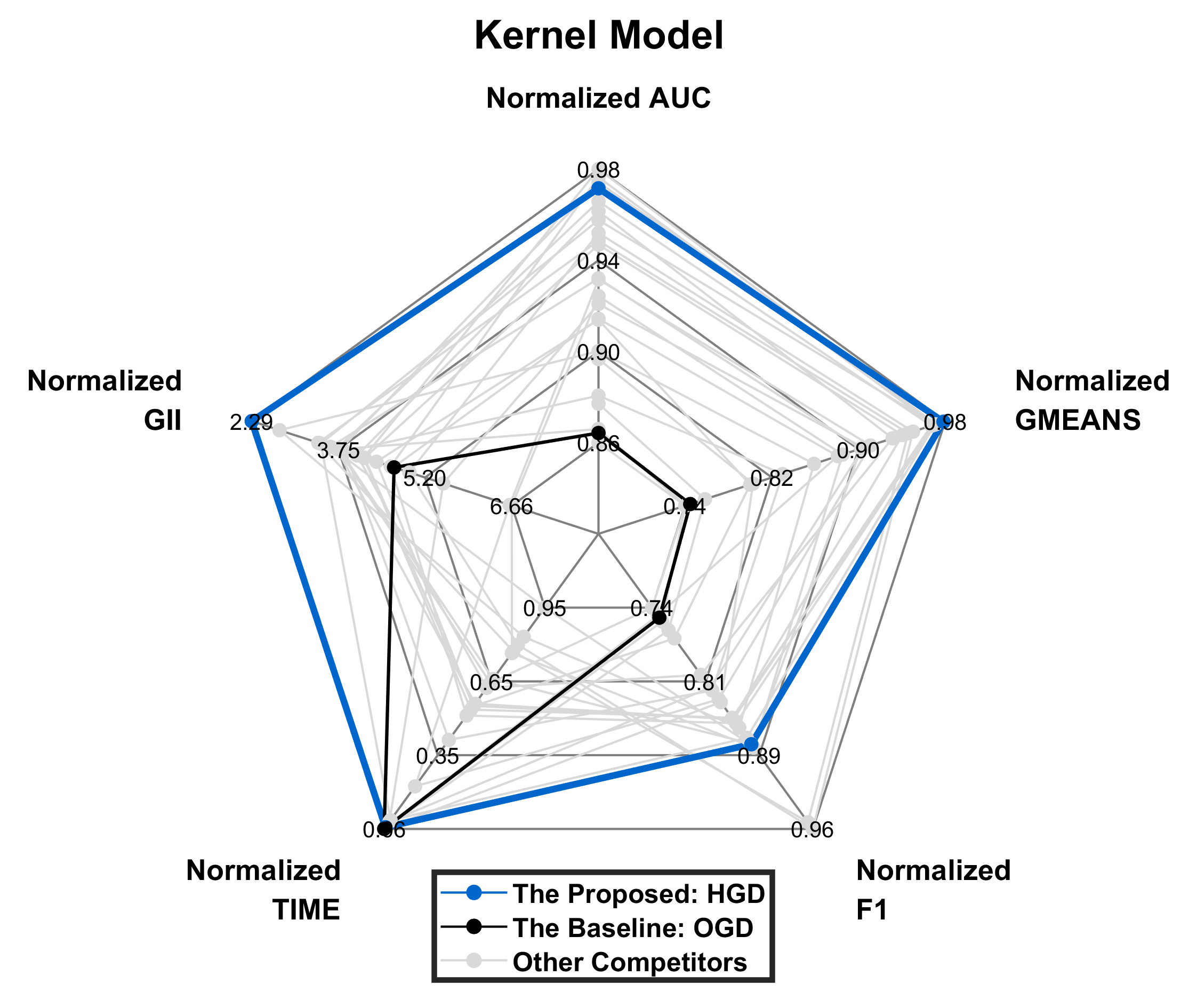}
    \end{minipage}} \\
    \caption{An Systematic Performance Comparison Across Multiple Metrics: AUC, GMEANS, F1 Score, GII, and Computational Time. (a) Perceptron, (b) Linear SVM, (c) Kernel Model as the base learner.}
    \label{fig:static_Spider}
\end{figure*}
%===============================================

%========================================
\begin{table*}[]
\scriptsize
\centering
\renewcommand\arraystretch{1.05}
\caption{The Averaged Performance (Normalized) and Ranks Attained by Different Methods Over All Datasets.}
\resizebox{\linewidth}{!}{
\begin{tabular}{lcccccccccccccccccccc}
\hline
\textbf{Base Learner} & \multicolumn{6}{c}{\textbf{Perceptron}} & \textbf{} & \multicolumn{6}{c}{\textbf{Linear SVM}} & \textbf{} & \multicolumn{6}{c}{\textbf{Kernel Model}} \\ \hline
\multicolumn{1}{l|}{\textbf{Metric}} & \textbf{AUC} & \textbf{Rank} & \textbf{GMEANS} & \textbf{Rank} & \textbf{F1} & \textbf{Rank} & \multicolumn{1}{c|}{\textbf{}} & \textbf{AUC} & \textbf{Rank} & \textbf{GMEANS} & \textbf{Rank} & \textbf{F1} & \textbf{Rank} & \multicolumn{1}{c|}{\textbf{}} & \textbf{AUC} & \textbf{Rank} & \textbf{GMEANS} & \textbf{Rank} & \textbf{F1} & \textbf{Rank} \\ \hline
\multicolumn{21}{l}{\cellcolor[HTML]{C0C0C0}\textbf{Baseline}} \\
\multicolumn{1}{l|}{OGD} & \cellcolor[HTML]{FCFCFF}0.913 & 14.6 & \cellcolor[HTML]{FCFCFF}0.857 & 15.1 & \cellcolor[HTML]{FCFCFF}0.868 & 12.5 & \multicolumn{1}{c|}{} & \cellcolor[HTML]{FCFCFF}0.863 & 17.8 & \cellcolor[HTML]{FCFCFF}0.662 & 17.5 & \cellcolor[HTML]{FCFCFF}0.636 & 16.1 & \multicolumn{1}{c|}{} & \cellcolor[HTML]{FCF7FA}0.861 & 17.0 & \cellcolor[HTML]{FCFAFD}0.743 & 15.6 & \cellcolor[HTML]{FCF6F9}0.748 & 12.4 \\
\multicolumn{21}{l}{\cellcolor[HTML]{C0C0C0}\textbf{Data Level Approaches}} \\
\multicolumn{1}{l|}{OSMOTE} & \cellcolor[HTML]{FCF5F8}0.915 & 13.4 & \cellcolor[HTML]{FCF9FC}0.860 & 14.2 & \cellcolor[HTML]{D5E0F1}0.853 & 11.6 & \multicolumn{1}{c|}{} & \cellcolor[HTML]{FCDBDE}0.889 & 16.4 & \cellcolor[HTML]{FBBBBE}0.802 & 16.1 & \cellcolor[HTML]{FBB3B6}0.785 & 14.4 & \multicolumn{1}{c|}{} & \cellcolor[HTML]{FAA5A7}0.930 & 13.9 & \cellcolor[HTML]{FA9496}0.911 & 12.8 & \cellcolor[HTML]{FBC0C2}0.831 & 13.9 \\
\multicolumn{1}{l|}{OUR} & \cellcolor[HTML]{FAA2A5}0.944 & 11.6 & \cellcolor[HTML]{F98689}0.951 & 11.1 & \cellcolor[HTML]{A8C0E1}0.835 & 13.5 & \multicolumn{1}{c|}{} & \cellcolor[HTML]{FAB2B4}0.921 & 15.4 & \cellcolor[HTML]{F9878A}0.913 & 13.8 & \cellcolor[HTML]{FAA5A8}0.813 & 16.1 & \multicolumn{1}{c|}{} & \cellcolor[HTML]{FAB2B4}0.919 & 15.5 & \cellcolor[HTML]{FA989A}0.905 & 15.8 & \cellcolor[HTML]{FBC7CA}0.819 & 16.5 \\
\multicolumn{1}{l|}{OOR} & \cellcolor[HTML]{F98183}0.956 & \textbf{9.8} & \cellcolor[HTML]{F9797B}0.962 & \textbf{9.2} & \cellcolor[HTML]{D3DFF0}0.852 & 12.2 & \multicolumn{1}{c|}{} & \cellcolor[HTML]{FA8F91}0.947 & 12.6 & \cellcolor[HTML]{F97274}0.958 & 11.3 & \cellcolor[HTML]{FA9395}0.850 & 14.2 & \multicolumn{1}{c|}{} & \cellcolor[HTML]{F98587}0.956 & 11.1 & \cellcolor[HTML]{F96F71}0.972 & 9.9 & \cellcolor[HTML]{FAA6A8}0.870 & 11.7 \\
\multicolumn{1}{l|}{OHR} & \cellcolor[HTML]{F98789}0.954 & \textbf{9.8} & \cellcolor[HTML]{F9898B}0.950 & 10.1 & \cellcolor[HTML]{F8696B}0.904 & \textbf{8.6} & \multicolumn{1}{c|}{} & \cellcolor[HTML]{FAA3A5}0.932 & 13.7 & \cellcolor[HTML]{F98284}0.924 & 13.9 & \cellcolor[HTML]{F97D7F}0.895 & 12.6 & \multicolumn{1}{c|}{} & \cellcolor[HTML]{FBBABC}0.912 & 15.7 & \cellcolor[HTML]{FAA6A9}0.881 & 16.8 & \cellcolor[HTML]{FCF7FA}0.746 & 18.3 \\
\multicolumn{21}{l}{\cellcolor[HTML]{C0C0C0}\textbf{Cost Sensitive Approaches}} \\
\multicolumn{1}{l|}{CSRDA$_{I}$} & \cellcolor[HTML]{F8696B}0.964 & \textbf{7.8} & \cellcolor[HTML]{F96E70}0.971 & \textbf{6.5} & \cellcolor[HTML]{E7EDF7}0.860 & \textbf{9.7} & \multicolumn{1}{c|}{} & \cellcolor[HTML]{FCEEF1}0.875 & 21.4 & \cellcolor[HTML]{FAAEB1}0.829 & 20.5 & \cellcolor[HTML]{FAA1A4}0.822 & 18.9 & \multicolumn{1}{c|}{} & - & - & - & - & - & - \\
\multicolumn{1}{l|}{CSRDA$_{II}$} & \cellcolor[HTML]{F96A6C}0.964 & \textbf{7.4} & \cellcolor[HTML]{F8696B}0.974 & \textbf{6.3} & \cellcolor[HTML]{FCE6E9}0.874 & \textbf{9.0} & \multicolumn{1}{c|}{} & \cellcolor[HTML]{FAAEB1}0.923 & 17.3 & \cellcolor[HTML]{F97D7F}0.935 & 15.9 & \cellcolor[HTML]{FAA4A7}0.815 & 19.4 & \multicolumn{1}{c|}{} & - & - & - & - & - & - \\
\multicolumn{1}{l|}{CSRDA$_{III}$} & \cellcolor[HTML]{FCF0F3}0.917 & 12.0 & \cellcolor[HTML]{FCDDE0}0.882 & 10.7 & \cellcolor[HTML]{7FA4D3}0.819 & 12.7 & \multicolumn{1}{c|}{} & \cellcolor[HTML]{F97B7D}0.963 & \textbf{8.1} & \cellcolor[HTML]{F96A6C}0.975 & \textbf{6.7} & \cellcolor[HTML]{F98789}0.875 & \textbf{9.8} & \multicolumn{1}{c|}{} & - & - & - & - & \textbf{-} & - \\
\multicolumn{1}{l|}{CSRDA$_{IV}$} & - & - & - & - & - & - & \multicolumn{1}{c|}{} & \cellcolor[HTML]{FCD8DB}0.891 & 13.2 & \cellcolor[HTML]{FAA3A5}0.854 & 11.5 & \cellcolor[HTML]{FAB1B4}0.789 & 13.3 & \multicolumn{1}{c|}{} & - & - & - & - & - & - \\
\multicolumn{1}{l|}{CSOGD$_{C_I}$} & - & - & - & - & - & - & \multicolumn{1}{c|}{} & \cellcolor[HTML]{5A8AC6}0.832 & 21.8 & \cellcolor[HTML]{5A8AC6}0.610 & 22.2 & \cellcolor[HTML]{5A8AC6}0.613 & 19.0 & \multicolumn{1}{c|}{} & \cellcolor[HTML]{FCFCFF}0.857 & 18.1 & \cellcolor[HTML]{FCFCFF}0.738 & 17.2 & \cellcolor[HTML]{FCF3F6}0.752 & 13.7 \\
\multicolumn{1}{l|}{CSOGD$_{C_{II}}$} & \cellcolor[HTML]{FCF9FC}0.914 & 16.3 & \cellcolor[HTML]{5A8AC6}0.815 & 18.7 & \cellcolor[HTML]{CDDBEE}0.850 & 14.3 & \multicolumn{1}{c|}{} & \cellcolor[HTML]{FCEAED}0.878 & 19.2 & \cellcolor[HTML]{FCD9DC}0.738 & 21.2 & \cellcolor[HTML]{FAA5A7}0.815 & 19.1 & \multicolumn{1}{c|}{} & \cellcolor[HTML]{FBD0D2}0.894 & 16.4 & \cellcolor[HTML]{FBD6D9}0.801 & 16.8 & \cellcolor[HTML]{FBBEC0}0.834 & 17.1 \\
\multicolumn{1}{l|}{CSOGD$_{S_{I}}$} & - & - & - & - & - & - & \multicolumn{1}{c|}{} & \cellcolor[HTML]{F9FAFE}0.863 & 16.9 & \cellcolor[HTML]{FCFAFD}0.668 & 16.9 & \cellcolor[HTML]{D7E2F2}0.631 & 15.9 & \multicolumn{1}{c|}{} & \cellcolor[HTML]{FCE8EA}0.874 & 15.6 & \cellcolor[HTML]{FCF1F4}0.756 & 14.5 & \cellcolor[HTML]{FCEDF0}0.761 & 11.9 \\
\multicolumn{1}{l|}{CSOGD$_{S_{II}}$} & \cellcolor[HTML]{F96C6E}0.964 & \textbf{7.5} & \cellcolor[HTML]{F97173}0.969 & \textbf{6.8} & \cellcolor[HTML]{E8EEF8}0.860 & 10.1 & \multicolumn{1}{c|}{} & \cellcolor[HTML]{F98183}0.958 & \textbf{8.6} & \cellcolor[HTML]{F96D6F}0.969 & \textbf{7.1} & \cellcolor[HTML]{FA8F91}0.859 & 10.9 & \multicolumn{1}{c|}{} & \cellcolor[HTML]{F97375}0.971 & \textbf{6.5} & \cellcolor[HTML]{F8696B}0.982 & \textbf{4.8} & \cellcolor[HTML]{FAA2A5}0.876 & \textbf{7.6} \\
\multicolumn{21}{l}{\cellcolor[HTML]{C0C0C0}\textbf{Ensemble Learning Approaches}} \\
\multicolumn{1}{l|}{OB} & \cellcolor[HTML]{5B8BC6}0.894 & 18.3 & \cellcolor[HTML]{BCCFE8}0.841 & 18.7 & \cellcolor[HTML]{B4C9E5}0.840 & 16.0 & \multicolumn{1}{c|}{} & \cellcolor[HTML]{B6CBE6}0.850 & 19.6 & \cellcolor[HTML]{D0DDEF}0.648 & 19.0 & \cellcolor[HTML]{608EC8}0.614 & 18.8 & \multicolumn{1}{c|}{} & \cellcolor[HTML]{FCF5F8}0.863 & 17.8 & \cellcolor[HTML]{FCFCFF}0.738 & 16.6 & \cellcolor[HTML]{FCFCFF}0.738 & 13.9 \\
\multicolumn{1}{l|}{OAdaB} & \cellcolor[HTML]{A0BBDE}0.902 & 18.3 & \cellcolor[HTML]{E0E8F5}0.850 & 18.6 & \cellcolor[HTML]{ADC4E3}0.837 & 16.0 & \multicolumn{1}{c|}{} & \cellcolor[HTML]{FCE4E7}0.882 & 20.0 & \cellcolor[HTML]{FBB7BA}0.810 & 19.4 & \cellcolor[HTML]{FAB2B5}0.787 & 18.1 & \multicolumn{1}{c|}{} & \cellcolor[HTML]{FBCBCE}0.898 & 16.5 & \cellcolor[HTML]{FBB4B6}0.859 & 16.7 & \cellcolor[HTML]{FAADAF}0.860 & 13.6 \\
\multicolumn{1}{l|}{OAdaC2} & \cellcolor[HTML]{FCE6E9}0.921 & 15.8 & \cellcolor[HTML]{FCE5E8}0.876 & 17.7 & \cellcolor[HTML]{FAA6A9}0.889 & 11.5 & \multicolumn{1}{c|}{} & \cellcolor[HTML]{FBC7CA}0.904 & 18.4 & \cellcolor[HTML]{FAACAF}0.833 & 19.4 & \cellcolor[HTML]{F9878A}0.874 & 15.4 & \multicolumn{1}{c|}{} & \cellcolor[HTML]{FAAEB0}0.922 & 15.4 & \cellcolor[HTML]{FAA4A6}0.885 & 16.1 & \cellcolor[HTML]{FAA0A2}0.880 & 14.1 \\
\multicolumn{1}{l|}{OCSB2} & \cellcolor[HTML]{FCEAED}0.919 & 16.1 & \cellcolor[HTML]{FCFAFD}0.859 & 18.6 & \cellcolor[HTML]{DBE5F3}0.855 & 13.4 & \multicolumn{1}{c|}{} & \cellcolor[HTML]{FBBFC2}0.911 & 18.0 & \cellcolor[HTML]{FAAAAD}0.838 & 18.8 & \cellcolor[HTML]{F98C8E}0.865 & 14.5 & \multicolumn{1}{c|}{} & \cellcolor[HTML]{FAA4A7}0.930 & 14.3 & \cellcolor[HTML]{FA999B}0.903 & 15.1 & \cellcolor[HTML]{FAA1A3}0.878 & 12.6 \\
\multicolumn{1}{l|}{OKB} & \cellcolor[HTML]{5A8AC6}0.894 & 18.5 & \cellcolor[HTML]{B6CBE6}0.839 & 18.7 & \cellcolor[HTML]{7DA2D2}0.818 & 16.7 & \multicolumn{1}{c|}{} & \cellcolor[HTML]{FCF7FA}0.868 & 20.4 & \cellcolor[HTML]{FBC7CA}0.776 & 19.8 & \cellcolor[HTML]{FBC7CA}0.745 & 19.2 & \multicolumn{1}{c|}{} & \cellcolor[HTML]{FCE3E6}0.878 & 17.0 & \cellcolor[HTML]{FBC5C8}0.829 & 16.1 & \cellcolor[HTML]{FBC5C8}0.822 & 13.5 \\
\multicolumn{1}{l|}{OUOB} & \cellcolor[HTML]{FA9D9F}0.946 & 13.0 & \cellcolor[HTML]{F98688}0.952 & 12.0 & \cellcolor[HTML]{AAC2E2}0.836 & 15.5 & \multicolumn{1}{c|}{} & \cellcolor[HTML]{F97A7C}0.964 & \textbf{7.1} & \cellcolor[HTML]{F96C6E}0.972 & \textbf{6.7} & \cellcolor[HTML]{F98C8E}0.865 & 10.7 & \multicolumn{1}{c|}{} & \cellcolor[HTML]{F97173}0.973 & \textbf{7.1} & \cellcolor[HTML]{F96A6C}0.980 & \textbf{6.6} & \cellcolor[HTML]{FAA2A5}0.876 & \textbf{9.8} \\
\multicolumn{1}{l|}{ORUSB1} & \cellcolor[HTML]{FBCCCF}0.930 & 14.3 & \cellcolor[HTML]{FBB8BB}0.912 & 15.1 & \cellcolor[HTML]{5A8AC6}0.804 & 14.7 & \multicolumn{1}{c|}{} & \cellcolor[HTML]{FCE5E8}0.881 & 17.4 & \cellcolor[HTML]{F6F7FC}0.660 & 20.6 & \cellcolor[HTML]{FBC1C4}0.756 & 17.6 & \multicolumn{1}{c|}{} & \cellcolor[HTML]{FBBABD}0.912 & 11.3 & \cellcolor[HTML]{FCD8DB}0.799 & 15.0 & \cellcolor[HTML]{FCE8EB}0.769 & 14.5 \\
\multicolumn{1}{l|}{ORUSB2} & \cellcolor[HTML]{FBBFC1}0.934 & 14.8 & \cellcolor[HTML]{FA9EA1}0.932 & 14.8 & \cellcolor[HTML]{AAC2E2}0.836 & 14.1 & \multicolumn{1}{c|}{} & \cellcolor[HTML]{FA9597}0.943 & 12.4 & \cellcolor[HTML]{F98588}0.917 & 15.1 & \cellcolor[HTML]{F98789}0.874 & 12.4 & \multicolumn{1}{c|}{} & \cellcolor[HTML]{FA9294}0.945 & 12.4 & \cellcolor[HTML]{F98284}0.942 & 13.6 & \cellcolor[HTML]{FAB3B5}0.851 & 12.5 \\
\multicolumn{1}{l|}{ORUSB3} & \cellcolor[HTML]{FBBABC}0.936 & 13.7 & \cellcolor[HTML]{FAA5A8}0.927 & 13.5 & \cellcolor[HTML]{5C8BC6}0.805 & 16.4 & \multicolumn{1}{c|}{} & \cellcolor[HTML]{FAADB0}0.924 & 13.6 & \cellcolor[HTML]{FBB8BB}0.808 & 16.0 & \cellcolor[HTML]{FAB1B3}0.790 & 17.5 & \multicolumn{1}{c|}{} & \cellcolor[HTML]{F98082}0.960 & 9.4 & \cellcolor[HTML]{F9878A}0.932 & 11.3 & \cellcolor[HTML]{FBD0D3}0.806 & 14.6 \\
\multicolumn{1}{l|}{OOB} & \cellcolor[HTML]{FA9698}0.949 & 10.9 & \cellcolor[HTML]{FA8F92}0.944 & 11.2 & \cellcolor[HTML]{F98588}0.897 & \textbf{10.0} & \multicolumn{1}{c|}{} & \cellcolor[HTML]{FA9496}0.944 & 9.8 & \cellcolor[HTML]{F98083}0.927 & 10.3 & \cellcolor[HTML]{F8696B}0.934 & \textbf{7.6} & \multicolumn{1}{c|}{} & \cellcolor[HTML]{FA9092}0.947 & 10.8 & \cellcolor[HTML]{F97F81}0.947 & 9.4 & \cellcolor[HTML]{F96E70}0.955 & \textbf{7.9} \\
\multicolumn{1}{l|}{OUB} & \cellcolor[HTML]{FAA0A3}0.945 & 13.7 & \cellcolor[HTML]{F98B8D}0.948 & 13.6 & \cellcolor[HTML]{C8D8ED}0.848 & 14.3 & \multicolumn{1}{c|}{} & \cellcolor[HTML]{FAA8AB}0.928 & 12.9 & \cellcolor[HTML]{FA9294}0.890 & 14.0 & \cellcolor[HTML]{F98D90}0.862 & 13.1 & \multicolumn{1}{c|}{} & \cellcolor[HTML]{F8696B}0.978 & \textbf{5.4} & \cellcolor[HTML]{F96F71}0.973 & \textbf{7.2} & \cellcolor[HTML]{FBB3B6}0.850 & \textbf{10.8} \\
\multicolumn{1}{l|}{OWOB} & \cellcolor[HTML]{FA9B9D}0.947 & 11.3 & \cellcolor[HTML]{F98B8D}0.948 & 10.6 & \cellcolor[HTML]{F98385}0.898 & 10.3 & \multicolumn{1}{c|}{} & \cellcolor[HTML]{FA9DA0}0.936 & 9.7 & \cellcolor[HTML]{F97F81}0.930 & 8.4 & \cellcolor[HTML]{F96F71}0.922 & \textbf{7.1} & \multicolumn{1}{c|}{} & \cellcolor[HTML]{F98C8E}0.950 & 9.5 & \cellcolor[HTML]{F97C7E}0.951 & \textbf{8.2} & \cellcolor[HTML]{F8696B}0.962 & \textbf{6.3} \\
\multicolumn{1}{l|}{OWUB} & \cellcolor[HTML]{FA9C9E}0.947 & 13.6 & \cellcolor[HTML]{F98385}0.954 & 12.7 & \cellcolor[HTML]{B8CCE7}0.841 & 15.5 & \multicolumn{1}{c|}{} & \cellcolor[HTML]{F8696B}0.976 & \textbf{5.4} & \cellcolor[HTML]{F96B6D}0.973 & \textbf{6.2} & \cellcolor[HTML]{FA8F91}0.858 & \textbf{10.5} & \multicolumn{1}{c|}{} & \cellcolor[HTML]{F96E70}0.975 & \textbf{6.9} & \cellcolor[HTML]{F97072}0.971 & 8.4 & \cellcolor[HTML]{FAB0B2}0.855 & 11.5 \\
\multicolumn{1}{l|}{OEB} & \cellcolor[HTML]{FBB6B9}0.937 & 14.8 & \cellcolor[HTML]{FA9396}0.941 & 13.9 & \cellcolor[HTML]{89ABD6}0.823 & 16.7 & \multicolumn{1}{c|}{} & \cellcolor[HTML]{FAA1A3}0.934 & 11.7 & \cellcolor[HTML]{F98082}0.928 & 11.1 & \cellcolor[HTML]{FAA0A2}0.824 & 14.2 & \multicolumn{1}{c|}{} & \cellcolor[HTML]{F97B7D}0.964 & 9.5 & \cellcolor[HTML]{F97173}0.968 & 9.5 & \cellcolor[HTML]{FAB2B5}0.851 & 12.7 \\
\multicolumn{21}{l}{\cellcolor[HTML]{C0C0C0}\textbf{The Proposed Approach}} \\
\multicolumn{1}{l|}{HGD} & \cellcolor[HTML]{F96C6E}0.964 & \textbf{7.5} & \cellcolor[HTML]{F96E70}0.971 & \textbf{6.7} & \cellcolor[HTML]{ECF1F9}0.862 & \textbf{9.8} & \multicolumn{1}{c|}{} & \cellcolor[HTML]{F97274}0.970 & \textbf{7.2} & \cellcolor[HTML]{F8696B}0.976 & \textbf{6.5} & \cellcolor[HTML]{F9898B}0.871 & \textbf{9.6} & \multicolumn{1}{c|}{} & \cellcolor[HTML]{F97476}0.970 & \textbf{7.0} & \cellcolor[HTML]{F96B6D}0.980 & \textbf{6.2} & \cellcolor[HTML]{FAA2A4}0.876 & \textbf{8.6} \\ \hline
\end{tabular}
}
\begin{tablenotes}
    \item{$\cdot$} The color of cells shows the performance increments (in red) or decrements (in blue) w.r.t. the BaseLine (OGD). 
    \item{$\cdot$} The rank was calculated by averaging ranks in each datasets. 
    \item{$\cdot$} The bold entries stand for the top five highest rankings.
\end{tablenotes}
\label{tab:cmp_static}
\end{table*}
%========================================

\subsection{Learning imbalanced data streams}

\subsubsection{An overview of the comparison}

An overview of the results achieved by 28 methods with 3 base learner on 72 datasets is visualized in the spider plots (Figure~\ref{fig:static_Spider}), in terms of the normalized AUC, G-Means, F1, GII and time. Here, we highlighted the performance attained by the proposed method and the baseline OGD. We have the following observations form these figures.
\begin{itemize}
    \item The baseline, OGD (denoted in black), unsurprisingly fails to achieve satisfactory performance in terms of AUC, GMEANS, F1, and GII. This is primarily due to the performance bias incurred by OGD. OGD updates the model parameters based on the gradients of the loss function, and in imbalanced data streams, these gradients are dominated by the majority class. As a result, the model focuses more on optimizing performance for the majority class, leading to poor performance on the minority class. In contrast, the proposed HGD demonstrates significant performance improvements w.r.t. the baseline, underscoring its effectiveness in handling imbalanced data stream learning.
    \item In terms of computational time, the proposed method is quite efficient compared to the baseline. This efficiency is attributed to the fact that the proposed HGD only requires a simple weight factor calculation. In contrast, many competitors exhibit less efficiency because they require additional learning strategies to accommodate imbalanced data streams, such as resampling and ensemble learning.
    \item Comparing with all the competitors, the proposed method achieves highly competitive performance in terms of most metrics with all different base learners. However, an exception can be found in the results in terms of F1, where the results achieved by HGD are slightly less competitive. This can be attributed to the tendency oh HGD to identify the minority class more often, which may reduce precision and subsequently result in a lower F1 score.
\end{itemize}

%===============================================
\begin{figure*}[t]
    \centering
        \subfloat[Perceptron]{
        \includegraphics[width=5.5cm]{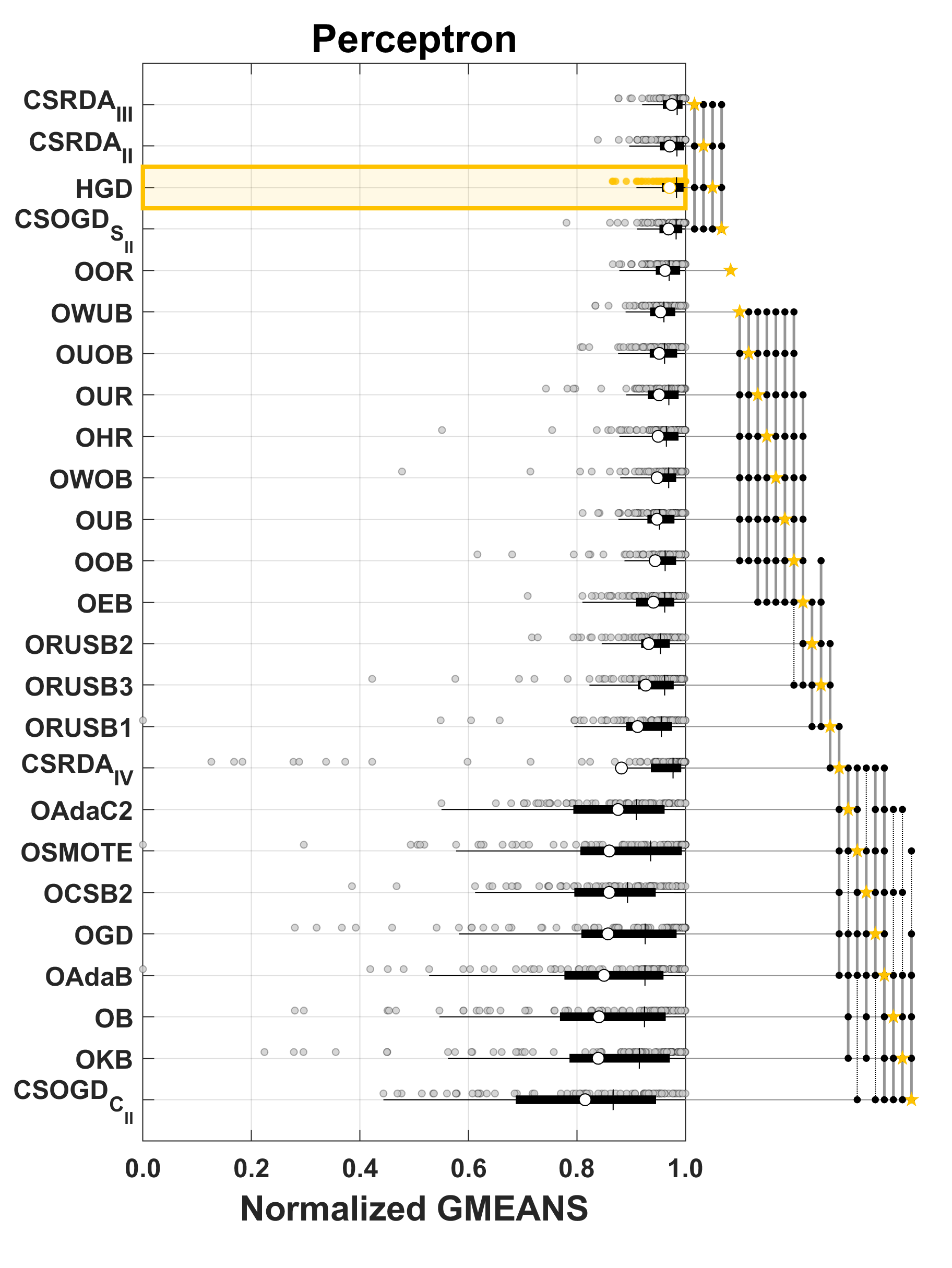}
    }
    \subfloat[Linear SVM]{
        \includegraphics[width=5.5cm]{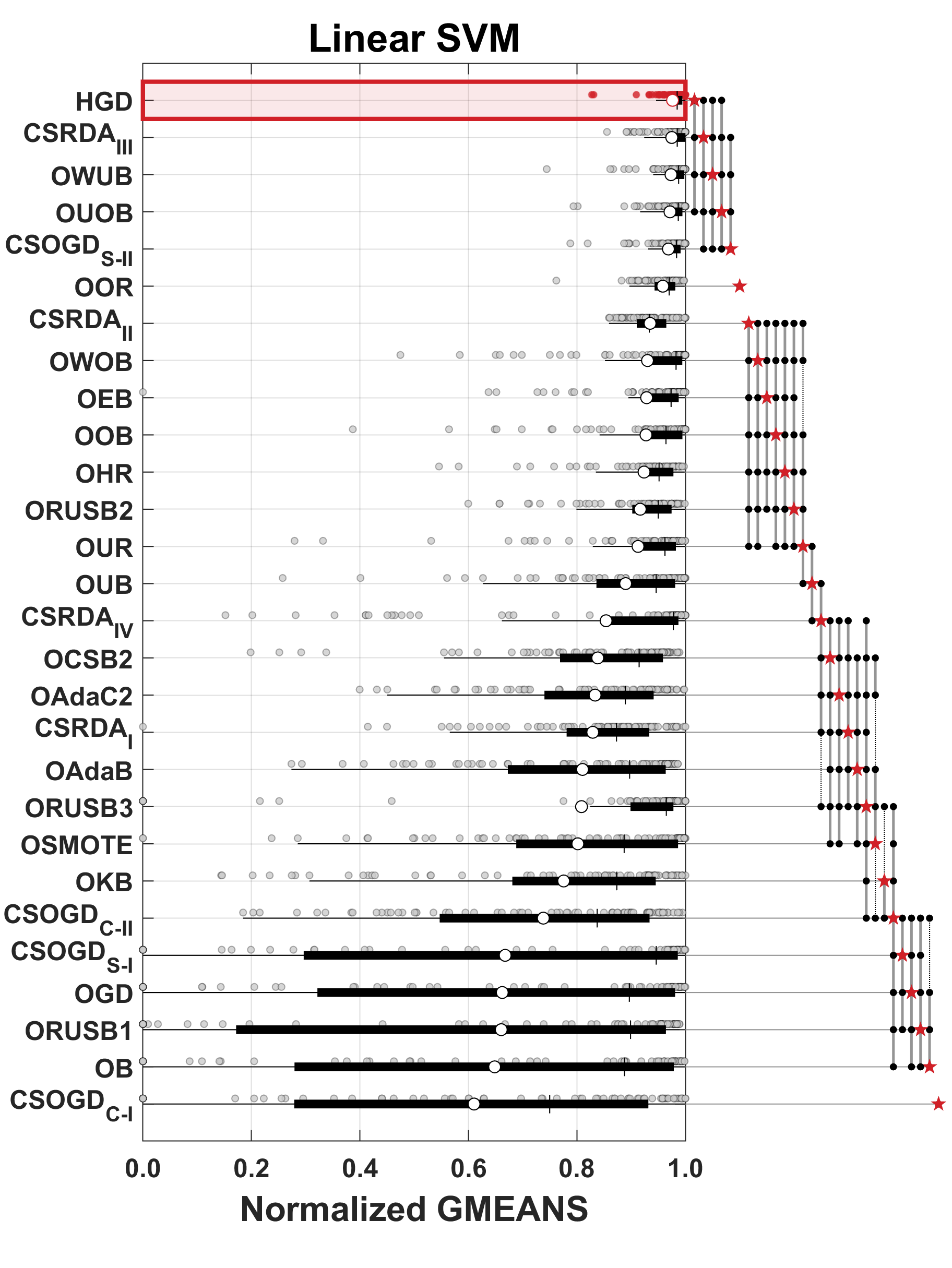}
    }
    \subfloat[Kernel Model]{
        \includegraphics[width=5.5cm]{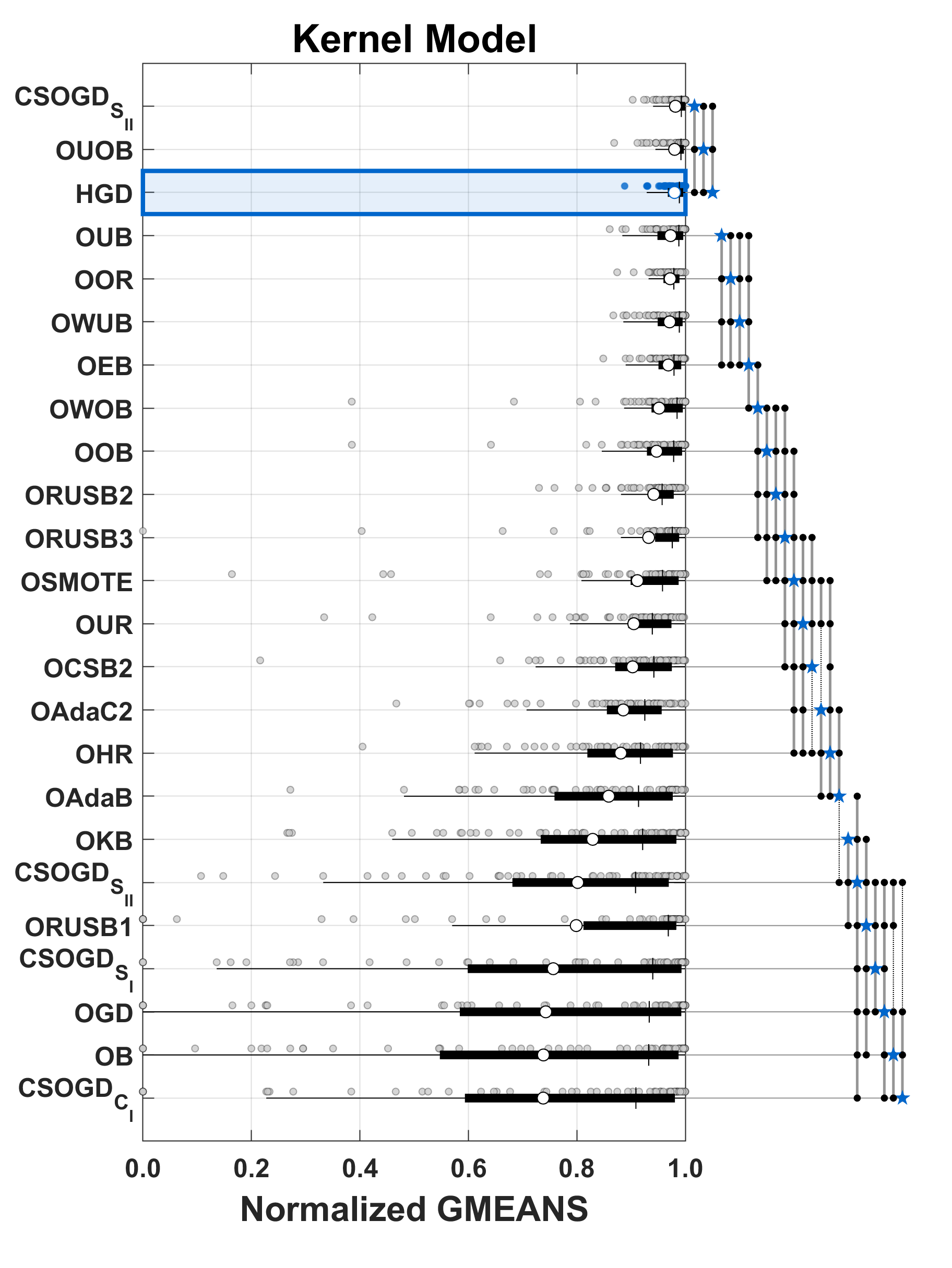}
    } 
    \caption{The performance of different methods on all datasets, in terms of GMEANS. Mean performance is denoted in encircled dot. The boxes are presented in the order of descending from top to bottom. Methods without significant performance difference are connected in the right, tested by the t-test at the significance level of 0.05.}
    \label{fig:static_AllBoxPlot_GMEANS}
\end{figure*}
%===============================================

%===============================================
\begin{figure*}[th]
    \centering
    \subfloat[Perceptron]{
        \includegraphics[width=6cm]{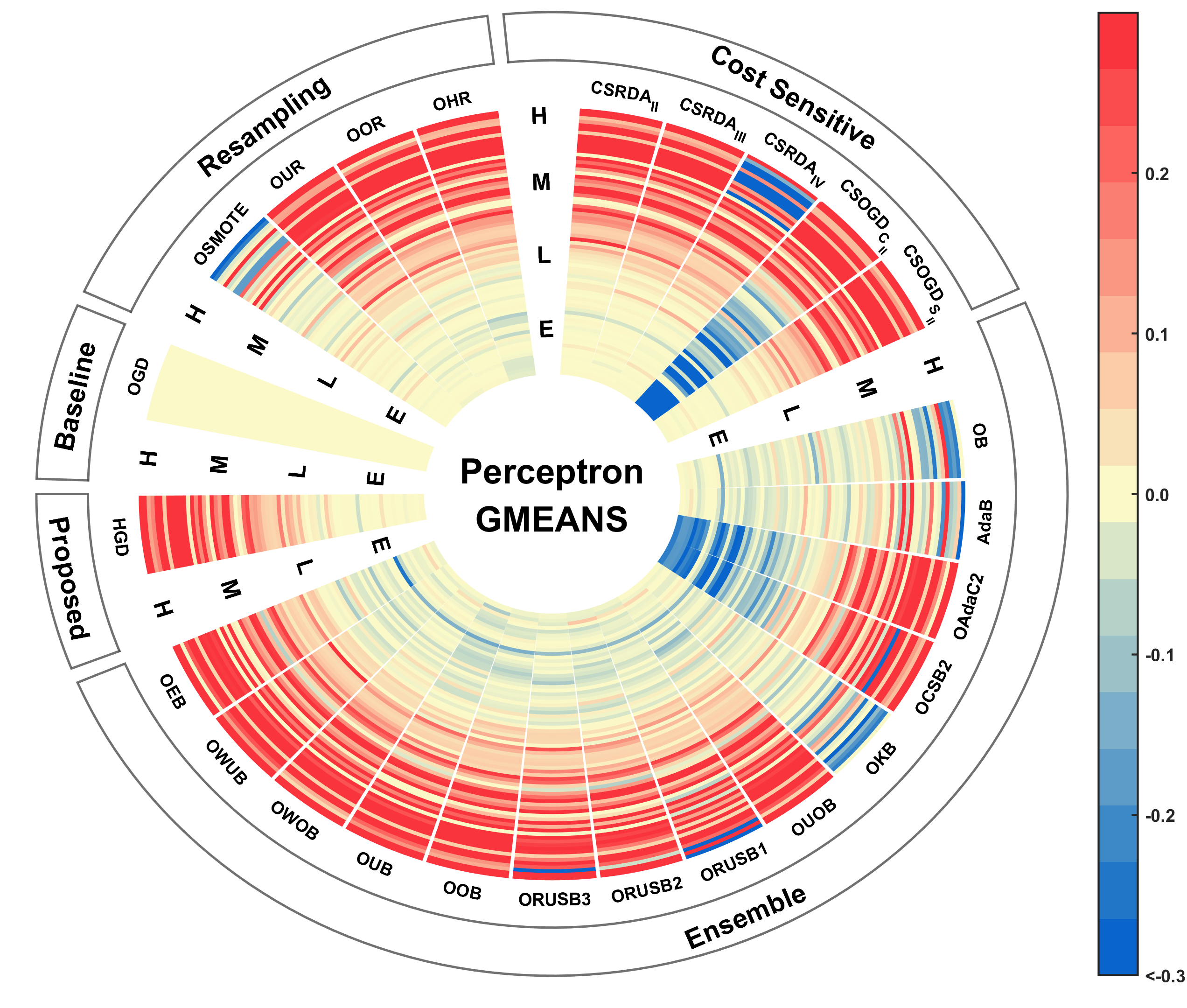}
    }
    \subfloat[Linear SVM]{
        \includegraphics[width=6cm]{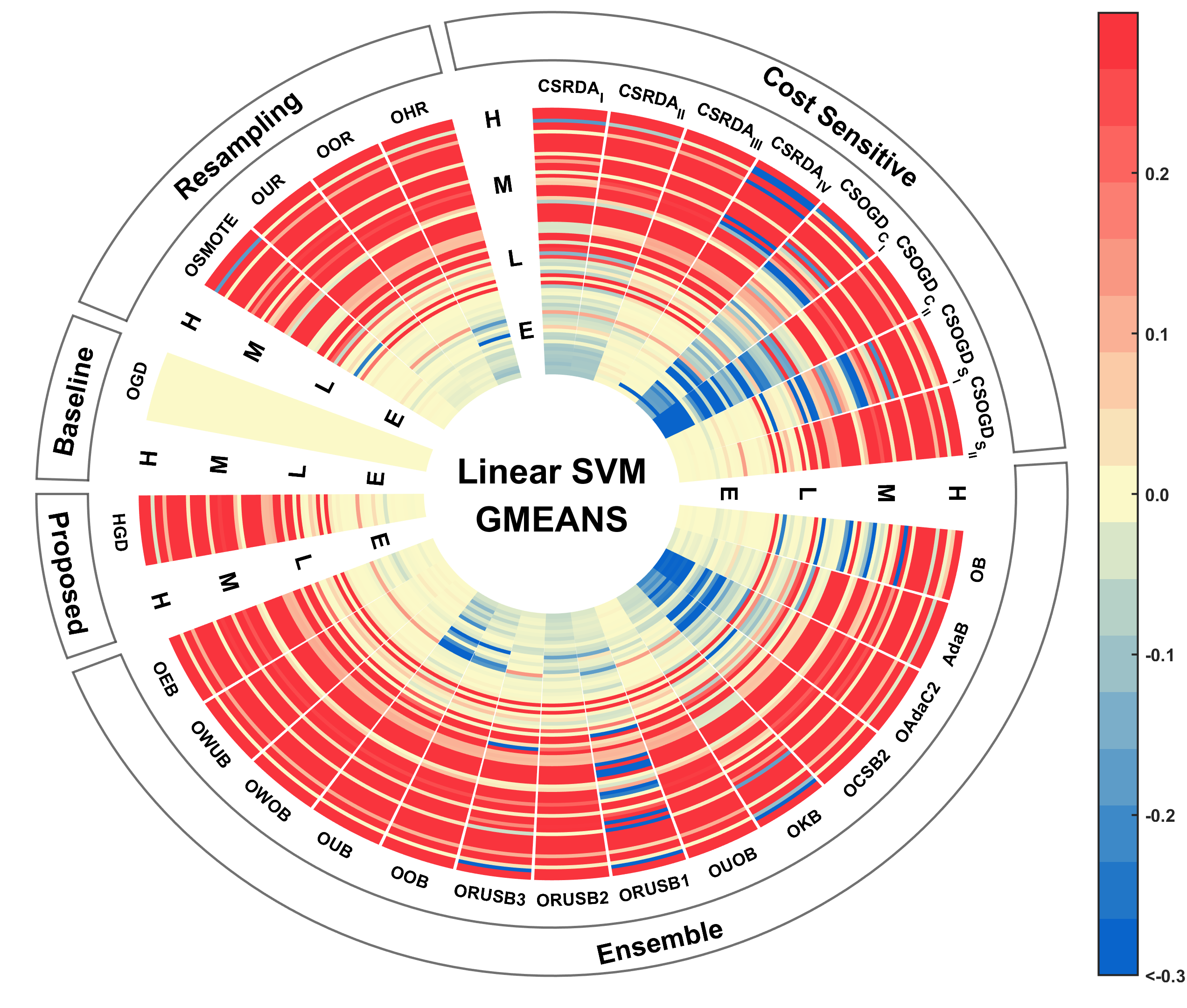}
    }
    \subfloat[Kernel Model]{
        \includegraphics[width=6cm]{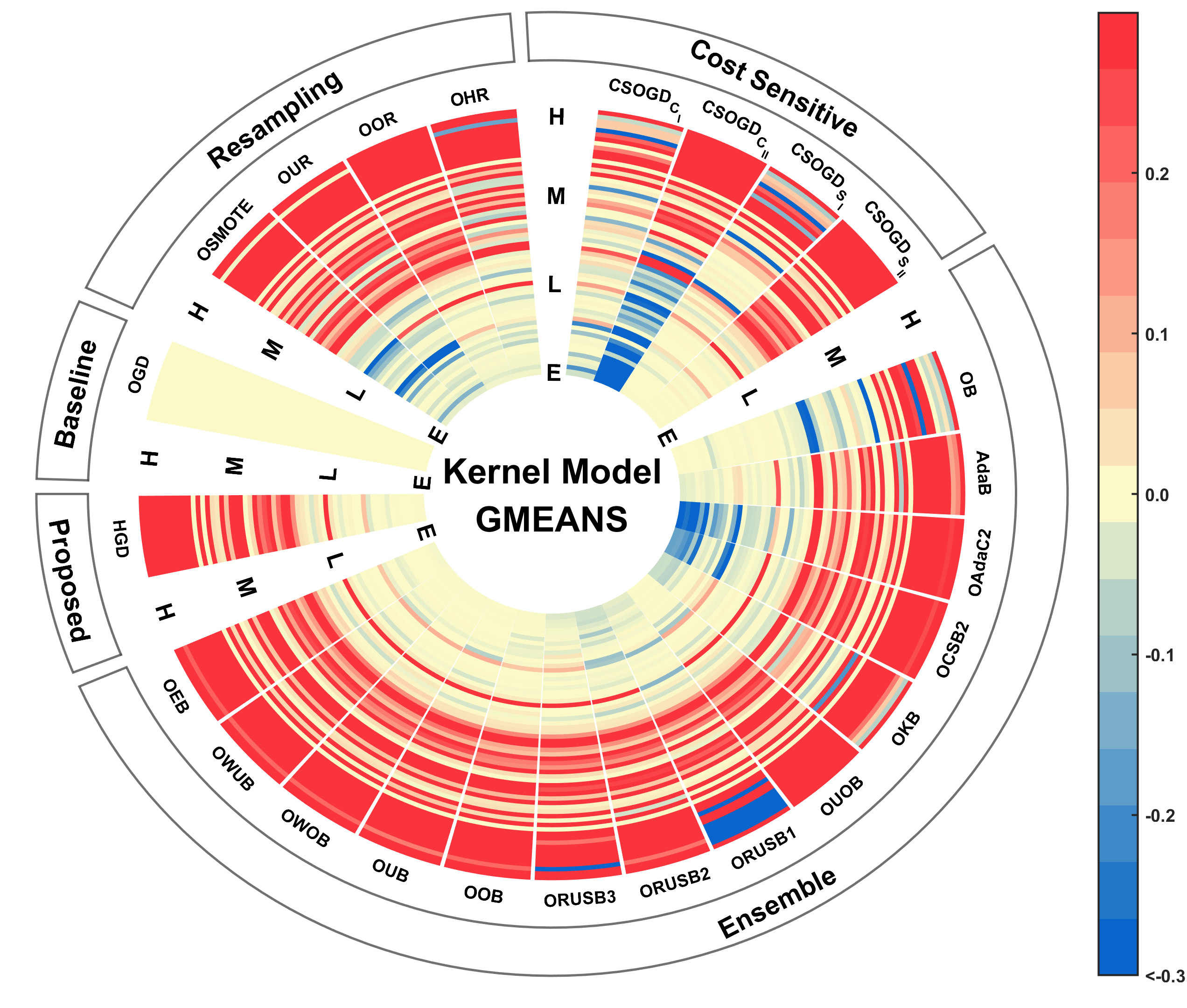}
    }
    \caption{The circular heat maps. Each wedge represents the performance (normalized) of a method on all datasets. The inner segments denote results on datasets with lower imbalance ratios, while the outer segments denote results on datasets with higher imbalance ratios. Blue shades indicate decreasing performance and red shades indicate increasing performance, w.r.t the baseline OGD.}
    \label{fig:static_CirHeatMap}
\end{figure*}
%===============================================

\subsubsection{The effectiveness of HGD} 

We provide the detailed average performance in Table~\ref{tab:cmp_static}, including numerical values for three normalized metrics and the average ranks obtained by different methods across all datasets. The box plot visualizations of the GMEANS results are depicted in Figure~\ref{fig:static_AllBoxPlot_GMEANS}. In each box plot, the left and right edges of the black box represent the 25th and 75th percentiles, respectively. The encircled dot within the box indicates the mean performance. Whiskers extend to the most extreme data points, excluding outliers. The arrangement of the boxes follows a descending order based on mean performance, from top to bottom. We highlighted the proposed HGD in yellow, red and blue when applying the perceptron, the linear SVM and the kernel model as the base learner, respectively. Additionally, the right part of each figure illustrates the performance difference, assessed via the t-test at a significance level of 0.05. Methods exhibiting no significant performance difference are connected by black lines and dots. Additional results can be found in our supplementary material.

Observations from the above results are similar to those in the previous subsection. The proposed HGD is among one of the most effective methods across all three different base learners. However, it is not surprising that the performance of HGD in terms of F1 is slightly less effective, resulting from a performance compromise between positive and negative aspects.

%===============================================
\begin{figure*}[t]
    \centering
    \subfloat[]{
        \includegraphics[width=4.2cm]{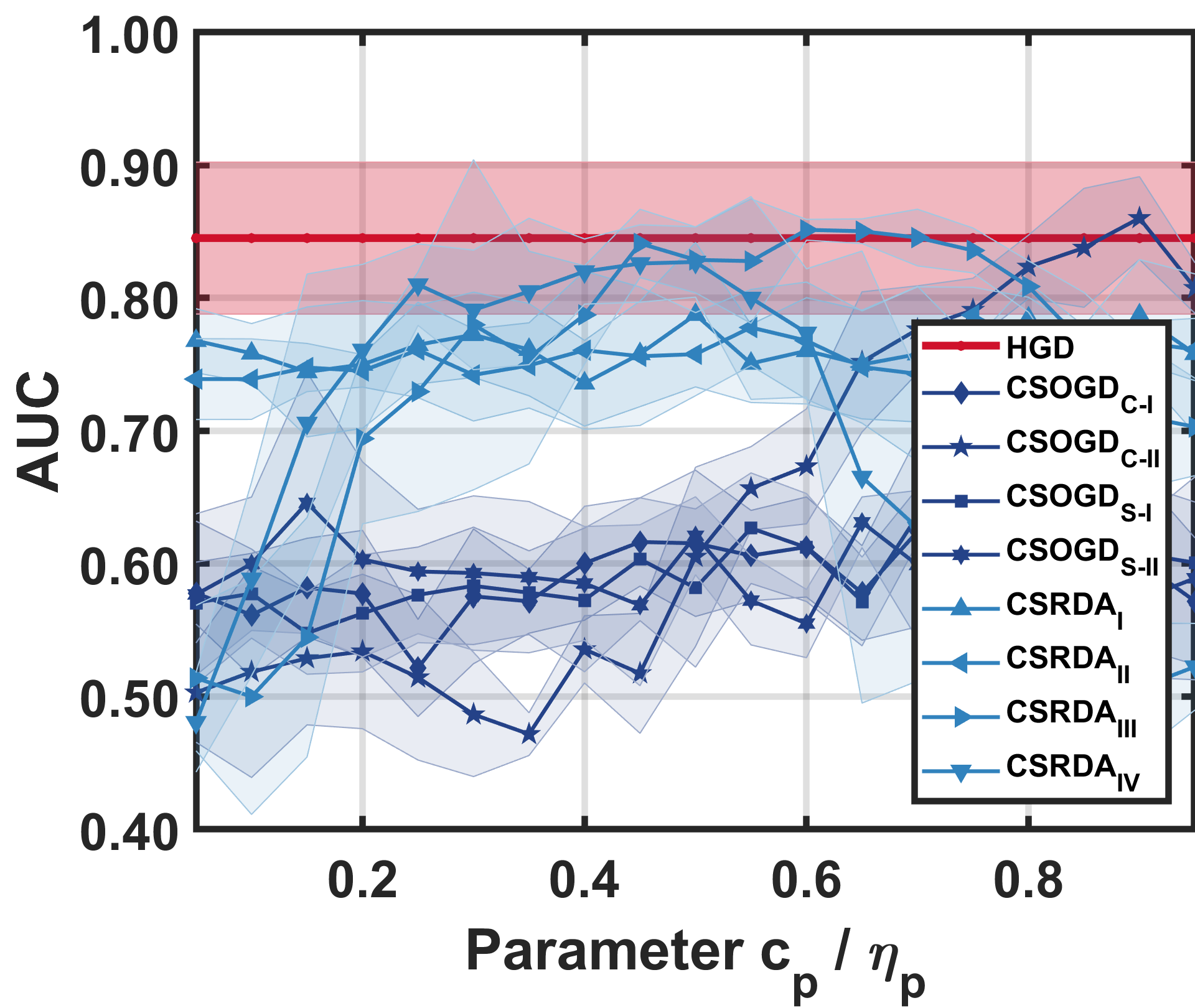}
    } \hspace{-2mm}
    \subfloat[]{
        \includegraphics[width=4.2cm]{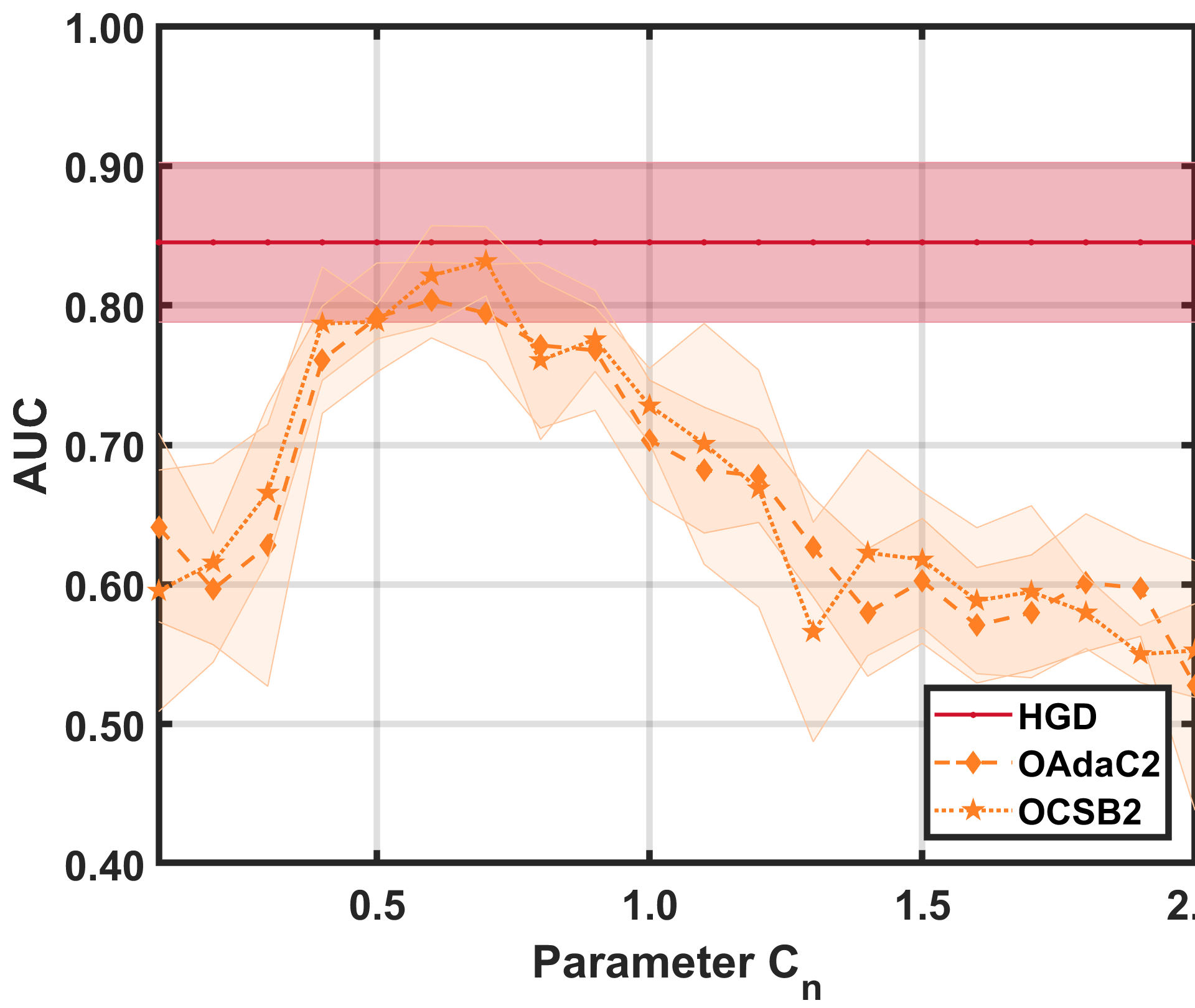}
    } \hspace{-2mm}
    \subfloat[]{
        \includegraphics[width=4.2cm]{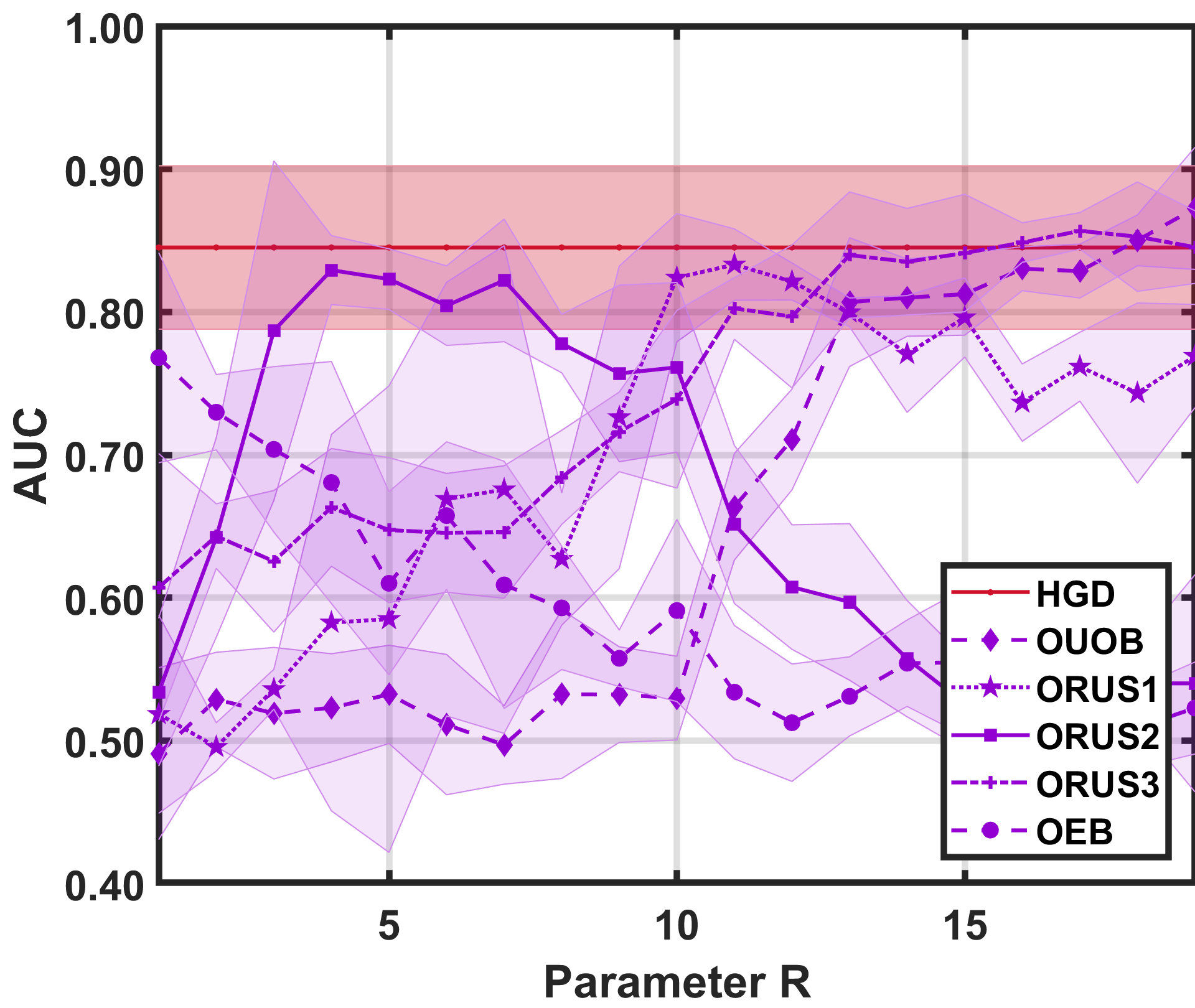}
    } \hspace{-2mm}
    \subfloat[]{
        \includegraphics[width=4.2cm]{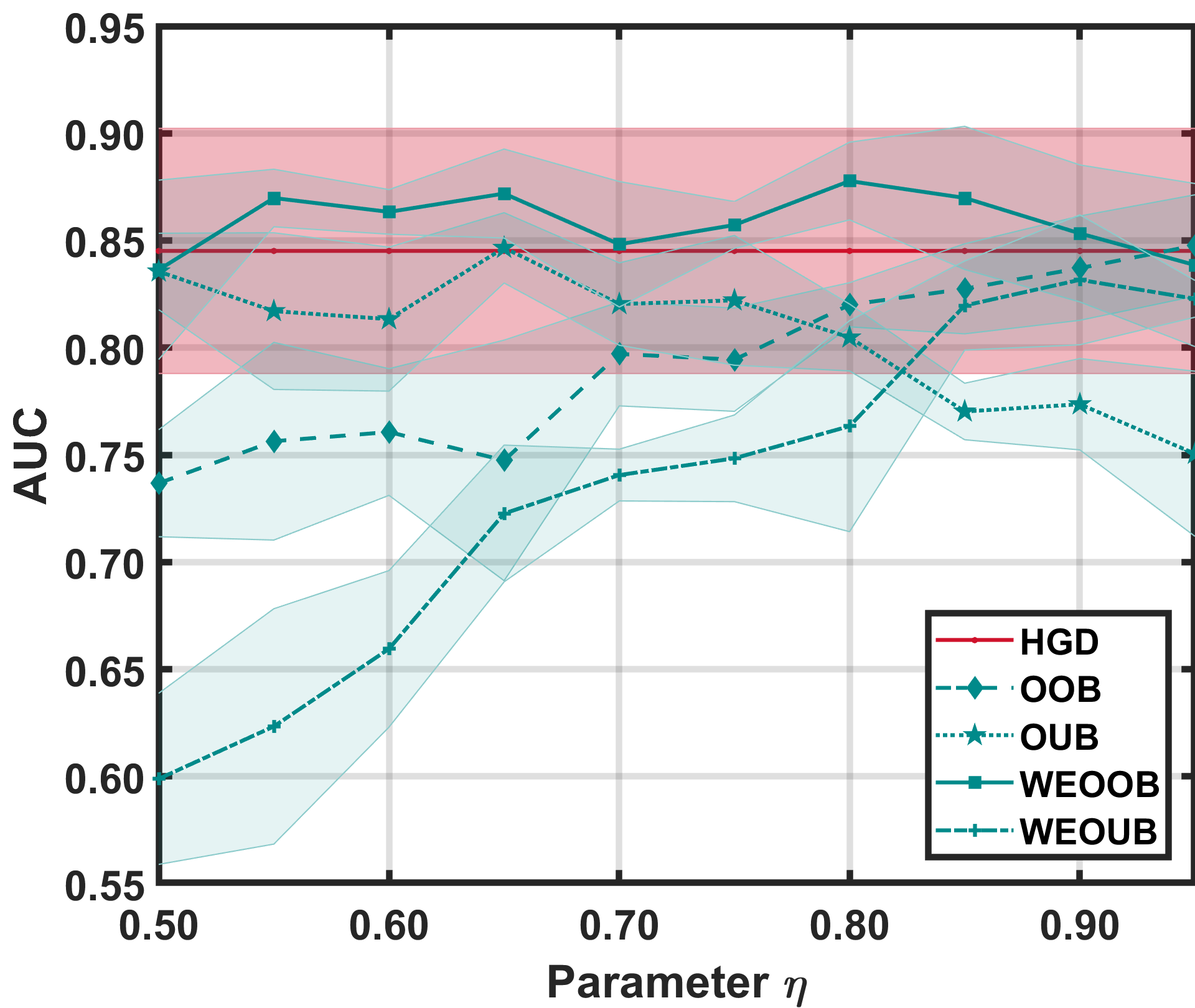}
    }
    \caption{The performance of different methods with varying parameter, in terms of AUC, on the dataset \textit{ecoli}.}
    \label{fig:static_Para}
\end{figure*}
%===============================================

%===============================================
\begin{figure}[th]
    \centering
    \subfloat[Perceptron]{
        \includegraphics[width=5.5cm]{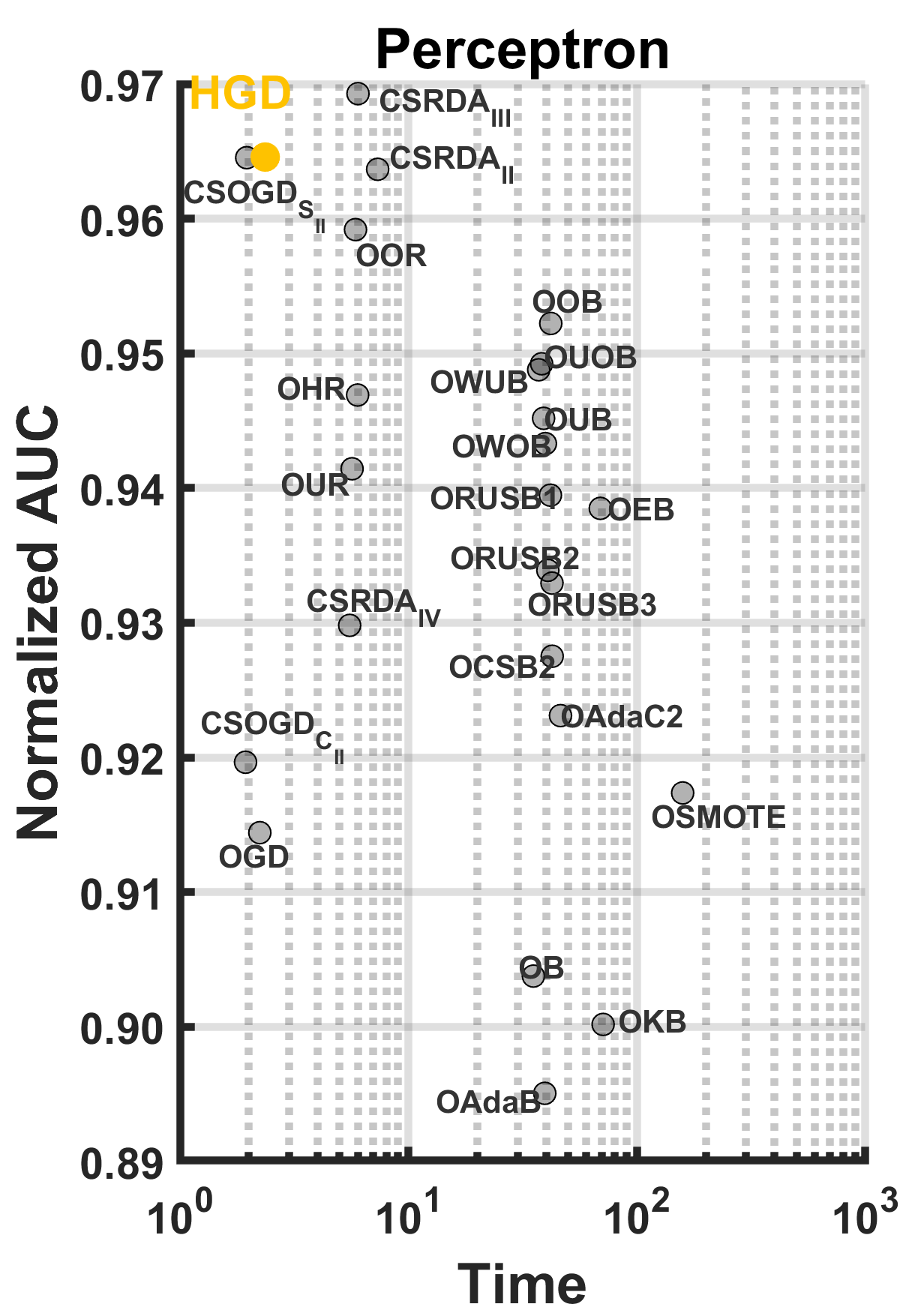}
    }
    \subfloat[Linear SVM]{
        \includegraphics[width=5.5cm]{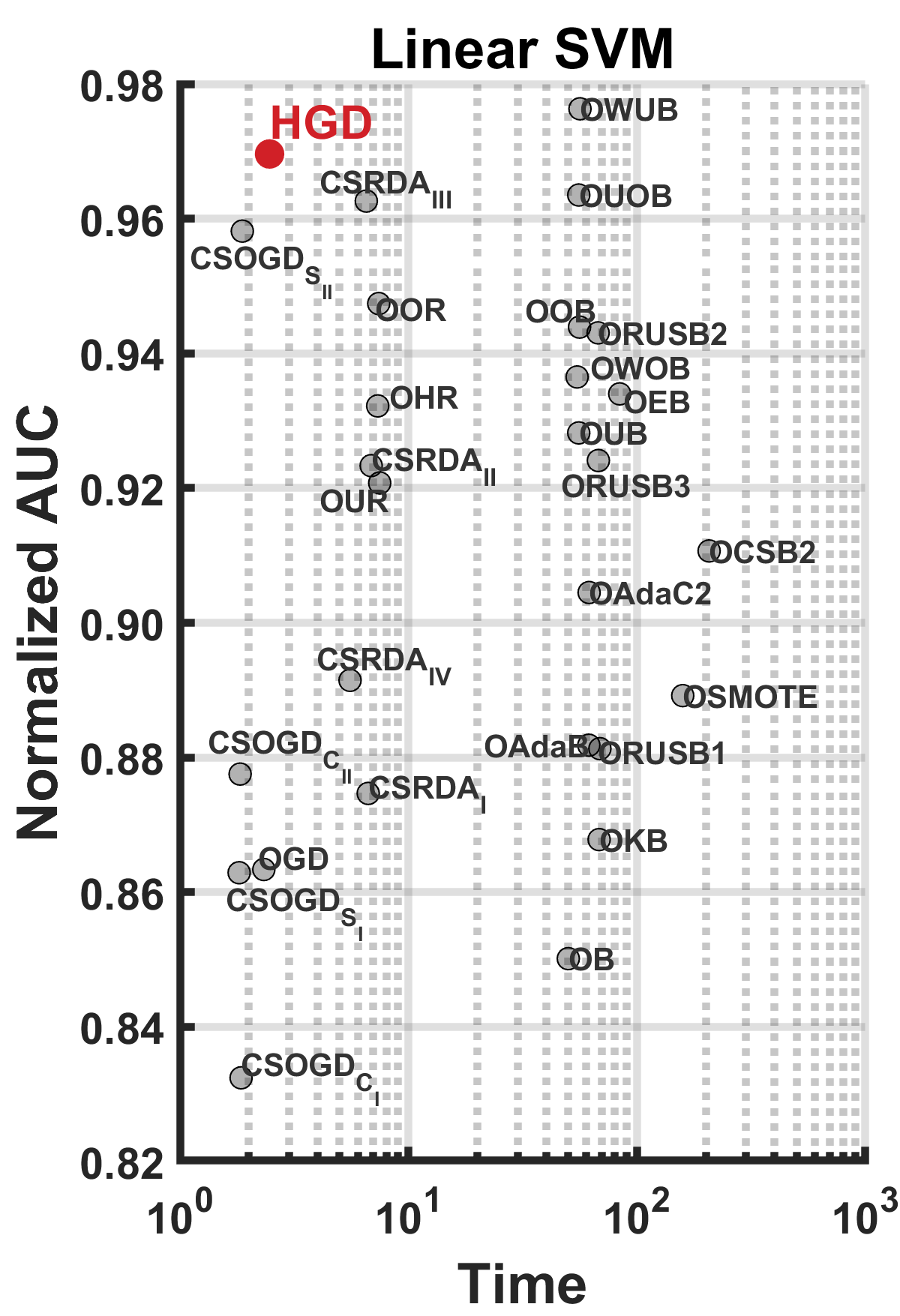}
    }
    \subfloat[Kernel Model]{
        \includegraphics[width=5.5cm]{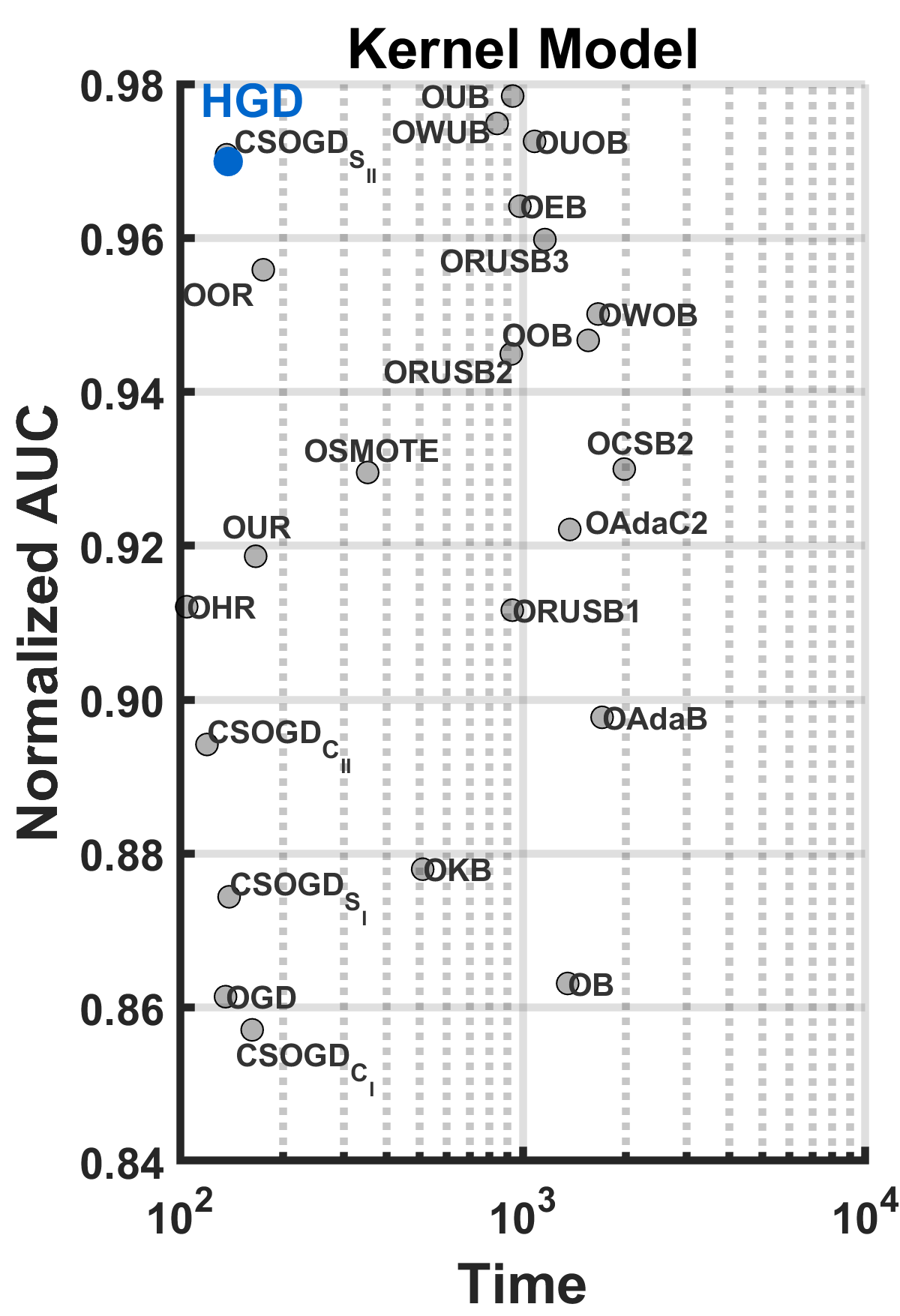}
    }
    \caption{The performance and computational time comparison of all the methods, in terms of AUC. Up-left denotes higher performance with less computational cost.}
    \label{fig:static_TvP_AUC}
\end{figure}
%===============================================

Many techniques may require intricate hyper-parameter tuning to adapt to different imbalance ratios. Instead, our algorithm requires no parameter settings, and thus facilitate its easy implementation on streams with different imbalance ratio. Here, we provide the steady performance achieved by HGD. 
Circular heat maps were generated to visualize the performance, as shown in Figure~\ref{fig:static_CirHeatMap}. In these figures, each wedge represents the performance of a method on all datasets. The inner segments denote results on datasets with lower imbalance ratios, while the outer segments denote results on datasets with higher imbalance ratios. The color of each segment in the circular heat map corresponds to performance variations, with blue shades indicating lower performance and redder shades indicating higher performance. This visualization leads to several interesting observations.
\begin{itemize}
    \item Generally, the performance improvements relative to the baseline increase as the imbalance ratio becomes larger, especially in terms of AUC and GMEANS. This trend demonstrates the effectiveness of these methods in imbalanced data stream learning.
    \item However, the performance of certain methods may exhibit decrements relative to the baseline. This phenomenon arises due to the necessity for precise parameter tuning; if parameters are not well-optimized, performance may suffer. In contrast, the proposed method consistently achieves good performance without any parameter tuning.
\end{itemize}
Additionally, we show the performance attained by different methods over all datasets w.r.t. their Imbalance Ratios (IR, i.e., Equal IR, Low IR, Medium IR and High IR) in Table~\ref{S:tab:cmp_static_p}-\ref{S:tab:cmp_static_v} and Figure~\ref{S:fig:static_BoxPlot_P}-\ref{S:fig:static_BoxPlot_K} in supplementary material.

\subsubsection{The efficiency of HGD}

To comprehensively evaluate the effectiveness and efficiency of the proposed HGD, we conducted a comparative analysis that examines the performance metrics in relation to computational time across all methods considered. The results are depicted in Figure~\ref{fig:static_TvP_AUC}. More results can be found in the supplementary material.
Methods positioned in the upper-left part of the figures denote those achieving superior performance while requiring fewer computational resources. Our findings reveal that HGD is consistently positioned favorably in the upper-left part across all the base learner settings, demonstrating both performance efficacy and computational efficiency.

\subsubsection{The simplicity of HGD}

The fact highlighting the ease of implementation of HGD is its parameter-free nature. 
In Figure~\ref{fig:static_Para}, we present a parameter study comparing various methods. Many competing methods inevitably require fine-tuning of parameters to adapt to data streams with varying degrees of imbalance. Improper parameter configuration may degrade performance. 
Notably, the performance curve of HGD appears as a horizontal red line, indicating that HGD ensures consistent and highly competitive performance without any parameter tuning. This parameter-free characteristic enables the learning strategy of HGD to remain consistent across different scenarios, including both balanced and imbalanced data streams.

%=========================================
\begin{table*}[]
\scriptsize
\centering
\renewcommand\arraystretch{1.05}
\caption{The Averaged Performance (Normalized AUC, G-means and F1-Score) Attained by Different Methods Over All Datasets with Dynamic Imbalance Ratio.}
\resizebox{\linewidth}{!}{
\begin{tabular}{lcccccccccccccccccccc}
\hline
\textbf{Base Learner} & \multicolumn{6}{c}{\textbf{Perceptron}} & \textbf{} & \multicolumn{6}{c}{\textbf{Linear SVM}} & \textbf{} & \multicolumn{6}{c}{\textbf{Kernel Model}} \\ \hline
\multicolumn{1}{l|}{\textbf{Metric}} & \textbf{AUC} & \textbf{Rank} & \textbf{GMEANS} & \textbf{Rank} & \textbf{F1} & \textbf{Rank} & \multicolumn{1}{c|}{\textbf{}} & \textbf{AUC} & \textbf{Rank} & \textbf{GMEANS} & \textbf{Rank} & \textbf{F1} & \textbf{Rank} & \multicolumn{1}{c|}{\textbf{}} & \textbf{AUC} & \textbf{Rank} & \textbf{GMEANS} & \textbf{Rank} & \textbf{F1} & \textbf{Rank} \\ \hline
\multicolumn{21}{l}{\cellcolor[HTML]{C0C0C0}\textbf{Baseline}} \\
\multicolumn{1}{l|}{OGD} & \cellcolor[HTML]{FCFCFF}0.967 & 11.1 & \cellcolor[HTML]{FCFCFF}0.935 & 11.0 & \cellcolor[HTML]{FCFCFF}0.934 & \textbf{9.1} & \multicolumn{1}{c|}{} & \cellcolor[HTML]{FCFCFF}0.930 & 14.3 & \cellcolor[HTML]{FCFCFF}0.770 & 13.0 & \cellcolor[HTML]{FCFCFF}0.749 & \textbf{11.7} & \multicolumn{1}{c|}{} & \cellcolor[HTML]{FCFCFF}0.937 & 15.7 & \cellcolor[HTML]{FCFCFF}0.850 & 13.0 & \cellcolor[HTML]{FCFCFF}0.856 & \textbf{9.7} \\
\multicolumn{21}{l}{\cellcolor[HTML]{C0C0C0}\textbf{Data Level Approaches}} \\
\multicolumn{1}{l|}{OSMOTE} & \cellcolor[HTML]{EFF2FA}0.962 & 12.4 & \cellcolor[HTML]{F5F7FC}0.932 & 12.1 & \cellcolor[HTML]{EEF2FA}0.923 & 9.5 & \multicolumn{1}{c|}{} & \cellcolor[HTML]{FAB1B4}0.955 & 14.2 & \cellcolor[HTML]{FA9A9C}0.912 & 12.0 & \cellcolor[HTML]{F9797B}0.899 & \textbf{10.8} & \multicolumn{1}{c|}{} & \cellcolor[HTML]{FBCDD0}0.953 & 15.4 & \cellcolor[HTML]{FA9597}0.949 & 13.5 & \cellcolor[HTML]{F97072}0.962 & 11.6 \\
\multicolumn{1}{l|}{OUR} & \cellcolor[HTML]{FBC2C5}0.970 & \textbf{8.6} & \cellcolor[HTML]{F98082}0.973 & \textbf{8.5} & \cellcolor[HTML]{C5D5EB}0.892 & 9.4 & \multicolumn{1}{c|}{} & \cellcolor[HTML]{FBD2D5}0.944 & 16.7 & \cellcolor[HTML]{F98587}0.941 & 16.6 & \cellcolor[HTML]{FAAEB0}0.838 & 16.4 & \multicolumn{1}{c|}{} & \cellcolor[HTML]{8BACD7}0.925 & 16.8 & \cellcolor[HTML]{FCF8FB}0.854 & 18.7 & \cellcolor[HTML]{5A8AC6}0.825 & 19.4 \\
\multicolumn{1}{l|}{OOR} & \cellcolor[HTML]{ACC3E2}0.937 & 19.6 & \cellcolor[HTML]{C1D3EA}0.906 & 18.8 & \cellcolor[HTML]{BBCEE8}0.885 & 16.0 & \multicolumn{1}{c|}{} & \cellcolor[HTML]{FA9295}0.965 & 13.8 & \cellcolor[HTML]{F97678}0.963 & 13.0 & \cellcolor[HTML]{FAA9AB}0.844 & 14.9 & \multicolumn{1}{c|}{} & \cellcolor[HTML]{FCF4F7}0.940 & 17.8 & \cellcolor[HTML]{F2F5FB}0.846 & 14.5 & \cellcolor[HTML]{E6ECF7}0.851 & 10.8 \\
\multicolumn{1}{l|}{OHR} & \cellcolor[HTML]{B0C6E4}0.939 & 17.8 & \cellcolor[HTML]{C0D1E9}0.905 & 17.9 & \cellcolor[HTML]{AAC2E2}0.872 & 15.2 & \multicolumn{1}{c|}{} & \cellcolor[HTML]{FA9699}0.963 & 14.6 & \cellcolor[HTML]{F97577}0.964 & 13.2 & \cellcolor[HTML]{F97F82}0.891 & 14.3 & \multicolumn{1}{c|}{} & \cellcolor[HTML]{FBD7DA}0.950 & 17.3 & \cellcolor[HTML]{FBC2C5}0.905 & 16.1 & \cellcolor[HTML]{FBD6D9}0.884 & 14.1 \\
\multicolumn{21}{l}{\cellcolor[HTML]{C0C0C0}\textbf{Cost Sensitive Approaches:}} \\
\multicolumn{1}{l|}{CSRDA$_{I}$} & \cellcolor[HTML]{FA9DA0}0.973 & \textbf{7.0} & \cellcolor[HTML]{F97E80}0.974 & \textbf{7.0} & \cellcolor[HTML]{9AB7DC}0.860 & \textbf{9.0} & \multicolumn{1}{c|}{} & \cellcolor[HTML]{FBD7DA}0.942 & 14.3 & \cellcolor[HTML]{FA9B9D}0.911 & 14.2 & \cellcolor[HTML]{F97C7E}0.895 & 13.1 & \multicolumn{1}{c|}{} & - & - & - & - & - & - \\
\multicolumn{1}{l|}{CSRDA$_{II}$} & \cellcolor[HTML]{F8696B}0.976 & \textbf{7.3} & \cellcolor[HTML]{F8696B}0.980 & \textbf{6.1} & \cellcolor[HTML]{A3BDDF}0.867 & \textbf{8.4} & \multicolumn{1}{c|}{} & \cellcolor[HTML]{FAB2B5}0.954 & 14.5 & \cellcolor[HTML]{F97C7E}0.955 & 13.3 & \cellcolor[HTML]{FBB9BC}0.825 & 15.6 & \multicolumn{1}{c|}{} & - & - & - & - & - & -  \\
\multicolumn{1}{l|}{CSRDA$_{III}$} & \cellcolor[HTML]{9EBADE}0.932 & 10.6 & \cellcolor[HTML]{A3BDDF}0.891 & 10.0 & \cellcolor[HTML]{82A6D4}0.841 & 10.8 & \multicolumn{1}{c|}{} & \cellcolor[HTML]{FA9396}0.964 & 14.5 & \cellcolor[HTML]{F97B7D}0.956 & 16.5 & \cellcolor[HTML]{FAA8AB}0.845 & 16.8 & \multicolumn{1}{c|}{} & - & - & - & - & - & -  \\
\multicolumn{1}{l|}{CSRDA$_{IV}$} & - & - & - & - & - & - & \multicolumn{1}{c|}{} & \cellcolor[HTML]{FCF7FA}0.932 & 17.2 & \cellcolor[HTML]{FBCDD0}0.838 & 16.6 & \cellcolor[HTML]{FAB2B4}0.834 & 16.3 & \multicolumn{1}{c|}{} & - & - & - & - & - & -  \\
\multicolumn{1}{l|}{CSOGD$_{C_I}$} & - & - & - & - & - & - & \multicolumn{1}{c|}{} & \cellcolor[HTML]{5A8AC6}0.876 & 22.0 & \cellcolor[HTML]{5A8AC6}0.711 & 22.9 & \cellcolor[HTML]{5A8AC6}0.681 & 22.0 & \multicolumn{1}{c|}{} & \cellcolor[HTML]{FCFCFF}0.937 & 17.1 & \cellcolor[HTML]{FAFAFE}0.849 & 14.6 & \cellcolor[HTML]{DCE6F4}0.850 & 11.9 \\
\multicolumn{1}{l|}{CSOGD$_{C_{II}}$} & \cellcolor[HTML]{D2DFF0}0.952 & 12.5 & \cellcolor[HTML]{D0DDEF}0.913 & 15.6 & \cellcolor[HTML]{97B5DB}0.857 & 13.7 & \multicolumn{1}{c|}{} & \cellcolor[HTML]{F8F9FD}0.929 & 20.6 & \cellcolor[HTML]{FAAEB1}0.882 & 21.3 & \cellcolor[HTML]{FBBCBF}0.822 & 22.3 & \multicolumn{1}{c|}{} & \cellcolor[HTML]{FBBDC0}0.959 & 12.8 & \cellcolor[HTML]{FAB0B2}0.923 & 16.4 & \cellcolor[HTML]{FCE8EB}0.871 & 18.6 \\
\multicolumn{1}{l|}{CSOGD$_{S_{I}}$} & - & - & - & - & - & - & \multicolumn{1}{c|}{} & \cellcolor[HTML]{F8F9FD}0.929 & 16.0 & \cellcolor[HTML]{FCFBFE}0.772 & 15.6 & \cellcolor[HTML]{FAFAFE}0.748 & 13.8 & \multicolumn{1}{c|}{} & \cellcolor[HTML]{EDF2FA}0.935 & 18.2 & \cellcolor[HTML]{FCFAFD}0.852 & 13.3 & \cellcolor[HTML]{FCFCFF}0.856 & 10.0 \\
\multicolumn{1}{l|}{CSOGD$_{S_{II}}$} & \cellcolor[HTML]{FBBCBF}0.971 & \textbf{7.5} & \cellcolor[HTML]{F9797B}0.975 & \textbf{7.0} & \cellcolor[HTML]{9CB8DD}0.861 & \textbf{8.9} & \multicolumn{1}{c|}{} & \cellcolor[HTML]{F8696B}0.978 & \textbf{10.4} & \cellcolor[HTML]{F96F71}0.974 & \textbf{10.2} & \cellcolor[HTML]{FA9FA1}0.855 & \textbf{11.7} & \multicolumn{1}{c|}{} & \cellcolor[HTML]{F96C6E}0.988 & \textbf{5.5} & \cellcolor[HTML]{F96A6C}0.990 & \textbf{4.4} & \cellcolor[HTML]{FBD5D8}0.885 & \textbf{7.7} \\
\multicolumn{21}{l}{\cellcolor[HTML]{C0C0C0}\textbf{Ensemble Learning Approaches}} \\
\multicolumn{1}{l|}{OB} & \cellcolor[HTML]{C6D6EC}0.947 & 13.9 & \cellcolor[HTML]{FBC0C3}0.953 & 13.1 & \cellcolor[HTML]{7BA1D1}0.836 & 14.3 & \multicolumn{1}{c|}{} & \cellcolor[HTML]{F0F4FB}0.926 & 15.8 & \cellcolor[HTML]{F8F9FD}0.769 & 13.9 & \cellcolor[HTML]{F5F7FC}0.746 & 12.6 & \multicolumn{1}{c|}{} & \cellcolor[HTML]{FCF0F3}0.941 & 18.1 & \cellcolor[HTML]{FAA7A9}0.931 & 19.0 & \cellcolor[HTML]{FA9092}0.938 & 17.7 \\
\multicolumn{1}{l|}{OAdaB} & \cellcolor[HTML]{F8F9FD}0.965 & 10.1 & \cellcolor[HTML]{F98B8D}0.970 & 9.2 & \cellcolor[HTML]{94B2DA}0.855 & 11.0 & \multicolumn{1}{c|}{} & \cellcolor[HTML]{FAAFB1}0.955 & 15.4 & \cellcolor[HTML]{FAA1A3}0.902 & 15.8 & \cellcolor[HTML]{F9898B}0.881 & 13.3 & \multicolumn{1}{c|}{} & \cellcolor[HTML]{F98285}0.980 & 10.7 & \cellcolor[HTML]{F97375}0.981 & 10.3 & \cellcolor[HTML]{FCE2E4}0.876 & 13.5 \\
\multicolumn{1}{l|}{OAdaC2} & \cellcolor[HTML]{F5F7FC}0.964 & 12.2 & \cellcolor[HTML]{FBCBCE}0.950 & 14.5 & \cellcolor[HTML]{ACC4E3}0.873 & 13.8 & \multicolumn{1}{c|}{} & \cellcolor[HTML]{FAAFB1}0.955 & 16.0 & \cellcolor[HTML]{F98184}0.947 & 17.5 & \cellcolor[HTML]{F9888B}0.881 & 17.8 & \multicolumn{1}{c|}{} & \cellcolor[HTML]{FA9496}0.974 & 10.1 & \cellcolor[HTML]{F98486}0.965 & 15.2 & \cellcolor[HTML]{FBBABD}0.906 & 16.9 \\
\multicolumn{1}{l|}{OCSB2} & \cellcolor[HTML]{CFDCEF}0.950 & 13.8 & \cellcolor[HTML]{FCEBEE}0.940 & 15.3 & \cellcolor[HTML]{94B3DA}0.855 & 13.5 & \multicolumn{1}{c|}{} & \cellcolor[HTML]{FA9597}0.964 & 15.3 & \cellcolor[HTML]{F97F81}0.950 & 17.7 & \cellcolor[HTML]{FA9D9F}0.858 & 19.9 & \multicolumn{1}{c|}{} & \cellcolor[HTML]{F9787A}0.984 & 11.0 & \cellcolor[HTML]{F98385}0.965 & 13.6 & \cellcolor[HTML]{FCE3E6}0.875 & 16.1 \\
\multicolumn{1}{l|}{OKB} & \cellcolor[HTML]{5A8AC6}0.907 & 20.6 & \cellcolor[HTML]{5A8AC6}0.854 & 19.7 & \cellcolor[HTML]{5A8AC6}0.811 & 17.4 & \multicolumn{1}{c|}{} & \cellcolor[HTML]{F2F5FB}0.927 & 14.3 & \cellcolor[HTML]{FBD6D9}0.826 & 12.9 & \cellcolor[HTML]{FBD1D4}0.798 & 12.5 & \multicolumn{1}{c|}{} & \cellcolor[HTML]{E4EBF6}0.934 & 13.6 & \cellcolor[HTML]{FCDCDF}0.880 & 11.9 & \cellcolor[HTML]{F2F5FB}0.854 & 10.1 \\
\multicolumn{1}{l|}{OUOB} & \cellcolor[HTML]{D8E2F2}0.954 & 15.9 & \cellcolor[HTML]{FAB3B5}0.958 & 14.6 & \cellcolor[HTML]{7DA2D2}0.837 & 16.3 & \multicolumn{1}{c|}{} & \cellcolor[HTML]{F96A6C}0.978 & \textbf{11.5} & \cellcolor[HTML]{F8696B}0.981 & \textbf{11.5} & \cellcolor[HTML]{FA979A}0.864 & 13.5 & \multicolumn{1}{c|}{} & \cellcolor[HTML]{F8696B}0.989 & \textbf{8.0} & \cellcolor[HTML]{F8696B}0.990 & \textbf{8.7} & \cellcolor[HTML]{FBCFD1}0.890 & 11.5 \\
\multicolumn{1}{l|}{ORUSB1} & \cellcolor[HTML]{BED0E9}0.944 & 15.3 & \cellcolor[HTML]{BBCEE8}0.903 & 17.0 & \cellcolor[HTML]{759DCF}0.832 & 15.9 & \multicolumn{1}{c|}{} & \cellcolor[HTML]{FCE8EB}0.937 & 14.2 & \cellcolor[HTML]{7AA1D1}0.723 & 17.0 & \cellcolor[HTML]{FBC6C8}0.811 & 15.6 & \multicolumn{1}{c|}{} & \cellcolor[HTML]{5A8AC6}0.920 & 9.7 & \cellcolor[HTML]{5A8AC6}0.786 & 15.7 & \cellcolor[HTML]{88AAD6}0.834 & 16.2 \\
\multicolumn{1}{l|}{ORUSB2} & \cellcolor[HTML]{E1E9F5}0.957 & 14.1 & \cellcolor[HTML]{FAB3B5}0.958 & 14.6 & \cellcolor[HTML]{90B0D9}0.852 & 13.9 & \multicolumn{1}{c|}{} & \cellcolor[HTML]{FBB4B6}0.954 & 13.2 & \cellcolor[HTML]{F9898B}0.936 & 14.5 & \cellcolor[HTML]{FA9193}0.872 & 13.4 & \multicolumn{1}{c|}{} & \cellcolor[HTML]{F98C8F}0.976 & 10.1 & \cellcolor[HTML]{F98184}0.967 & 13.0 & \cellcolor[HTML]{FBC1C4}0.900 & 13.1 \\
\multicolumn{1}{l|}{ORUSB3} & \cellcolor[HTML]{A9C1E1}0.936 & 16.1 & \cellcolor[HTML]{CBD9ED}0.911 & 15.9 & \cellcolor[HTML]{6B96CC}0.824 & 16.9 & \multicolumn{1}{c|}{} & \cellcolor[HTML]{FBD4D6}0.943 & \textbf{11.1} & \cellcolor[HTML]{FBC6C8}0.849 & 13.0 & \cellcolor[HTML]{FAA6A8}0.847 & 12.7 & \multicolumn{1}{c|}{} & \cellcolor[HTML]{FCECEF}0.943 & 12.9 & \cellcolor[HTML]{FBD7DA}0.885 & 14.0 & \cellcolor[HTML]{D7E2F2}0.849 & 13.6 \\
\multicolumn{1}{l|}{OOB} & \cellcolor[HTML]{FCF6F9}0.967 & 10.9 & \cellcolor[HTML]{FA9799}0.966 & 12.0 & \cellcolor[HTML]{B6CAE6}0.881 & 12.2 & \multicolumn{1}{c|}{} & \cellcolor[HTML]{F97779}0.973 & \textbf{10.4} & \cellcolor[HTML]{F96E70}0.974 & \textbf{9.6} & \cellcolor[HTML]{F96A6C}0.916 & \textbf{10.1} & \multicolumn{1}{c|}{} & \cellcolor[HTML]{F9888A}0.978 & 10.5 & \cellcolor[HTML]{F98082}0.968 & \textbf{8.2} & \cellcolor[HTML]{F8696B}0.967 & \textbf{7.3} \\
\multicolumn{1}{l|}{OUB} & \cellcolor[HTML]{D8E3F2}0.954 & 14.1 & \cellcolor[HTML]{FBBABD}0.955 & 14.2 & \cellcolor[HTML]{86A9D5}0.844 & 14.7 & \multicolumn{1}{c|}{} & \cellcolor[HTML]{FBC3C6}0.949 & 16.1 & \cellcolor[HTML]{F9898C}0.935 & 17.6 & \cellcolor[HTML]{FA9294}0.870 & 15.8 & \multicolumn{1}{c|}{} & \cellcolor[HTML]{F98C8E}0.976 & \textbf{8.7} & \cellcolor[HTML]{F9787A}0.976 & 10.6 & \cellcolor[HTML]{FAAFB2}0.914 & 11.9 \\
\multicolumn{1}{l|}{OWOB} & \cellcolor[HTML]{F2F5FB}0.963 & 11.3 & \cellcolor[HTML]{FAA1A3}0.963 & 12.1 & \cellcolor[HTML]{B3C8E5}0.878 & 12.4 & \multicolumn{1}{c|}{} & \cellcolor[HTML]{F97F81}0.971 & \textbf{11.0} & \cellcolor[HTML]{F97678}0.964 & \textbf{10.0} & \cellcolor[HTML]{F8696B}0.916 & \textbf{9.3} & \multicolumn{1}{c|}{} & \cellcolor[HTML]{FA9194}0.975 & \textbf{8.9} & \cellcolor[HTML]{F97F81}0.970 & \textbf{7.0} & \cellcolor[HTML]{F97375}0.960 & \textbf{6.7} \\
\multicolumn{1}{l|}{OWUB} & \cellcolor[HTML]{E0E8F5}0.957 & 16.1 & \cellcolor[HTML]{FAA9AC}0.960 & 15.1 & \cellcolor[HTML]{82A6D4}0.842 & 16.7 & \multicolumn{1}{c|}{} & \cellcolor[HTML]{F98C8E}0.967 & 12.9 & \cellcolor[HTML]{F9787A}0.960 & 14.1 & \cellcolor[HTML]{FA9C9E}0.859 & 14.6 & \multicolumn{1}{c|}{} & \cellcolor[HTML]{F97B7D}0.983 & 11.6 & \cellcolor[HTML]{F97678}0.978 & 11.2 & \cellcolor[HTML]{FAA2A5}0.924 & 12.1 \\
\multicolumn{1}{l|}{OEB} & \cellcolor[HTML]{C2D3EA}0.945 & 17.9 & \cellcolor[HTML]{FBD2D5}0.948 & 16.8 & \cellcolor[HTML]{729BCE}0.830 & 17.8 & \multicolumn{1}{c|}{} & \cellcolor[HTML]{FAA0A3}0.960 & 12.8 & \cellcolor[HTML]{F9797B}0.958 & 12.2 & \cellcolor[HTML]{FA9B9D}0.860 & 12.6 & \multicolumn{1}{c|}{} & \cellcolor[HTML]{FA8F91}0.975 & 13.0 & \cellcolor[HTML]{F9797B}0.975 & 11.9 & \cellcolor[HTML]{FAA5A7}0.922 & 11.9 \\
\multicolumn{21}{l}{\cellcolor[HTML]{C0C0C0}\textbf{The Proposed Approach}} \\
\multicolumn{1}{l|}{HGD} & \cellcolor[HTML]{FCF8FB}0.967 & \textbf{8.4} & \cellcolor[HTML]{F97A7C}0.975 & \textbf{7.0} & \cellcolor[HTML]{A3BDDF}0.866 & \textbf{8.4} & \multicolumn{1}{c|}{} & \cellcolor[HTML]{FA8F91}0.966 & 12.8 & \cellcolor[HTML]{F97072}0.972 & \textbf{10.1} & \cellcolor[HTML]{FA9FA1}0.855 & 12.8 & \multicolumn{1}{c|}{} & \cellcolor[HTML]{F97B7D}0.983 & \textbf{6.4} & \cellcolor[HTML]{F96C6E}0.988 & \textbf{5.1} & \cellcolor[HTML]{FBBFC2}0.902 & \textbf{7.5} \\ \hline
\end{tabular}
}
\begin{tablenotes}
    \item{$\cdot$} The color of cells shows the performance increments (in red) or decrements (in blue) w.r.t. the BaseLine (OGD). 
    \item{$\cdot$} The bold entries stand for the top five highest rankings.
\end{tablenotes}
\label{tab:cmp_dy}
\end{table*}
%============================================

\subsection{Learning in Dynamic Settings}

To further study the performance of HGD under dynamic imbalanced data streams, we selected several datasets from Table~\ref{tab:datasets} and simulated data streams featuring dynamic imbalance ratios. A comprehensive details of the dynamic data streams is provided in Table~\ref{S:tab:datasets_dy} in supplementary material. Specifically, our analysis encompassed three types of dynamic imbalance ratios: sudden increase, sudden decrease, and gradual variation, thereby covering distinct characteristics of dynamic imbalance phenomena.

%===============================================
\begin{figure}[t]
    \centering
    \subfloat[Perceptron]{
        \includegraphics[width=5cm]{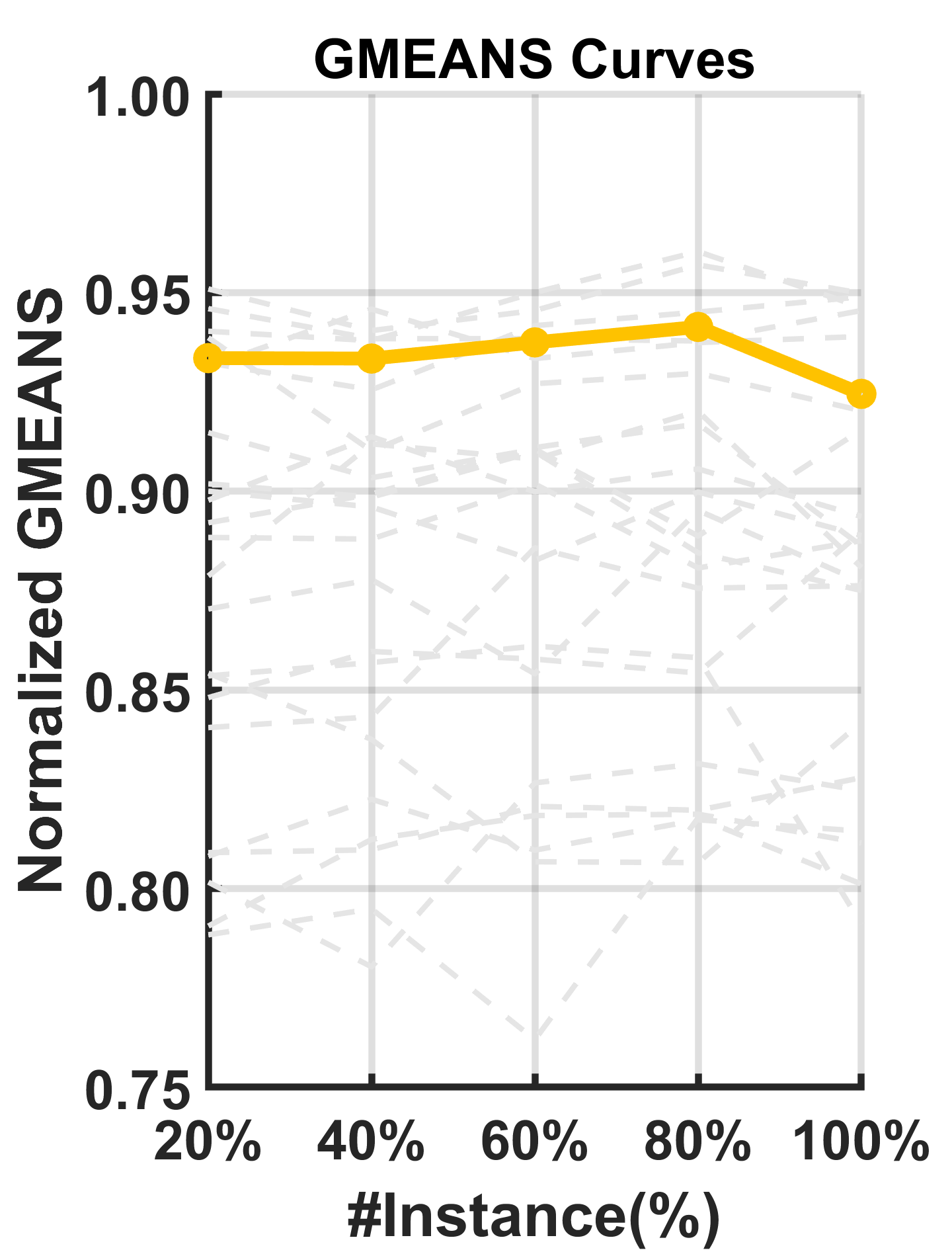}
    }
    \subfloat[Linear SVM]{
        \includegraphics[width=5cm]{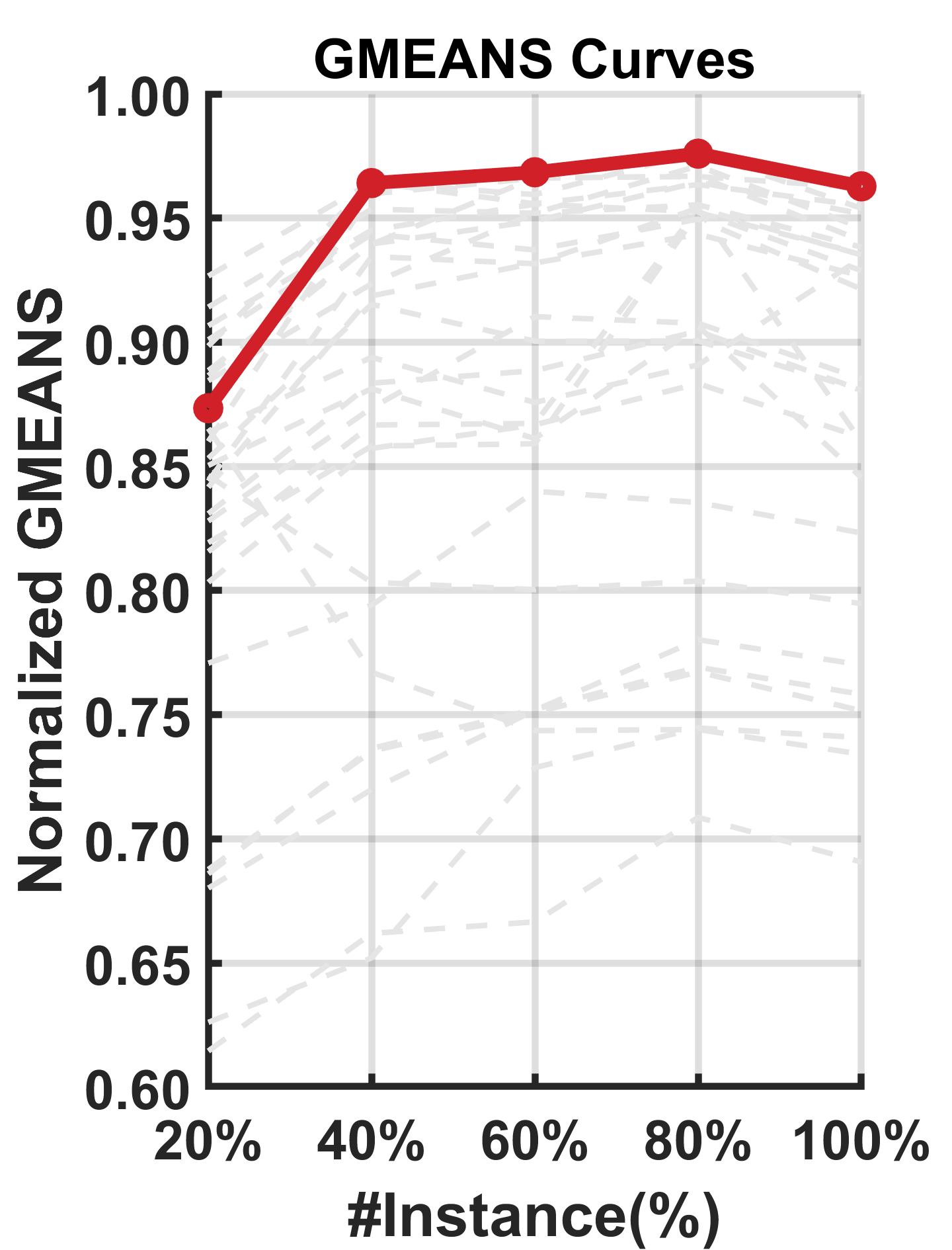}
    }
    \subfloat[Kernel Model]{
        \includegraphics[width=5cm]{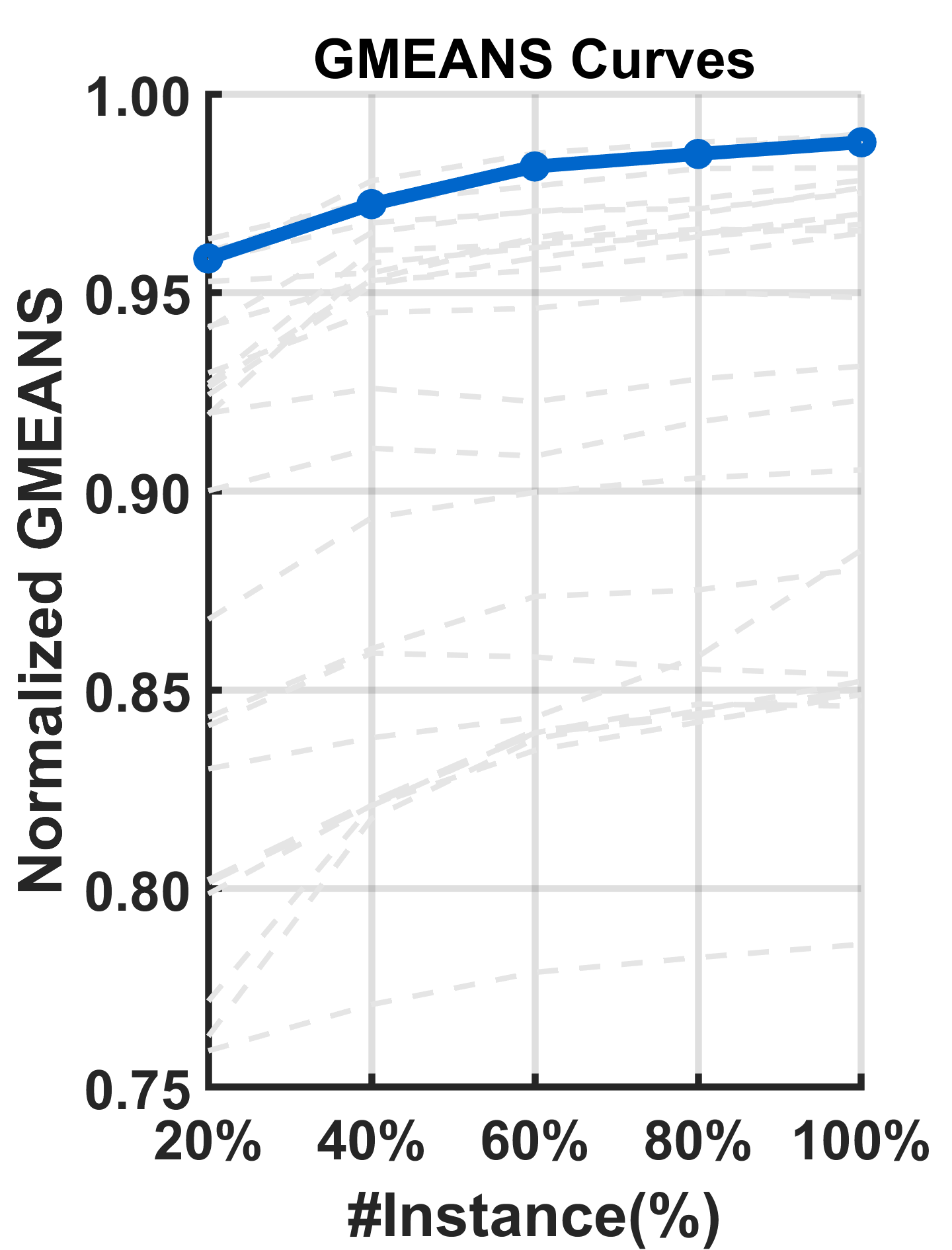}
    }
    \caption{The performance curves of all the methods under dynamic imbalance ratio scenarios, in terms of GMEANS. The proposed HGD is highlighted in yellow, red, and blue when utilizing perceptron, linear SVM and kernel model as the base learner. The gray curves denote competitors.}
    \label{fig:dy_curves_GMEANS}
\end{figure}
%===============================================

The experimental results in dynamic imbalance ratio scenarios are presented in Table~\ref{tab:cmp_dy}. The performance curves throughout the online learning process are depicted in Figure~\ref{fig:dy_curves_GMEANS}. As we can see, HGD demonstrates noteworthy improvements over the baseline and exhibits highly competitive performance compared with competitors across all base learners and metrics. Exceptions can be found in the results evaluated by the F1 metric when utilizing the perceptron as the base learner. This performance decrements arises due to the method's attempt towards identifying the minority class, resulting in an inevitable trade-off.

%===============================================
\begin{figure}[t]
    \centering
    \includegraphics[width=4.25cm]{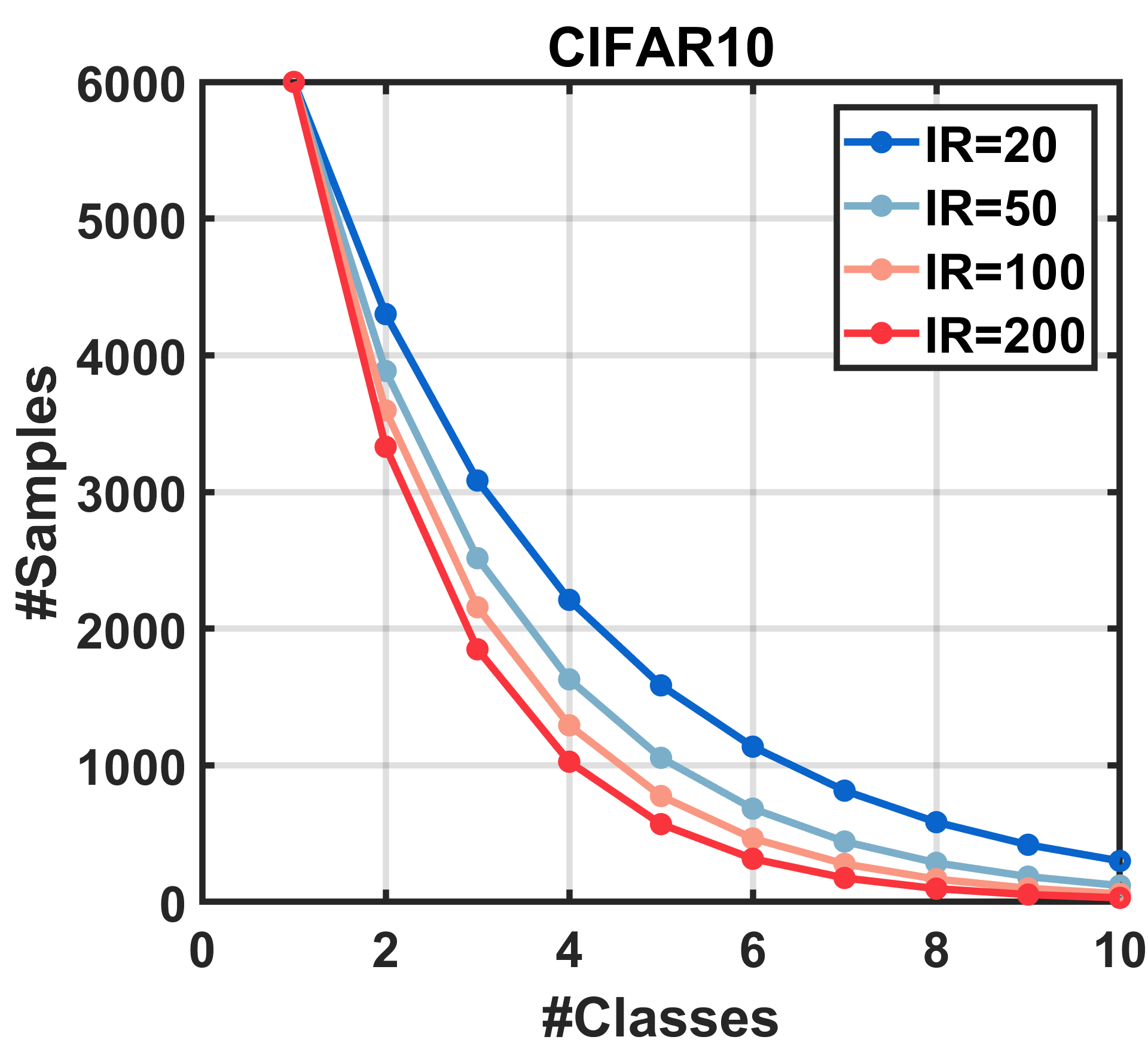} 
    \includegraphics[width=4.25cm]{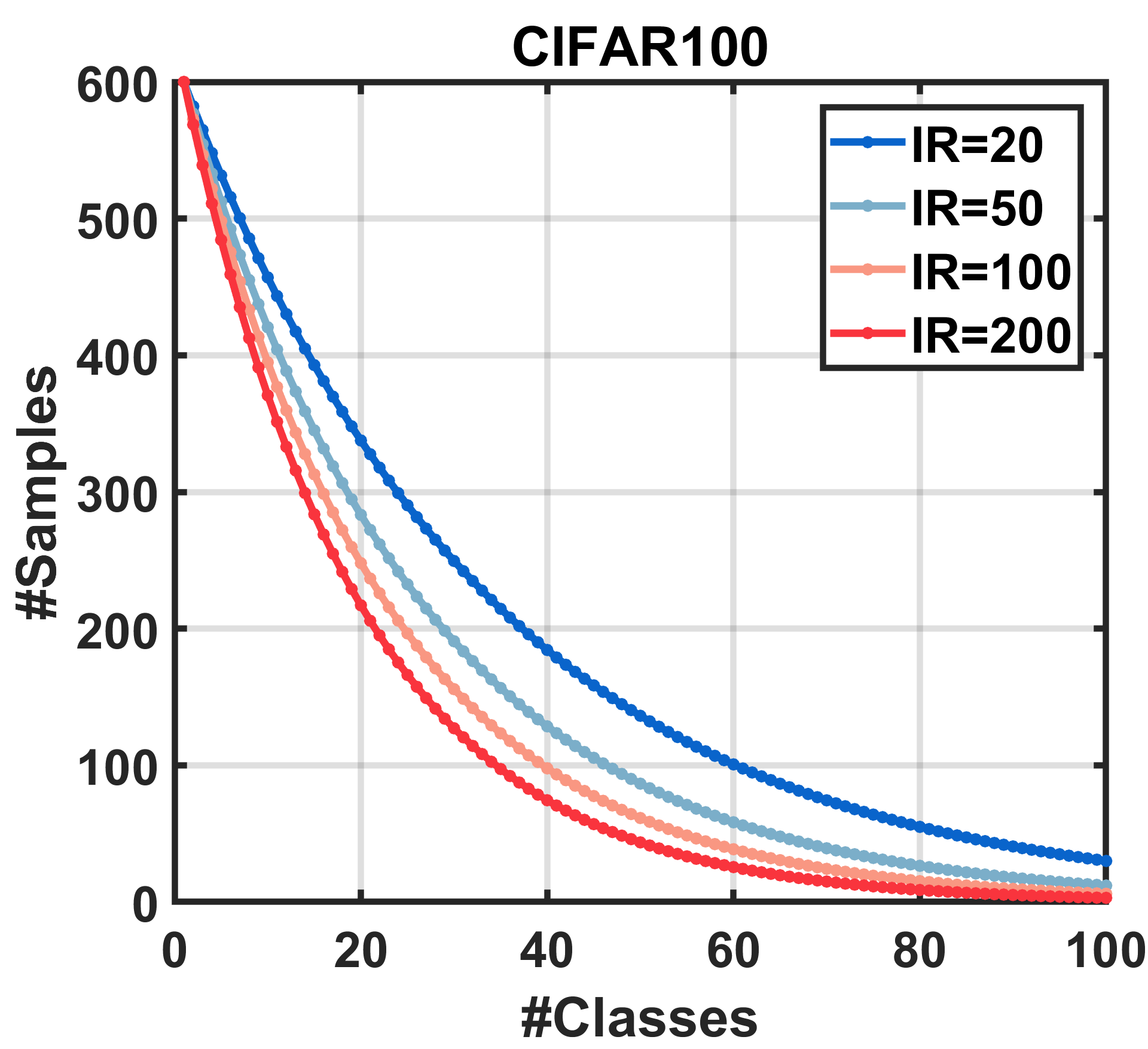}
    \caption{The distributions of training samples per class in long-tailed CIFAR 10/100 data streams under different imbalance ratio.}
    \label{fig:dl_cifar}
\end{figure}
%===============================================

%=======================================
\begin{table*}[t]
\centering
\scriptsize
\renewcommand\arraystretch{1.1}
\caption{The Top-1 Test Accuracy on Long-Tailed CIFAR10/CIFAR100 Data Streams under Different Imbalance Ratios. The Bold Entries Stand for the Best Performance.}
\begin{tabular}{lcccccccc}
\hline
\multicolumn{1}{l|}{\textbf{Datasets}} & \multicolumn{4}{c|}{\textbf{CIFAR10}} & \multicolumn{4}{c}{\textbf{CIFAR100}} \\ \hline
\multicolumn{1}{l|}{\textbf{IR}} & \textbf{200} & \textbf{100} & \textbf{50} & \multicolumn{1}{c|}{\textbf{20}} & \textbf{200} & \textbf{100} & \textbf{50} & \textbf{20} \\ \hline
\multicolumn{9}{l}{\cellcolor[HTML]{C0C0C0}\textbf{Baseline}} \\
\multicolumn{1}{l|}{ResNet32} & 0.635 & 0.699 & 0.747 & \multicolumn{1}{c|}{0.813} & 0.342 & 0.387 & 0.432 & 0.497 \\
\multicolumn{9}{l}{\cellcolor[HTML]{C0C0C0}\textbf{Cost Sensitive Approach}} \\
\multicolumn{1}{l|}{Focal Loss} & 0.641 & 0.691 & 0.747 & \multicolumn{1}{c|}{0.822} & 0.339 & 0.383 & 0.423 & 0.510 \\
\multicolumn{9}{l}{\cellcolor[HTML]{C0C0C0}\textbf{Data Level Approches}} \\
\multicolumn{1}{l|}{ROS} & 0.634 & 0.701 & 0.754 & \multicolumn{1}{c|}{0.822} & 0.341 & 0.397 & 0.441 & 0.517 \\
\multicolumn{1}{l|}{OMCQ} & 0.637 & 0.693 & 0.758 & \multicolumn{1}{c|}{0.824} & 0.323 & 0.366 & 0.419 & 0.487 \\
\multicolumn{1}{l|}{CSMOTE} & 0.649 & 0.702 & 0.750 & \multicolumn{1}{c|}{0.815} & 0.352 & 0.394 & 0.437 & 0.505 \\
\multicolumn{9}{l}{\cellcolor[HTML]{C0C0C0}\textbf{Training Modification Approaches}} \\
\multicolumn{1}{l|}{MixGradient} & \textbf{0.727} & 0.755 & 0.794 & \multicolumn{1}{c|}{0.844} & \textbf{0.381} & \textbf{0.425} & 0.474 & \textbf{0.536} \\
\multicolumn{9}{l}{\cellcolor[HTML]{C0C0C0}\textbf{The Proposed Approach}} \\
\multicolumn{1}{l|}{HGD} & 0.713 & \textbf{0.763} & \textbf{0.795} & \multicolumn{1}{c|}{\textbf{0.862}} & 0.363 & 0.414 & \textbf{0.484} & 0.531 \\ \hline
\end{tabular}
\begin{tablenotes}
    \item{$\cdot$} The accuracy is suitable for performance measurement in this experiment since the test set is balanced.
\end{tablenotes}
\label{tab:cmp_dl}
\end{table*}
%=======================================

\subsection{Extension to Deep Neural Networks}

We proceed to extend the application of HGD to deep neural network training, showing its straightforward implementation across various machine learning models that utilize gradient descent optimization.

We selected CIFAR10/100 as evaluation datasets. The original datasets are balanced, containing 60000 images evenly distributed across classes (6000 for each class in CIFAR10; 600 for each class in CIFAR100). Following the approach outlined in previous research \cite{Cui2019Longtail}, long-tail CIFAR10/100 datasets are generated by removing samples per class from the training set. We assigned the imbalance ratio to range from 20 to 200, with the distributions of training samples per class depicted in Figure~\ref{fig:dl_cifar}. Subsequently, the long-tailed CIFAR datasets were divided into 10 sequential data chunks to simulate the long-tailed CIFAR data streams.

We selected ResNet32 \cite{He2016Resnet} as the backbone to learn the data streams. In addition to the backbone serving as a competitor, other competitive methods include Focal Loss \cite{Lin2017Focal}, Random Oversampling \cite{Chen2013RO}, Online MC-Queue (OMCQ) \cite{Sadeghi2021MCQueue}, Continuous SMOTE \cite{Bernardo2020CSMOTE}, and MixGradient \cite{peng2023mixgradient}. Their performance metrics were adopted directly from prior research \cite{peng2023mixgradient}. Since the samples in the test set are evenly categorized, we chose top-1 accuracy as the metric to evaluate the performance. The results are provided in Table~\ref{tab:cmp_dl}. Notably, HGD demonstrates an ability to enhance the baseline performance across imbalanced data streams under all imbalance ratios. Furthermore, we do not claim that our method always achieves the best performance; however, it consistently demonstrates competitive results across multiple imbalance ratios on the two datasets analyzed.

\section{Conclusion}
\label{sec:Conclusion}

This study tackles the challenge of imbalanced data stream learning by refining the training process, particularly within the gradient descent framework. Our findings reveal that the biased performance of traditional methods stems from gradient steps dominated by the majority class. This dominance leads to classifier weights favoring configurations that align with the majority class, consequently neglecting adequate fitting for the minority class. In this situation, we develop a weighted gradient descent step to harmonize gradient imbalances, resulting in the HGD algorithm. The proposed HGD algorithm operates without the need for data buffers, additional parameters, or prior knowledge, and treats imbalanced data streams equivalently to balanced ones. This enables its straightforward implementation for any learning model that utilizes gradient descent for optimization. By theoretical analysis, we prove that the algorithm demonstrates a satisfactory sublinear regret of $\mathcal{O}(\sqrt{T})$. Extensive experimental results across various scenarios highlight its competitive performance in online imbalanced data stream learning.

\bibliographystyle{unsrt}
\bibliography{ref}

\clearpage
\newpage

\begin{center}
{\fontsize{22}{14}\selectfont \textbf{SUPPLEMENTARY MATERIAL}}
\end{center}

\section{Results: Learning Under Static Imbalanced Data Streams}

\subsection{Datasets and Competitors}

Seventy-two public datasets with different imbalance ratios were selected to evaluate the performance of the proposed method. 
Each instance in these datasets were provided sequentially, one at a time, in a randomly shuffled order, to stimulate one-pass data streams. Table~\ref{S:tab:datasets} summarizes the details of the datasets, including the number of instances, the number of features, and the imbalance ratio. In particular, we empirically divided the datasets into four groups regarding their imbalance ratio, encompassing Equal (E), Low (L), Medium (M) and High (H) imbalance groups.

%========================================
\begin{table*}[h]
\centering
\renewcommand\arraystretch{0.9}
\caption{The Selected Datasets for Experiments}
\resizebox{\linewidth}{!}{
\begin{tabular}{cccccc|cccccc|cccccc}
\hline
\textbf{No.} & \textbf{Dataset} & \textbf{\#Instances} & \textbf{\#Fea} & \textbf{IR} & \textbf{Group} & \textbf{No.} & \textbf{Dataset} & \textbf{\#Instances} & \textbf{\#Fea} & \textbf{IR} & \textbf{Group} & \textbf{No.} & \textbf{Dataset} & \textbf{\#Instances} & \textbf{\#Fea} & \textbf{IR} & \multicolumn{1}{c}{\textbf{Group}} \\ \hline
\textbf{D1} & bank & 10578 & 7 & 1 & E & \textbf{D25} & diabetes & 768 & 8 & 1.87 & L & \textbf{D49} & pc4 & 1458 & 37 & 7.19 & M \\
\textbf{D2} & credit & 16714 & 10 & 1 & E & \textbf{D26} & steel-plates-fault & 1941 & 33 & 1.88 & L & \textbf{D50} & yeast3 & 1484 & 8 & 8.1 & M \\
\textbf{D3} & default & 13272 & 20 & 1 & E & \textbf{D27} & breast-w & 699 & 9 & 1.9 & L & \textbf{D51} & pc3 & 1563 & 37 & 8.77 & M \\
\textbf{D4} & electricity & 38474 & 7 & 1 & E & \textbf{D28} & BNG & 39366 & 9 & 1.91 & L & \textbf{D52} & pageblocks0 & 5472 & 10 & 8.79 & M \\
\textbf{D5} & heloc & 10000 & 22 & 1 & E & \textbf{D29} & QSAR & 1055 & 41 & 1.96 & L & \textbf{D53} & ijcnn1 & 49990 & 22 & 9.3 & M \\
\textbf{D6} & house16H & 13488 & 16 & 1 & E & \textbf{D30} & pol & 15000 & 48 & 1.98 & L & \textbf{D54} & kc3 & 458 & 39 & 9.65 & M \\
\textbf{D7} & jannis & 57580 & 54 & 1 & E & \textbf{D31} & rna & 59535 & 8 & 2 & L & \textbf{D55} & HTRU2 & 17898 & 8 & 9.92 & M \\
\textbf{D8} & jannisSample & 2000 & 54 & 1 & E & \textbf{D32} & vertebra & 310 & 6 & 2.1 & M & \textbf{D56} & vowel0 & 988 & 13 & 9.98 & M \\
\textbf{D9} & twonorm & 7400 & 20 & 1 & E & \textbf{D33} & phoneme & 5404 & 5 & 2.41 & M & \textbf{D57} & modelUQ & 540 & 20 & 10.74 & H \\
\textbf{D10} & Friedman & 1000 & 25 & 1.01 & L & \textbf{D34} & yeast1 & 1484 & 8 & 2.46 & M & \textbf{D58} & led7digit & 443 & 7 & 10.97 & H \\
\textbf{D11} & equity & 96320 & 21 & 1.02 & L & \textbf{D35} & ILPD & 583 & 10 & 2.49 & M & \textbf{D59} & mw1 & 403 & 37 & 12 & H \\
\textbf{D12} & ringnorm & 7400 & 20 & 1.02 & L & \textbf{D36} & MiniBooNE & 130064 & 50 & 2.56 & M & \textbf{D60} & shuttle04 & 1829 & 9 & 13.87 & H \\
\textbf{D13} & stock & 950 & 9 & 1.06 & L & \textbf{D37} & SPECTF & 349 & 44 & 2.67 & M & \textbf{D61} & Ozone & 2534 & 72 & 14.84 & H \\
\textbf{D14} & mushroom & 8124 & 21 & 1.07 & L & \textbf{D38} & AutoUniv & 1000 & 20 & 2.86 & M & \textbf{D62} & Sick & 3772 & 29 & 15.33 & H \\
\textbf{D15} & splice & 3175 & 60 & 1.08 & L & \textbf{D39} & ada & 4147 & 48 & 3.03 & M & \textbf{D63} & ecoli4 & 336 & 7 & 15.8 & H \\
\textbf{D16} & higgs & 98050 & 28 & 1.12 & L & \textbf{D40} & a8a & 32561 & 123 & 3.15 & M & \textbf{D64} & spills & 937 & 49 & 21.85 & H \\
\textbf{D17} & eye & 14980 & 14 & 1.23 & L & \textbf{D41} & a9a & 48842 & 123 & 3.18 & M & \textbf{D65} & yeast4 & 1484 & 8 & 28.1 & H \\
\textbf{D18} & banknote & 1372 & 4 & 1.25 & L & \textbf{D42} & svmguide3 & 1243 & 21 & 3.2 & M & \textbf{D66} & w8a & 64700 & 300 & 32.47 & H \\
\textbf{D19} & svmguide1 & 7089 & 4 & 1.29 & L & \textbf{D43} & FRMTC & 748 & 4 & 3.2 & M & \textbf{D67} & yeast5 & 1484 & 8 & 32.73 & H \\
\textbf{D20} & bank8FM & 8192 & 8 & 1.48 & L & \textbf{D44} & vehicle0 & 846 & 18 & 3.25 & M & \textbf{D68} & yeast6 & 1484 & 8 & 41.4 & H \\
\textbf{D21} & spambase & 4601 & 57 & 1.54 & L & \textbf{D45} & Click & 39948 & 11 & 4.94 & M & \textbf{D69} & mammography & 11183 & 6 & 42.01 & H \\
\textbf{D22} & analcatdata & 841 & 70 & 1.65 & L & \textbf{D46} & kc1 & 2109 & 21 & 5.47 & M & \textbf{D70} & Satellite & 5100 & 36 & 67 & H \\
\textbf{D23} & magic04 & 19020 & 10 & 1.84 & L & \textbf{D47} & musk2 & 6598 & 166 & 5.49 & M & \textbf{D71} & mc1 & 9466 & 38 & 138.21 & H \\
\textbf{D24} & MagicTelescope & 19020 & 11 & 1.84 & L & \textbf{D48} & segment0 & 2308 & 19 & 6.02 & M & \textbf{D72} & pc2 & 5589 & 36 & 242 & H \\ \hline
\end{tabular}
}
\begin{tablenotes}
    \item{$\cdot$} We empirically divide datasets into four groups regarding their imbalance ratio. 
    \item{$\cdot$} Equal: $IR=1$; Low $1< IR\leq 2$; Medium $2<IR\leq 10$; High $IR>10$.
    \item{$\cdot$} Due to computational memory and time constraints, we utilized datasets with fewer than 20,000 instances for the kernel model.
\end{tablenotes}
\label{S:tab:datasets}
\end{table*}
%========================================

We conducted a comprehensive comparison between HGD and several popular strategies for learning from imbalanced data streams, encompassing the baseline method (OGD), data-level approaches, cost-sensitive approaches, and ensemble learning techniques. Details of these methods are summarized in Table~\ref{S:tab:methods}. Given that our method aims to enhance the performance of gradient descent-based models in imbalanced data streams, we selected three widely-used gradient descent-based classifiers as base learners: perceptron, linear support vector machine, and kernel models. The learning rate $\eta_t$ was uniformly set to 0.3 across all methods. While many techniques may necessitate intricate hyper-parameter tuning to accommodate varying imbalance ratios, the specific settings utilized are provided in Table~\ref{S:tab:methods}.

%========================================
\begin{table*}[h]
\centering
\renewcommand\arraystretch{0.9}
\caption{The Details of The Various Methods for Imbalanced Data Stream Learning.}
\resizebox{\linewidth}{!}{
\begin{tabular}{clcccccc}
\hline
\multicolumn{1}{c|}{\textbf{No.}} & \textbf{Methods} & \textbf{Subscript} & \textbf{Abbr.} & \textbf{\begin{tabular}[c]{@{}c@{}}Classifier\\ Agnostic\end{tabular}} & \textbf{\begin{tabular}[c]{@{}c@{}}Loss\\ Agnostic\end{tabular}} & \textbf{Imbalanced Strategy} & \textbf{Parameters for Imbalance Learning} \\ \hline
\multicolumn{8}{l}{\cellcolor[HTML]{C0C0C0}\textbf{Baseline}} \\
\multicolumn{1}{c|}{\textbf{M1}} & \textbf{Online Gradient Descent} & $-$ & OGD & $\circ$ & $\surd$ & $-$ & - \\
\multicolumn{8}{l}{\cellcolor[HTML]{C0C0C0}\textbf{Data-Level Approaches}} \\
\multicolumn{1}{c|}{\textbf{M2}} & \textbf{One Pass Online SMOTE} & $-$ & OSMOTE & $\surd$ & $\surd$ & RO & $K = 3$ \\
\multicolumn{1}{c|}{\textbf{M3}} & \textbf{Online Over Resampling} & $-$ & OOR & $\surd$ & $\surd$ & RO & $R=IR$ \\
\multicolumn{1}{c|}{\textbf{M4}} & \textbf{Online Under Resampling} & $-$ & OUR & $\surd$ & $\surd$ & RU & $R=1/IR$ \\
\multicolumn{1}{c|}{\textbf{M5}} & \textbf{Online Hybrid Resampling} & $-$ & OHR & $\surd$ & $\surd$ & RU,RO & $R_1=IR, R_2=1/IR$ \\
\multicolumn{8}{l}{\cellcolor[HTML]{C0C0C0}\textbf{Cost Sensitive Approaches}} \\
\multicolumn{1}{c|}{\textbf{M6}} &  & I & CSRDA$_{I}$ & $\times$ & $\times$ &  &  \\
\multicolumn{1}{c|}{\textbf{M7}} &  & II & CSRDA$_{II}$ & $\times$ & $\surd$ &  &  \\
\multicolumn{1}{c|}{\textbf{M8}} &  & III & CSRDA$_{III}$ & $\times$ & $\surd$ &  &  \\
\multicolumn{1}{c|}{\textbf{M9}} & \multirow{-4}{*}{\textbf{Cost Sensitive Regularized Dual Averaging}} & IV & CSRDA$_{IV}$ & $\times$ & $\surd$ &  & \multirow{-4}{*}{\begin{tabular}[c]{@{}c@{}}$n_p=0.5, n_n=0.5$\\ $\lambda=0.1,\gamma=1e^{-3}$\end{tabular}} \\ \cline{8-8} 
\multicolumn{1}{c|}{\textbf{M10}} &  & Cost$_I$ & CSOGD$_{C_I}$ & $\circ$ & $\times$ &  &  \\
\multicolumn{1}{c|}{\textbf{M11}} &  & Cost$_{II}$ & CSOGD$_{C_{II}}$ & $\circ$ & $\surd$ &  & \multirow{-2}{*}{$c_p=0.95,c_n=0.05$} \\ \cline{8-8} 
\multicolumn{1}{c|}{\textbf{M12}} &  & Sum$_I$ & CSOGD$_{S_I}$ & $\circ$ & $\times$ &  &  \\
\multicolumn{1}{c|}{\textbf{M13}} & \multirow{-4}{*}{\textbf{Cost Sensitive Online Gradient Descent}} & Sum$_{II}$ & CSOGD$_{S_{II}}$ & $\circ$ & $\surd$ & \multirow{-8}{*}{CS} & \multirow{-2}{*}{$n_p=0.5,n_n=0.5$} \\
\multicolumn{8}{l}{\cellcolor[HTML]{9B9B9B}\textbf{Ensemble Learning Approaches}} \\
\multicolumn{1}{c|}{\textbf{M14}} & \textbf{Online Bagging} & $-$ & OB & $\surd$ & $\surd$ & $-$ &  \\
\multicolumn{1}{c|}{\textbf{M15}} & \textbf{Online Adaptive Boosting} & $-$ & OAdB & $\surd$ & $\surd$ & $-$ &  \\
\multicolumn{1}{c|}{\textbf{M16}} & \textbf{Online Kappa Bagging} & $-$ & OKB & $\surd$ & $\surd$ & CS & \multirow{-3}{*}{$M=10$} \\ \cline{8-8} 
\multicolumn{1}{c|}{\textbf{M17}} & \textbf{Online Under Bagging} & $-$ & OUB & $\surd$ & $\surd$ & RU &  \\
\multicolumn{1}{c|}{\textbf{M18}} & \textbf{Online Weighted Under Bagging} & $-$ & OWUB & $\surd$ & $\surd$ & RU &  \\
\multicolumn{1}{c|}{\textbf{M19}} & \textbf{Online Over Bagging} & $-$ & OOB & $\surd$ & $\surd$ & RO &  \\
\multicolumn{1}{c|}{\textbf{M20}} & \textbf{Online Weighted Over Bagging} & $-$ & OWOB & $\surd$ & $\surd$ & RO & \multirow{-4}{*}{$M=10,\eta=0.9$} \\ \cline{8-8} 
\multicolumn{1}{c|}{\textbf{M21}} & \textbf{Online Cost Sensitive Adaptive Boosting} & 2 & OAdaC2 & $\surd$ & $\surd$ & CS, RU, RO &  \\
\multicolumn{1}{c|}{\textbf{M22}} & \textbf{Online Cost Sensitive Boosting} & 2 & OCSB2 & $\surd$ & $\surd$ & CS, RU, RO & \multirow{-2}{*}{$M=10,c_p=1, c_n=0.8$} \\ \cline{8-8} 
\multicolumn{1}{c|}{\textbf{M23}} &  & 1 & ORUSB1 & $\surd$ & $\surd$ & CS, RU, RO &  \\
\multicolumn{1}{c|}{\textbf{M24}} &  & 2 & ORUSB2 & $\surd$ & $\surd$ & CS, RU, RO & \multirow{-2}{*}{$M=10,R=0.7$} \\ \cline{8-8} 
\multicolumn{1}{c|}{\textbf{M25}} & \multirow{-3}{*}{\textbf{Online RUSBoosting}} & 3 & ORUSB3 & $\surd$ & $\surd$ & RU, RO &  \\
\multicolumn{1}{c|}{\textbf{M26}} & \textbf{Online Under Over Bagging} & $-$ & OUOB & $\surd$ & $\surd$ & RU, RO & \multirow{-2}{*}{$M=10,R=IR$} \\ \cline{8-8} 
\multicolumn{1}{c|}{\textbf{M27}} & \textbf{Online Effective Bagging} & $-$ & OEB & $\surd$ & $\surd$ & CS, RU, RO & $M=10,R=1/IR$ \\
\multicolumn{8}{l}{\cellcolor[HTML]{C0C0C0}\textbf{The Proposed Approach}} \\
\multicolumn{1}{c|}{\textbf{M28}} & \textbf{Harmonizing Gradient Descent} & $-$ & HGD & $\circ$ & $\surd$ & TM & $-$ \\ \hline
\end{tabular}
}
\begin{tablenotes}
    \item{$\cdot$} CS: Cost Sensitive; RU: Under Resampling; RO: Over Resampling; TM: Training-Modification.
\end{tablenotes}
\label{S:tab:methods}
\end{table*}
%========================================

\clearpage
\newpage 

\subsection{Performance Overview}

\begin{figure*}[th]
    \centering
    \subfloat[Perceptron]{
    \begin{minipage}[c]{0.3\textwidth}
        \includegraphics[width=5.5cm]{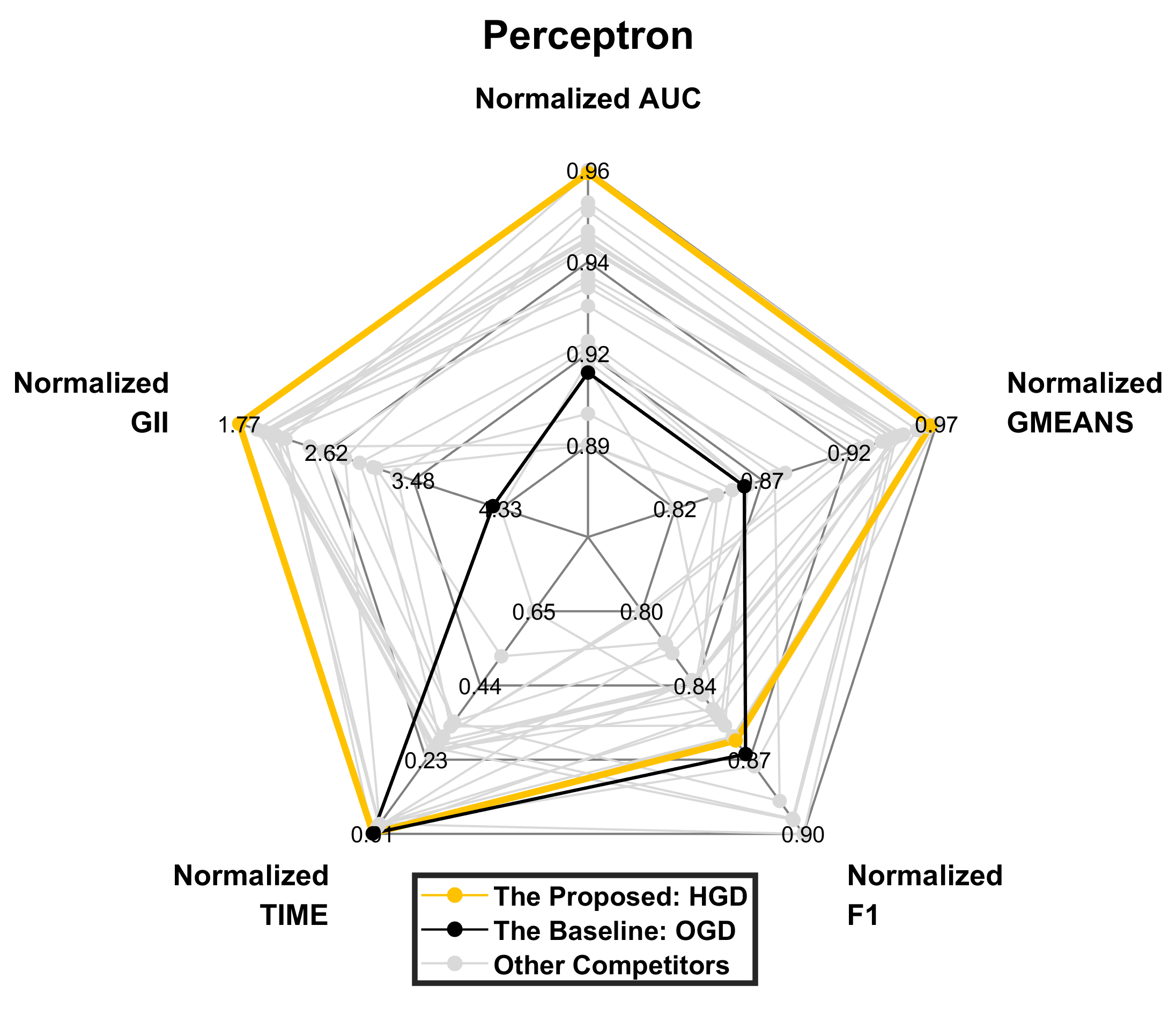}
    \end{minipage}}
    \subfloat[Linear SVM]{
    \begin{minipage}[c]{0.3\textwidth}
        \includegraphics[width=5.5cm]{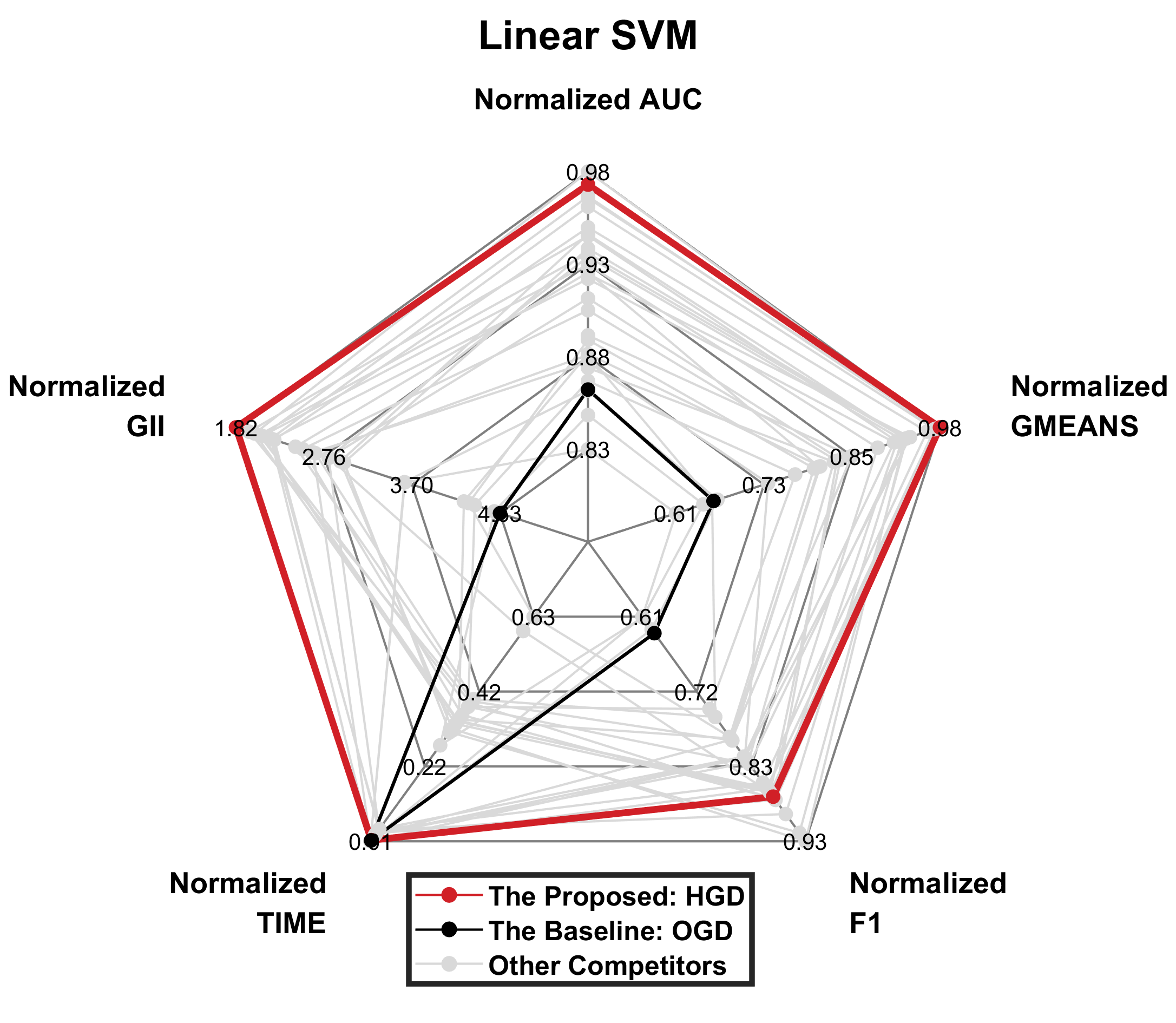}
    \end{minipage}}
        \subfloat[Kernel Model]{
    \begin{minipage}[c]{0.3\textwidth}
        \includegraphics[width=5.5cm]{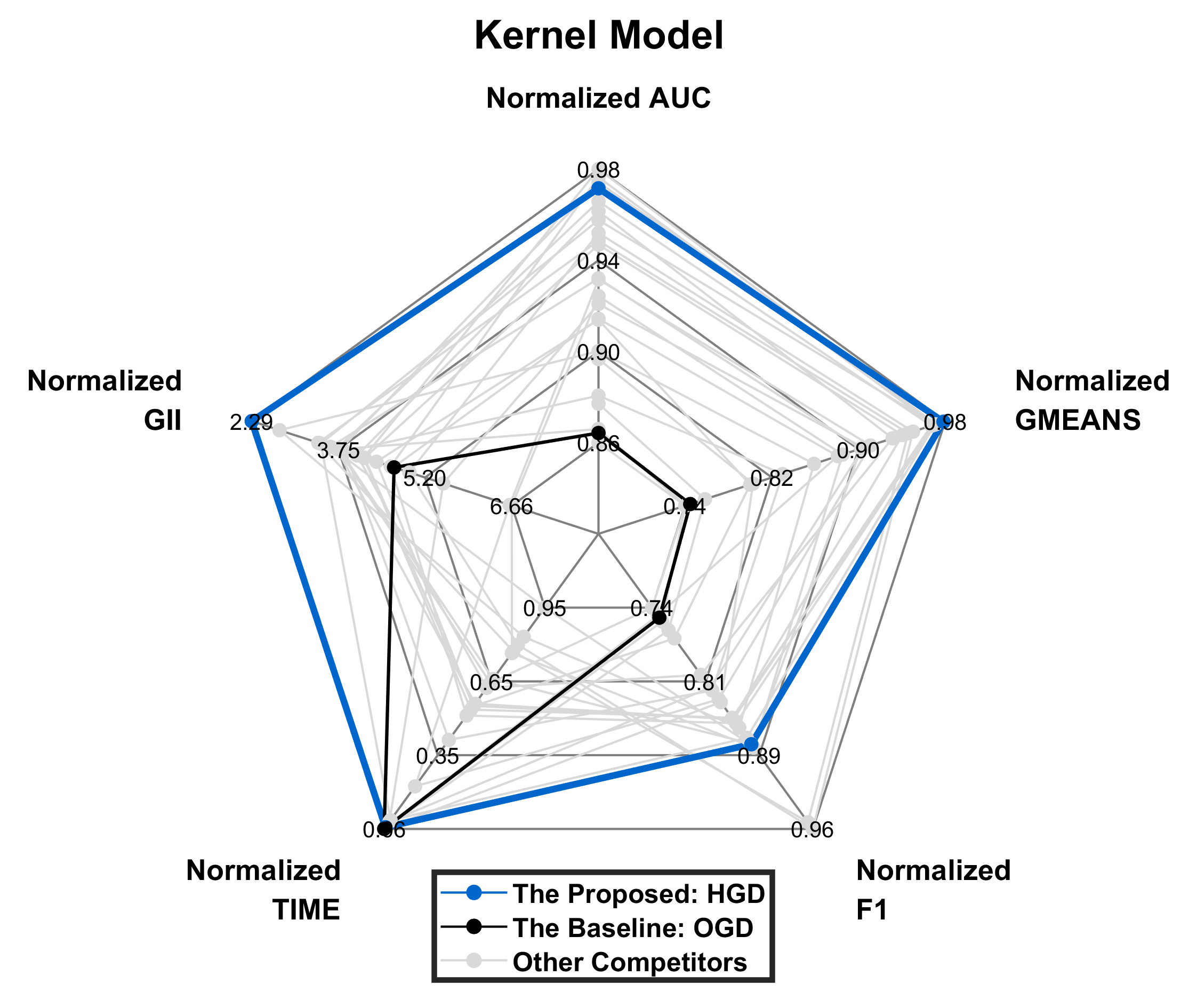}
    \end{minipage}} \\
    \caption{An Systematic Performance Comparison Across Multiple Metrics: AUC, GMEANS, F1 Score, GII, and Computational Time. (a) Perceptron, (b) Linear SVM, (c) Kernel Model as the base learner.}
    \label{S:fig:static_Spider}
\end{figure*}

An overview of the results achieved by 28 methods with 3 base learner on 72 datasets is visualized in the spider plots (Figure~\ref{S:fig:static_Spider}), in terms of the normalized AUC, G-Means, F1, GII and time. Here, we highlighted the performance attained by the proposed method and the baseline OGD. We have the following observations form these figures.
\begin{itemize}
    \item The baseline, OGD (denoted in black), unsurprisingly fails to achieve satisfactory performance in terms of AUC, GMEANS, F1, and GII. This is primarily due to the performance bias incurred by OGD. OGD updates the model parameters based on the gradients of the loss function, and in imbalanced data streams, these gradients are dominated by the majority class. As a result, the model focuses more on optimizing performance for the majority class, leading to poor performance on the minority class. In contrast, the proposed HGD demonstrates significant performance improvements w.r.t. the baseline, underscoring its effectiveness in handling imbalanced data stream learning.
    \item In terms of computational time, the proposed method is quite efficient compared to the baseline. This efficiency is attributed to the fact that the proposed HGD only requires a simple weight factor calculation. In contrast, many competitors exhibit less efficiency because they require additional learning strategies to accommodate imbalanced data streams, such as resampling and ensemble learning.
    \item Comparing with all the competitors, the proposed method achieves highly competitive performance in terms of most metrics with all different base learners. However, an exception can be found in the results in terms of F1, where the results achieved by HGD are slightly less competitive. This can be attributed to the tendency oh HGD to identify the minority class more often, which may reduce precision and subsequently result in a lower F1 score.
\end{itemize}

%===============================================
%===============================================

\clearpage
\newpage

\subsection{Performance Analysis}

\begin{figure*}[th]
    \centering
    \subfloat[AUC]{
        \includegraphics[width=2.21cm]{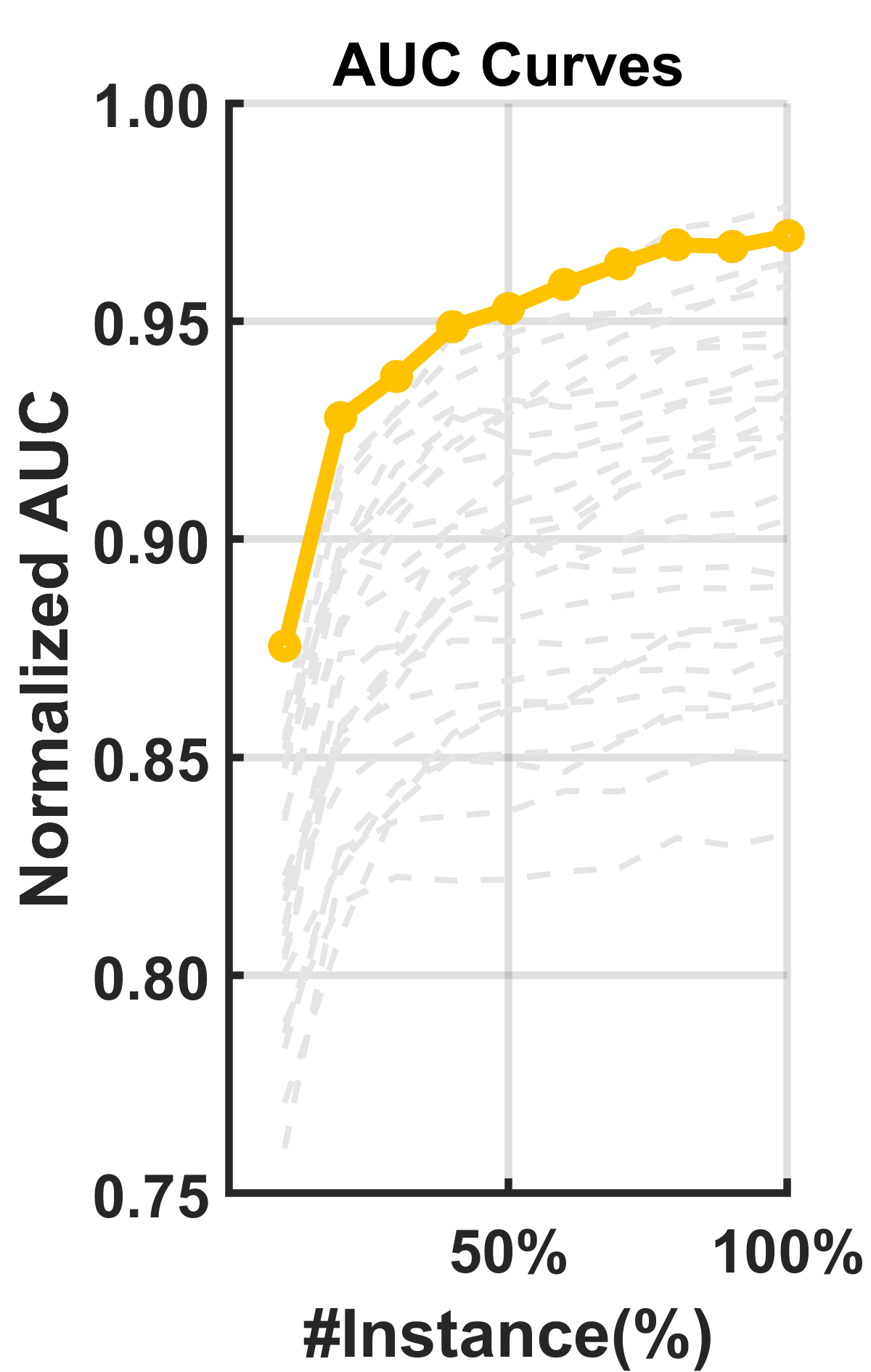}
    }\hspace{-6mm}
    \subfloat[AUC]{
        \includegraphics[width=2.21cm]{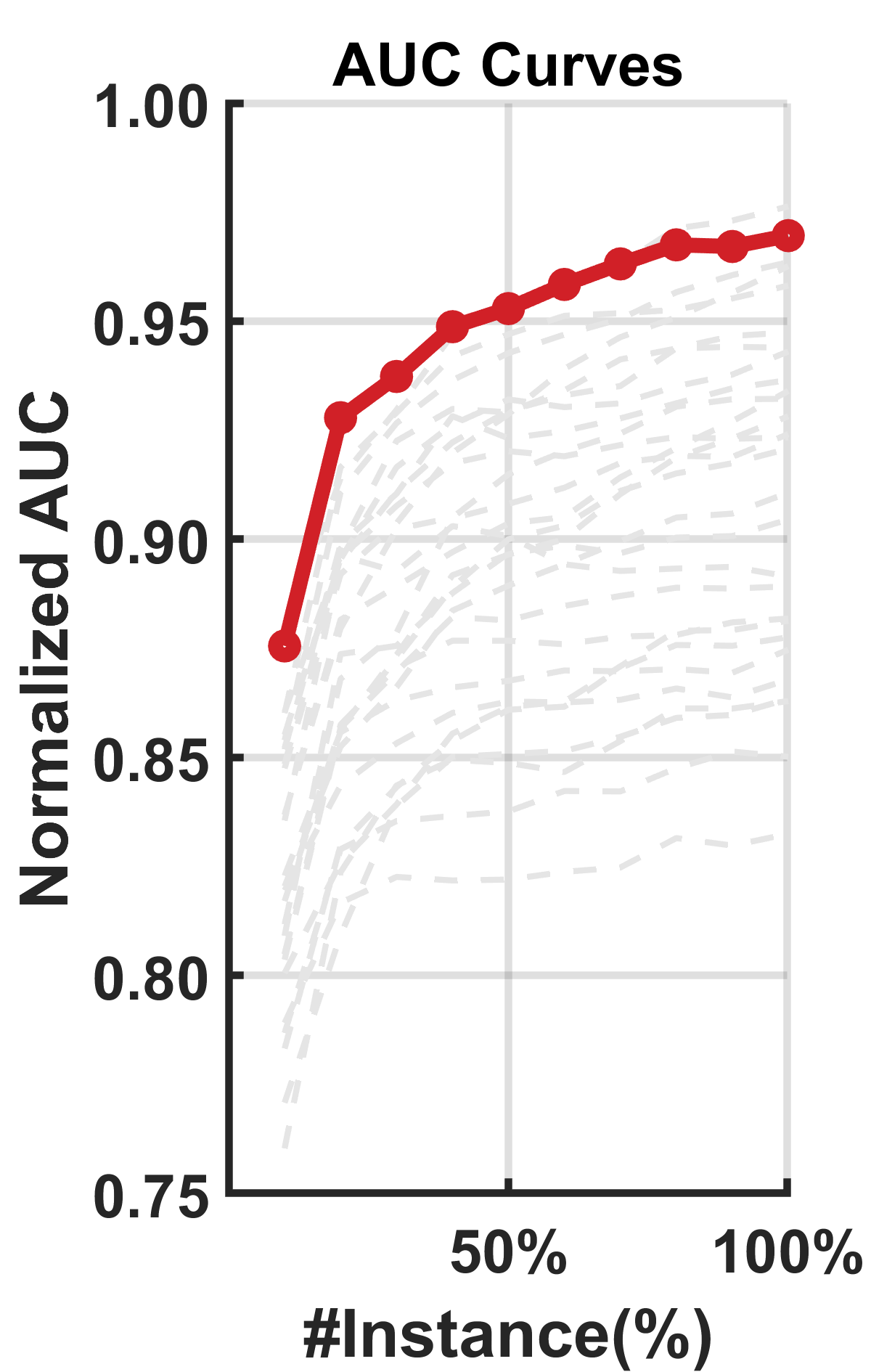}
    }\hspace{-6mm}
        \subfloat[AUC]{
        \includegraphics[width=2.21cm]{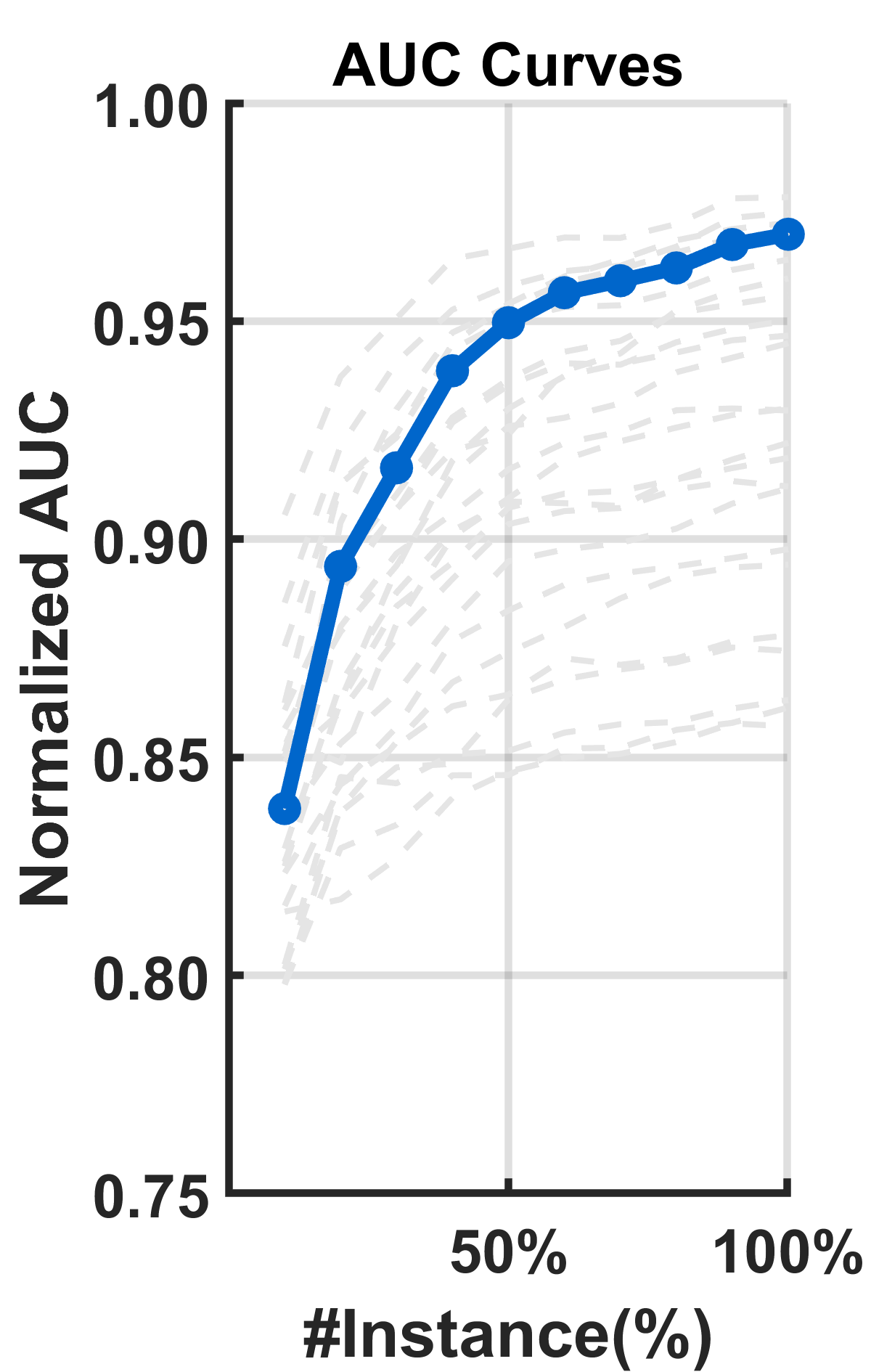}
    }\hspace{-6mm}
    \subfloat[GMEANS]{
        \includegraphics[width=2.21cm]{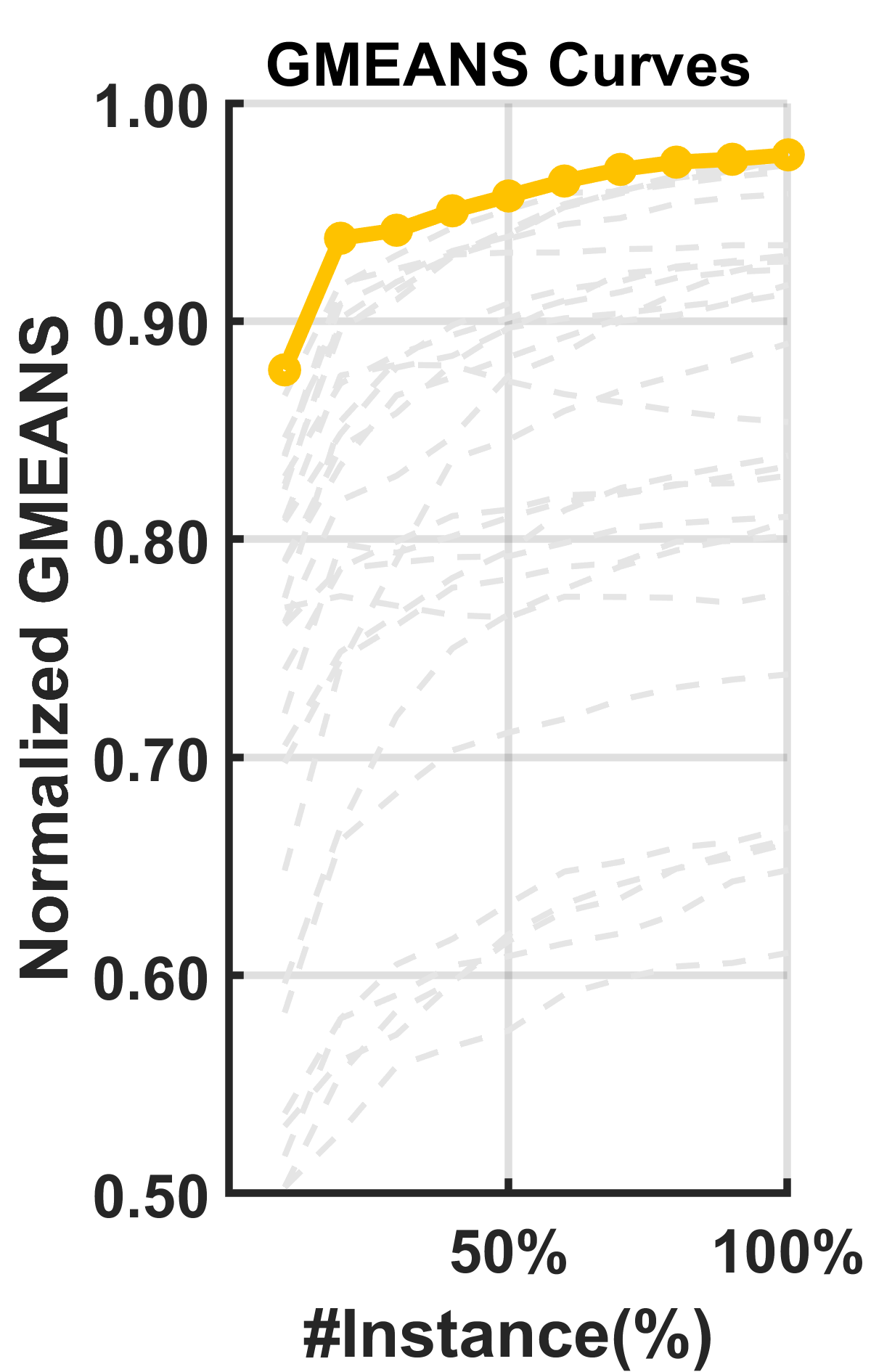}
    } \hspace{-6mm}
    \subfloat[GMEANS]{
        \includegraphics[width=2.21cm]{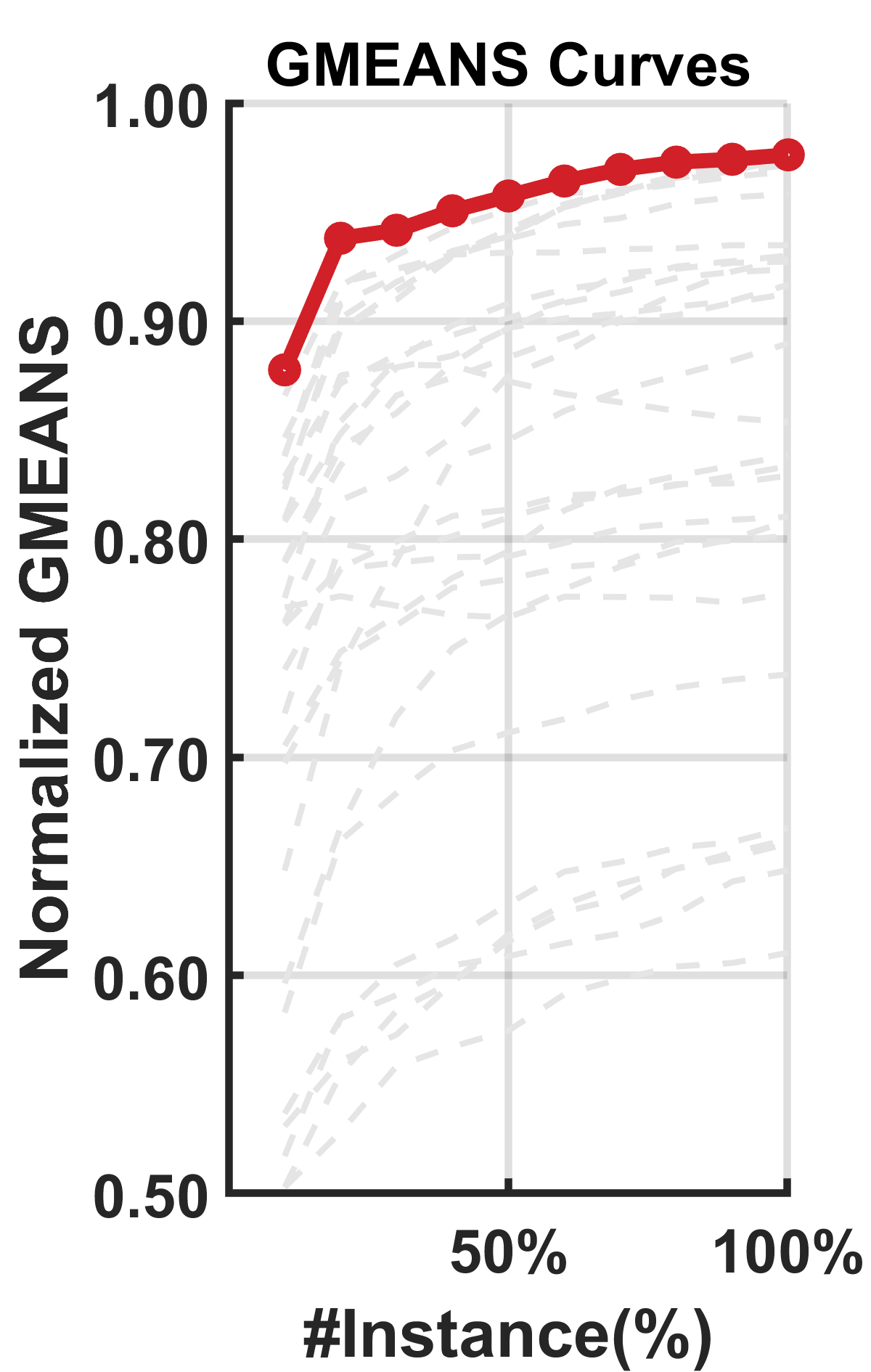}
    } \hspace{-6mm}
    \subfloat[GMEANS]{
        \includegraphics[width=2.21cm]{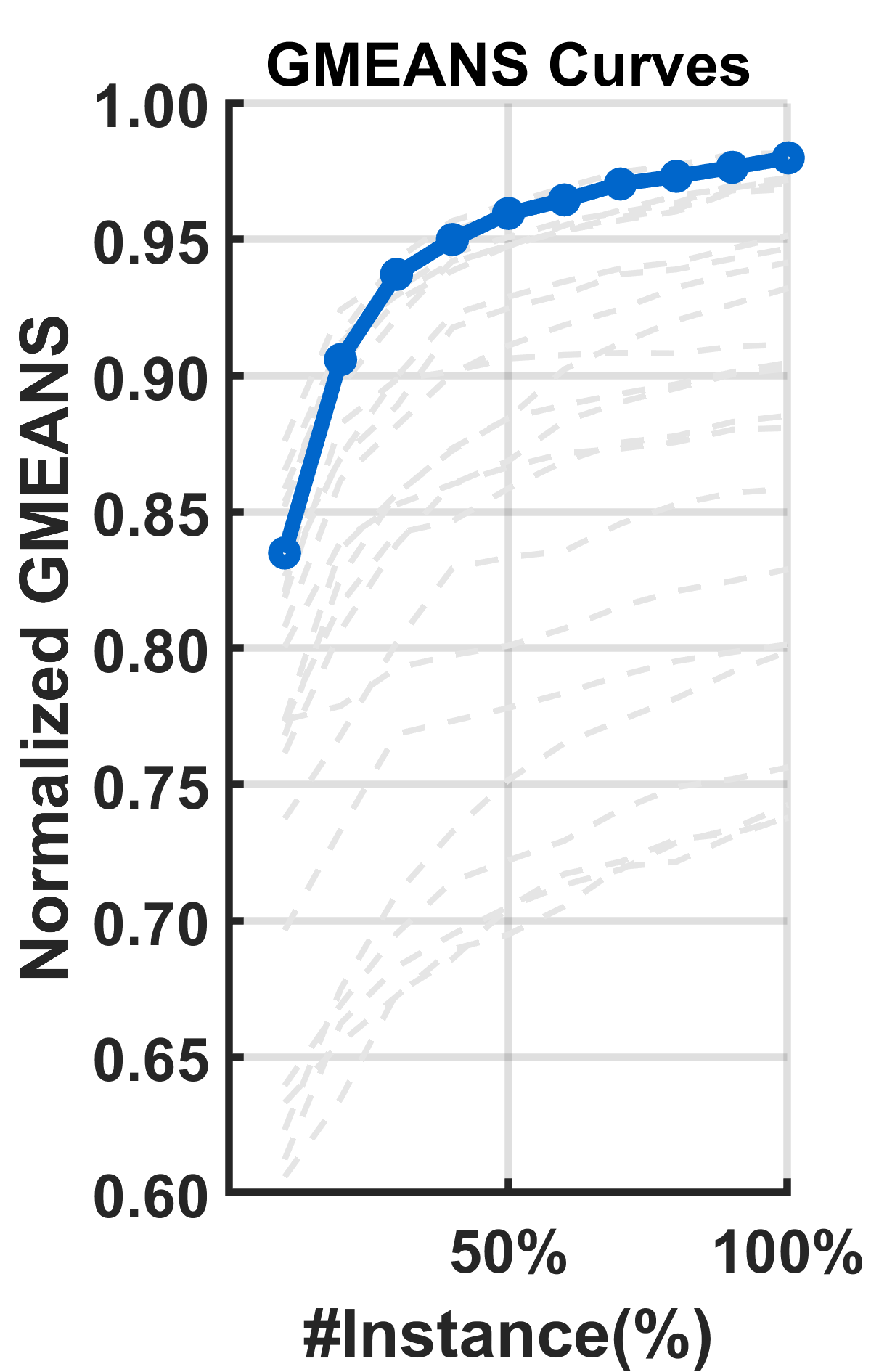}
    } \hspace{-6mm}
    \subfloat[F1]{
        \includegraphics[width=2.21cm]{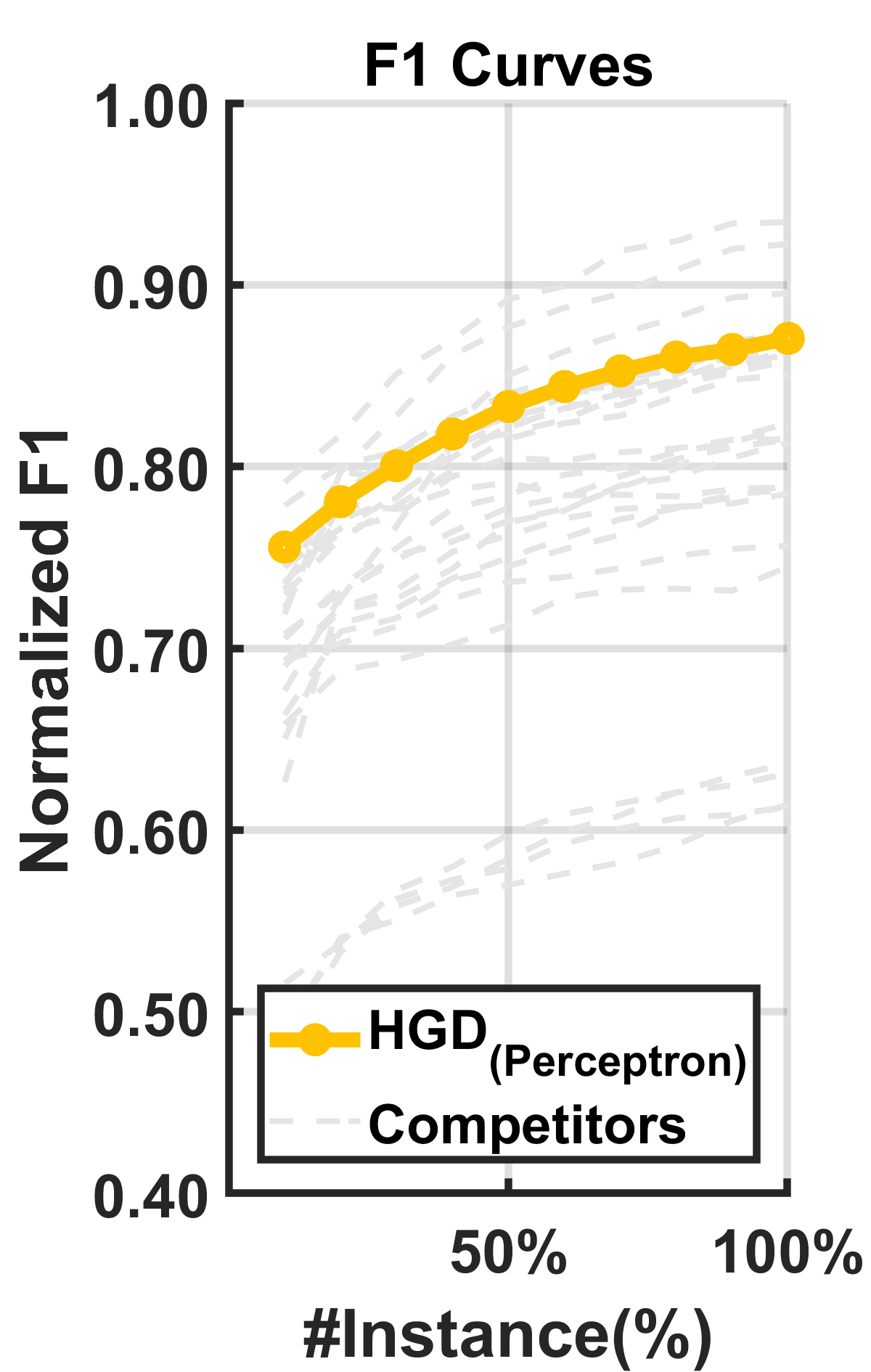}
    } \hspace{-6mm}
    \subfloat[F1]{
        \includegraphics[width=2.21cm]{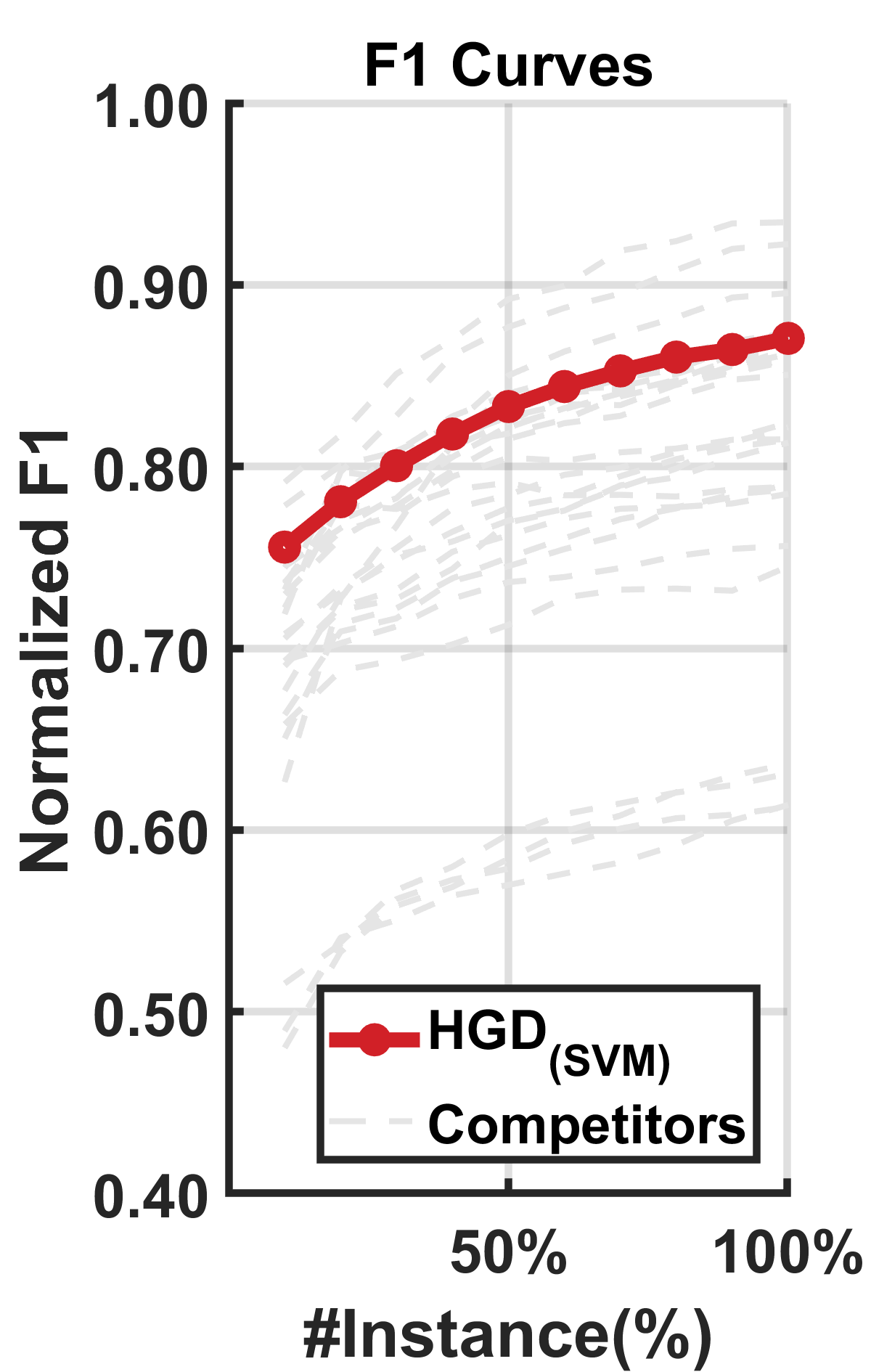}
    } \hspace{-6mm}
    \subfloat[F1]{
        \includegraphics[width=2.21cm]{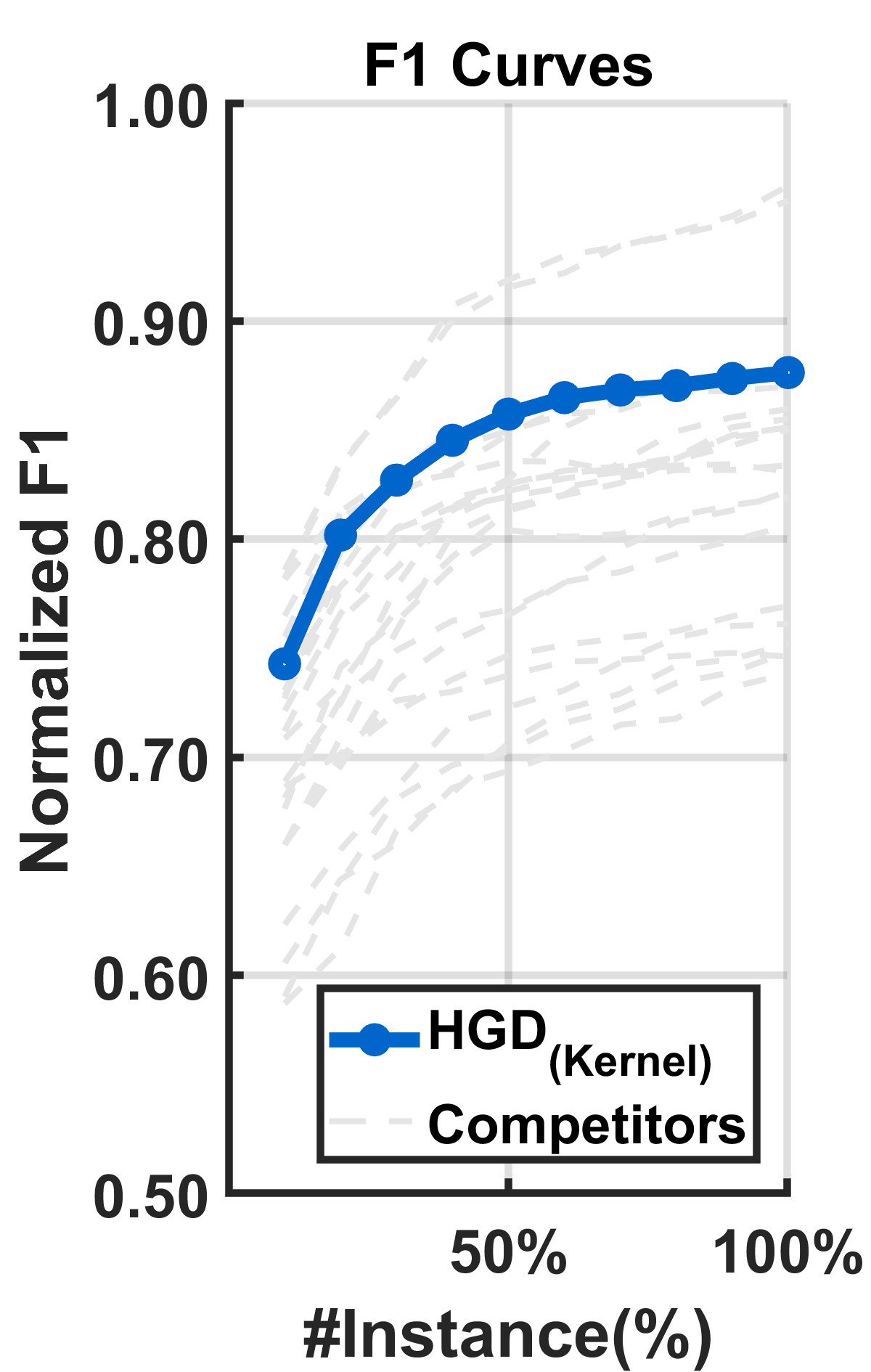}
    } 
    \caption{The performance curves of all the methods under static imbalance ratio scenarios. The proposed HGD is highlighted in yellow, red and blue when utilizing perceptron, linear SVM and kernel model as the base learner.}
    \label{S:fig:static_curves}
\end{figure*}

\begin{figure*}[th]
    \centering
    \subfloat[Perceptron]{
        \includegraphics[width=6cm]{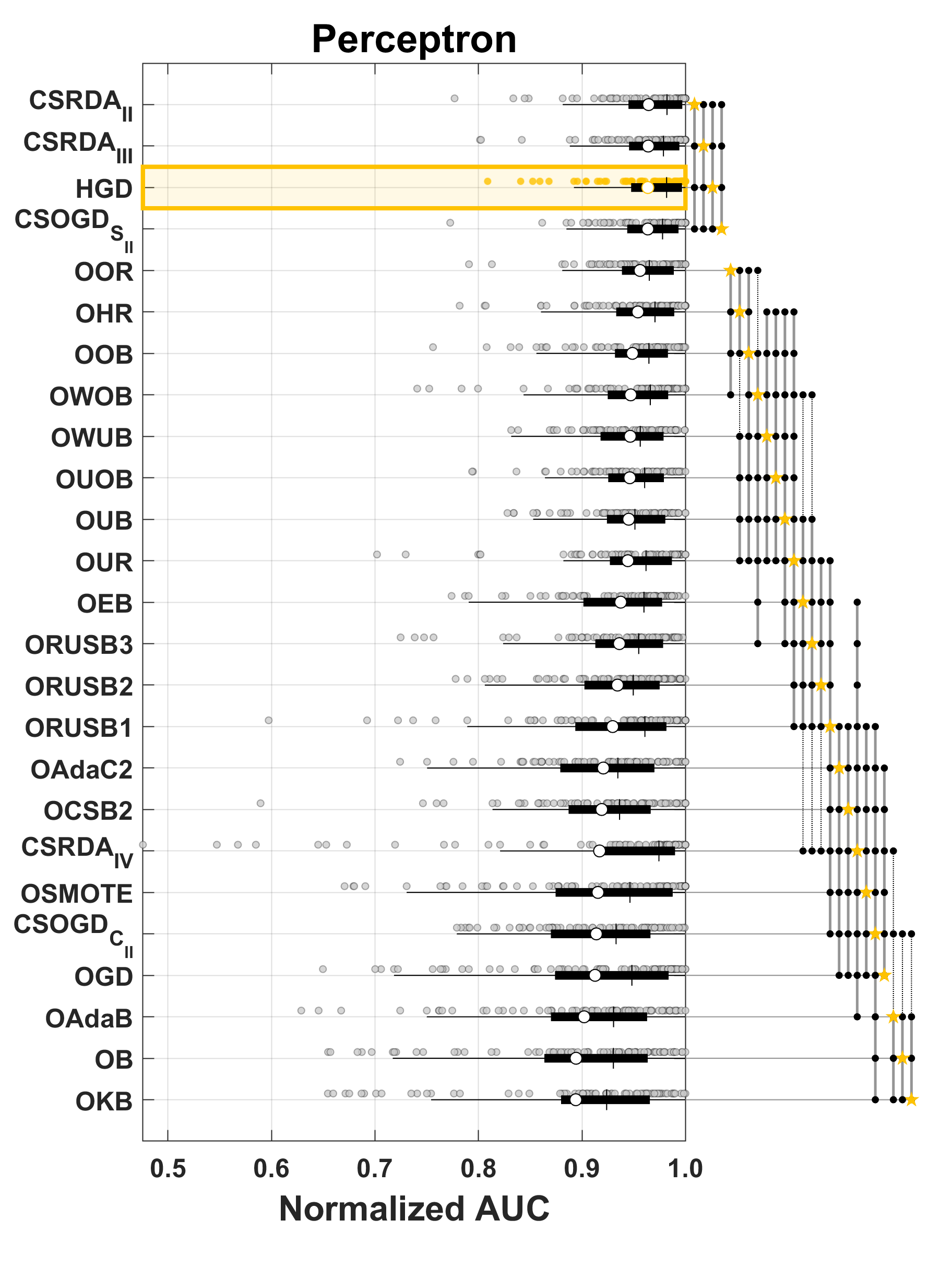}
    }
    \subfloat[Linear SVM]{
        \includegraphics[width=6cm]{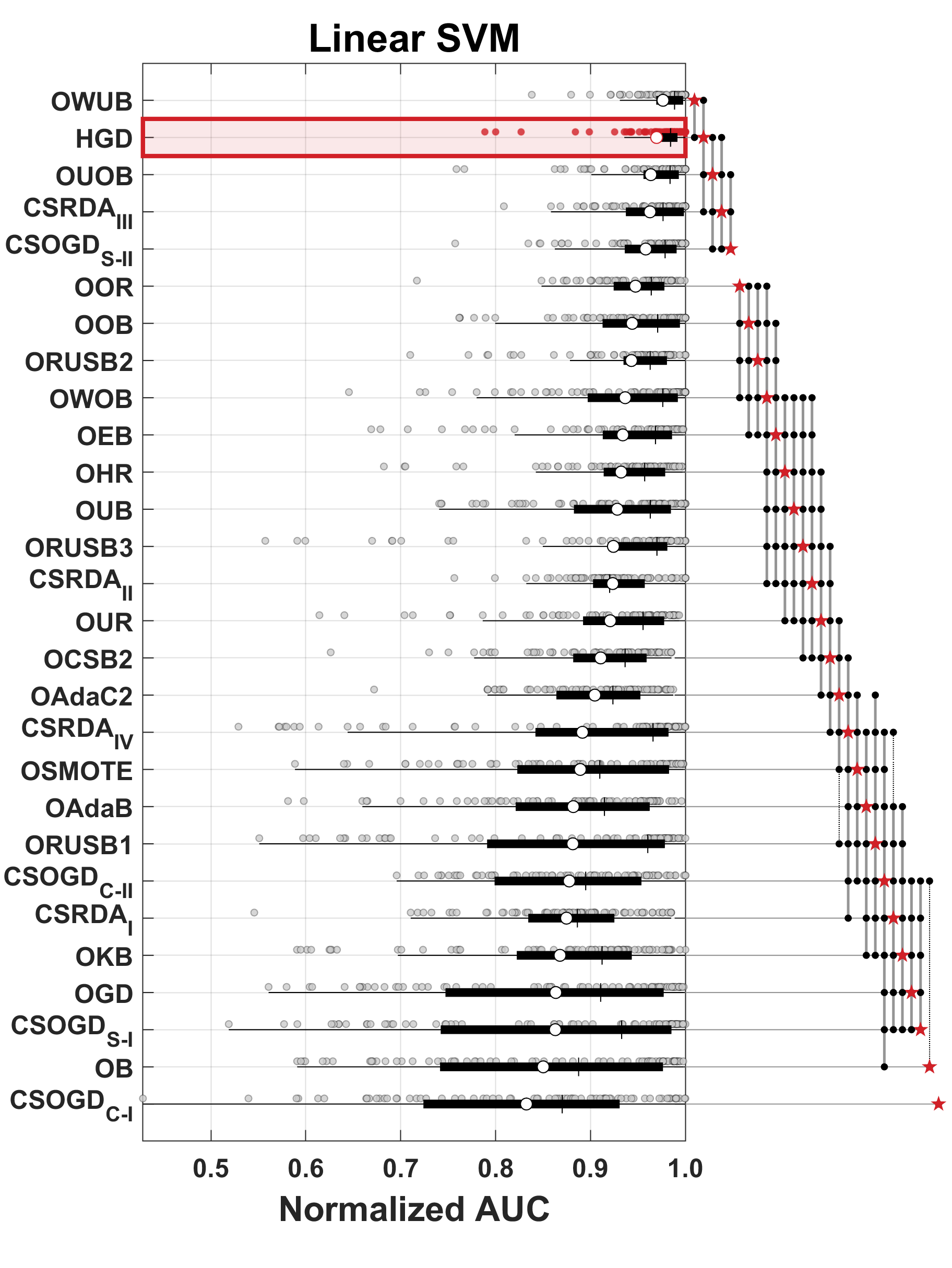}
    }
    \subfloat[Kernel Model]{
        \includegraphics[width=6cm]{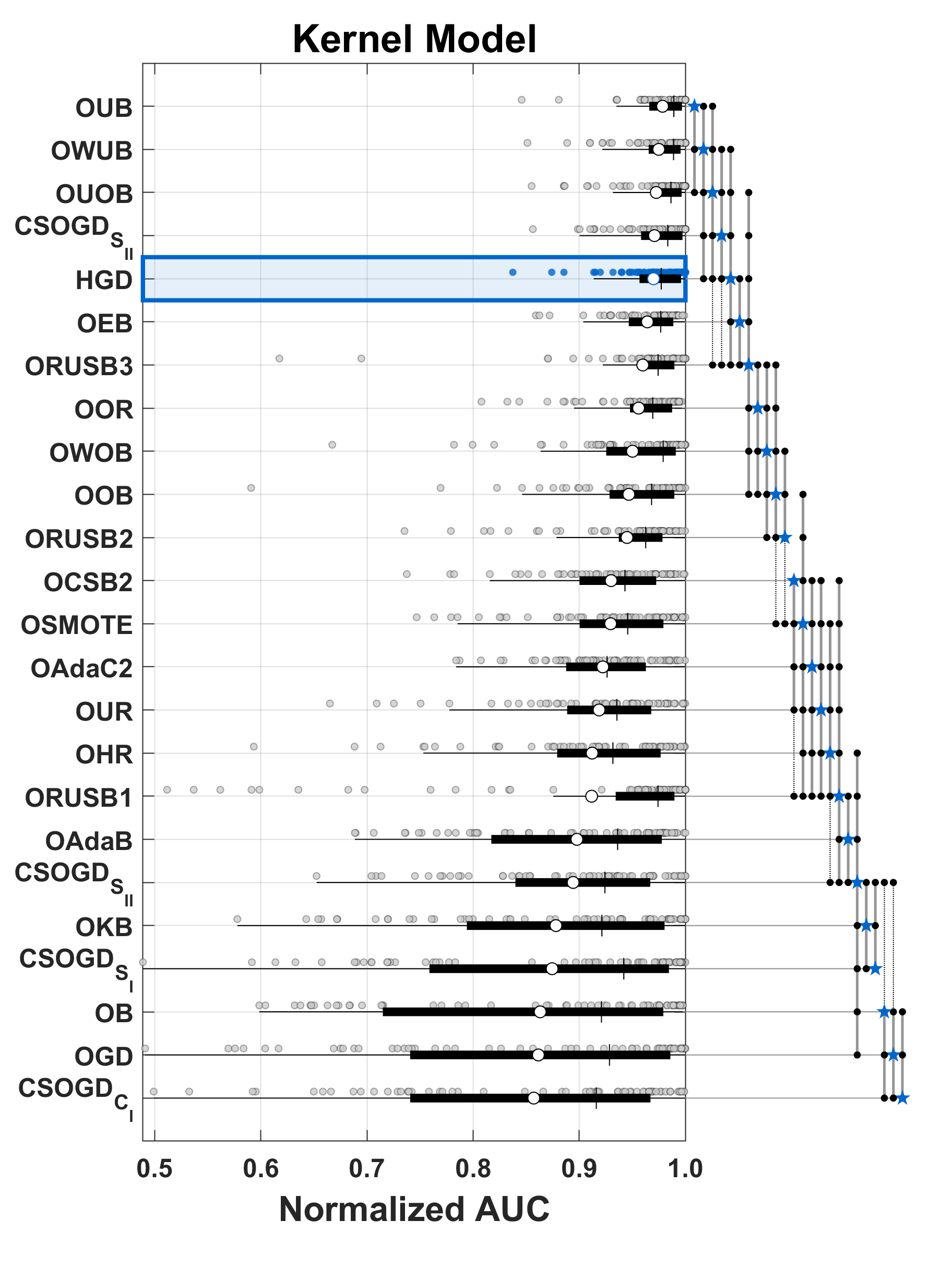}
    } 
    \caption{The performance of different methods on all datasets, in terms of AUC. Mean performance is denoted in encircled dot. The boxes are presented in the order of descending from top to bottom. Methods without significant performance difference are connected in the right, tested by the t-test at the significance level of 0.05.}
    \label{S:fig:static_AllBoxPlot_AUC}
\end{figure*}

\begin{figure*}[th]
    \centering
    \subfloat[Perceptron]{
        \includegraphics[width=6cm]{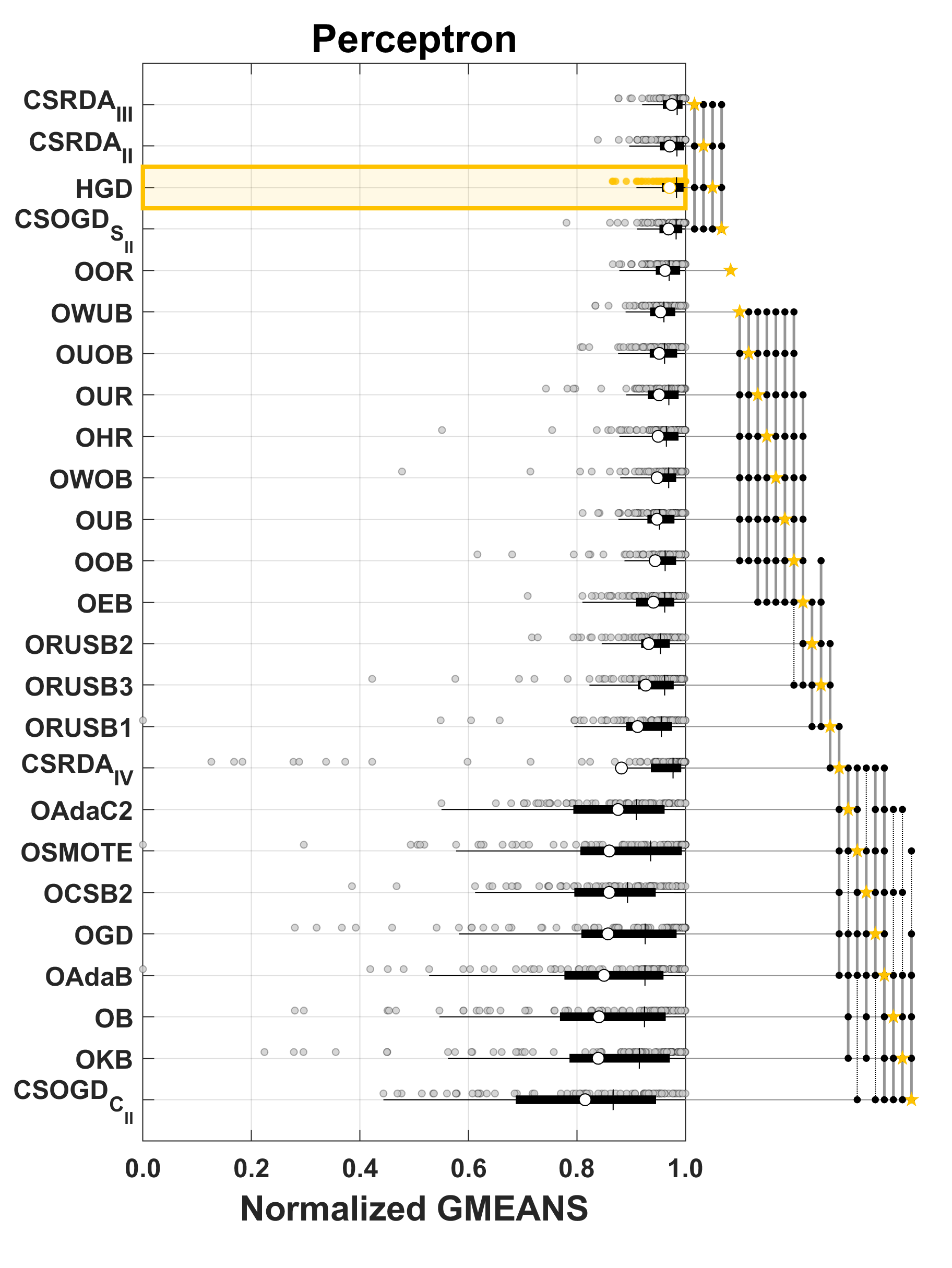}
    }
    \subfloat[Linear SVM]{
        \includegraphics[width=6cm]{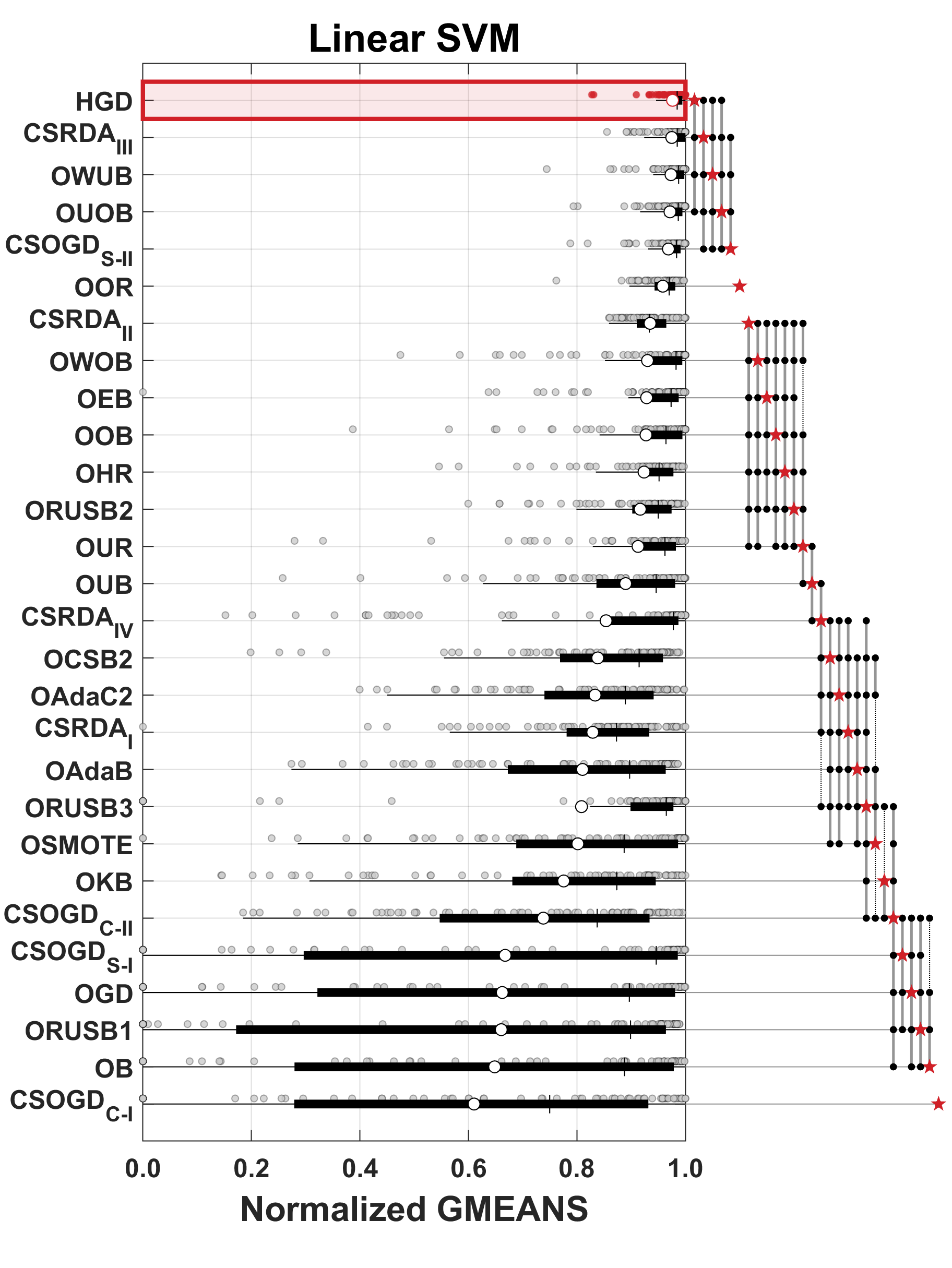}
    }
    \subfloat[Kernel Model]{
        \includegraphics[width=6cm]{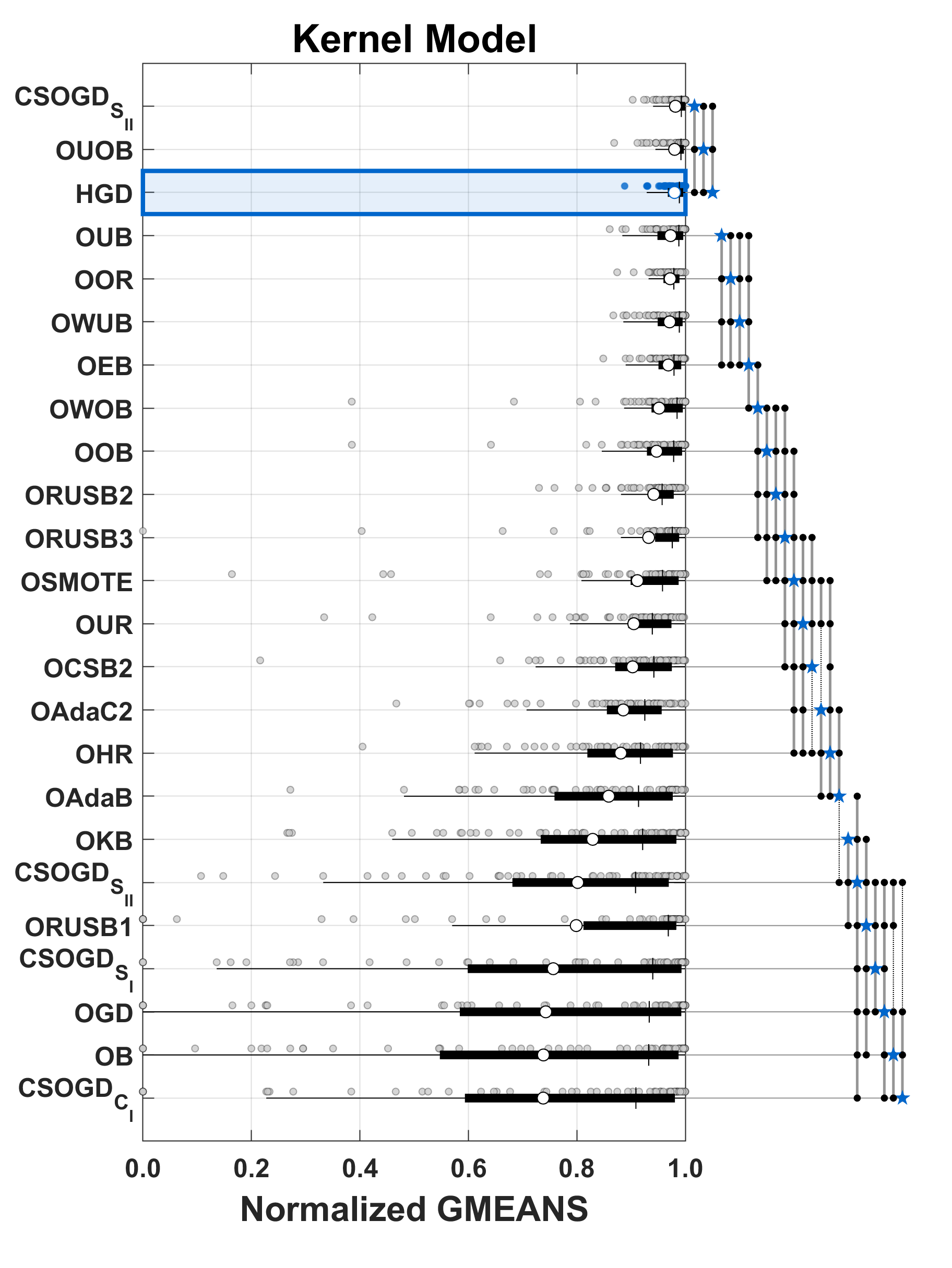}
    } 
    \caption{The performance of different methods on all datasets, in terms of GMEANS. Mean performance is denoted in encircled dot. The boxes are presented in the order of descending from top to bottom. Methods without significant performance difference are connected in the right, tested by the t-test at the significance level of 0.05.}
    \label{S:fig:static_AllBoxPlot_GMEANS}
\end{figure*}

\begin{figure*}[th]
    \centering
    \subfloat[Perceptron]{
        \includegraphics[width=6cm]{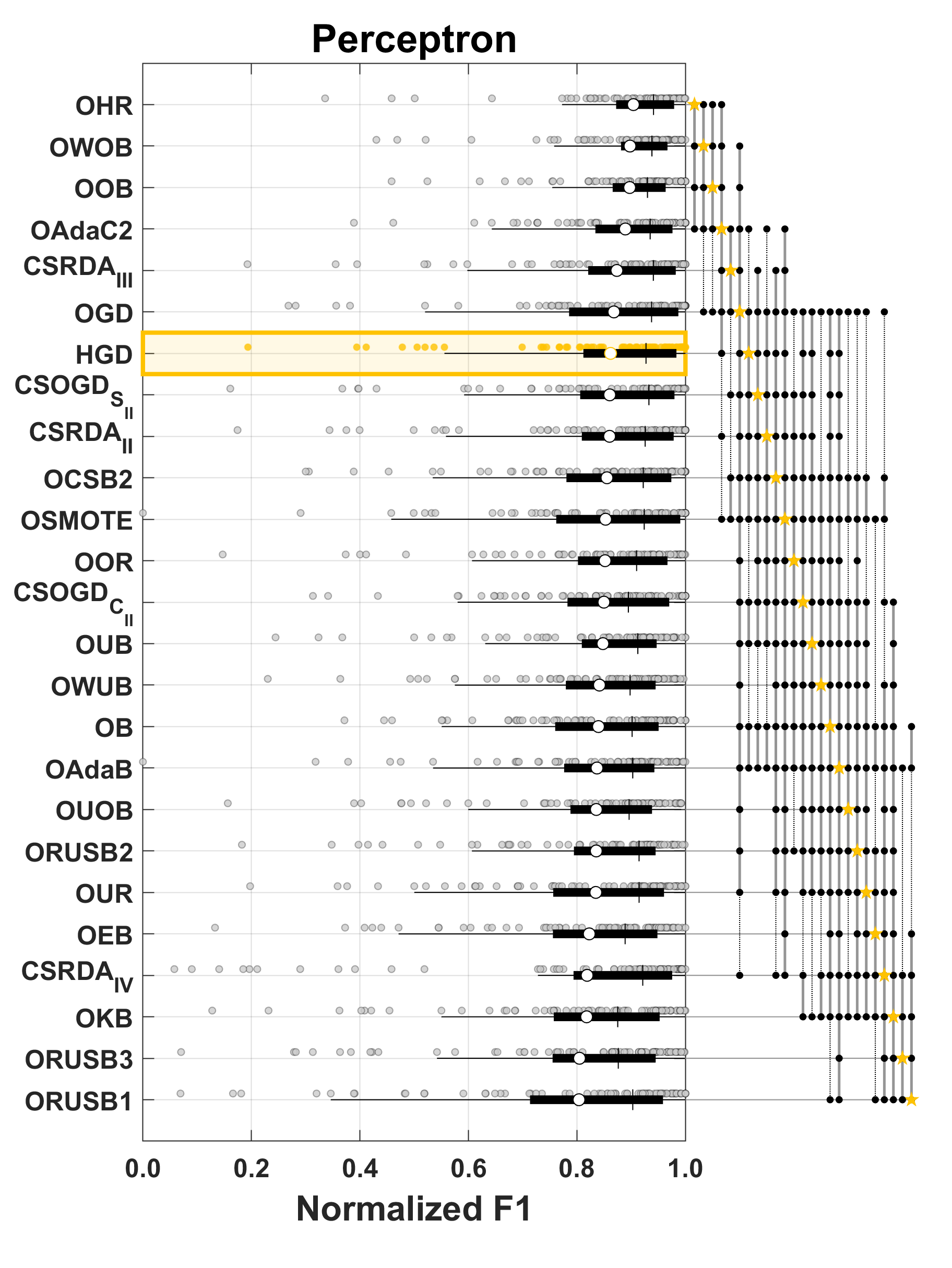}
    }
    \subfloat[Linear SVM]{
        \includegraphics[width=6cm]{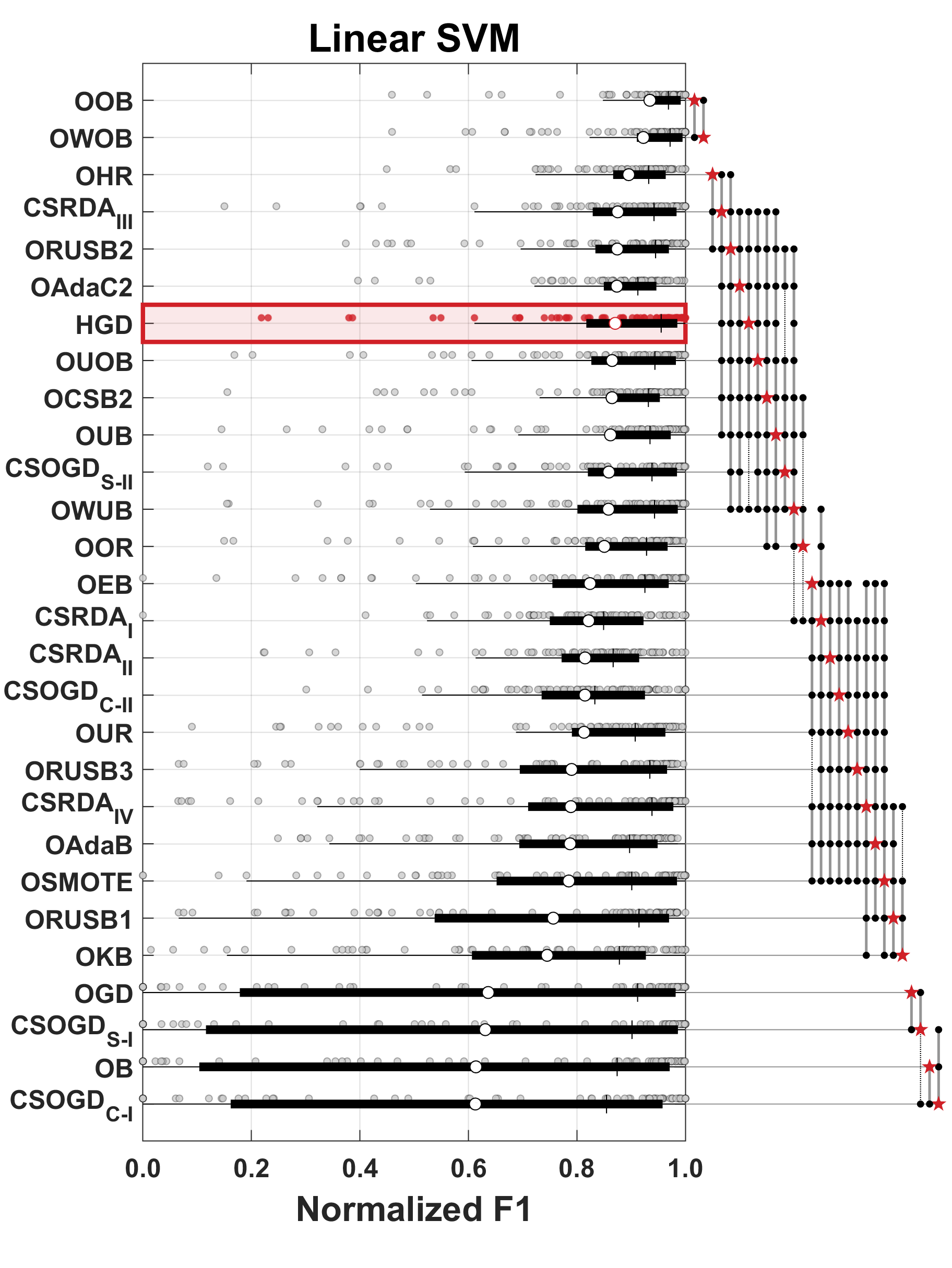}
    }
    \subfloat[Kernel Model]{
        \includegraphics[width=6cm]{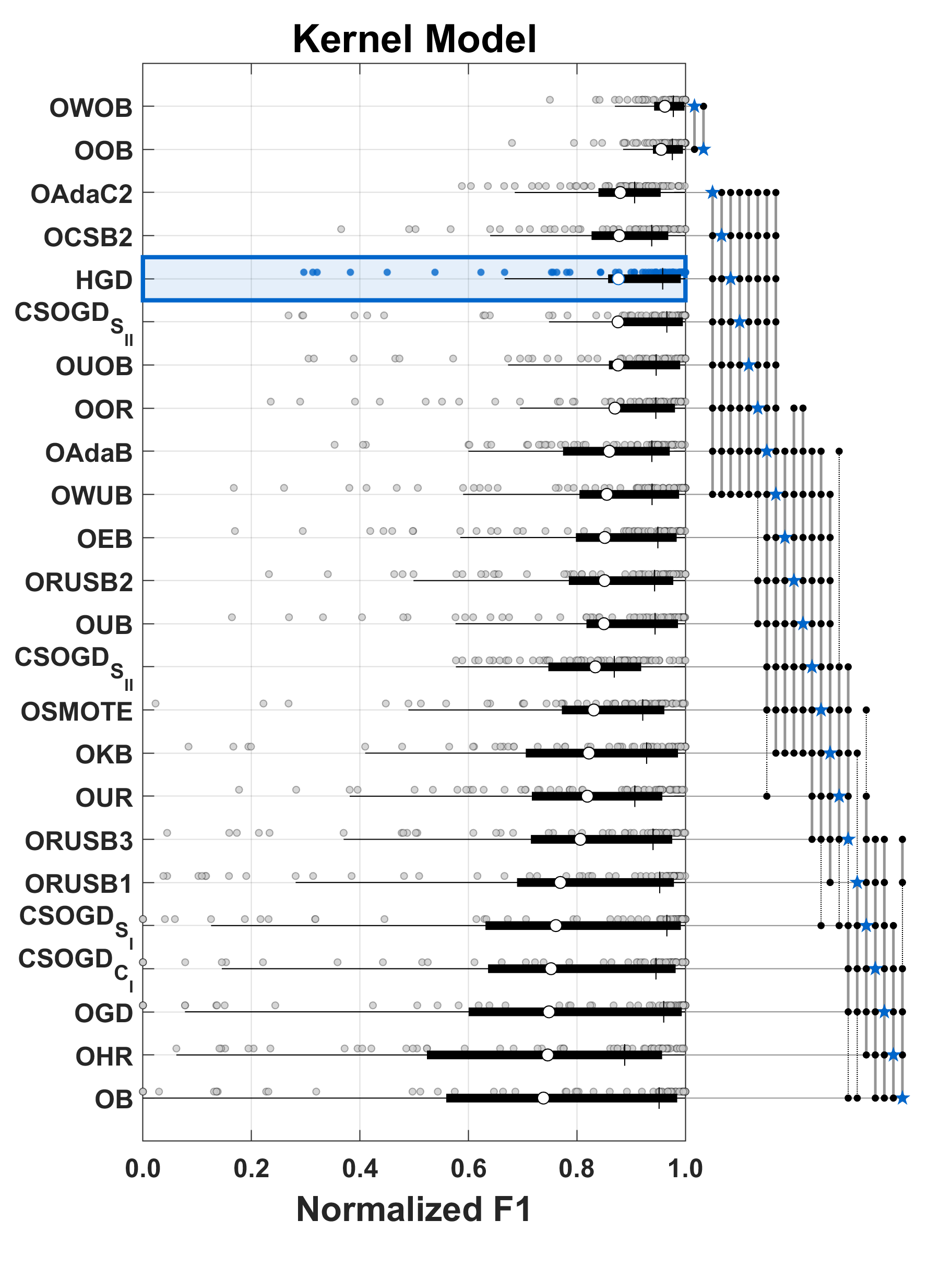}
    } 
    \caption{The performance of different methods on all datasets, in terms of F1. Mean performance is denoted in encircled dot. The boxes are presented in the order of descending from top to bottom. Methods without significant performance difference are connected in the right, tested by the t-test at the significance level of 0.05.}
    \label{S:fig:static_AllBoxPlot_F1}
\end{figure*}

We provide the detailed average performance in Table~\ref{S:tab:cmp_static}, including numerical values for three normalized metrics and the average ranks obtained by different methods across all datasets. The performance curves throughout the online learning process are depicted in Figure~\ref{S:fig:static_curves}. The box plot visualizations of the results are depicted in from Figure~\ref{S:fig:static_AllBoxPlot_AUC} to Figure~\ref{S:fig:static_AllBoxPlot_F1}. In each box plot, the left and right edges of the black box represent the 25th and 75th percentiles, respectively. The encircled dot within the box indicates the mean performance. Whiskers extend to the most extreme data points, excluding outliers. The arrangement of the boxes follows a descending order based on mean performance, from top to bottom. We highlighted the proposed HGD in yellow, red and blue when applying the perceptron, the linear SVM and the kernel model as the base learner, respectively. Additionally, the right part of each figure illustrates the performance difference, assessed via the t-test at a significance level of 0.05. Methods exhibiting no significant performance difference are connected by black lines and dots.  Clearly, the proposed HGD is among one of the most effective methods across all three different base learners. However, it is not surprising that the performance of HGD in terms of F1 is slightly less effective, resulting from a performance compromise between positive and negative aspects.

%========================================
\begin{table*}[h]
\scriptsize
\centering
\renewcommand\arraystretch{1.2}
\caption{The Averaged Performance (Normalized AUC, G-means and F1-Score) Attained by Different Methods Over All Datasets.}
\resizebox{\linewidth}{!}{
\begin{tabular}{lcccccccccccccccccccc}
\hline
\textbf{Base Learner} & \multicolumn{6}{c}{\textbf{Perceptron}} & \textbf{} & \multicolumn{6}{c}{\textbf{Linear SVM}} & \textbf{} & \multicolumn{6}{c}{\textbf{Kernel Model}} \\ \hline
\multicolumn{1}{l|}{\textbf{Metric}} & \textbf{AUC} & \textbf{Rank} & \textbf{GMEANS} & \textbf{Rank} & \textbf{F1} & \textbf{Rank} & \multicolumn{1}{c|}{\textbf{}} & \textbf{AUC} & \textbf{Rank} & \textbf{GMEANS} & \textbf{Rank} & \textbf{F1} & \textbf{Rank} & \multicolumn{1}{c|}{\textbf{}} & \textbf{AUC} & \textbf{Rank} & \textbf{GMEANS} & \textbf{Rank} & \textbf{F1} & \textbf{Rank} \\ \hline
\multicolumn{21}{l}{\cellcolor[HTML]{C0C0C0}\textbf{Baseline}} \\
\multicolumn{1}{l|}{OGD} & \cellcolor[HTML]{FCFCFF}0.913 & 14.6 & \cellcolor[HTML]{FCFCFF}0.857 & 15.1 & \cellcolor[HTML]{FCFCFF}0.868 & 12.5 & \multicolumn{1}{c|}{} & \cellcolor[HTML]{FCFCFF}0.863 & 17.8 & \cellcolor[HTML]{FCFCFF}0.662 & 17.5 & \cellcolor[HTML]{FCFCFF}0.636 & 16.1 & \multicolumn{1}{c|}{} & \cellcolor[HTML]{FCF7FA}0.861 & 17.0 & \cellcolor[HTML]{FCFAFD}0.743 & 15.6 & \cellcolor[HTML]{FCF6F9}0.748 & 12.4 \\
\multicolumn{21}{l}{\cellcolor[HTML]{C0C0C0}\textbf{Data Level Approaches}} \\
\multicolumn{1}{l|}{OSMOTE} & \cellcolor[HTML]{FCF5F8}0.915 & 13.4 & \cellcolor[HTML]{FCF9FC}0.860 & 14.2 & \cellcolor[HTML]{D5E0F1}0.853 & 11.6 & \multicolumn{1}{c|}{} & \cellcolor[HTML]{FCDBDE}0.889 & 16.4 & \cellcolor[HTML]{FBBBBE}0.802 & 16.1 & \cellcolor[HTML]{FBB3B6}0.785 & 14.4 & \multicolumn{1}{c|}{} & \cellcolor[HTML]{FAA5A7}0.930 & 13.9 & \cellcolor[HTML]{FA9496}0.911 & 12.8 & \cellcolor[HTML]{FBC0C2}0.831 & 13.9 \\
\multicolumn{1}{l|}{OUR} & \cellcolor[HTML]{FAA2A5}0.944 & 11.6 & \cellcolor[HTML]{F98689}0.951 & 11.1 & \cellcolor[HTML]{A8C0E1}0.835 & 13.5 & \multicolumn{1}{c|}{} & \cellcolor[HTML]{FAB2B4}0.921 & 15.4 & \cellcolor[HTML]{F9878A}0.913 & 13.8 & \cellcolor[HTML]{FAA5A8}0.813 & 16.1 & \multicolumn{1}{c|}{} & \cellcolor[HTML]{FAB2B4}0.919 & 15.5 & \cellcolor[HTML]{FA989A}0.905 & 15.8 & \cellcolor[HTML]{FBC7CA}0.819 & 16.5 \\
\multicolumn{1}{l|}{OOR} & \cellcolor[HTML]{F98183}0.956 & \textbf{9.8} & \cellcolor[HTML]{F9797B}0.962 & \textbf{9.2} & \cellcolor[HTML]{D3DFF0}0.852 & 12.2 & \multicolumn{1}{c|}{} & \cellcolor[HTML]{FA8F91}0.947 & 12.6 & \cellcolor[HTML]{F97274}0.958 & 11.3 & \cellcolor[HTML]{FA9395}0.850 & 14.2 & \multicolumn{1}{c|}{} & \cellcolor[HTML]{F98587}0.956 & 11.1 & \cellcolor[HTML]{F96F71}0.972 & 9.9 & \cellcolor[HTML]{FAA6A8}0.870 & 11.7 \\
\multicolumn{1}{l|}{OHR} & \cellcolor[HTML]{F98789}0.954 & \textbf{9.8} & \cellcolor[HTML]{F9898B}0.950 & 10.1 & \cellcolor[HTML]{F8696B}0.904 & \textbf{8.6} & \multicolumn{1}{c|}{} & \cellcolor[HTML]{FAA3A5}0.932 & 13.7 & \cellcolor[HTML]{F98284}0.924 & 13.9 & \cellcolor[HTML]{F97D7F}0.895 & 12.6 & \multicolumn{1}{c|}{} & \cellcolor[HTML]{FBBABC}0.912 & 15.7 & \cellcolor[HTML]{FAA6A9}0.881 & 16.8 & \cellcolor[HTML]{FCF7FA}0.746 & 18.3 \\
\multicolumn{21}{l}{\cellcolor[HTML]{C0C0C0}\textbf{Cost Sensitive Approaches}} \\
\multicolumn{1}{l|}{CSRDA$_{I}$} & \cellcolor[HTML]{F8696B}0.964 & \textbf{7.8} & \cellcolor[HTML]{F96E70}0.971 & \textbf{6.5} & \cellcolor[HTML]{E7EDF7}0.860 & \textbf{9.7} & \multicolumn{1}{c|}{} & \cellcolor[HTML]{FCEEF1}0.875 & 21.4 & \cellcolor[HTML]{FAAEB1}0.829 & 20.5 & \cellcolor[HTML]{FAA1A4}0.822 & 18.9 & \multicolumn{1}{c|}{} & - & - & - & - & - & - \\
\multicolumn{1}{l|}{CSRDA$_{II}$} & \cellcolor[HTML]{F96A6C}0.964 & \textbf{7.4} & \cellcolor[HTML]{F8696B}0.974 & \textbf{6.3} & \cellcolor[HTML]{FCE6E9}0.874 & \textbf{9.0} & \multicolumn{1}{c|}{} & \cellcolor[HTML]{FAAEB1}0.923 & 17.3 & \cellcolor[HTML]{F97D7F}0.935 & 15.9 & \cellcolor[HTML]{FAA4A7}0.815 & 19.4 & \multicolumn{1}{c|}{} & - & - & - & - & - & - \\
\multicolumn{1}{l|}{CSRDA$_{III}$} & \cellcolor[HTML]{FCF0F3}0.917 & 12.0 & \cellcolor[HTML]{FCDDE0}0.882 & 10.7 & \cellcolor[HTML]{7FA4D3}0.819 & 12.7 & \multicolumn{1}{c|}{} & \cellcolor[HTML]{F97B7D}0.963 & \textbf{8.1} & \cellcolor[HTML]{F96A6C}0.975 & \textbf{6.7} & \cellcolor[HTML]{F98789}0.875 & \textbf{9.8} & \multicolumn{1}{c|}{} & - & - & - & - & \textbf{-} & - \\
\multicolumn{1}{l|}{CSRDA$_{IV}$} & - & - & - & - & - & - & \multicolumn{1}{c|}{} & \cellcolor[HTML]{FCD8DB}0.891 & 13.2 & \cellcolor[HTML]{FAA3A5}0.854 & 11.5 & \cellcolor[HTML]{FAB1B4}0.789 & 13.3 & \multicolumn{1}{c|}{} & - & - & - & - & - & - \\
\multicolumn{1}{l|}{CSOGD$_{C_I}$} & - & - & - & - & - & - & \multicolumn{1}{c|}{} & \cellcolor[HTML]{5A8AC6}0.832 & 21.8 & \cellcolor[HTML]{5A8AC6}0.610 & 22.2 & \cellcolor[HTML]{5A8AC6}0.613 & 19.0 & \multicolumn{1}{c|}{} & \cellcolor[HTML]{FCFCFF}0.857 & 18.1 & \cellcolor[HTML]{FCFCFF}0.738 & 17.2 & \cellcolor[HTML]{FCF3F6}0.752 & 13.7 \\
\multicolumn{1}{l|}{CSOGD$_{C_{II}}$} & \cellcolor[HTML]{FCF9FC}0.914 & 16.3 & \cellcolor[HTML]{5A8AC6}0.815 & 18.7 & \cellcolor[HTML]{CDDBEE}0.850 & 14.3 & \multicolumn{1}{c|}{} & \cellcolor[HTML]{FCEAED}0.878 & 19.2 & \cellcolor[HTML]{FCD9DC}0.738 & 21.2 & \cellcolor[HTML]{FAA5A7}0.815 & 19.1 & \multicolumn{1}{c|}{} & \cellcolor[HTML]{FBD0D2}0.894 & 16.4 & \cellcolor[HTML]{FBD6D9}0.801 & 16.8 & \cellcolor[HTML]{FBBEC0}0.834 & 17.1 \\
\multicolumn{1}{l|}{CSOGD$_{S_{I}}$} & - & - & - & - & - & - & \multicolumn{1}{c|}{} & \cellcolor[HTML]{F9FAFE}0.863 & 16.9 & \cellcolor[HTML]{FCFAFD}0.668 & 16.9 & \cellcolor[HTML]{D7E2F2}0.631 & 15.9 & \multicolumn{1}{c|}{} & \cellcolor[HTML]{FCE8EA}0.874 & 15.6 & \cellcolor[HTML]{FCF1F4}0.756 & 14.5 & \cellcolor[HTML]{FCEDF0}0.761 & 11.9 \\
\multicolumn{1}{l|}{CSOGD$_{S_{II}}$} & \cellcolor[HTML]{F96C6E}0.964 & \textbf{7.5} & \cellcolor[HTML]{F97173}0.969 & \textbf{6.8} & \cellcolor[HTML]{E8EEF8}0.860 & 10.1 & \multicolumn{1}{c|}{} & \cellcolor[HTML]{F98183}0.958 & \textbf{8.6} & \cellcolor[HTML]{F96D6F}0.969 & \textbf{7.1} & \cellcolor[HTML]{FA8F91}0.859 & 10.9 & \multicolumn{1}{c|}{} & \cellcolor[HTML]{F97375}0.971 & \textbf{6.5} & \cellcolor[HTML]{F8696B}0.982 & \textbf{4.8} & \cellcolor[HTML]{FAA2A5}0.876 & \textbf{7.6} \\
\multicolumn{21}{l}{\cellcolor[HTML]{C0C0C0}\textbf{Ensemble Learning Approaches}} \\
\multicolumn{1}{l|}{OB} & \cellcolor[HTML]{5B8BC6}0.894 & 18.3 & \cellcolor[HTML]{BCCFE8}0.841 & 18.7 & \cellcolor[HTML]{B4C9E5}0.840 & 16.0 & \multicolumn{1}{c|}{} & \cellcolor[HTML]{B6CBE6}0.850 & 19.6 & \cellcolor[HTML]{D0DDEF}0.648 & 19.0 & \cellcolor[HTML]{608EC8}0.614 & 18.8 & \multicolumn{1}{c|}{} & \cellcolor[HTML]{FCF5F8}0.863 & 17.8 & \cellcolor[HTML]{FCFCFF}0.738 & 16.6 & \cellcolor[HTML]{FCFCFF}0.738 & 13.9 \\
\multicolumn{1}{l|}{OAdaB} & \cellcolor[HTML]{A0BBDE}0.902 & 18.3 & \cellcolor[HTML]{E0E8F5}0.850 & 18.6 & \cellcolor[HTML]{ADC4E3}0.837 & 16.0 & \multicolumn{1}{c|}{} & \cellcolor[HTML]{FCE4E7}0.882 & 20.0 & \cellcolor[HTML]{FBB7BA}0.810 & 19.4 & \cellcolor[HTML]{FAB2B5}0.787 & 18.1 & \multicolumn{1}{c|}{} & \cellcolor[HTML]{FBCBCE}0.898 & 16.5 & \cellcolor[HTML]{FBB4B6}0.859 & 16.7 & \cellcolor[HTML]{FAADAF}0.860 & 13.6 \\
\multicolumn{1}{l|}{OAdaC2} & \cellcolor[HTML]{FCE6E9}0.921 & 15.8 & \cellcolor[HTML]{FCE5E8}0.876 & 17.7 & \cellcolor[HTML]{FAA6A9}0.889 & 11.5 & \multicolumn{1}{c|}{} & \cellcolor[HTML]{FBC7CA}0.904 & 18.4 & \cellcolor[HTML]{FAACAF}0.833 & 19.4 & \cellcolor[HTML]{F9878A}0.874 & 15.4 & \multicolumn{1}{c|}{} & \cellcolor[HTML]{FAAEB0}0.922 & 15.4 & \cellcolor[HTML]{FAA4A6}0.885 & 16.1 & \cellcolor[HTML]{FAA0A2}0.880 & 14.1 \\
\multicolumn{1}{l|}{OCSB2} & \cellcolor[HTML]{FCEAED}0.919 & 16.1 & \cellcolor[HTML]{FCFAFD}0.859 & 18.6 & \cellcolor[HTML]{DBE5F3}0.855 & 13.4 & \multicolumn{1}{c|}{} & \cellcolor[HTML]{FBBFC2}0.911 & 18.0 & \cellcolor[HTML]{FAAAAD}0.838 & 18.8 & \cellcolor[HTML]{F98C8E}0.865 & 14.5 & \multicolumn{1}{c|}{} & \cellcolor[HTML]{FAA4A7}0.930 & 14.3 & \cellcolor[HTML]{FA999B}0.903 & 15.1 & \cellcolor[HTML]{FAA1A3}0.878 & 12.6 \\
\multicolumn{1}{l|}{OKB} & \cellcolor[HTML]{5A8AC6}0.894 & 18.5 & \cellcolor[HTML]{B6CBE6}0.839 & 18.7 & \cellcolor[HTML]{7DA2D2}0.818 & 16.7 & \multicolumn{1}{c|}{} & \cellcolor[HTML]{FCF7FA}0.868 & 20.4 & \cellcolor[HTML]{FBC7CA}0.776 & 19.8 & \cellcolor[HTML]{FBC7CA}0.745 & 19.2 & \multicolumn{1}{c|}{} & \cellcolor[HTML]{FCE3E6}0.878 & 17.0 & \cellcolor[HTML]{FBC5C8}0.829 & 16.1 & \cellcolor[HTML]{FBC5C8}0.822 & 13.5 \\
\multicolumn{1}{l|}{OUOB} & \cellcolor[HTML]{FA9D9F}0.946 & 13.0 & \cellcolor[HTML]{F98688}0.952 & 12.0 & \cellcolor[HTML]{AAC2E2}0.836 & 15.5 & \multicolumn{1}{c|}{} & \cellcolor[HTML]{F97A7C}0.964 & \textbf{7.1} & \cellcolor[HTML]{F96C6E}0.972 & \textbf{6.7} & \cellcolor[HTML]{F98C8E}0.865 & 10.7 & \multicolumn{1}{c|}{} & \cellcolor[HTML]{F97173}0.973 & \textbf{7.1} & \cellcolor[HTML]{F96A6C}0.980 & \textbf{6.6} & \cellcolor[HTML]{FAA2A5}0.876 & \textbf{9.8} \\
\multicolumn{1}{l|}{ORUSB1} & \cellcolor[HTML]{FBCCCF}0.930 & 14.3 & \cellcolor[HTML]{FBB8BB}0.912 & 15.1 & \cellcolor[HTML]{5A8AC6}0.804 & 14.7 & \multicolumn{1}{c|}{} & \cellcolor[HTML]{FCE5E8}0.881 & 17.4 & \cellcolor[HTML]{F6F7FC}0.660 & 20.6 & \cellcolor[HTML]{FBC1C4}0.756 & 17.6 & \multicolumn{1}{c|}{} & \cellcolor[HTML]{FBBABD}0.912 & 11.3 & \cellcolor[HTML]{FCD8DB}0.799 & 15.0 & \cellcolor[HTML]{FCE8EB}0.769 & 14.5 \\
\multicolumn{1}{l|}{ORUSB2} & \cellcolor[HTML]{FBBFC1}0.934 & 14.8 & \cellcolor[HTML]{FA9EA1}0.932 & 14.8 & \cellcolor[HTML]{AAC2E2}0.836 & 14.1 & \multicolumn{1}{c|}{} & \cellcolor[HTML]{FA9597}0.943 & 12.4 & \cellcolor[HTML]{F98588}0.917 & 15.1 & \cellcolor[HTML]{F98789}0.874 & 12.4 & \multicolumn{1}{c|}{} & \cellcolor[HTML]{FA9294}0.945 & 12.4 & \cellcolor[HTML]{F98284}0.942 & 13.6 & \cellcolor[HTML]{FAB3B5}0.851 & 12.5 \\
\multicolumn{1}{l|}{ORUSB3} & \cellcolor[HTML]{FBBABC}0.936 & 13.7 & \cellcolor[HTML]{FAA5A8}0.927 & 13.5 & \cellcolor[HTML]{5C8BC6}0.805 & 16.4 & \multicolumn{1}{c|}{} & \cellcolor[HTML]{FAADB0}0.924 & 13.6 & \cellcolor[HTML]{FBB8BB}0.808 & 16.0 & \cellcolor[HTML]{FAB1B3}0.790 & 17.5 & \multicolumn{1}{c|}{} & \cellcolor[HTML]{F98082}0.960 & 9.4 & \cellcolor[HTML]{F9878A}0.932 & 11.3 & \cellcolor[HTML]{FBD0D3}0.806 & 14.6 \\
\multicolumn{1}{l|}{OOB} & \cellcolor[HTML]{FA9698}0.949 & 10.9 & \cellcolor[HTML]{FA8F92}0.944 & 11.2 & \cellcolor[HTML]{F98588}0.897 & \textbf{10.0} & \multicolumn{1}{c|}{} & \cellcolor[HTML]{FA9496}0.944 & 9.8 & \cellcolor[HTML]{F98083}0.927 & 10.3 & \cellcolor[HTML]{F8696B}0.934 & \textbf{7.6} & \multicolumn{1}{c|}{} & \cellcolor[HTML]{FA9092}0.947 & 10.8 & \cellcolor[HTML]{F97F81}0.947 & 9.4 & \cellcolor[HTML]{F96E70}0.955 & \textbf{7.9} \\
\multicolumn{1}{l|}{OUB} & \cellcolor[HTML]{FAA0A3}0.945 & 13.7 & \cellcolor[HTML]{F98B8D}0.948 & 13.6 & \cellcolor[HTML]{C8D8ED}0.848 & 14.3 & \multicolumn{1}{c|}{} & \cellcolor[HTML]{FAA8AB}0.928 & 12.9 & \cellcolor[HTML]{FA9294}0.890 & 14.0 & \cellcolor[HTML]{F98D90}0.862 & 13.1 & \multicolumn{1}{c|}{} & \cellcolor[HTML]{F8696B}0.978 & \textbf{5.4} & \cellcolor[HTML]{F96F71}0.973 & \textbf{7.2} & \cellcolor[HTML]{FBB3B6}0.850 & \textbf{10.8} \\
\multicolumn{1}{l|}{OWOB} & \cellcolor[HTML]{FA9B9D}0.947 & 11.3 & \cellcolor[HTML]{F98B8D}0.948 & 10.6 & \cellcolor[HTML]{F98385}0.898 & 10.3 & \multicolumn{1}{c|}{} & \cellcolor[HTML]{FA9DA0}0.936 & 9.7 & \cellcolor[HTML]{F97F81}0.930 & 8.4 & \cellcolor[HTML]{F96F71}0.922 & \textbf{7.1} & \multicolumn{1}{c|}{} & \cellcolor[HTML]{F98C8E}0.950 & 9.5 & \cellcolor[HTML]{F97C7E}0.951 & \textbf{8.2} & \cellcolor[HTML]{F8696B}0.962 & \textbf{6.3} \\
\multicolumn{1}{l|}{OWUB} & \cellcolor[HTML]{FA9C9E}0.947 & 13.6 & \cellcolor[HTML]{F98385}0.954 & 12.7 & \cellcolor[HTML]{B8CCE7}0.841 & 15.5 & \multicolumn{1}{c|}{} & \cellcolor[HTML]{F8696B}0.976 & \textbf{5.4} & \cellcolor[HTML]{F96B6D}0.973 & \textbf{6.2} & \cellcolor[HTML]{FA8F91}0.858 & \textbf{10.5} & \multicolumn{1}{c|}{} & \cellcolor[HTML]{F96E70}0.975 & \textbf{6.9} & \cellcolor[HTML]{F97072}0.971 & 8.4 & \cellcolor[HTML]{FAB0B2}0.855 & 11.5 \\
\multicolumn{1}{l|}{OEB} & \cellcolor[HTML]{FBB6B9}0.937 & 14.8 & \cellcolor[HTML]{FA9396}0.941 & 13.9 & \cellcolor[HTML]{89ABD6}0.823 & 16.7 & \multicolumn{1}{c|}{} & \cellcolor[HTML]{FAA1A3}0.934 & 11.7 & \cellcolor[HTML]{F98082}0.928 & 11.1 & \cellcolor[HTML]{FAA0A2}0.824 & 14.2 & \multicolumn{1}{c|}{} & \cellcolor[HTML]{F97B7D}0.964 & 9.5 & \cellcolor[HTML]{F97173}0.968 & 9.5 & \cellcolor[HTML]{FAB2B5}0.851 & 12.7 \\
\multicolumn{21}{l}{\cellcolor[HTML]{C0C0C0}\textbf{The Proposed Approach}} \\
\multicolumn{1}{l|}{HGD} & \cellcolor[HTML]{F96C6E}0.964 & \textbf{7.5} & \cellcolor[HTML]{F96E70}0.971 & \textbf{6.7} & \cellcolor[HTML]{ECF1F9}0.862 & \textbf{9.8} & \multicolumn{1}{c|}{} & \cellcolor[HTML]{F97274}0.970 & \textbf{7.2} & \cellcolor[HTML]{F8696B}0.976 & \textbf{6.5} & \cellcolor[HTML]{F9898B}0.871 & \textbf{9.6} & \multicolumn{1}{c|}{} & \cellcolor[HTML]{F97476}0.970 & \textbf{7.0} & \cellcolor[HTML]{F96B6D}0.980 & \textbf{6.2} & \cellcolor[HTML]{FAA2A4}0.876 & \textbf{8.6} \\ \hline
\end{tabular}
}
\begin{tablenotes}
    \item{$\cdot$} The color of cells shows the performance increments (in red) or decrements (in blue) w.r.t. the BaseLine (OGD). 
    \item{$\cdot$} Rank was calculated by averaging ranks in all datasets. 
    \item{$\cdot$} The bold entries stand for the top five highest rankings.
\end{tablenotes}
\label{S:tab:cmp_static}
\end{table*}
%========================================

%===============================================
%===============================================

%===============================================
%===============================================
%===============================================
%===============================================
%===============================================
%===============================================

\clearpage
\newpage

\subsection{Parameter Sensitivity}

\begin{figure}[h]
    \centering
    \subfloat[AUC]{
        \includegraphics[width=15cm]{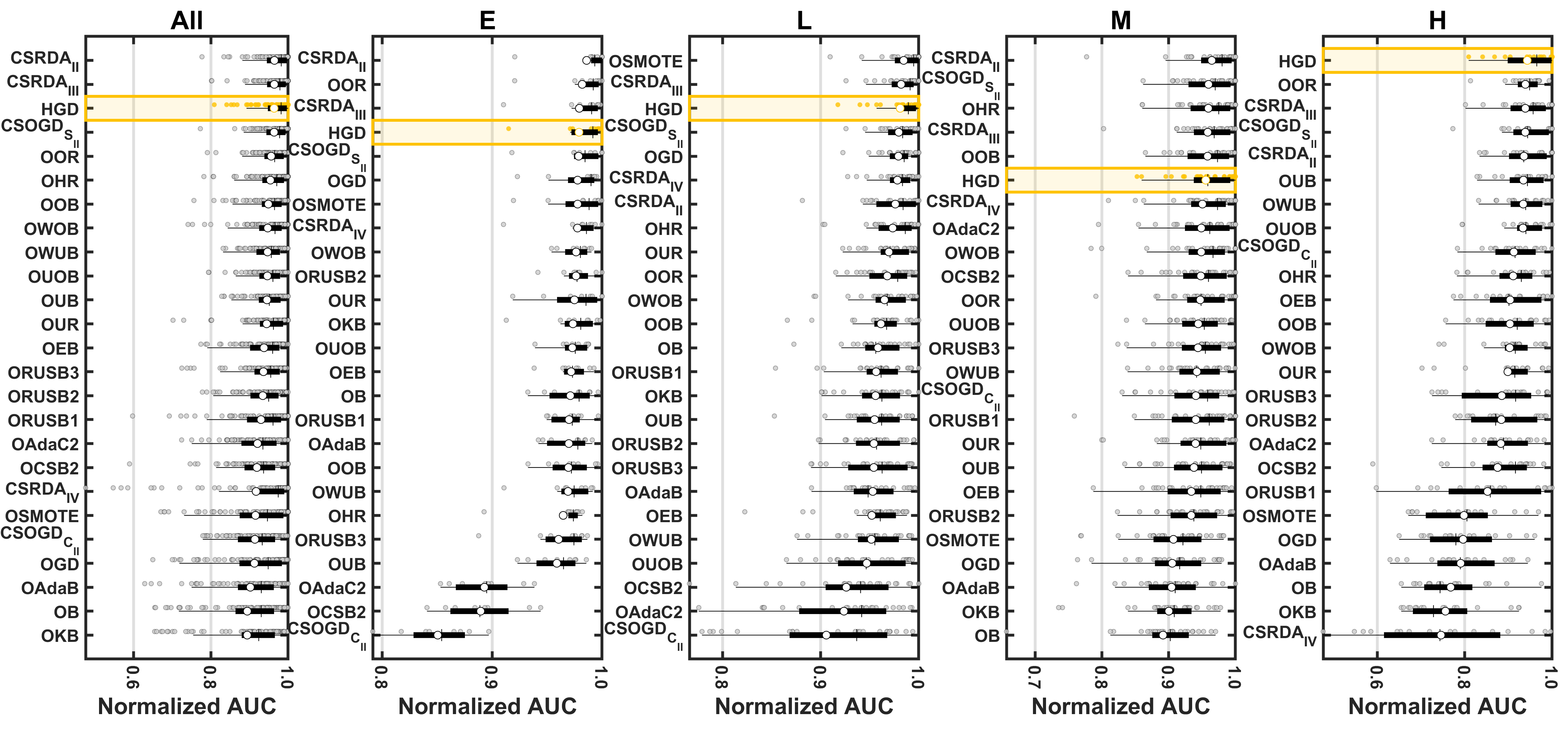}
    }\\
    \subfloat[GMEANS]{
        \includegraphics[width=15cm]{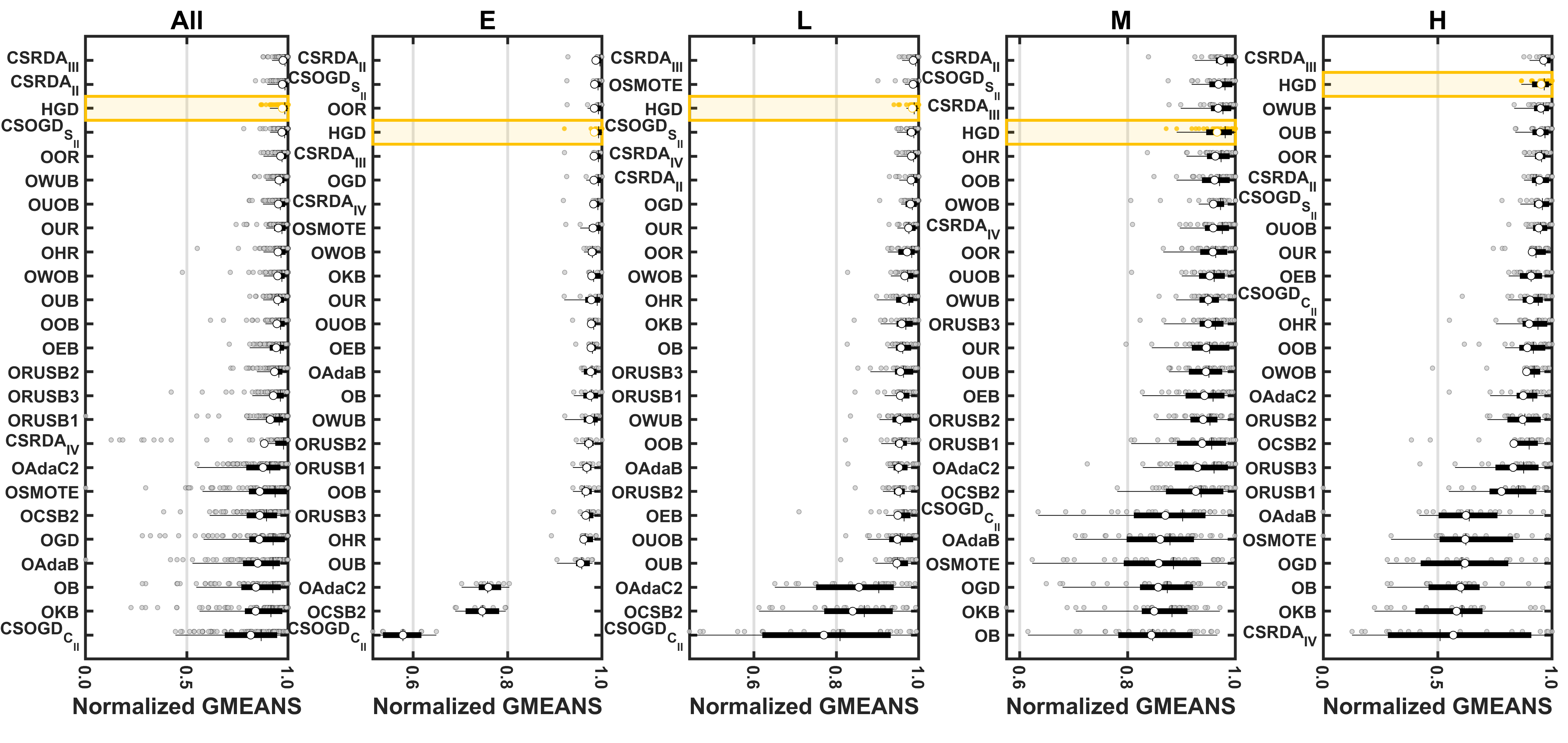}
    }\\
    \subfloat[F1]{
        \includegraphics[width=15cm]{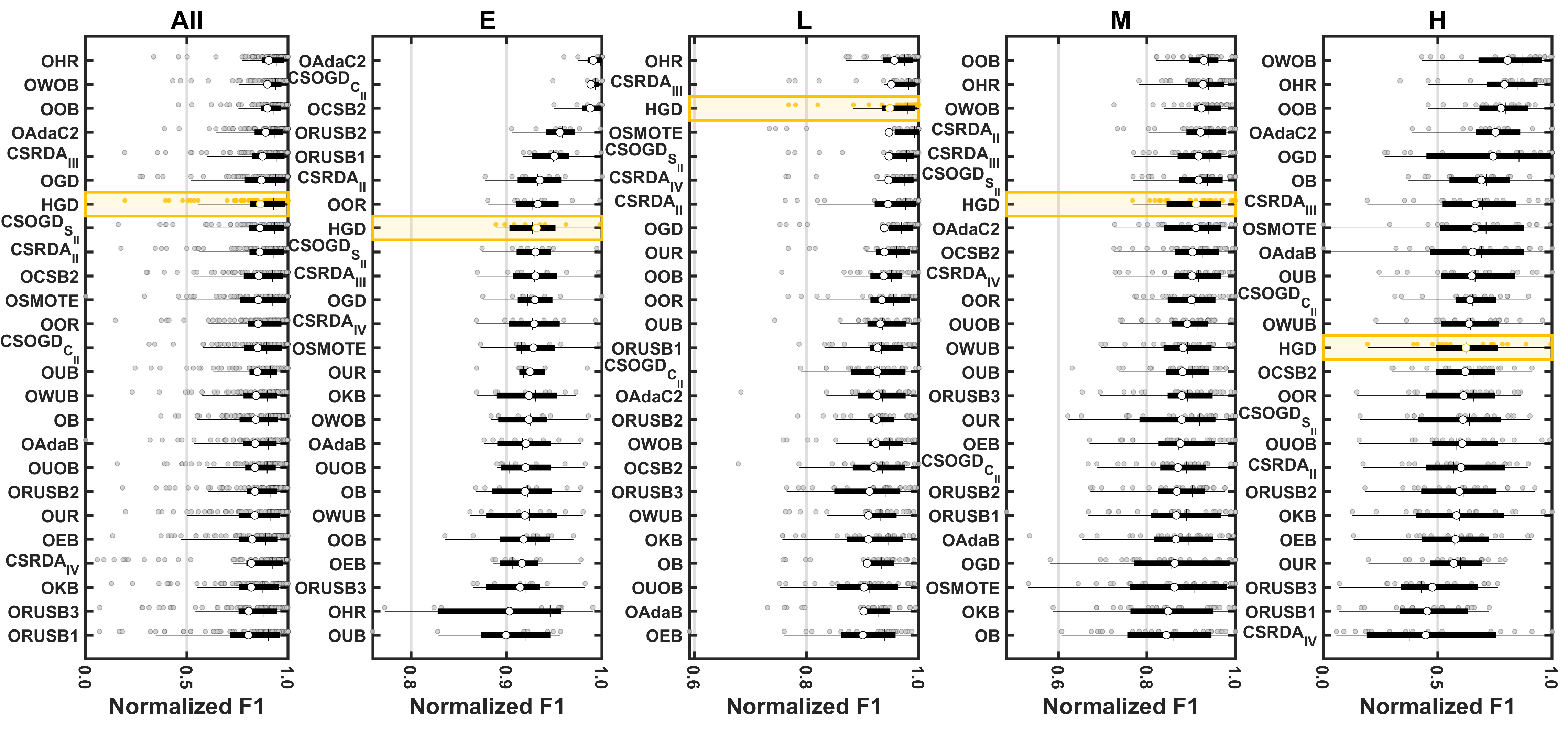}
    } \\
    \caption{The performance of different methods w.r.t. all, equal, low, medium, high imbalance ratios datasets. Perceptron as the base learner. }
    \label{S:fig:static_BoxPlot_P}
\end{figure}

\begin{figure}[h]
    \centering
    \subfloat[AUC]{
        \includegraphics[width=15cm]{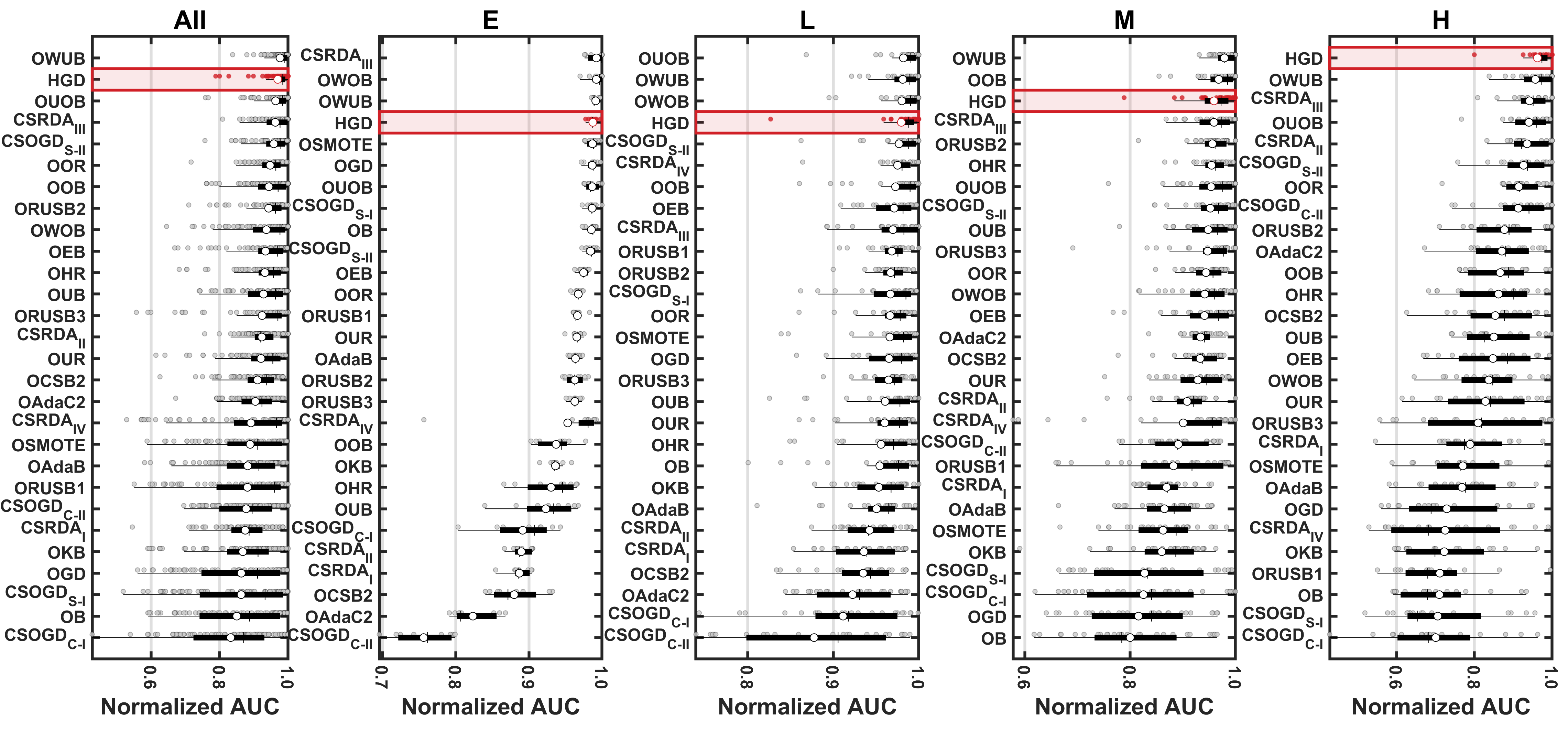}
    }\\
    \subfloat[GMEANS]{
        \includegraphics[width=15cm]{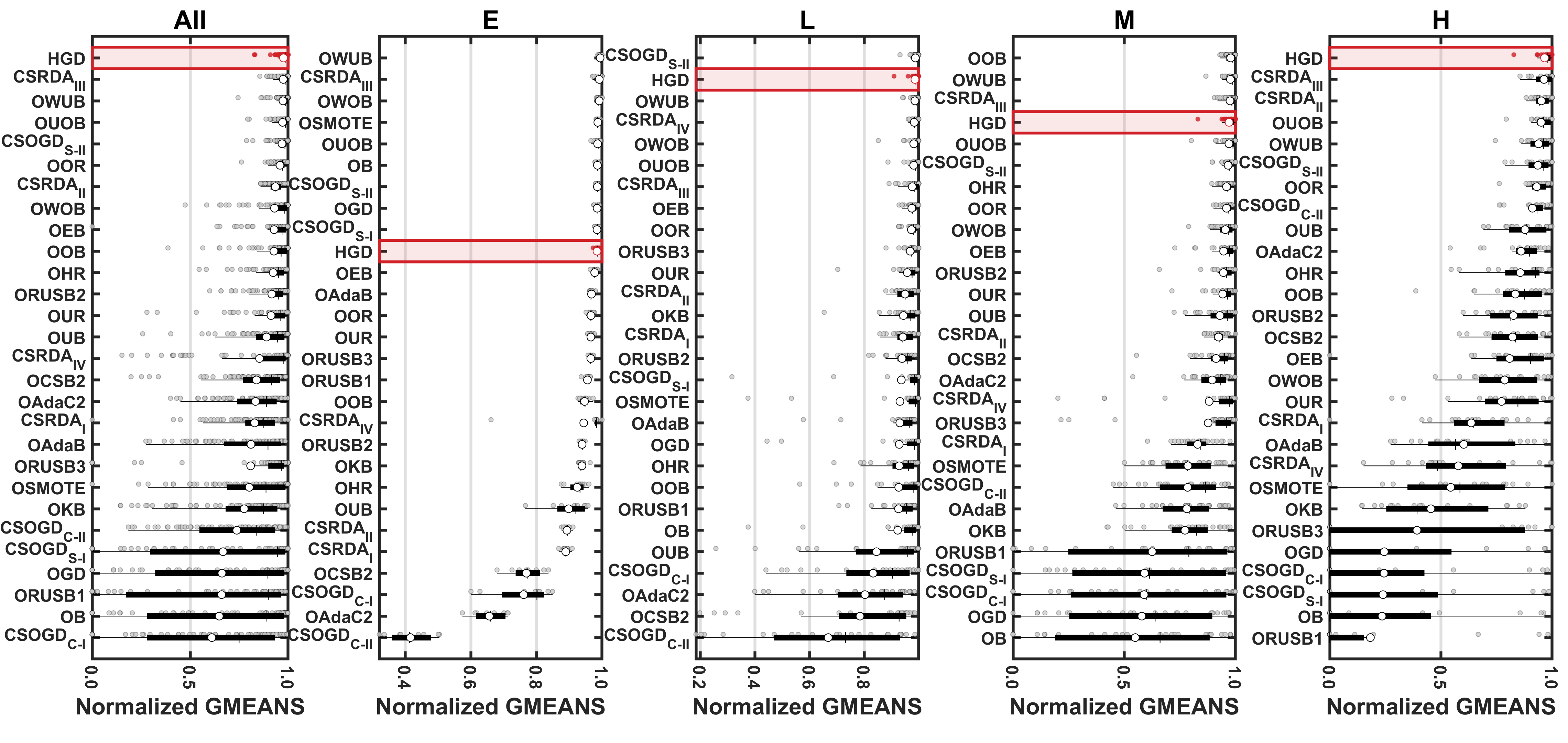}
    }\\
    \subfloat[F1]{
        \includegraphics[width=15cm]{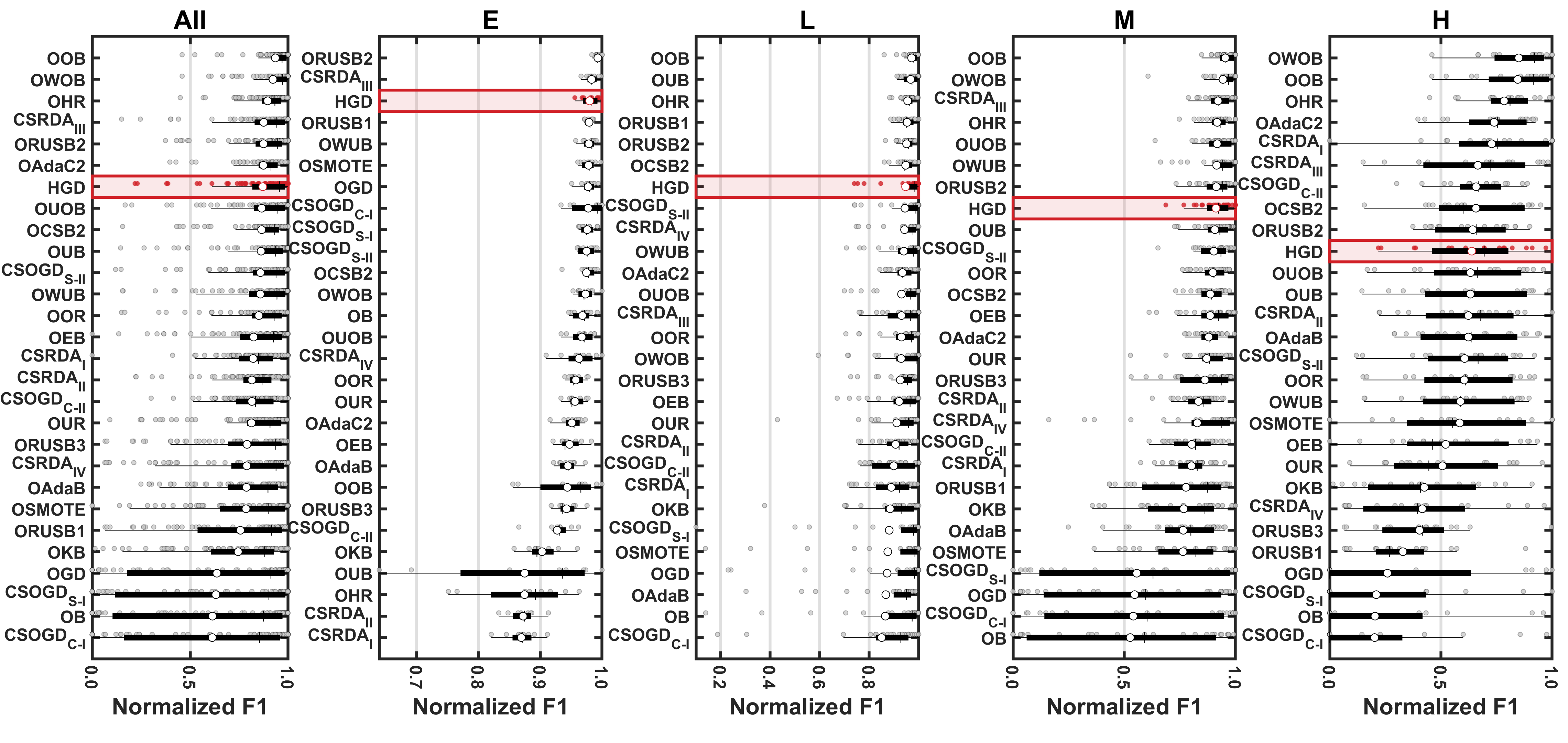}
    } \\
    \caption{The performance of different methods w.r.t. all, equal, low, medium, high imbalance ratios datasets. Linear SVM as the base learner. }
    \label{S:fig:static_BoxPlot_V}
\end{figure}

\begin{figure}[h]
    \centering
    \subfloat[AUC]{
        \includegraphics[width=15cm]{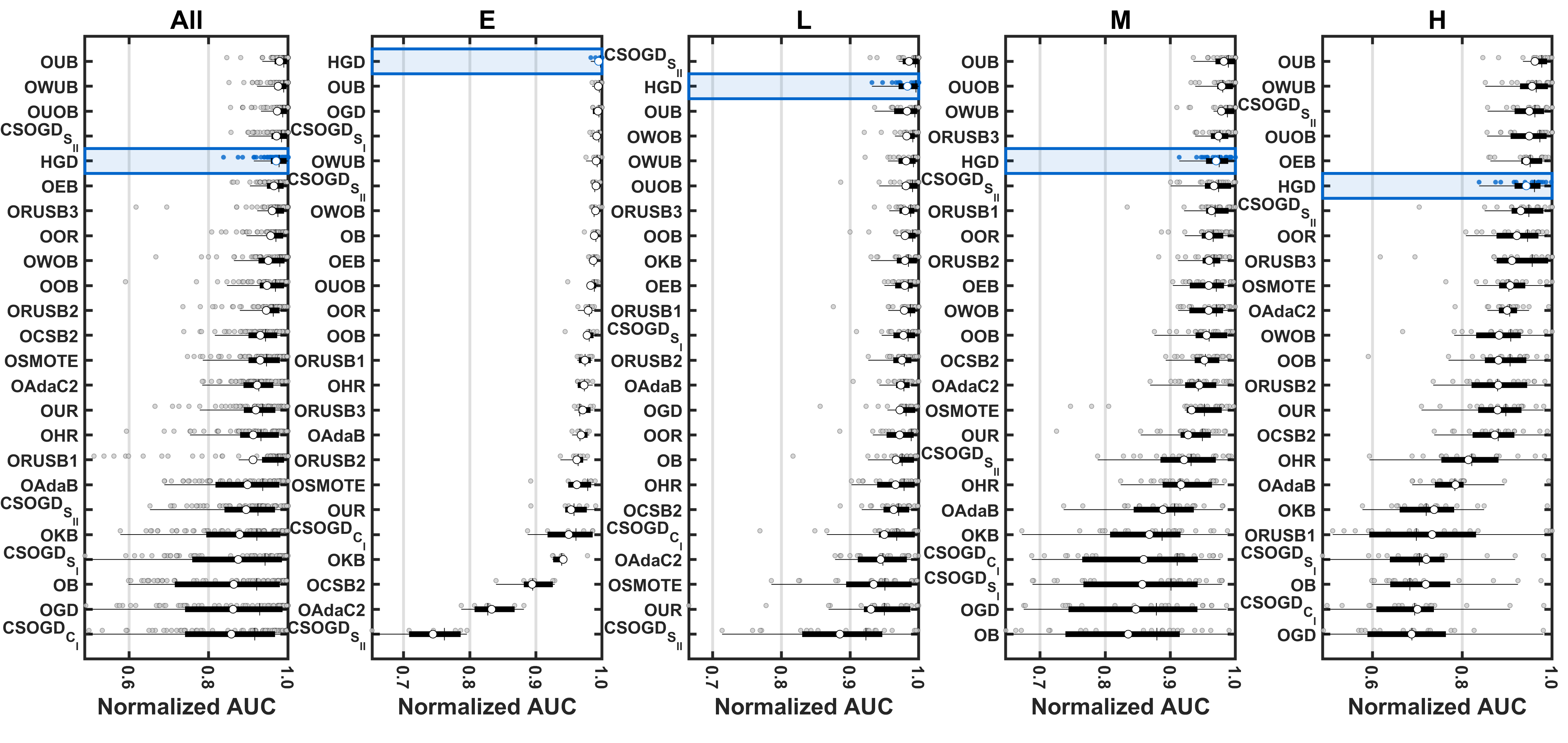}
    }\\
    \subfloat[GMEANS]{
        \includegraphics[width=15cm]{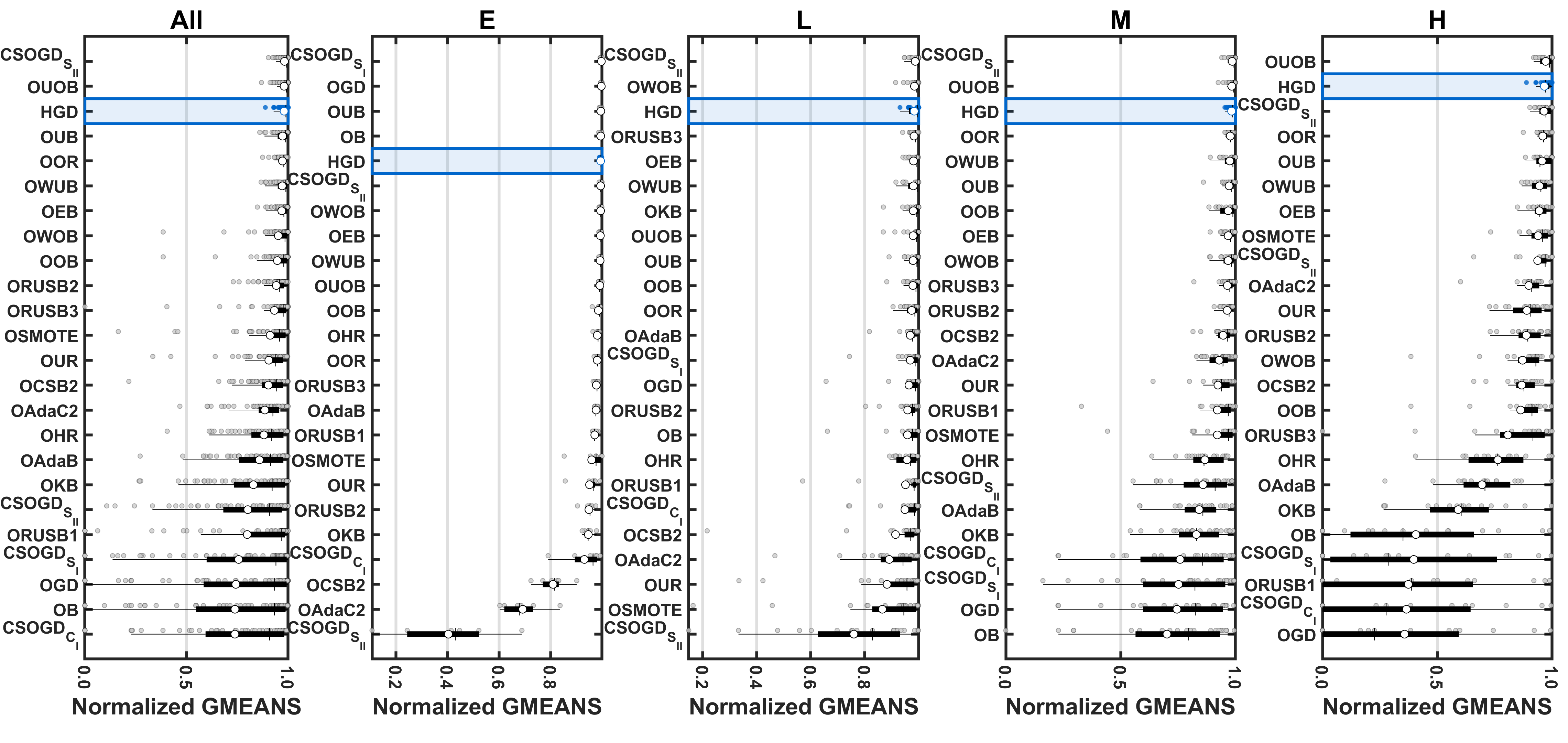}
    }\\
    \subfloat[F1]{
        \includegraphics[width=15cm]{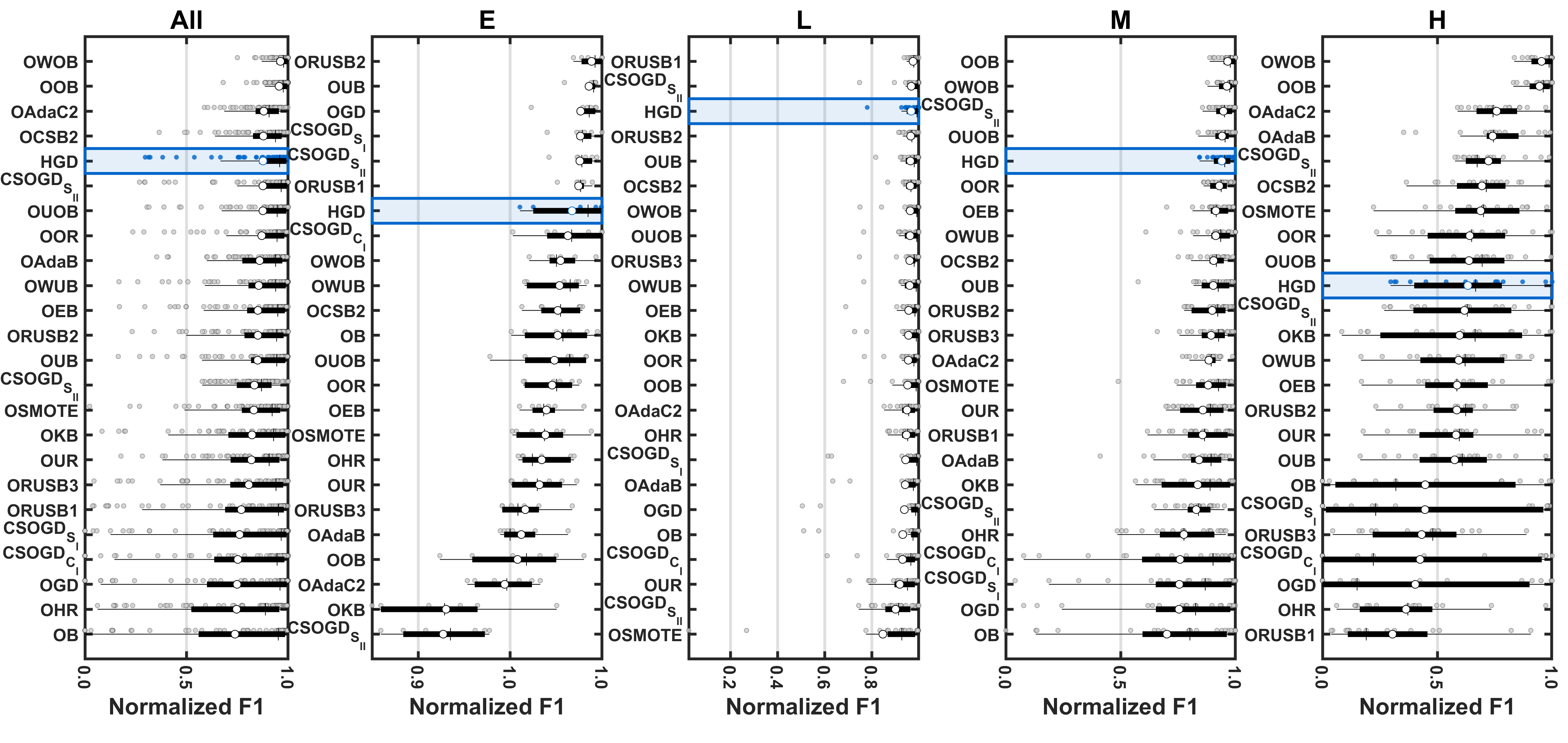}
    } \\
    \caption{The performance of different methods w.r.t. all, equal, low, medium, high imbalance ratios datasets. Kernel model the base learner. }
    \label{S:fig:static_BoxPlot_K}
\end{figure}

\begin{figure*}[h]
    \centering
    \subfloat[AUC]{
    \begin{minipage}[c]{0.3\textwidth}
        \includegraphics[width=5.5cm]{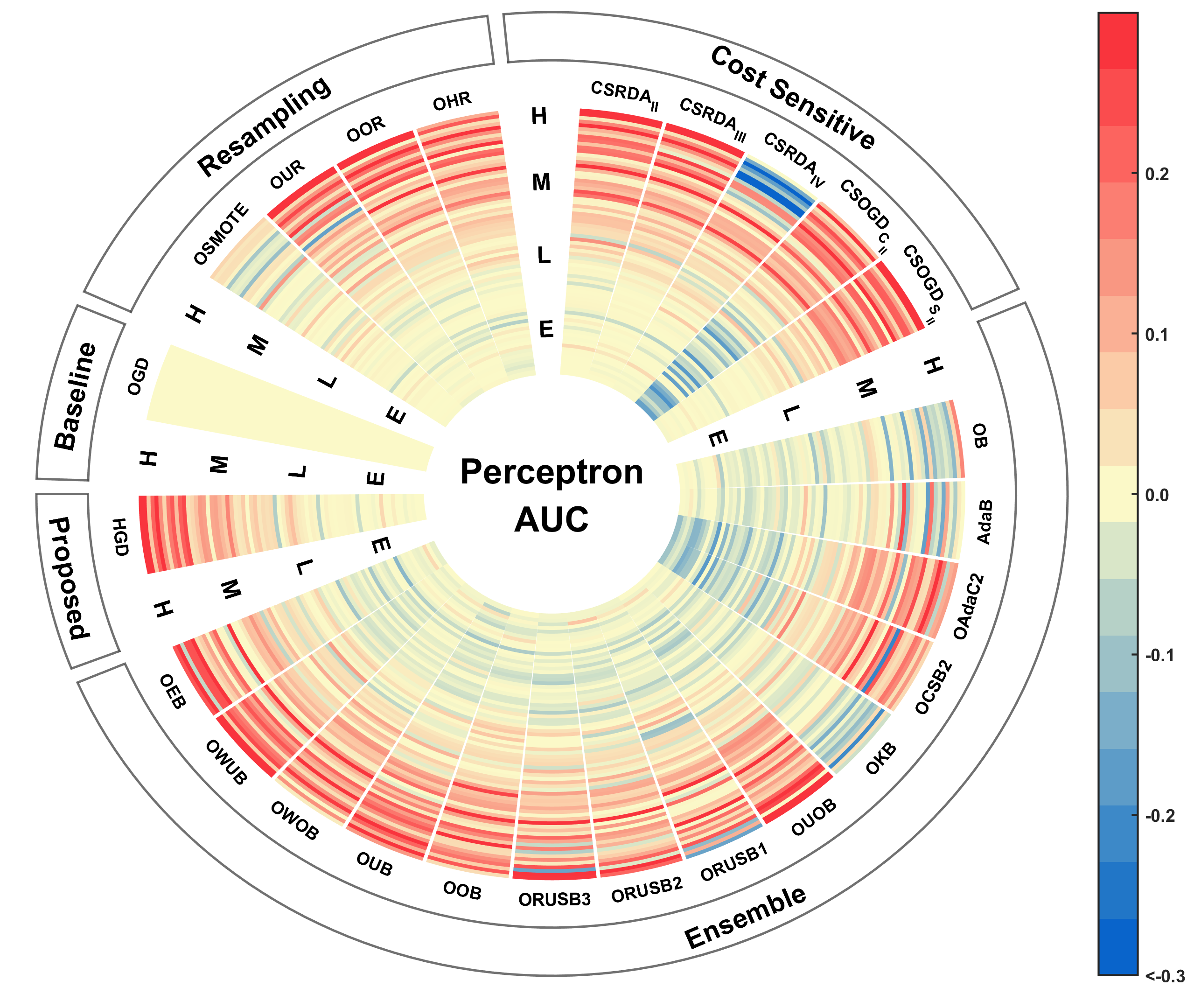}
    \end{minipage}}
    \subfloat[GMEANS]{
    \begin{minipage}[c]{0.3\textwidth}
        \includegraphics[width=5.5cm]{Figs/Exp_Static/Static_HeatMap_Dataset_P_GMEANS.png}
    \end{minipage}}
        \subfloat[F1]{
    \begin{minipage}[c]{0.3\textwidth}
        \includegraphics[width=5.5cm]{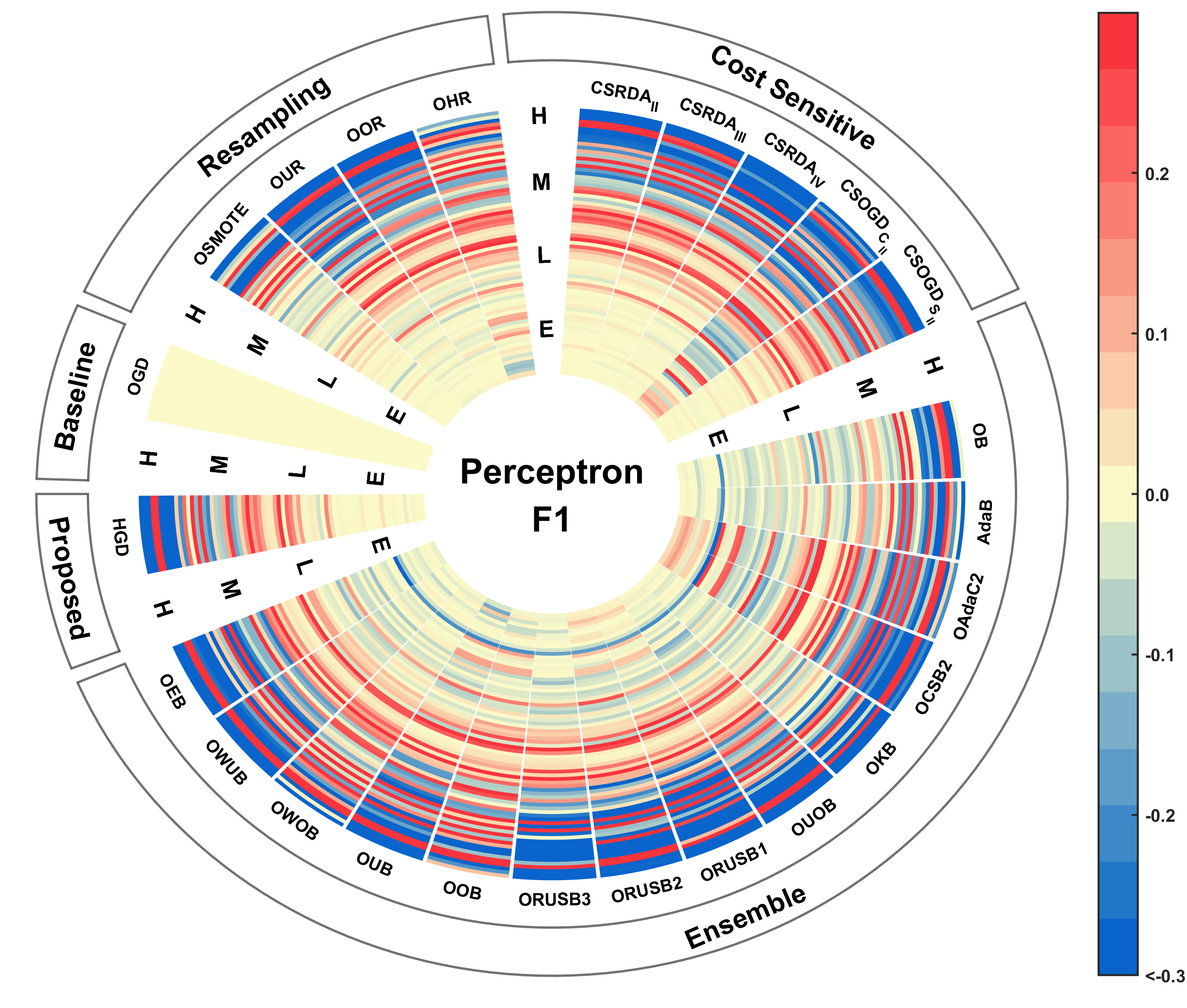}
    \end{minipage}} \\
    \subfloat[AUC]{
    \begin{minipage}[c]{0.3\textwidth}
        \includegraphics[width=5.5cm]{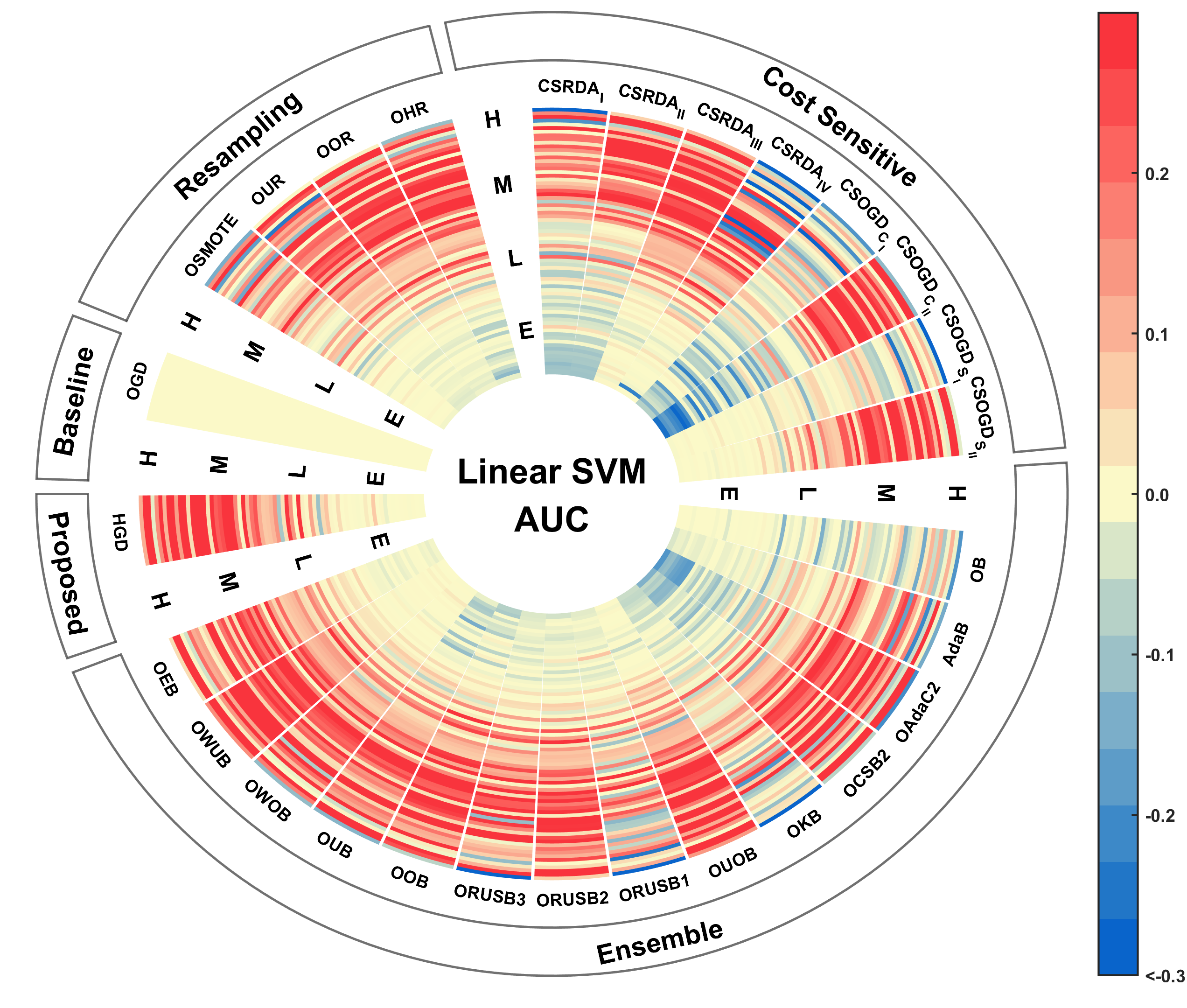}
    \end{minipage}}
    \subfloat[GMEANS]{
    \begin{minipage}[c]{0.3\textwidth}
        \includegraphics[width=5.5cm]{Figs/Exp_Static/Static_HeatMap_Dataset_V_GMEANS.png}
    \end{minipage}}
        \subfloat[F1]{
    \begin{minipage}[c]{0.3\textwidth}
        \includegraphics[width=5.5cm]{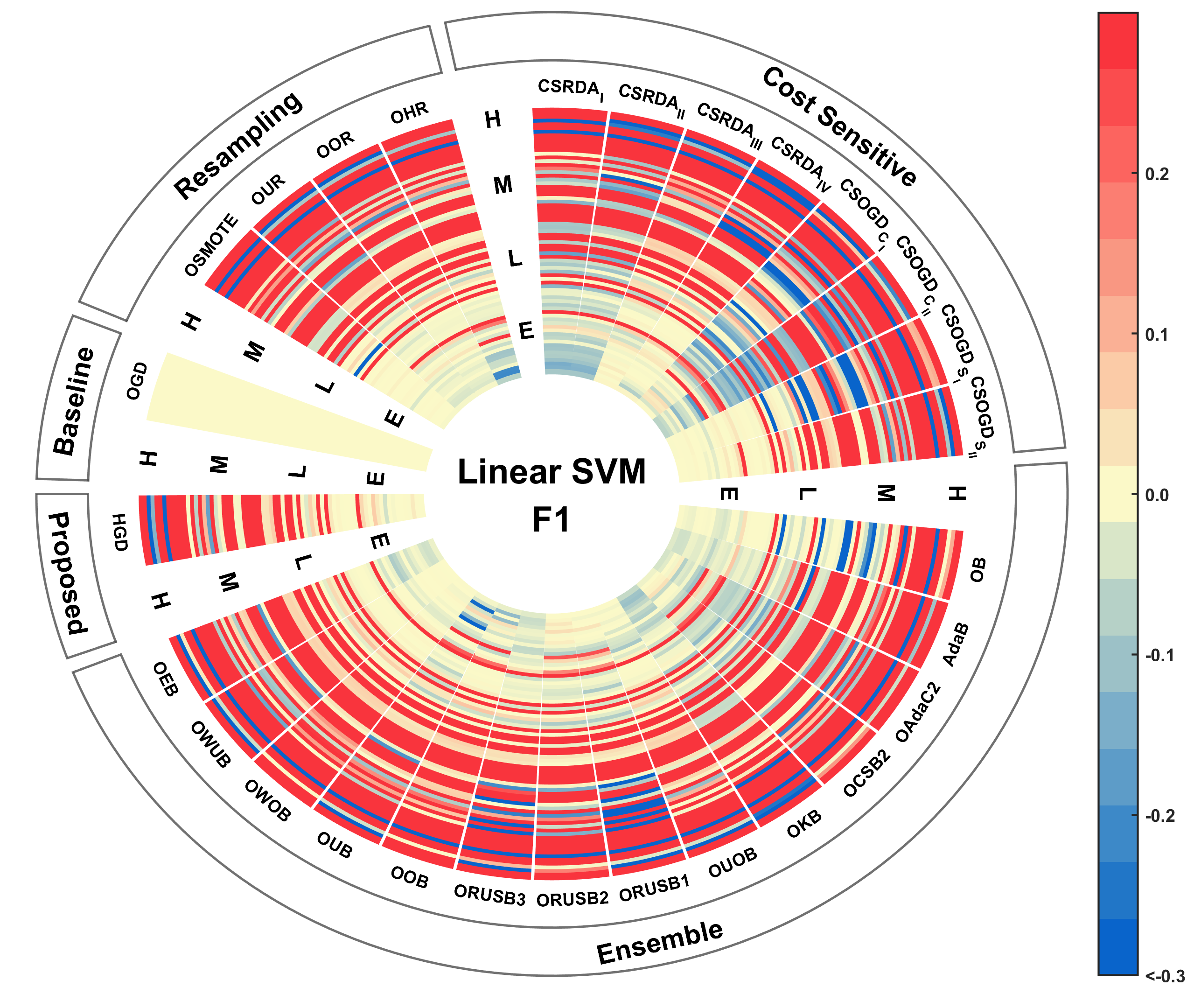}
    \end{minipage}} \\
        \subfloat[AUC]{
    \begin{minipage}[c]{0.3\textwidth}
        \includegraphics[width=5.5cm]{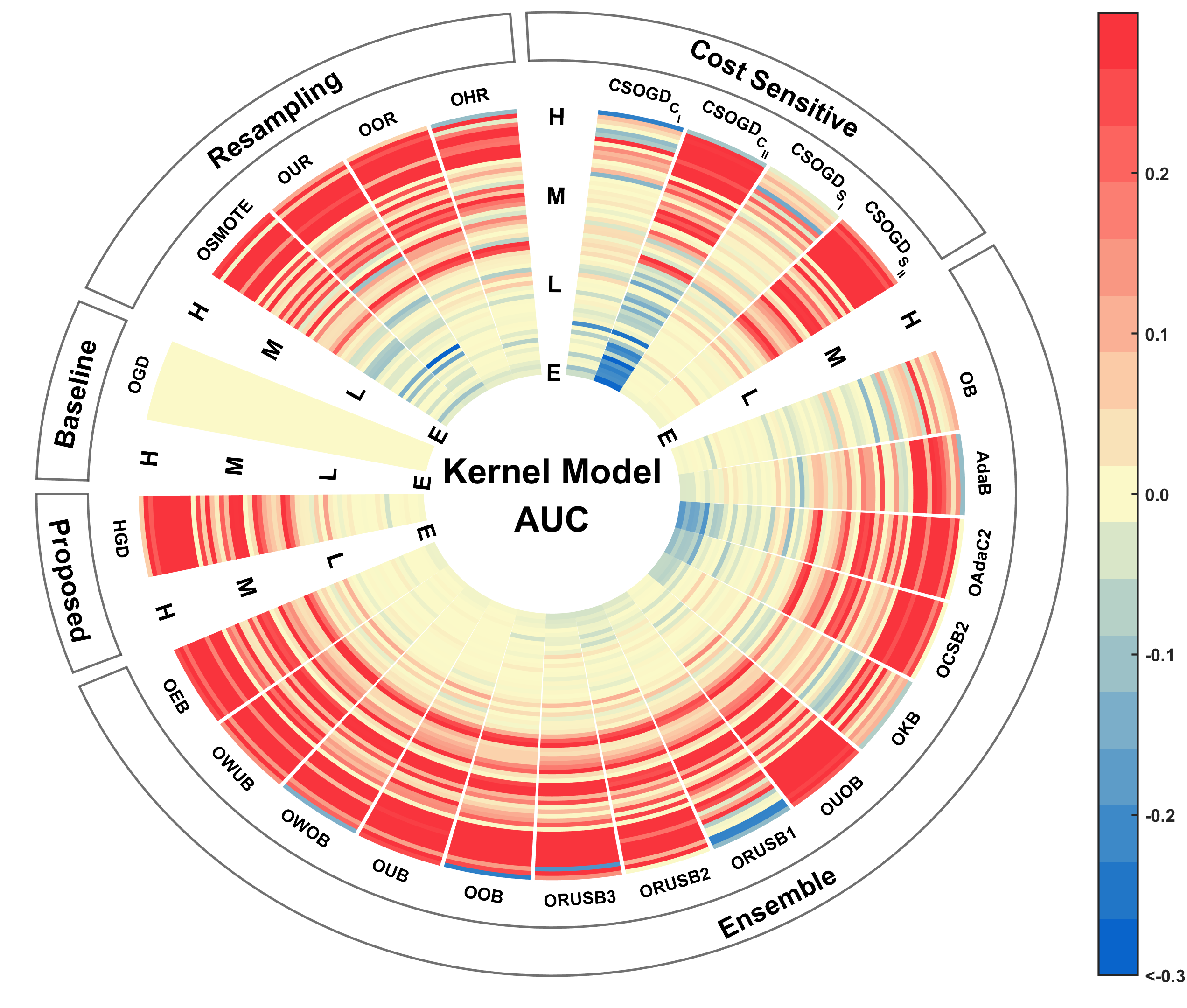}
    \end{minipage}}
    \subfloat[GMEANS]{
    \begin{minipage}[c]{0.3\textwidth}
        \includegraphics[width=5.5cm]{Figs/Exp_Static/Static_HeatMap_Dataset_K_GMEANS.png}
    \end{minipage}}
        \subfloat[F1]{
    \begin{minipage}[c]{0.3\textwidth}
        \includegraphics[width=5.5cm]{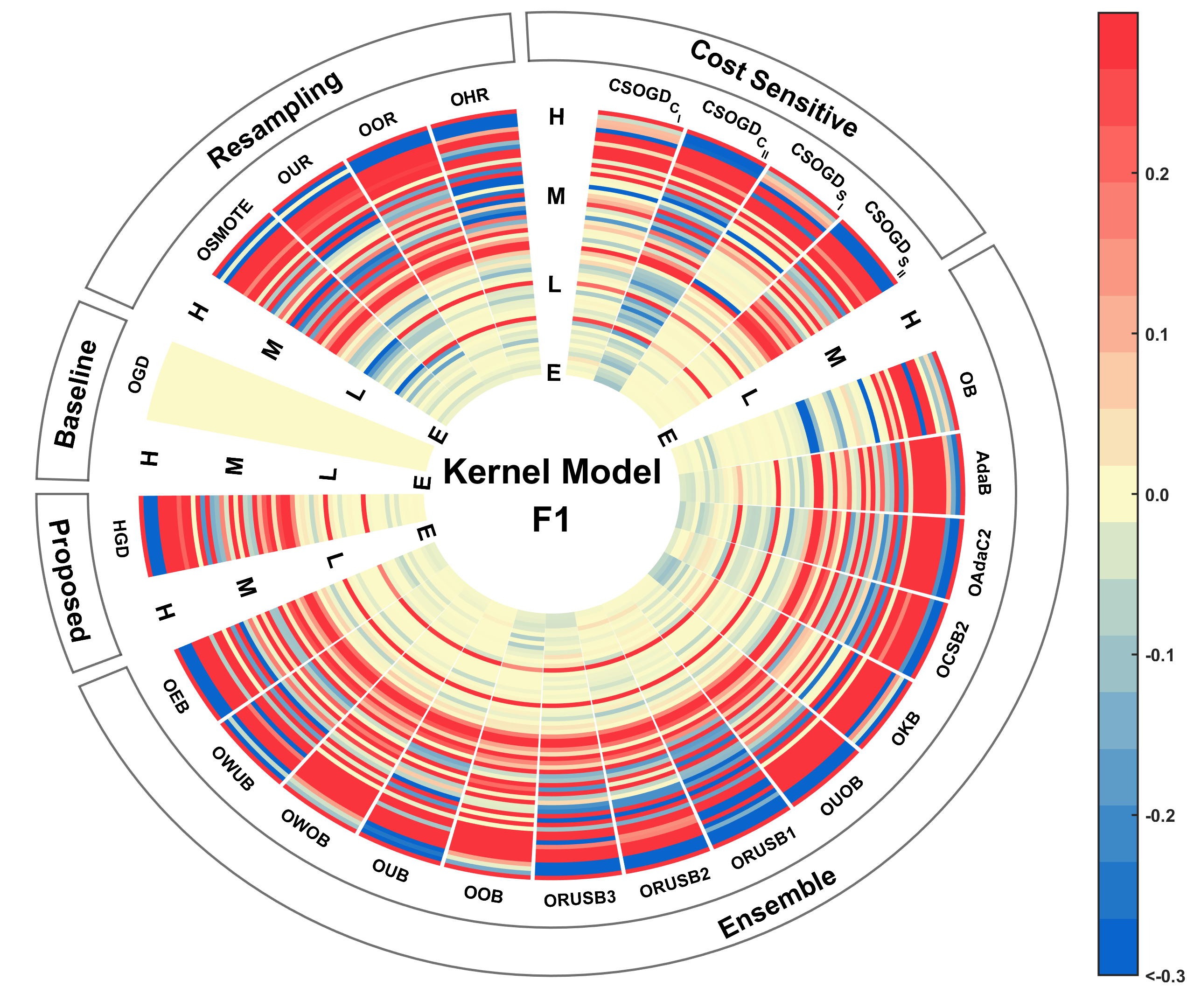}
    \end{minipage}} \\
    \caption{The circular heat maps. Each wedge represents the performance of a method on all datasets. The inner segments denote results on datasets with lower imbalance ratios, while the outer segments denote results on datasets with higher imbalance ratios. Blue shades indicate decreasing performance and red shades indicate increasing performance, w.r.t the baseline OGD.}
    \label{S:fig:static_CirHeatMap}
\end{figure*}

Many techniques may require intricate hyper-parameter tuning to adapt to different imbalance ratios, whereas our algorithm requires no parameter settings for imbalance learning. We show the performance attained by different methods over all datasets w.r.t. their Imbalance Ratios (IR, i.e., Equal IR, Low IR, Medium IR and High IR) in Table~\ref{S:tab:cmp_static_p}-\ref{S:tab:cmp_static_v} and Figure~\ref{S:fig:static_BoxPlot_P}-\ref{S:fig:static_BoxPlot_K}. 
Meanwhile, circular heat maps were generated to visualize the performance, as shown in Figure~\ref{S:fig:static_CirHeatMap}. In these figures, each wedge represents the performance of a method on all datasets. The inner segments denote results on datasets with lower imbalance ratios, while the outer segments denote results on datasets with higher imbalance ratios. The color of each segment in the circular heat map corresponds to performance variations, with blue shades indicating lower performance and redder shades indicating higher performance. This visualization leads to several interesting observations.
\begin{itemize}
    \item Generally, the performance improvements of most methods relative to the baseline increase as the imbalance ratio becomes larger, especially in terms of AUC and GMEANS. This trend demonstrates the effectiveness of these methods in imbalanced data stream learning.
    \item However, the performance of certain methods may exhibit decrements relative to the baseline. This phenomenon arises due to the necessity for precise parameter tuning; if parameters are not well-optimized, performance may suffer. In contrast, the proposed method consistently achieves good performance without any parameter tuning.
\end{itemize}

%===============================================
\begin{table*}[h]
\scriptsize
\centering
\renewcommand\arraystretch{1.2}
\caption{The Averaged Performance (AUC, G-means and F1-Score) Attained by Different Methods Over All Datasets w.r.t.
Their Imbalance Ratios (IR, i.e., Equal IR, Low IR, Medium IR and High IR). Perceptron as the Base Learner.}
\resizebox{\linewidth}{!}{
\begin{tabular}{lccccccccccccccccccc}
\hline
\multicolumn{1}{l|}{\textbf{Metric}} & \multicolumn{4}{c|}{\textbf{AUC}} & \multicolumn{1}{c|}{} & \textbf{} & \multicolumn{4}{c|}{\textbf{GMEANS}} & \multicolumn{1}{c|}{} & \textbf{} & \multicolumn{4}{c|}{\textbf{F1}} & \multicolumn{1}{c|}{} & \textbf{} &  \\ \cline{1-5} \cline{8-11} \cline{14-17}
\multicolumn{1}{l|}{\textbf{IR}} & \textbf{Equal} & \textbf{Low} & \textbf{Medium} & \multicolumn{1}{c|}{\textbf{High}} & \multicolumn{1}{c|}{\multirow{-2}{*}{\textbf{Rank}}} & \textbf{} & \textbf{Equal} & \textbf{Low} & \textbf{Medium} & \multicolumn{1}{c|}{\textbf{High}} & \multicolumn{1}{c|}{\multirow{-2}{*}{\textbf{Rank}}} & \textbf{} & \textbf{Equal} & \textbf{Low} & \textbf{Medium} & \multicolumn{1}{c|}{\textbf{High}} & \multicolumn{1}{c|}{\multirow{-2}{*}{\textbf{Rank}}} & \textbf{} & \multirow{-2}{*}{\textbf{\begin{tabular}[c]{@{}c@{}}Overall\\ Rank\end{tabular}}} \\ \cline{1-6} \cline{8-12} \cline{14-18} \cline{20-20} 
\multicolumn{20}{l}{\cellcolor[HTML]{C0C0C0}\textbf{Baseline:}} \\
\multicolumn{1}{l|}{OGD} & \cellcolor[HTML]{FCFCFF}0.000 & \cellcolor[HTML]{FCFCFF}0.000 & \cellcolor[HTML]{FCFCFF}0.000 & \multicolumn{1}{c|}{\cellcolor[HTML]{FCFCFF}0.000} & \multicolumn{1}{c|}{14.6} &  & \cellcolor[HTML]{FCFCFF}0.000 & \cellcolor[HTML]{FCFCFF}0.000 & \cellcolor[HTML]{FCFCFF}0.000 & \multicolumn{1}{c|}{\cellcolor[HTML]{FCFCFF}0.000} & \multicolumn{1}{c|}{15.1} &  & \cellcolor[HTML]{FCFCFF}0.000 & \cellcolor[HTML]{FCFCFF}0.000 & \cellcolor[HTML]{FCFCFF}0.000 & \multicolumn{1}{c|}{\cellcolor[HTML]{FCFCFF}0.000} & \multicolumn{1}{c|}{12.5} &  & 9.2 \\
\multicolumn{20}{l}{\cellcolor[HTML]{C0C0C0}\textbf{Data Level Approaches}} \\
\multicolumn{1}{l|}{OSMOTE} & \cellcolor[HTML]{FBFBFE}0.000 & \cellcolor[HTML]{FCF8FB}0.005 & \cellcolor[HTML]{FCFBFE}0.002 & \multicolumn{1}{c|}{\cellcolor[HTML]{FCFAFD}0.003} & \multicolumn{1}{c|}{13.4} &  & \cellcolor[HTML]{FBFBFE}-0.002 & \cellcolor[HTML]{FCFBFE}0.006 & \cellcolor[HTML]{FCFCFF}0.001 & \multicolumn{1}{c|}{\cellcolor[HTML]{FCFCFF}0.002} & \multicolumn{1}{c|}{14.2} &  & \cellcolor[HTML]{FBFBFE}-0.002 & \cellcolor[HTML]{FCEDF0}0.009 & \cellcolor[HTML]{FBFBFE}-0.001 & \multicolumn{1}{c|}{\cellcolor[HTML]{D0DDEF}-0.107} & \multicolumn{1}{c|}{11.6} &  & 8.6 \\
\multicolumn{1}{l|}{OUR} & \cellcolor[HTML]{F8F9FD}-0.003 & \cellcolor[HTML]{EFF3FA}-0.010 & \cellcolor[HTML]{FCDEE1}0.038 & \multicolumn{1}{c|}{\cellcolor[HTML]{FA9698}0.129} & \multicolumn{1}{c|}{11.6} &  & \cellcolor[HTML]{F9FAFE}-0.005 & \cellcolor[HTML]{FAFAFE}-0.005 & \cellcolor[HTML]{FCE1E4}0.104 & \multicolumn{1}{c|}{\cellcolor[HTML]{F97F81}0.474} & \multicolumn{1}{c|}{11.1} &  & \cellcolor[HTML]{F9FAFE}-0.006 & \cellcolor[HTML]{FBFBFE}0.000 & \cellcolor[HTML]{FCDCDF}0.019 & \multicolumn{1}{c|}{\cellcolor[HTML]{9DB9DD}-0.232} & \multicolumn{1}{c|}{13.5} &  & 8.2 \\
\multicolumn{1}{l|}{OOR} & \cellcolor[HTML]{FCF9FC}0.004 & \cellcolor[HTML]{EDF1F9}-0.012 & \cellcolor[HTML]{FBD8DA}0.046 & \multicolumn{1}{c|}{\cellcolor[HTML]{F96E70}0.179} & \multicolumn{1}{c|}{\textbf{9.8}} &  & \cellcolor[HTML]{FCFCFF}0.001 & \cellcolor[HTML]{F8F9FD}-0.008 & \cellcolor[HTML]{FCDDE0}0.118 & \multicolumn{1}{c|}{\cellcolor[HTML]{F97173}0.526} & \multicolumn{1}{c|}{\textbf{9.2}} &  & \cellcolor[HTML]{FCF7FA}0.003 & \cellcolor[HTML]{FAFAFE}-0.004 & \cellcolor[HTML]{FAB0B2}0.044 & \multicolumn{1}{c|}{\cellcolor[HTML]{B5CAE6}-0.174} & \multicolumn{1}{c|}{12.2} &  & 7.1 \\
\multicolumn{1}{l|}{OHR} & \cellcolor[HTML]{EBF0F9}-0.013 & \cellcolor[HTML]{F4F6FC}-0.006 & \cellcolor[HTML]{FBCDD0}0.060 & \multicolumn{1}{c|}{\cellcolor[HTML]{F98A8C}0.144} & \multicolumn{1}{c|}{\textbf{9.8}} &  & \cellcolor[HTML]{F3F5FB}-0.022 & \cellcolor[HTML]{F6F7FC}-0.015 & \cellcolor[HTML]{FCDCDF}0.124 & \multicolumn{1}{c|}{\cellcolor[HTML]{F98487}0.454} & \multicolumn{1}{c|}{10.1} &  & \cellcolor[HTML]{F0F3FA}-0.029 & \cellcolor[HTML]{FCDBDE}0.019 & \cellcolor[HTML]{F97C7E}0.074 & \multicolumn{1}{c|}{\cellcolor[HTML]{F9888A}0.067} & \multicolumn{1}{c|}{\textbf{8.6}} &  & \textbf{6.2} \\
\multicolumn{20}{l}{\cellcolor[HTML]{C0C0C0}\textbf{Cost Sensitive Approaches}} \\
\multicolumn{1}{l|}{CSRDA$_{II}$} & \cellcolor[HTML]{FCF6F9}0.008 & \cellcolor[HTML]{F7F9FD}-0.003 & \cellcolor[HTML]{FBC8CB}0.065 & \multicolumn{1}{c|}{\cellcolor[HTML]{F97173}0.175} & \multicolumn{1}{c|}{\textbf{7.8}} &  & \cellcolor[HTML]{FCFBFE}0.004 & \cellcolor[HTML]{FCFCFF}0.001 & \cellcolor[HTML]{FCD8DB}0.136 & \multicolumn{1}{c|}{\cellcolor[HTML]{F97173}0.526} & \multicolumn{1}{c|}{\textbf{6.5}} &  & \cellcolor[HTML]{FCF3F6}0.006 & \cellcolor[HTML]{FCF0F3}0.007 & \cellcolor[HTML]{F98789}0.068 & \multicolumn{1}{c|}{\cellcolor[HTML]{AFC5E3}-0.188} & \multicolumn{1}{c|}{\textbf{9.7}} &  & \textbf{5.4} \\
\multicolumn{1}{l|}{CSRDA$_{III}$} & \cellcolor[HTML]{FCFBFE}0.002 & \cellcolor[HTML]{FCFAFD}0.003 & \cellcolor[HTML]{FBCED0}0.059 & \multicolumn{1}{c|}{\cellcolor[HTML]{F96E70}0.179} & \multicolumn{1}{c|}{\textbf{7.4}} &  & \cellcolor[HTML]{FCFCFF}0.000 & \cellcolor[HTML]{FCFBFE}0.007 & \cellcolor[HTML]{FCDADD}0.130 & \multicolumn{1}{c|}{\cellcolor[HTML]{F8696B}0.555} & \multicolumn{1}{c|}{\textbf{6.3}} &  & \cellcolor[HTML]{FCFBFE}0.001 & \cellcolor[HTML]{FCE6E8}0.013 & \cellcolor[HTML]{FA8F92}0.063 & \multicolumn{1}{c|}{\cellcolor[HTML]{D1DDEF}-0.105} & \multicolumn{1}{c|}{\textbf{9.0}} &  & \textbf{5.1} \\
\multicolumn{1}{l|}{CSRDA$_{IV}$} & \cellcolor[HTML]{FBFBFE}0.000 & \cellcolor[HTML]{FAFBFE}-0.001 & \cellcolor[HTML]{FBD5D7}0.050 & \multicolumn{1}{c|}{\cellcolor[HTML]{ACC4E3}-0.063} & \multicolumn{1}{c|}{12.0} &  & \cellcolor[HTML]{FBFBFE}-0.001 & \cellcolor[HTML]{FCFCFF}0.002 & \cellcolor[HTML]{FCDDE0}0.119 & \multicolumn{1}{c|}{\cellcolor[HTML]{DBE5F3}-0.082} & \multicolumn{1}{c|}{10.7} &  & \cellcolor[HTML]{FBFBFE}-0.001 & \cellcolor[HTML]{FCEEF1}0.008 & \cellcolor[HTML]{FAADB0}0.046 & \multicolumn{1}{c|}{\cellcolor[HTML]{5A8AC6}-0.397} & \multicolumn{1}{c|}{12.7} &  & 7.8 \\
\multicolumn{1}{l|}{CSOGD$_{C_{II}}$} & \cellcolor[HTML]{5A8AC6}-0.130 & \cellcolor[HTML]{9EBADE}-0.075 & \cellcolor[HTML]{FCDDE0}0.039 & \multicolumn{1}{c|}{\cellcolor[HTML]{F9888A}0.146} & \multicolumn{1}{c|}{16.3} &  & \cellcolor[HTML]{5A8AC6}-0.412 & \cellcolor[HTML]{A7C0E1}-0.215 & \cellcolor[HTML]{FCF8FB}0.015 & \multicolumn{1}{c|}{\cellcolor[HTML]{F98385}0.459} & \multicolumn{1}{c|}{18.7} &  & \cellcolor[HTML]{FA9092}0.063 & \cellcolor[HTML]{F6F8FD}-0.013 & \cellcolor[HTML]{FCEDF0}0.009 & \multicolumn{1}{c|}{\cellcolor[HTML]{C4D4EB}-0.136} & \multicolumn{1}{c|}{14.3} &  & 11.0 \\
\multicolumn{1}{l|}{CSOGD$_{S_{II}}$} & \cellcolor[HTML]{FCFCFF}0.001 & \cellcolor[HTML]{FCFCFF}0.000 & \cellcolor[HTML]{FBCDCF}0.060 & \multicolumn{1}{c|}{\cellcolor[HTML]{F96E70}0.179} & \multicolumn{1}{c|}{\textbf{7.5}} &  & \cellcolor[HTML]{FCFCFF}0.001 & \cellcolor[HTML]{FCFCFF}0.002 & \cellcolor[HTML]{FCDADD}0.131 & \multicolumn{1}{c|}{\cellcolor[HTML]{F97274}0.522} & \multicolumn{1}{c|}{\textbf{6.8}} &  & \cellcolor[HTML]{FCFBFE}0.001 & \cellcolor[HTML]{FCEDF0}0.009 & \cellcolor[HTML]{FA9092}0.063 & \multicolumn{1}{c|}{\cellcolor[HTML]{B3C9E5}-0.177} & \multicolumn{1}{c|}{\textbf{10.1}} &  & \textbf{5.6} \\
\multicolumn{20}{l}{\cellcolor[HTML]{C0C0C0}\textbf{Ensemble Learning Approaches}} \\
\multicolumn{1}{l|}{OB} & \cellcolor[HTML]{F3F6FC}-0.007 & \cellcolor[HTML]{E1E9F5}-0.021 & \cellcolor[HTML]{E9EEF8}-0.015 & \multicolumn{1}{c|}{\cellcolor[HTML]{CFDCEF}-0.036} & \multicolumn{1}{c|}{18.3} &  & \cellcolor[HTML]{F8F9FD}-0.008 & \cellcolor[HTML]{F2F5FB}-0.024 & \cellcolor[HTML]{F6F7FC}-0.015 & \multicolumn{1}{c|}{\cellcolor[HTML]{F0F3FA}-0.030} & \multicolumn{1}{c|}{18.7} &  & \cellcolor[HTML]{F7F8FD}-0.011 & \cellcolor[HTML]{EEF2FA}-0.032 & \cellcolor[HTML]{F3F5FB}-0.021 & \multicolumn{1}{c|}{\cellcolor[HTML]{DFE8F5}-0.069} & \multicolumn{1}{c|}{16.0} &  & 11.6 \\
\multicolumn{1}{l|}{OAdaB} & \cellcolor[HTML]{F1F4FB}-0.008 & \cellcolor[HTML]{DAE4F3}-0.027 & \cellcolor[HTML]{FAFBFE}-0.001 & \multicolumn{1}{c|}{\cellcolor[HTML]{F3F5FB}-0.007} & \multicolumn{1}{c|}{18.3} &  & \cellcolor[HTML]{F9FAFE}-0.007 & \cellcolor[HTML]{F0F3FA}-0.029 & \cellcolor[HTML]{FCFBFE}0.005 & \multicolumn{1}{c|}{\cellcolor[HTML]{FCFBFE}0.007} & \multicolumn{1}{c|}{18.6} &  & \cellcolor[HTML]{F7F9FD}-0.010 & \cellcolor[HTML]{ECF0F9}-0.039 & \cellcolor[HTML]{FCF6F9}0.003 & \multicolumn{1}{c|}{\cellcolor[HTML]{CBD9ED}-0.120} & \multicolumn{1}{c|}{16.0} &  & 11.5 \\
\multicolumn{1}{l|}{OAdaC2} & \cellcolor[HTML]{8FAFD8}-0.087 & \cellcolor[HTML]{B5CAE6}-0.057 & \cellcolor[HTML]{FBD6D9}0.048 & \multicolumn{1}{c|}{\cellcolor[HTML]{FAA6A8}0.109} & \multicolumn{1}{c|}{15.8} &  & \cellcolor[HTML]{A2BCDF}-0.228 & \cellcolor[HTML]{C9D8ED}-0.128 & \cellcolor[HTML]{FCE6E9}0.085 & \multicolumn{1}{c|}{\cellcolor[HTML]{FA9092}0.412} & \multicolumn{1}{c|}{17.7} &  & \cellcolor[HTML]{F98B8D}0.066 & \cellcolor[HTML]{F6F7FC}-0.014 & \cellcolor[HTML]{FA9C9E}0.056 & \multicolumn{1}{c|}{\cellcolor[HTML]{FCE4E7}0.014} & \multicolumn{1}{c|}{11.5} &  & 9.7 \\
\multicolumn{1}{l|}{OCSB2} & \cellcolor[HTML]{8BACD7}-0.090 & \cellcolor[HTML]{B8CCE7}-0.054 & \cellcolor[HTML]{FBD7DA}0.047 & \multicolumn{1}{c|}{\cellcolor[HTML]{FAADB0}0.099} & \multicolumn{1}{c|}{16.1} &  & \cellcolor[HTML]{9DB9DD}-0.241 & \cellcolor[HTML]{C3D4EB}-0.143 & \cellcolor[HTML]{FCE7EA}0.081 & \multicolumn{1}{c|}{\cellcolor[HTML]{FAA1A3}0.345} & \multicolumn{1}{c|}{18.6} &  & \cellcolor[HTML]{FA9193}0.062 & \cellcolor[HTML]{F3F6FC}-0.020 & \cellcolor[HTML]{FAABAE}0.047 & \multicolumn{1}{c|}{\cellcolor[HTML]{B9CDE7}-0.164} & \multicolumn{1}{c|}{13.4} &  & 10.7 \\
\multicolumn{1}{l|}{OKB} & \cellcolor[HTML]{F6F8FD}-0.004 & \cellcolor[HTML]{DEE7F4}-0.024 & \cellcolor[HTML]{F5F7FC}-0.006 & \multicolumn{1}{c|}{\cellcolor[HTML]{BCCFE8}-0.051} & \multicolumn{1}{c|}{18.5} &  & \cellcolor[HTML]{FAFAFE}-0.005 & \cellcolor[HTML]{F3F5FB}-0.022 & \cellcolor[HTML]{F8F9FD}-0.009 & \multicolumn{1}{c|}{\cellcolor[HTML]{E4EBF6}-0.059} & \multicolumn{1}{c|}{18.7} &  & \cellcolor[HTML]{F9FAFE}-0.007 & \cellcolor[HTML]{EFF3FA}-0.030 & \cellcolor[HTML]{F5F7FC}-0.017 & \multicolumn{1}{c|}{\cellcolor[HTML]{A4BEE0}-0.215} & \multicolumn{1}{c|}{16.7} &  & 11.8 \\
\multicolumn{1}{l|}{OUOB} & \cellcolor[HTML]{F5F7FC}-0.005 & \cellcolor[HTML]{D2DFF0}-0.033 & \cellcolor[HTML]{FCDADD}0.043 & \multicolumn{1}{c|}{\cellcolor[HTML]{F97375}0.172} & \multicolumn{1}{c|}{13.0} &  & \cellcolor[HTML]{F9FAFE}-0.006 & \cellcolor[HTML]{EFF2FA}-0.033 & \cellcolor[HTML]{FCDFE2}0.112 & \multicolumn{1}{c|}{\cellcolor[HTML]{F97375}0.521} & \multicolumn{1}{c|}{12.0} &  & \cellcolor[HTML]{F7F8FD}-0.011 & \cellcolor[HTML]{ECF0F9}-0.038 & \cellcolor[HTML]{FBC3C6}0.033 & \multicolumn{1}{c|}{\cellcolor[HTML]{B1C7E4}-0.182} & \multicolumn{1}{c|}{15.5} &  & 9.2 \\
\multicolumn{1}{l|}{ORUSB1} & \cellcolor[HTML]{F2F4FB}-0.008 & \cellcolor[HTML]{DFE7F4}-0.023 & \cellcolor[HTML]{FCDEE0}0.039 & \multicolumn{1}{c|}{\cellcolor[HTML]{FBC4C7}0.071} & \multicolumn{1}{c|}{14.3} &  & \cellcolor[HTML]{F5F7FC}-0.015 & \cellcolor[HTML]{F2F4FB}-0.025 & \cellcolor[HTML]{FCE3E6}0.095 & \multicolumn{1}{c|}{\cellcolor[HTML]{FBB8BB}0.258} & \multicolumn{1}{c|}{15.1} &  & \cellcolor[HTML]{FBD7DA}0.022 & \cellcolor[HTML]{F6F8FD}-0.013 & \cellcolor[HTML]{FCF5F8}0.004 & \multicolumn{1}{c|}{\cellcolor[HTML]{5D8CC7}-0.388} & \multicolumn{1}{c|}{14.7} &  & 9.9 \\
\multicolumn{1}{l|}{ORUSB2} & \cellcolor[HTML]{FAFAFE}-0.001 & \cellcolor[HTML]{DCE5F3}-0.026 & \cellcolor[HTML]{FCE3E6}0.031 & \multicolumn{1}{c|}{\cellcolor[HTML]{FAA5A7}0.110} & \multicolumn{1}{c|}{14.8} &  & \cellcolor[HTML]{F7F9FD}-0.011 & \cellcolor[HTML]{F0F3FA}-0.029 & \cellcolor[HTML]{FCE3E6}0.097 & \multicolumn{1}{c|}{\cellcolor[HTML]{FA9193}0.408} & \multicolumn{1}{c|}{14.8} &  & \cellcolor[HTML]{FBCCCE}0.028 & \cellcolor[HTML]{F5F7FC}-0.015 & \cellcolor[HTML]{FCF2F5}0.006 & \multicolumn{1}{c|}{\cellcolor[HTML]{ABC3E2}-0.199} & \multicolumn{1}{c|}{14.1} &  & 9.6 \\
\multicolumn{1}{l|}{ORUSB3} & \cellcolor[HTML]{E5ECF7}-0.018 & \cellcolor[HTML]{DCE5F3}-0.026 & \cellcolor[HTML]{FCDBDD}0.043 & \multicolumn{1}{c|}{\cellcolor[HTML]{FAA4A6}0.111} & \multicolumn{1}{c|}{13.7} &  & \cellcolor[HTML]{F5F7FC}-0.017 & \cellcolor[HTML]{F2F5FB}-0.025 & \cellcolor[HTML]{FCE0E3}0.108 & \multicolumn{1}{c|}{\cellcolor[HTML]{FAA2A5}0.340} & \multicolumn{1}{c|}{13.5} &  & \cellcolor[HTML]{F5F7FC}-0.017 & \cellcolor[HTML]{F0F3FA}-0.028 & \cellcolor[HTML]{FCDCDF}0.019 & \multicolumn{1}{c|}{\cellcolor[HTML]{6A95CB}-0.358} & \multicolumn{1}{c|}{16.4} &  & 10.0 \\
\multicolumn{1}{l|}{OOB} & \cellcolor[HTML]{F1F4FB}-0.009 & \cellcolor[HTML]{E5ECF7}-0.018 & \cellcolor[HTML]{FBCED1}0.058 & \multicolumn{1}{c|}{\cellcolor[HTML]{FA9193}0.135} & \multicolumn{1}{c|}{10.9} &  & \cellcolor[HTML]{F5F7FC}-0.017 & \cellcolor[HTML]{F1F4FB}-0.027 & \cellcolor[HTML]{FCDCDF}0.122 & \multicolumn{1}{c|}{\cellcolor[HTML]{F9888A}0.440} & \multicolumn{1}{c|}{11.2} &  & \cellcolor[HTML]{F6F8FD}-0.013 & \cellcolor[HTML]{FBFBFE}-0.001 & \cellcolor[HTML]{F9787A}0.077 & \multicolumn{1}{c|}{\cellcolor[HTML]{FAACAE}0.047} & \multicolumn{1}{c|}{\textbf{10.0}} &  & 7.1 \\
\multicolumn{1}{l|}{OUB} & \cellcolor[HTML]{E3EBF6}-0.019 & \cellcolor[HTML]{DDE6F4}-0.025 & \cellcolor[HTML]{FCE0E3}0.036 & \multicolumn{1}{c|}{\cellcolor[HTML]{F97274}0.174} & \multicolumn{1}{c|}{13.7} &  & \cellcolor[HTML]{F0F4FB}-0.028 & \cellcolor[HTML]{EEF2FA}-0.034 & \cellcolor[HTML]{FCE1E4}0.103 & \multicolumn{1}{c|}{\cellcolor[HTML]{F97072}0.531} & \multicolumn{1}{c|}{13.6} &  & \cellcolor[HTML]{EEF2FA}-0.033 & \cellcolor[HTML]{F8F9FD}-0.008 & \cellcolor[HTML]{FCD8DB}0.021 & \multicolumn{1}{c|}{\cellcolor[HTML]{C8D8ED}-0.125} & \multicolumn{1}{c|}{14.3} &  & 9.3 \\
\multicolumn{1}{l|}{OWOB} & \cellcolor[HTML]{FAFAFE}-0.001 & \cellcolor[HTML]{E9EFF8}-0.015 & \cellcolor[HTML]{FBD6D9}0.048 & \multicolumn{1}{c|}{\cellcolor[HTML]{FA9294}0.134} & \multicolumn{1}{c|}{11.3} &  & \cellcolor[HTML]{FAFAFE}-0.004 & \cellcolor[HTML]{F6F7FC}-0.014 & \cellcolor[HTML]{FCDDE0}0.119 & \multicolumn{1}{c|}{\cellcolor[HTML]{F9898C}0.435} & \multicolumn{1}{c|}{10.6} &  & \cellcolor[HTML]{F9FAFE}-0.007 & \cellcolor[HTML]{F5F7FC}-0.016 & \cellcolor[HTML]{F98385}0.070 & \multicolumn{1}{c|}{\cellcolor[HTML]{F8696B}0.085} & \multicolumn{1}{c|}{10.3} &  & 7.0 \\
\multicolumn{1}{l|}{OWUB} & \cellcolor[HTML]{F0F4FB}-0.009 & \cellcolor[HTML]{D8E3F2}-0.028 & \cellcolor[HTML]{FCDCDF}0.041 & \multicolumn{1}{c|}{\cellcolor[HTML]{F97375}0.173} & \multicolumn{1}{c|}{13.6} &  & \cellcolor[HTML]{F8F9FD}-0.009 & \cellcolor[HTML]{F1F4FB}-0.027 & \cellcolor[HTML]{FCE0E3}0.109 & \multicolumn{1}{c|}{\cellcolor[HTML]{F96F71}0.536} & \multicolumn{1}{c|}{12.7} &  & \cellcolor[HTML]{F7F8FD}-0.011 & \cellcolor[HTML]{EFF3FA}-0.030 & \cellcolor[HTML]{FBD6D9}0.022 & \multicolumn{1}{c|}{\cellcolor[HTML]{C2D3EA}-0.142} & \multicolumn{1}{c|}{15.5} &  & 9.4 \\
\multicolumn{1}{l|}{OEB} & \cellcolor[HTML]{F4F6FC}-0.006 & \cellcolor[HTML]{D9E3F2}-0.028 & \cellcolor[HTML]{FCE3E6}0.032 & \multicolumn{1}{c|}{\cellcolor[HTML]{FA9193}0.135} & \multicolumn{1}{c|}{14.8} &  & \cellcolor[HTML]{F9FAFE}-0.006 & \cellcolor[HTML]{EFF3FA}-0.031 & \cellcolor[HTML]{FCE2E5}0.100 & \multicolumn{1}{c|}{\cellcolor[HTML]{F98183}0.467} & \multicolumn{1}{c|}{13.9} &  & \cellcolor[HTML]{F6F7FC}-0.015 & \cellcolor[HTML]{EBF0F9}-0.041 & \cellcolor[HTML]{FCE2E5}0.015 & \multicolumn{1}{c|}{\cellcolor[HTML]{A1BCDF}-0.221} & \multicolumn{1}{c|}{16.7} &  & 10.2 \\
\multicolumn{20}{l}{\cellcolor[HTML]{C0C0C0}\textbf{The Proposed Approach:}} \\
\multicolumn{1}{l|}{HGD} & \cellcolor[HTML]{FCFBFE}0.001 & \cellcolor[HTML]{FCFBFE}0.002 & \cellcolor[HTML]{FBD0D3}0.055 & \multicolumn{1}{c|}{\cellcolor[HTML]{F8696B}0.185} & \multicolumn{1}{c|}{\textbf{7.5}} &  & \cellcolor[HTML]{FCFCFF}0.001 & \cellcolor[HTML]{FCFBFE}0.004 & \cellcolor[HTML]{FCDBDE}0.127 & \multicolumn{1}{c|}{\cellcolor[HTML]{F96E70}0.538} & \multicolumn{1}{c|}{\textbf{6.7}} &  & \cellcolor[HTML]{FCFAFD}0.002 & \cellcolor[HTML]{FCEAED}0.010 & \cellcolor[HTML]{FA9B9D}0.056 & \multicolumn{1}{c|}{\cellcolor[HTML]{BACEE8}-0.160} & \multicolumn{1}{c|}{\textbf{9.8}} &  & \textbf{5.5} \\ \hline
\end{tabular}
}
\begin{tablenotes}
    \item{$\cdot$} The results are shown as increments or decrements w.r.t. the BaseLine (OGD). The bold entries stand for the top three highest rankings.
\end{tablenotes}
\label{S:tab:cmp_static_p}
\end{table*}
%===============================================

%===============================================
\begin{table*}[h]
\scriptsize
\centering
\renewcommand\arraystretch{1.2}
\caption{The Averaged Performance (AUC, G-means and F1-Score) Attained by Different Methods Over All Datasets w.r.t.
Their Imbalance Ratios (IR, i.e., Equal IR, Low IR, Medium IR and High IR). Linear SVM as the Base Learner.}
\resizebox{\linewidth}{!}{
\begin{tabular}{lccccccccccccccccccc}
\hline
\multicolumn{1}{l|}{\textbf{Metric}} & \multicolumn{4}{c|}{\textbf{AUC}} & \multicolumn{1}{c|}{} & \textbf{} & \multicolumn{4}{c|}{\textbf{GMEANS}} & \multicolumn{1}{c|}{} & \textbf{} & \multicolumn{4}{c|}{\textbf{F1}} & \multicolumn{1}{c|}{} & \textbf{} &  \\ \cline{1-5} \cline{8-11} \cline{14-17}
\multicolumn{1}{l|}{\textbf{IR}} & \textbf{Equal} & \textbf{Low} & \textbf{Medium} & \multicolumn{1}{c|}{\textbf{High}} & \multicolumn{1}{c|}{\multirow{-2}{*}{\textbf{Rank}}} & \textbf{} & \textbf{Equal} & \textbf{Low} & \textbf{Medium} & \multicolumn{1}{c|}{\textbf{High}} & \multicolumn{1}{c|}{\multirow{-2}{*}{\textbf{Rank}}} & \textbf{} & \textbf{Equal} & \textbf{Low} & \textbf{Medium} & \multicolumn{1}{c|}{\textbf{High}} & \multicolumn{1}{c|}{\multirow{-2}{*}{\textbf{Rank}}} & \textbf{} & \multirow{-2}{*}{\textbf{\begin{tabular}[c]{@{}c@{}}Overall\\ Rank\end{tabular}}} \\ \cline{1-6} \cline{8-12} \cline{14-18} \cline{20-20} 
\multicolumn{20}{l}{\cellcolor[HTML]{C0C0C0}\textbf{Baseline:}} \\
\multicolumn{1}{l|}{OGD} & \cellcolor[HTML]{FCFCFF}0.000 & \cellcolor[HTML]{FCFCFF}0.000 & \cellcolor[HTML]{FCFCFF}0.000 & \multicolumn{1}{c|}{\cellcolor[HTML]{FCFCFF}0.000} & \multicolumn{1}{c|}{17.8} &  & \cellcolor[HTML]{FCFCFF}0.000 & \cellcolor[HTML]{FCFCFF}0.000 & \cellcolor[HTML]{FCFCFF}0.000 & \multicolumn{1}{c|}{\cellcolor[HTML]{FCFCFF}0.000} & \multicolumn{1}{c|}{17.5} &  & \cellcolor[HTML]{FCFCFF}0.000 & \cellcolor[HTML]{FCFCFF}0.000 & \cellcolor[HTML]{FCFCFF}0.000 & \multicolumn{1}{c|}{\cellcolor[HTML]{FCFCFF}0.000} & \multicolumn{1}{c|}{16.1} &  & 17.1 \\
\multicolumn{20}{l}{\cellcolor[HTML]{C0C0C0}\textbf{Data Level Approaches}} \\
\multicolumn{1}{l|}{OSMOTE} & \cellcolor[HTML]{FCFCFF}0.000 & \cellcolor[HTML]{FCFCFF}0.001 & \cellcolor[HTML]{FCE2E5}0.057 & \multicolumn{1}{c|}{\cellcolor[HTML]{FCE2E5}0.057} & \multicolumn{1}{c|}{16.4} &  & \cellcolor[HTML]{FCFCFF}0.002 & \cellcolor[HTML]{FCFCFF}0.003 & \cellcolor[HTML]{FCEBED}0.358 & \multicolumn{1}{c|}{\cellcolor[HTML]{FBC0C2}1.217} & \multicolumn{1}{c|}{16.1} &  & \cellcolor[HTML]{FCFCFF}0.001 & \cellcolor[HTML]{FCFCFF}0.003 & \cellcolor[HTML]{FCE3E6}0.398 & \multicolumn{1}{c|}{\cellcolor[HTML]{FAACAE}1.253} & \multicolumn{1}{c|}{14.4} &  & 15.6 \\
\multicolumn{1}{l|}{OUR} & \cellcolor[HTML]{EDF1F9}-0.021 & \cellcolor[HTML]{F8F9FD}-0.005 & \cellcolor[HTML]{FBBDC0}0.138 & \multicolumn{1}{c|}{\cellcolor[HTML]{FBBEC0}0.136} & \multicolumn{1}{c|}{15.4} &  & \cellcolor[HTML]{F6F8FD}-0.019 & \cellcolor[HTML]{FCFBFE}0.032 & \cellcolor[HTML]{FCDDE0}0.628 & \multicolumn{1}{c|}{\cellcolor[HTML]{FA9193}2.150} & \multicolumn{1}{c|}{13.8} &  & \cellcolor[HTML]{ECF0F9}-0.022 & \cellcolor[HTML]{FCFAFD}0.045 & \cellcolor[HTML]{FBD6D9}0.593 & \multicolumn{1}{c|}{\cellcolor[HTML]{FBBFC2}0.952} & \multicolumn{1}{c|}{16.1} &  & 15.1 \\
\multicolumn{1}{l|}{OOR} & \cellcolor[HTML]{EEF2FA}-0.019 & \cellcolor[HTML]{FCFCFF}0.001 & \cellcolor[HTML]{FBB4B7}0.157 & \multicolumn{1}{c|}{\cellcolor[HTML]{F9888A}0.254} & \multicolumn{1}{c|}{12.6} &  & \cellcolor[HTML]{F6F8FD}-0.019 & \cellcolor[HTML]{FCFAFD}0.048 & \cellcolor[HTML]{FCDCDE}0.657 & \multicolumn{1}{c|}{\cellcolor[HTML]{F97173}2.795} & \multicolumn{1}{c|}{11.3} &  & \cellcolor[HTML]{EDF1F9}-0.021 & \cellcolor[HTML]{FCF8FB}0.066 & \cellcolor[HTML]{FBD3D6}0.643 & \multicolumn{1}{c|}{\cellcolor[HTML]{FAA7A9}1.330} & \multicolumn{1}{c|}{14.2} &  & 12.7 \\
\multicolumn{1}{l|}{OHR} & \cellcolor[HTML]{D4E0F1}-0.057 & \cellcolor[HTML]{F5F7FC}-0.010 & \cellcolor[HTML]{FAAEB1}0.170 & \multicolumn{1}{c|}{\cellcolor[HTML]{FAA9AB}0.182} & \multicolumn{1}{c|}{13.7} &  & \cellcolor[HTML]{EAEFF8}-0.061 & \cellcolor[HTML]{FBFBFE}0.000 & \cellcolor[HTML]{FCDCDE}0.657 & \multicolumn{1}{c|}{\cellcolor[HTML]{F98082}2.494} & \multicolumn{1}{c|}{13.9} &  & \cellcolor[HTML]{AFC5E3}-0.105 & \cellcolor[HTML]{FCF6F9}0.095 & \cellcolor[HTML]{FBD1D3}0.678 & \multicolumn{1}{c|}{\cellcolor[HTML]{F97A7C}2.022} & \multicolumn{1}{c|}{12.6} &  & 13.4 \\
\multicolumn{20}{l}{\cellcolor[HTML]{C0C0C0}\textbf{Cost-Sensitive Approaches}} \\
\multicolumn{1}{l|}{CSRDA$_{I}$} & \cellcolor[HTML]{B6CAE6}-0.101 & \cellcolor[HTML]{E6EDF7}-0.030 & \cellcolor[HTML]{FCDEE1}0.065 & \multicolumn{1}{c|}{\cellcolor[HTML]{FBD7D9}0.082} & \multicolumn{1}{c|}{21.4} &  & \cellcolor[HTML]{E0E8F5}-0.097 & \cellcolor[HTML]{FCFCFF}0.015 & \cellcolor[HTML]{FCE7EA}0.433 & \multicolumn{1}{c|}{\cellcolor[HTML]{FAADAF}1.593} & \multicolumn{1}{c|}{20.5} &  & \cellcolor[HTML]{ABC3E2}-0.111 & \cellcolor[HTML]{FCFBFE}0.019 & \cellcolor[HTML]{FCDEE1}0.467 & \multicolumn{1}{c|}{\cellcolor[HTML]{F9888A}1.806} & \multicolumn{1}{c|}{18.9} &  & 20.3 \\
\multicolumn{1}{l|}{CSRDA$_{II}$} & \cellcolor[HTML]{B7CCE7}-0.098 & \cellcolor[HTML]{EBF0F9}-0.024 & \cellcolor[HTML]{FBC8CB}0.113 & \multicolumn{1}{c|}{\cellcolor[HTML]{F97A7C}0.283} & \multicolumn{1}{c|}{17.3} &  & \cellcolor[HTML]{E1E9F5}-0.093 & \cellcolor[HTML]{FCFBFE}0.023 & \cellcolor[HTML]{FCDFE2}0.595 & \multicolumn{1}{c|}{\cellcolor[HTML]{F96C6E}2.879} & \multicolumn{1}{c|}{15.9} &  & \cellcolor[HTML]{ADC4E3}-0.108 & \cellcolor[HTML]{FCFAFD}0.039 & \cellcolor[HTML]{FCDBDD}0.525 & \multicolumn{1}{c|}{\cellcolor[HTML]{FAA2A4}1.402} & \multicolumn{1}{c|}{19.4} &  & 17.6 \\
\multicolumn{1}{l|}{CSRDA$_{III}$} & \cellcolor[HTML]{FCFAFD}0.006 & \cellcolor[HTML]{FCFAFD}0.005 & \cellcolor[HTML]{FAACAE}0.175 & \multicolumn{1}{c|}{\cellcolor[HTML]{F97678}0.292} & \multicolumn{1}{c|}{\textbf{8.1}} &  & \cellcolor[HTML]{FCFCFF}0.007 & \cellcolor[HTML]{FCFAFD}0.051 & \cellcolor[HTML]{FCDADD}0.684 & \multicolumn{1}{c|}{\cellcolor[HTML]{F96A6C}2.927} & \multicolumn{1}{c|}{\textbf{6.7}} &  & \cellcolor[HTML]{FCFCFF}0.006 & \cellcolor[HTML]{FCF8FB}0.067 & \cellcolor[HTML]{FBD0D3}0.686 & \multicolumn{1}{c|}{\cellcolor[HTML]{FA979A}1.566} & \multicolumn{1}{c|}{\textbf{9.8}} &  & \textbf{8.2} \\
\multicolumn{1}{l|}{CSRDA$_{IV}$} & \cellcolor[HTML]{E4EBF6}-0.034 & \cellcolor[HTML]{FCF8FB}0.010 & \cellcolor[HTML]{FBCDCF}0.104 & \multicolumn{1}{c|}{\cellcolor[HTML]{F8F9FD}-0.006} & \multicolumn{1}{c|}{13.2} &  & \cellcolor[HTML]{F0F3FA}-0.041 & \cellcolor[HTML]{FCFAFD}0.059 & \cellcolor[HTML]{FCE2E5}0.523 & \multicolumn{1}{c|}{\cellcolor[HTML]{FBB9BB}1.357} & \multicolumn{1}{c|}{11.5} &  & \cellcolor[HTML]{F0F4FB}-0.015 & \cellcolor[HTML]{FCF7FA}0.081 & \cellcolor[HTML]{FCDBDE}0.515 & \multicolumn{1}{c|}{\cellcolor[HTML]{FBD5D8}0.604} & \multicolumn{1}{c|}{13.3} &  & 12.7 \\
\multicolumn{1}{l|}{CSOGD$_{C_I}$} & \cellcolor[HTML]{B9CCE7}-0.096 & \cellcolor[HTML]{D5E0F1}-0.056 & \cellcolor[HTML]{FCF7FA}0.011 & \multicolumn{1}{c|}{\cellcolor[HTML]{E0E8F5}-0.039} & \multicolumn{1}{c|}{21.8} &  & \cellcolor[HTML]{BCCFE8}-0.227 & \cellcolor[HTML]{DFE7F4}-0.103 & \cellcolor[HTML]{FCFCFF}0.018 & \multicolumn{1}{c|}{\cellcolor[HTML]{FAFBFE}-0.004} & \multicolumn{1}{c|}{22.2} &  & \cellcolor[HTML]{FBFBFE}0.000 & \cellcolor[HTML]{E9EEF8}-0.025 & \cellcolor[HTML]{F3F6FC}-0.011 & \multicolumn{1}{c|}{\cellcolor[HTML]{5A8AC6}-0.222} & \multicolumn{1}{c|}{19.0} &  & 21.0 \\
\multicolumn{1}{l|}{CSOGD$_{C_{II}}$} & \cellcolor[HTML]{5A8AC6}-0.233 & \cellcolor[HTML]{BCCFE8}-0.091 & \cellcolor[HTML]{FBD2D5}0.093 & \multicolumn{1}{c|}{\cellcolor[HTML]{F9898B}0.252} & \multicolumn{1}{c|}{19.2} &  & \cellcolor[HTML]{5A8AC6}-0.577 & \cellcolor[HTML]{ADC4E3}-0.280 & \cellcolor[HTML]{FCEBEE}0.355 & \multicolumn{1}{c|}{\cellcolor[HTML]{F97577}2.716} & \multicolumn{1}{c|}{21.2} &  & \cellcolor[HTML]{D6E1F1}-0.051 & \cellcolor[HTML]{FCFBFD}0.031 & \cellcolor[HTML]{FCDEE1}0.469 & \multicolumn{1}{c|}{\cellcolor[HTML]{FA999C}1.536} & \multicolumn{1}{c|}{19.1} &  & 19.8 \\
\multicolumn{1}{l|}{CSOGD$_{S_{I}}$} & \cellcolor[HTML]{FBFBFE}0.000 & \cellcolor[HTML]{FCFCFF}0.002 & \cellcolor[HTML]{FCF6F9}0.014 & \multicolumn{1}{c|}{\cellcolor[HTML]{E5ECF7}-0.032} & \multicolumn{1}{c|}{16.9} &  & \cellcolor[HTML]{FBFBFE}0.000 & \cellcolor[HTML]{FCFCFF}0.008 & \cellcolor[HTML]{FCFBFE}0.021 & \multicolumn{1}{c|}{\cellcolor[HTML]{F5F7FC}-0.022} & \multicolumn{1}{c|}{16.9} &  & \cellcolor[HTML]{FAFBFE}-0.002 & \cellcolor[HTML]{FCFCFF}0.010 & \cellcolor[HTML]{FCFBFE}0.018 & \multicolumn{1}{c|}{\cellcolor[HTML]{6F98CD}-0.193} & \multicolumn{1}{c|}{15.9} &  & 16.6 \\
\multicolumn{1}{l|}{CSOGD$_{S_{II}}$} & \cellcolor[HTML]{FAFAFE}-0.002 & \cellcolor[HTML]{FCF7FA}0.012 & \cellcolor[HTML]{FAB0B2}0.167 & \multicolumn{1}{c|}{\cellcolor[HTML]{F98082}0.271} & \multicolumn{1}{c|}{\textbf{8.6}} &  & \cellcolor[HTML]{FCFCFF}0.001 & \cellcolor[HTML]{FCF9FC}0.063 & \cellcolor[HTML]{FCDBDE}0.670 & \multicolumn{1}{c|}{\cellcolor[HTML]{F97072}2.815} & \multicolumn{1}{c|}{\textbf{7.1}} &  & \cellcolor[HTML]{FAFAFE}-0.003 & \cellcolor[HTML]{FCF7FA}0.082 & \cellcolor[HTML]{FBD2D5}0.651 & \multicolumn{1}{c|}{\cellcolor[HTML]{FAA6A9}1.336} & \multicolumn{1}{c|}{10.9} &  & \textbf{8.9} \\
\multicolumn{20}{l}{\cellcolor[HTML]{C0C0C0}\textbf{Ensemble Learning Approaches}} \\
\multicolumn{1}{l|}{OB} & \cellcolor[HTML]{FBFBFE}-0.001 & \cellcolor[HTML]{F4F6FC}-0.011 & \cellcolor[HTML]{EEF2FA}-0.019 & \multicolumn{1}{c|}{\cellcolor[HTML]{E9EFF8}-0.026} & \multicolumn{1}{c|}{19.6} &  & \cellcolor[HTML]{FCFCFF}0.001 & \cellcolor[HTML]{FAFAFE}-0.006 & \cellcolor[HTML]{EDF1F9}-0.051 & \multicolumn{1}{c|}{\cellcolor[HTML]{F0F4FB}-0.040} & \multicolumn{1}{c|}{19.0} &  & \cellcolor[HTML]{F4F7FC}-0.010 & \cellcolor[HTML]{F5F7FC}-0.008 & \cellcolor[HTML]{E1E9F5}-0.036 & \multicolumn{1}{c|}{\cellcolor[HTML]{5E8DC7}-0.216} & \multicolumn{1}{c|}{18.8} &  & 19.1 \\
\multicolumn{1}{l|}{OAdaB} & \cellcolor[HTML]{EBF0F9}-0.023 & \cellcolor[HTML]{F1F4FB}-0.015 & \cellcolor[HTML]{FCE1E4}0.060 & \multicolumn{1}{c|}{\cellcolor[HTML]{FCE4E7}0.054} & \multicolumn{1}{c|}{20.0} &  & \cellcolor[HTML]{F7F8FD}-0.017 & \cellcolor[HTML]{FCFCFF}0.002 & \cellcolor[HTML]{FCEBEE}0.349 & \multicolumn{1}{c|}{\cellcolor[HTML]{FBB4B6}1.457} & \multicolumn{1}{c|}{19.4} &  & \cellcolor[HTML]{E3EAF6}-0.034 & \cellcolor[HTML]{F7F8FD}-0.006 & \cellcolor[HTML]{FCE3E6}0.400 & \multicolumn{1}{c|}{\cellcolor[HTML]{FAA2A4}1.402} & \multicolumn{1}{c|}{18.1} &  & 19.2 \\
\multicolumn{1}{l|}{OAdaC2} & \cellcolor[HTML]{89ABD6}-0.165 & \cellcolor[HTML]{DDE6F4}-0.044 & \cellcolor[HTML]{FBBABD}0.145 & \multicolumn{1}{c|}{\cellcolor[HTML]{FAA3A6}0.194} & \multicolumn{1}{c|}{18.4} &  & \cellcolor[HTML]{9EBADE}-0.334 & \cellcolor[HTML]{D6E1F1}-0.135 & \cellcolor[HTML]{FCE1E4}0.546 & \multicolumn{1}{c|}{\cellcolor[HTML]{F97F81}2.507} & \multicolumn{1}{c|}{19.4} &  & \cellcolor[HTML]{E8EEF8}-0.027 & \cellcolor[HTML]{FCF8FB}0.071 & \cellcolor[HTML]{FBD5D8}0.609 & \multicolumn{1}{c|}{\cellcolor[HTML]{F98587}1.848} & \multicolumn{1}{c|}{15.4} &  & 17.8 \\
\multicolumn{1}{l|}{OCSB2} & \cellcolor[HTML]{B1C7E4}-0.108 & \cellcolor[HTML]{E6ECF7}-0.031 & \cellcolor[HTML]{FBBBBD}0.143 & \multicolumn{1}{c|}{\cellcolor[HTML]{FAAEB0}0.171} & \multicolumn{1}{c|}{18.0} &  & \cellcolor[HTML]{BED0E9}-0.219 & \cellcolor[HTML]{D0DDEF}-0.155 & \cellcolor[HTML]{FCE0E2}0.576 & \multicolumn{1}{c|}{\cellcolor[HTML]{F98789}2.352} & \multicolumn{1}{c|}{18.8} &  & \cellcolor[HTML]{F9FAFE}-0.003 & \cellcolor[HTML]{FCF7FA}0.086 & \cellcolor[HTML]{FBD4D7}0.622 & \multicolumn{1}{c|}{\cellcolor[HTML]{FA9A9C}1.532} & \multicolumn{1}{c|}{14.5} &  & 17.1 \\
\multicolumn{1}{l|}{OKB} & \cellcolor[HTML]{D8E3F2}-0.051 & \cellcolor[HTML]{F3F5FB}-0.012 & \cellcolor[HTML]{FCE4E6}0.054 & \multicolumn{1}{c|}{\cellcolor[HTML]{F5F7FC}-0.009} & \multicolumn{1}{c|}{20.4} &  & \cellcolor[HTML]{EEF2FA}-0.047 & \cellcolor[HTML]{FCFCFF}0.017 & \cellcolor[HTML]{FCECEF}0.335 & \multicolumn{1}{c|}{\cellcolor[HTML]{FBD2D4}0.854} & \multicolumn{1}{c|}{19.8} &  & \cellcolor[HTML]{C4D4EB}-0.076 & \cellcolor[HTML]{FCFCFF}0.014 & \cellcolor[HTML]{FCE2E5}0.404 & \multicolumn{1}{c|}{\cellcolor[HTML]{FBD3D6}0.640} & \multicolumn{1}{c|}{19.2} &  & 19.8 \\
\multicolumn{1}{l|}{OUOB} & \cellcolor[HTML]{FBFBFE}0.000 & \cellcolor[HTML]{FCF5F7}0.017 & \cellcolor[HTML]{FAAFB1}0.169 & \multicolumn{1}{c|}{\cellcolor[HTML]{F97779}0.290} & \multicolumn{1}{c|}{\textbf{7.1}} &  & \cellcolor[HTML]{FCFCFF}0.002 & \cellcolor[HTML]{FCFAFD}0.057 & \cellcolor[HTML]{FCDBDD}0.676 & \multicolumn{1}{c|}{\cellcolor[HTML]{F96D6F}2.875} & \multicolumn{1}{c|}{\textbf{6.7}} &  & \cellcolor[HTML]{F4F6FC}-0.010 & \cellcolor[HTML]{FCF8FB}0.067 & \cellcolor[HTML]{FBD1D3}0.678 & \multicolumn{1}{c|}{\cellcolor[HTML]{FAA0A2}1.439} & \multicolumn{1}{c|}{10.7} &  & \textbf{8.1} \\
\multicolumn{1}{l|}{ORUSB1} & \cellcolor[HTML]{EDF1F9}-0.020 & \cellcolor[HTML]{FCFBFE}0.003 & \cellcolor[HTML]{FBD7DA}0.081 & \multicolumn{1}{c|}{\cellcolor[HTML]{EAF0F9}-0.024} & \multicolumn{1}{c|}{17.4} &  & \cellcolor[HTML]{F3F6FC}-0.030 & \cellcolor[HTML]{FBFBFE}-0.003 & \cellcolor[HTML]{FCF8FB}0.080 & \multicolumn{1}{c|}{\cellcolor[HTML]{B4C9E5}-0.254} & \multicolumn{1}{c|}{20.6} &  & \cellcolor[HTML]{FCFCFF}0.001 & \cellcolor[HTML]{FCF7F9}0.093 & \cellcolor[HTML]{FCE1E4}0.422 & \multicolumn{1}{c|}{\cellcolor[HTML]{FCEBEE}0.268} & \multicolumn{1}{c|}{17.6} &  & 18.5 \\
\multicolumn{1}{l|}{ORUSB2} & \cellcolor[HTML]{EBF0F9}-0.024 & \cellcolor[HTML]{FCFCFF}0.002 & \cellcolor[HTML]{FAADB0}0.172 & \multicolumn{1}{c|}{\cellcolor[HTML]{FA9FA1}0.203} & \multicolumn{1}{c|}{12.4} &  & \cellcolor[HTML]{EEF2FA}-0.046 & \cellcolor[HTML]{FCFCFF}0.011 & \cellcolor[HTML]{FCDDE0}0.633 & \multicolumn{1}{c|}{\cellcolor[HTML]{F98688}2.369} & \multicolumn{1}{c|}{15.1} &  & \cellcolor[HTML]{FCFBFE}0.016 & \cellcolor[HTML]{FCF7FA}0.090 & \cellcolor[HTML]{FBD1D4}0.672 & \multicolumn{1}{c|}{\cellcolor[HTML]{FA9D9F}1.477} & \multicolumn{1}{c|}{12.4} &  & 13.3 \\
\multicolumn{1}{l|}{ORUSB3} & \cellcolor[HTML]{EBF0F9}-0.024 & \cellcolor[HTML]{FBFBFE}-0.001 & \cellcolor[HTML]{FAB2B5}0.161 & \multicolumn{1}{c|}{\cellcolor[HTML]{FBCACC}0.111} & \multicolumn{1}{c|}{13.6} &  & \cellcolor[HTML]{F6F8FD}-0.020 & \cellcolor[HTML]{FCFAFD}0.043 & \cellcolor[HTML]{FCE3E6}0.515 & \multicolumn{1}{c|}{\cellcolor[HTML]{FCDEE1}0.600} & \multicolumn{1}{c|}{16.0} &  & \cellcolor[HTML]{E1E9F5}-0.036 & \cellcolor[HTML]{FCF9FC}0.061 & \cellcolor[HTML]{FBD7DA}0.578 & \multicolumn{1}{c|}{\cellcolor[HTML]{FCD9DB}0.556} & \multicolumn{1}{c|}{17.5} &  & 15.7 \\
\multicolumn{1}{l|}{OOB} & \cellcolor[HTML]{D9E3F2}-0.050 & \cellcolor[HTML]{FCF9FC}0.008 & \cellcolor[HTML]{FAA7A9}0.186 & \multicolumn{1}{c|}{\cellcolor[HTML]{FAA6A8}0.188} & \multicolumn{1}{c|}{9.8} &  & \cellcolor[HTML]{F1F4FB}-0.039 & \cellcolor[HTML]{FBFBFE}-0.002 & \cellcolor[HTML]{FCDADD}0.690 & \multicolumn{1}{c|}{\cellcolor[HTML]{F98487}2.401} & \multicolumn{1}{c|}{10.3} &  & \cellcolor[HTML]{E2EAF6}-0.035 & \cellcolor[HTML]{FCF5F8}0.113 & \cellcolor[HTML]{FBCCCF}0.745 & \multicolumn{1}{c|}{\cellcolor[HTML]{F96B6D}2.257} & \multicolumn{1}{c|}{\textbf{7.6}} &  & 9.2 \\
\multicolumn{1}{l|}{OUB} & \cellcolor[HTML]{CFDCEF}-0.064 & \cellcolor[HTML]{F8F9FD}-0.005 & \cellcolor[HTML]{FAB2B4}0.162 & \multicolumn{1}{c|}{\cellcolor[HTML]{FAB0B3}0.166} & \multicolumn{1}{c|}{12.9} &  & \cellcolor[HTML]{E3EAF6}-0.088 & \cellcolor[HTML]{E2EAF6}-0.090 & \cellcolor[HTML]{FCDEE1}0.606 & \multicolumn{1}{c|}{\cellcolor[HTML]{F97B7D}2.581} & \multicolumn{1}{c|}{14.0} &  & \cellcolor[HTML]{AFC5E3}-0.105 & \cellcolor[HTML]{FCF5F8}0.109 & \cellcolor[HTML]{FBD2D5}0.659 & \multicolumn{1}{c|}{\cellcolor[HTML]{FAA0A2}1.437} & \multicolumn{1}{c|}{13.1} &  & 13.3 \\
\multicolumn{1}{l|}{OWOB} & \cellcolor[HTML]{FCFAFD}0.006 & \cellcolor[HTML]{FCF5F8}0.016 & \cellcolor[HTML]{FBB6B8}0.154 & \multicolumn{1}{c|}{\cellcolor[HTML]{FBB8BB}0.148} & \multicolumn{1}{c|}{9.7} &  & \cellcolor[HTML]{FCFCFF}0.006 & \cellcolor[HTML]{FCFAFD}0.057 & \cellcolor[HTML]{FCDCDF}0.648 & \multicolumn{1}{c|}{\cellcolor[HTML]{F98E90}2.207} & \multicolumn{1}{c|}{8.4} &  & \cellcolor[HTML]{F8F9FD}-0.005 & \cellcolor[HTML]{FCF8FB}0.065 & \cellcolor[HTML]{FBCED0}0.726 & \multicolumn{1}{c|}{\cellcolor[HTML]{F8696B}2.275} & \multicolumn{1}{c|}{\textbf{7.1}} &  & 8.4 \\
\multicolumn{1}{l|}{OWUB} & \cellcolor[HTML]{FCFAFD}0.005 & \cellcolor[HTML]{FCF5F8}0.017 & \cellcolor[HTML]{FAA1A3}0.200 & \multicolumn{1}{c|}{\cellcolor[HTML]{F96D6F}0.312} & \multicolumn{1}{c|}{\textbf{5.4}} &  & \cellcolor[HTML]{FCFCFF}0.009 & \cellcolor[HTML]{FCF9FC}0.062 & \cellcolor[HTML]{FCDADD}0.687 & \multicolumn{1}{c|}{\cellcolor[HTML]{F96F71}2.828} & \multicolumn{1}{c|}{\textbf{6.2}} &  & \cellcolor[HTML]{FCFCFF}0.001 & \cellcolor[HTML]{FCF8FA}0.077 & \cellcolor[HTML]{FBD1D4}0.674 & \multicolumn{1}{c|}{\cellcolor[HTML]{FAABAD}1.265} & \multicolumn{1}{c|}{\textbf{10.5}} &  & \textbf{7.4} \\
\multicolumn{1}{l|}{OEB} & \cellcolor[HTML]{F3F6FC}-0.012 & \cellcolor[HTML]{FCFAFD}0.006 & \cellcolor[HTML]{FBB6B8}0.154 & \multicolumn{1}{c|}{\cellcolor[HTML]{FAB2B4}0.163} & \multicolumn{1}{c|}{11.7} &  & \cellcolor[HTML]{F9FAFE}-0.007 & \cellcolor[HTML]{FCFAFD}0.049 & \cellcolor[HTML]{FCDDE0}0.635 & \multicolumn{1}{c|}{\cellcolor[HTML]{F98A8C}2.297} & \multicolumn{1}{c|}{11.1} &  & \cellcolor[HTML]{E5ECF7}-0.031 & \cellcolor[HTML]{FCF9FC}0.056 & \cellcolor[HTML]{FBD4D7}0.621 & \multicolumn{1}{c|}{\cellcolor[HTML]{FBBBBE}1.009} & \multicolumn{1}{c|}{14.2} &  & 12.3 \\
\multicolumn{20}{l}{\cellcolor[HTML]{C0C0C0}\textbf{The Proposed Approach:}} \\
\multicolumn{1}{l|}{HGD} & \cellcolor[HTML]{FCFCFF}0.001 & \cellcolor[HTML]{FCF6F9}0.015 & \cellcolor[HTML]{FAACAE}0.176 & \multicolumn{1}{c|}{\cellcolor[HTML]{F8696B}0.320} & \multicolumn{1}{c|}{\textbf{7.2}} &  & \cellcolor[HTML]{FBFBFE}0.000 & \cellcolor[HTML]{FCF9FC}0.062 & \cellcolor[HTML]{FCDBDD}0.677 & \multicolumn{1}{c|}{\cellcolor[HTML]{F8696B}2.937} & \multicolumn{1}{c|}{\textbf{6.5}} &  & \cellcolor[HTML]{FCFCFF}0.002 & \cellcolor[HTML]{FCF7FA}0.085 & \cellcolor[HTML]{FBD1D4}0.668 & \multicolumn{1}{c|}{\cellcolor[HTML]{FA9EA1}1.461} & \multicolumn{1}{c|}{\textbf{9.6}} &  & \textbf{7.8} \\ \hline
\end{tabular}
}
\begin{tablenotes}
    \item{$\cdot$} The results are shown as increments or decrements w.r.t. the BaseLine (OGD). The bold entries stand for the top three highest rankings.
\end{tablenotes}
\label{S:tab:cmp_static_v}
\end{table*}
%===============================================

%===============================================
\begin{table*}[h]
\scriptsize
\centering
\renewcommand\arraystretch{1.2}
\caption{The Averaged Performance (AUC, G-means and F1-Score) Attained by Different Methods Over All Datasets w.r.t.
Their Imbalance Ratios (IR, i.e., Equal IR, Low IR, Medium IR and High IR). Kernel Model as the Base Learner.}
\resizebox{\linewidth}{!}{
\begin{tabular}{lccccccccccccccccccc}
\hline
\multicolumn{1}{l|}{\textbf{Metric}} & \multicolumn{4}{c|}{\textbf{AUC}} & \multicolumn{1}{c|}{} & \textbf{} & \multicolumn{4}{c|}{\textbf{GMEANS}} & \multicolumn{1}{c|}{} & \textbf{} & \multicolumn{4}{c|}{\textbf{F1}} & \multicolumn{1}{c|}{} & \textbf{} &  \\ \cline{1-5} \cline{8-11} \cline{14-17}
\multicolumn{1}{l|}{\textbf{IR}} & \textbf{Equal} & \textbf{Low} & \textbf{Medium} & \multicolumn{1}{c|}{\textbf{High}} & \multicolumn{1}{c|}{\multirow{-2}{*}{\textbf{Rank}}} & \textbf{} & \textbf{Equal} & \textbf{Low} & \textbf{Medium} & \multicolumn{1}{c|}{\textbf{High}} & \multicolumn{1}{c|}{\multirow{-2}{*}{\textbf{Rank}}} & \textbf{} & \textbf{Equal} & \textbf{Low} & \textbf{Medium} & \multicolumn{1}{c|}{\textbf{High}} & \multicolumn{1}{c|}{\multirow{-2}{*}{\textbf{Rank}}} & \textbf{} & \multirow{-2}{*}{\textbf{\begin{tabular}[c]{@{}c@{}}Overall\\ Rank\end{tabular}}} \\ \cline{1-6} \cline{8-12} \cline{14-18} \cline{20-20} 
\multicolumn{20}{l}{\cellcolor[HTML]{C0C0C0}\textbf{Baseline}} \\
\multicolumn{1}{l|}{OGD} & \cellcolor[HTML]{FCFCFF}0.000 & \cellcolor[HTML]{FCFCFF}0.000 & \cellcolor[HTML]{FCFCFF}0.000 & \multicolumn{1}{c|}{\cellcolor[HTML]{FCFCFF}0.000} & \multicolumn{1}{c|}{17.0} &  & \cellcolor[HTML]{FCFCFF}0.000 & \cellcolor[HTML]{FCFCFF}0.000 & \cellcolor[HTML]{FCFCFF}0.000 & \multicolumn{1}{c|}{\cellcolor[HTML]{FCFCFF}0.000} & \multicolumn{1}{c|}{15.6} &  & \cellcolor[HTML]{FCFCFF}0.000 & \cellcolor[HTML]{FCFCFF}0.000 & \cellcolor[HTML]{FCFCFF}0.000 & \multicolumn{1}{c|}{\cellcolor[HTML]{FCFCFF}0.000} & \multicolumn{1}{c|}{12.4} &  & 15.0 \\
\multicolumn{20}{l}{\cellcolor[HTML]{C0C0C0}\textbf{Data-Level Approaches}}  \\
\multicolumn{1}{l|}{OSMOTE} & \cellcolor[HTML]{E7EDF7}-0.032 & \cellcolor[HTML]{E2EAF6}-0.039 & \cellcolor[HTML]{FBD7DA}0.101 & \multicolumn{1}{c|}{\cellcolor[HTML]{F98789}0.319} & \multicolumn{1}{c|}{13.9} &  & \cellcolor[HTML]{F1F4FB}-0.037 & \cellcolor[HTML]{E0E8F5}-0.102 & \cellcolor[HTML]{FCE8EB}0.236 & \multicolumn{1}{c|}{\cellcolor[HTML]{F97174}1.625} & \multicolumn{1}{c|}{12.8} &  & \cellcolor[HTML]{EFF2FA}-0.019 & \cellcolor[HTML]{BBCEE8}-0.097 & \cellcolor[HTML]{FCEAED}0.168 & \multicolumn{1}{c|}{\cellcolor[HTML]{FAB0B3}0.705} & \multicolumn{1}{c|}{13.9} &  & 13.5 \\
\multicolumn{1}{l|}{OUR} & \cellcolor[HTML]{E1E9F5}-0.041 & \cellcolor[HTML]{E0E8F5}-0.043 & \cellcolor[HTML]{FCDADC}0.095 & \multicolumn{1}{c|}{\cellcolor[HTML]{FA9698}0.278} & \multicolumn{1}{c|}{15.5} &  & \cellcolor[HTML]{EFF3FA}-0.046 & \cellcolor[HTML]{E5EBF6}-0.084 & \cellcolor[HTML]{FCE8EB}0.239 & \multicolumn{1}{c|}{\cellcolor[HTML]{F97D7F}1.488} & \multicolumn{1}{c|}{15.8} &  & \cellcolor[HTML]{EDF1F9}-0.023 & \cellcolor[HTML]{EEF2FA}-0.021 & \cellcolor[HTML]{FCEEF1}0.138 & \multicolumn{1}{c|}{\cellcolor[HTML]{FBCDD0}0.440} & \multicolumn{1}{c|}{16.5} &  & 15.9 \\
\multicolumn{1}{l|}{OOR} & \cellcolor[HTML]{F2F5FB}-0.015 & \cellcolor[HTML]{FBFBFE}0.000 & \cellcolor[HTML]{FBCCCE}0.133 & \multicolumn{1}{c|}{\cellcolor[HTML]{F97F81}0.341} & \multicolumn{1}{c|}{11.1} &  & \cellcolor[HTML]{F7F8FD}-0.016 & \cellcolor[HTML]{FCFCFF}0.010 & \cellcolor[HTML]{FCE2E5}0.312 & \multicolumn{1}{c|}{\cellcolor[HTML]{F96D6F}1.682} & \multicolumn{1}{c|}{9.9} &  & \cellcolor[HTML]{F1F4FB}-0.016 & \cellcolor[HTML]{FCFBFE}0.018 & \cellcolor[HTML]{FCE3E6}0.233 & \multicolumn{1}{c|}{\cellcolor[HTML]{FBBDC0}0.587} & \multicolumn{1}{c|}{11.7} &  & 10.9 \\
\multicolumn{1}{l|}{OHR} & \cellcolor[HTML]{EEF2FA}-0.021 & \cellcolor[HTML]{F7F9FD}-0.006 & \cellcolor[HTML]{FCDEE1}0.082 & \multicolumn{1}{c|}{\cellcolor[HTML]{FBB9BB}0.184} & \multicolumn{1}{c|}{15.7} &  & \cellcolor[HTML]{F7F9FD}-0.016 & \cellcolor[HTML]{F9FAFE}-0.008 & \cellcolor[HTML]{FCEFF1}0.163 & \multicolumn{1}{c|}{\cellcolor[HTML]{FA9B9E}1.136} & \multicolumn{1}{c|}{16.8} &  & \cellcolor[HTML]{EDF2FA}-0.021 & \cellcolor[HTML]{FCFBFE}0.010 & \cellcolor[HTML]{FCF9FC}0.028 & \multicolumn{1}{c|}{\cellcolor[HTML]{BACDE7}-0.100} & \multicolumn{1}{c|}{18.3} &  & 16.9 \\
\multicolumn{20}{l}{\cellcolor[HTML]{C0C0C0}\textbf{Cost-Sensitive Approaches}} \\
\multicolumn{1}{l|}{CSOGD$_{C_I}$} & \cellcolor[HTML]{DFE7F4}-0.044 & \cellcolor[HTML]{ECF1F9}-0.023 & \cellcolor[HTML]{FCF7FA}0.014 & \multicolumn{1}{c|}{\cellcolor[HTML]{FCF5F8}0.019} & \multicolumn{1}{c|}{18.1} &  & \cellcolor[HTML]{EAEFF8}-0.066 & \cellcolor[HTML]{F7F9FD}-0.016 & \cellcolor[HTML]{FCFBFE}0.018 & \multicolumn{1}{c|}{\cellcolor[HTML]{FCFBFE}0.023} & \multicolumn{1}{c|}{17.2} &  & \cellcolor[HTML]{F7F8FD}-0.007 & \cellcolor[HTML]{F6F8FD}-0.008 & \cellcolor[HTML]{FCFCFF}0.006 & \multicolumn{1}{c|}{\cellcolor[HTML]{FCF7FA}0.052} & \multicolumn{1}{c|}{13.7} &  & 16.3 \\
\multicolumn{1}{l|}{CSOGD$_{C_{II}}$} & \cellcolor[HTML]{5A8AC6}-0.251 & \cellcolor[HTML]{C2D3EA}-0.090 & \cellcolor[HTML]{FCDCDF}0.087 & \multicolumn{1}{c|}{\cellcolor[HTML]{F97A7D}0.353} & \multicolumn{1}{c|}{16.4} &  & \cellcolor[HTML]{5A8AC6}-0.595 & \cellcolor[HTML]{C2D3EA}-0.212 & \cellcolor[HTML]{FCEFF2}0.153 & \multicolumn{1}{c|}{\cellcolor[HTML]{F97274}1.620} & \multicolumn{1}{c|}{16.8} &  & \cellcolor[HTML]{CAD8ED}-0.076 & \cellcolor[HTML]{E1E9F5}-0.040 & \cellcolor[HTML]{FCF2F5}0.099 & \multicolumn{1}{c|}{\cellcolor[HTML]{FAA7AA}0.790} & \multicolumn{1}{c|}{17.1} &  & 16.7 \\
\multicolumn{1}{l|}{CSOGD$_{S_{I}}$} & \cellcolor[HTML]{FAFBFE}-0.002 & \cellcolor[HTML]{FCFBFE}0.005 & \cellcolor[HTML]{FCF8FB}0.012 & \multicolumn{1}{c|}{\cellcolor[HTML]{FCEBEE}0.048} & \multicolumn{1}{c|}{15.6} &  & \cellcolor[HTML]{FCFCFF}0.000 & \cellcolor[HTML]{FCFCFF}0.005 & \cellcolor[HTML]{FCFCFF}0.009 & \multicolumn{1}{c|}{\cellcolor[HTML]{FCF3F6}0.111} & \multicolumn{1}{c|}{14.5} &  & \cellcolor[HTML]{FBFBFE}0.000 & \cellcolor[HTML]{FCFCFF}0.006 & \cellcolor[HTML]{FCFCFF}0.002 & \multicolumn{1}{c|}{\cellcolor[HTML]{FCF1F4}0.105} & \multicolumn{1}{c|}{11.9} &  & 14.0 \\
\multicolumn{1}{l|}{CSOGD$_{S_{II}}$} & \cellcolor[HTML]{FAFAFE}-0.003 & \cellcolor[HTML]{FCF7FA}0.014 & \cellcolor[HTML]{FBC8CB}0.141 & \multicolumn{1}{c|}{\cellcolor[HTML]{F97072}0.381} & \multicolumn{1}{c|}{\textbf{6.5}} &  & \cellcolor[HTML]{FBFBFE}-0.003 & \cellcolor[HTML]{FCFAFD}0.024 & \cellcolor[HTML]{FCE1E4}0.323 & \multicolumn{1}{c|}{\cellcolor[HTML]{F96B6D}1.698} & \multicolumn{1}{c|}{\textbf{4.8}} &  & \cellcolor[HTML]{FBFBFE}0.000 & \cellcolor[HTML]{FCF9FC}0.031 & \cellcolor[HTML]{FCE1E4}0.257 & \multicolumn{1}{c|}{\cellcolor[HTML]{FBC3C6}0.530} & \multicolumn{1}{c|}{\textbf{7.6}} &  & \textbf{6.3} \\
\multicolumn{20}{l}{\cellcolor[HTML]{C0C0C0}\textbf{Ensemble Learning Approaches}} \\
\multicolumn{1}{l|}{OB} & \cellcolor[HTML]{F8F9FD}-0.006 & \cellcolor[HTML]{F8F9FD}-0.006 & \cellcolor[HTML]{F3F5FB}-0.014 & \multicolumn{1}{c|}{\cellcolor[HTML]{FCECEE}0.046} & \multicolumn{1}{c|}{17.8} &  & \cellcolor[HTML]{FBFBFE}-0.002 & \cellcolor[HTML]{FAFAFE}-0.006 & \cellcolor[HTML]{ECF0F9}-0.058 & \multicolumn{1}{c|}{\cellcolor[HTML]{FCF1F4}0.136} & \multicolumn{1}{c|}{16.6} &  & \cellcolor[HTML]{F3F6FC}-0.013 & \cellcolor[HTML]{F7F8FD}-0.007 & \cellcolor[HTML]{CDDBEE}-0.070 & \multicolumn{1}{c|}{\cellcolor[HTML]{FCF1F4}0.108} & \multicolumn{1}{c|}{13.9} &  & 16.1 \\
\multicolumn{1}{l|}{OAdaB} & \cellcolor[HTML]{EBF0F9}-0.026 & \cellcolor[HTML]{FCFCFF}0.001 & \cellcolor[HTML]{FCEAED}0.049 & \multicolumn{1}{c|}{\cellcolor[HTML]{FBC8CB}0.142} & \multicolumn{1}{c|}{16.5} &  & \cellcolor[HTML]{F6F8FD}-0.020 & \cellcolor[HTML]{FCFCFF}0.005 & \cellcolor[HTML]{FCF1F4}0.130 & \multicolumn{1}{c|}{\cellcolor[HTML]{FAACAE}0.943} & \multicolumn{1}{c|}{16.7} &  & \cellcolor[HTML]{E6ECF7}-0.033 & \cellcolor[HTML]{FCFCFF}0.004 & \cellcolor[HTML]{FCF0F3}0.115 & \multicolumn{1}{c|}{\cellcolor[HTML]{FAA2A5}0.834} & \multicolumn{1}{c|}{13.6} &  & 15.6 \\
\multicolumn{1}{l|}{OAdaC2} & \cellcolor[HTML]{93B2DA}-0.162 & \cellcolor[HTML]{E9EEF8}-0.029 & \cellcolor[HTML]{FBD2D5}0.115 & \multicolumn{1}{c|}{\cellcolor[HTML]{F98B8D}0.310} & \multicolumn{1}{c|}{15.4} &  & \cellcolor[HTML]{A8C1E1}-0.306 & \cellcolor[HTML]{E7EDF7}-0.076 & \cellcolor[HTML]{FCE7EA}0.247 & \multicolumn{1}{c|}{\cellcolor[HTML]{F97B7D}1.513} & \multicolumn{1}{c|}{16.1} &  & \cellcolor[HTML]{E0E8F5}-0.042 & \cellcolor[HTML]{FCFBFE}0.012 & \cellcolor[HTML]{FCEAED}0.173 & \multicolumn{1}{c|}{\cellcolor[HTML]{FA9EA0}0.877} & \multicolumn{1}{c|}{14.1} &  & 15.2 \\
\multicolumn{1}{l|}{OCSB2} & \cellcolor[HTML]{BBCEE8}-0.100 & \cellcolor[HTML]{F6F7FC}-0.009 & \cellcolor[HTML]{FBCFD2}0.124 & \multicolumn{1}{c|}{\cellcolor[HTML]{FA999C}0.269} & \multicolumn{1}{c|}{14.3} &  & \cellcolor[HTML]{C8D7EC}-0.188 & \cellcolor[HTML]{EDF2FA}-0.052 & \cellcolor[HTML]{FCE5E8}0.270 & \multicolumn{1}{c|}{\cellcolor[HTML]{F98285}1.426} & \multicolumn{1}{c|}{15.1} &  & \cellcolor[HTML]{F3F6FC}-0.013 & \cellcolor[HTML]{FCFAFD}0.027 & \cellcolor[HTML]{FCE7EA}0.200 & \multicolumn{1}{c|}{\cellcolor[HTML]{FAAFB1}0.719} & \multicolumn{1}{c|}{12.6} &  & 14.0 \\
\multicolumn{1}{l|}{OKB} & \cellcolor[HTML]{D9E4F3}-0.053 & \cellcolor[HTML]{FCFAFD}0.008 & \cellcolor[HTML]{FCF3F6}0.024 & \multicolumn{1}{c|}{\cellcolor[HTML]{FCE2E4}0.073} & \multicolumn{1}{c|}{17.0} &  & \cellcolor[HTML]{EEF2FA}-0.051 & \cellcolor[HTML]{FCFBFE}0.017 & \cellcolor[HTML]{FCF3F6}0.113 & \multicolumn{1}{c|}{\cellcolor[HTML]{FBC5C7}0.653} & \multicolumn{1}{c|}{16.1} &  & \cellcolor[HTML]{CBD9ED}-0.074 & \cellcolor[HTML]{FCFAFD}0.019 & \cellcolor[HTML]{FCF1F4}0.108 & \multicolumn{1}{c|}{\cellcolor[HTML]{FBC9CC}0.477} & \multicolumn{1}{c|}{13.5} &  & 15.5 \\
\multicolumn{1}{l|}{OUOB} & \cellcolor[HTML]{F4F6FC}-0.011 & \cellcolor[HTML]{FCF9FC}0.009 & \cellcolor[HTML]{FBC3C6}0.156 & \multicolumn{1}{c|}{\cellcolor[HTML]{F97072}0.381} & \multicolumn{1}{c|}{\textbf{7.1}} &  & \cellcolor[HTML]{FAFAFE}-0.006 & \cellcolor[HTML]{FCFBFE}0.016 & \cellcolor[HTML]{FCE1E4}0.320 & \multicolumn{1}{c|}{\cellcolor[HTML]{F8696B}1.718} & \multicolumn{1}{c|}{\textbf{6.6}} &  & \cellcolor[HTML]{F2F5FB}-0.014 & \cellcolor[HTML]{FCFAFD}0.026 & \cellcolor[HTML]{FCE2E4}0.249 & \multicolumn{1}{c|}{\cellcolor[HTML]{FBBEC0}0.580} & \multicolumn{1}{c|}{\textbf{9.8}} &  & \textbf{7.9} \\
\multicolumn{1}{l|}{ORUSB1} & \cellcolor[HTML]{EFF2FA}-0.020 & \cellcolor[HTML]{FCFAFD}0.007 & \cellcolor[HTML]{FBCACD}0.137 & \multicolumn{1}{c|}{\cellcolor[HTML]{FCE4E7}0.067} & \multicolumn{1}{c|}{11.3} &  & \cellcolor[HTML]{F5F7FC}-0.025 & \cellcolor[HTML]{F8F9FD}-0.013 & \cellcolor[HTML]{FCE8EB}0.236 & \multicolumn{1}{c|}{\cellcolor[HTML]{FCF9FC}0.046} & \multicolumn{1}{c|}{15.0} &  & \cellcolor[HTML]{FBFBFE}-0.001 & \cellcolor[HTML]{FCF8FB}0.039 & \cellcolor[HTML]{FCEEF1}0.136 & \multicolumn{1}{c|}{\cellcolor[HTML]{5A8AC6}-0.245} & \multicolumn{1}{c|}{14.5} &  & 13.6 \\
\multicolumn{1}{l|}{ORUSB2} & \cellcolor[HTML]{E7EDF7}-0.032 & \cellcolor[HTML]{FCFBFE}0.003; & \cellcolor[HTML]{FBCCCE}0.133 & \multicolumn{1}{c|}{\cellcolor[HTML]{FA9597}0.281} & \multicolumn{1}{c|}{12.4} &  & \cellcolor[HTML]{EEF2FA}-0.048 & \cellcolor[HTML]{FAFBFE}-0.005 & \cellcolor[HTML]{FCE3E6}0.293 & \multicolumn{1}{c|}{\cellcolor[HTML]{F97E80}1.482} & \multicolumn{1}{c|}{13.6} &  & \cellcolor[HTML]{FCFCFF}0.006 & \cellcolor[HTML]{FCF9FC}0.029 & \cellcolor[HTML]{FCE8EB}0.192 & \multicolumn{1}{c|}{\cellcolor[HTML]{FBCCCF}0.448} & \multicolumn{1}{c|}{12.5} &  & 12.8 \\
\multicolumn{1}{l|}{ORUSB3} & \cellcolor[HTML]{EDF1F9}-0.023 & \cellcolor[HTML]{FCF9FC}0.008 & \cellcolor[HTML]{FBC5C8}0.150 & \multicolumn{1}{c|}{\cellcolor[HTML]{F98587}0.325} & \multicolumn{1}{c|}{9.4} &  & \cellcolor[HTML]{F6F8FD}-0.019 & \cellcolor[HTML]{FCFBFE}0.019 & \cellcolor[HTML]{FCE3E6}0.295 & \multicolumn{1}{c|}{\cellcolor[HTML]{FA9193}1.258} & \multicolumn{1}{c|}{11.3} &  & \cellcolor[HTML]{E7EDF7}-0.030 & \cellcolor[HTML]{FCFAFD}0.024 & \cellcolor[HTML]{FCE8EB}0.185 & \multicolumn{1}{c|}{\cellcolor[HTML]{FCF5F8}0.069} & \multicolumn{1}{c|}{14.6} &  & 11.8 \\
\multicolumn{1}{l|}{OOB} & \cellcolor[HTML]{F1F4FB}-0.017 & \cellcolor[HTML]{FCFAFD}0.008 & \cellcolor[HTML]{FBCDD0}0.128 & \multicolumn{1}{c|}{\cellcolor[HTML]{FA9597}0.282} & \multicolumn{1}{c|}{10.8} &  & \cellcolor[HTML]{F8F9FD}-0.012 & \cellcolor[HTML]{FCFBFE}0.015 & \cellcolor[HTML]{FCE3E6}0.300 & \multicolumn{1}{c|}{\cellcolor[HTML]{F98486}1.413} & \multicolumn{1}{c|}{9.4} &  & \cellcolor[HTML]{E5EBF6}-0.035 & \cellcolor[HTML]{FCFBFE}0.015 & \cellcolor[HTML]{FCDEE1}0.281 & \multicolumn{1}{c|}{\cellcolor[HTML]{F96C6E}1.340} & \multicolumn{1}{c|}{\textbf{7.9}} &  & 9.4 \\
\multicolumn{1}{l|}{OUB} & \cellcolor[HTML]{FCFCFF}0.000 & \cellcolor[HTML]{FCF9FB}0.011 & \cellcolor[HTML]{FBC2C4}0.159 & \multicolumn{1}{c|}{\cellcolor[HTML]{F8696B}0.399} & \multicolumn{1}{c|}{\textbf{5.4}} &  & \cellcolor[HTML]{FBFBFE}-0.002 & \cellcolor[HTML]{FCFBFE}0.016 & \cellcolor[HTML]{FCE2E5}0.305 & \multicolumn{1}{c|}{\cellcolor[HTML]{F96D6F}1.674} & \multicolumn{1}{c|}{\textbf{7.2}} &  & \cellcolor[HTML]{FCFCFF}0.005 & \cellcolor[HTML]{FCFAFD}0.027 & \cellcolor[HTML]{FCE7EA}0.198 & \multicolumn{1}{c|}{\cellcolor[HTML]{FBCFD1}0.425} & \multicolumn{1}{c|}{10.8} &  & \textbf{7.8} \\
\multicolumn{1}{l|}{OWOB} & \cellcolor[HTML]{F9FAFE}-0.004 & \cellcolor[HTML]{FCF9FC}0.010 & \cellcolor[HTML]{FBCCCF}0.132 & \multicolumn{1}{c|}{\cellcolor[HTML]{FA9497}0.283} & \multicolumn{1}{c|}{9.5} &  & \cellcolor[HTML]{FBFBFE}-0.003 & \cellcolor[HTML]{FCFBFE}0.021 & \cellcolor[HTML]{FCE3E6}0.299 & \multicolumn{1}{c|}{\cellcolor[HTML]{F98284}1.437} & \multicolumn{1}{c|}{\textbf{8.2}} &  & \cellcolor[HTML]{F4F6FC}-0.011 & \cellcolor[HTML]{FCFAFD}0.026 & \cellcolor[HTML]{FCDFE1}0.276 & \multicolumn{1}{c|}{\cellcolor[HTML]{F8696B}1.361} & \multicolumn{1}{c|}{\textbf{6.3}} &  & \textbf{8.0} \\
\multicolumn{1}{l|}{OWUB} & \cellcolor[HTML]{FAFAFE}-0.003 & \cellcolor[HTML]{FCF9FC}0.009 & \cellcolor[HTML]{FBC3C6}0.155 & \multicolumn{1}{c|}{\cellcolor[HTML]{F96D6F}0.390} & \multicolumn{1}{c|}{\textbf{6.9}} &  & \cellcolor[HTML]{FAFBFE}-0.005 & \cellcolor[HTML]{FCFBFE}0.017 & \cellcolor[HTML]{FCE2E5}0.309 & \multicolumn{1}{c|}{\cellcolor[HTML]{F97072}1.643} & \multicolumn{1}{c|}{8.4} &  & \cellcolor[HTML]{F4F6FC}-0.012 & \cellcolor[HTML]{FCFAFD}0.024 & \cellcolor[HTML]{FCE6E8}0.212 & \multicolumn{1}{c|}{\cellcolor[HTML]{FBCACD}0.468} & \multicolumn{1}{c|}{11.5} &  & 8.9 \\
\multicolumn{1}{l|}{OEB} & \cellcolor[HTML]{F7F8FD}-0.007 & \cellcolor[HTML]{FCFAFD}0.007 & \cellcolor[HTML]{FBCCCF}0.132 & \multicolumn{1}{c|}{\cellcolor[HTML]{F97476}0.372} & \multicolumn{1}{c|}{9.5} &  & \cellcolor[HTML]{FBFBFE}-0.004 & \cellcolor[HTML]{FCFBFE}0.017 & \cellcolor[HTML]{FCE3E6}0.299 & \multicolumn{1}{c|}{\cellcolor[HTML]{F97072}1.640} & \multicolumn{1}{c|}{9.5} &  & \cellcolor[HTML]{EFF3FA}-0.019 & \cellcolor[HTML]{FCFAFD}0.020 & \cellcolor[HTML]{FCE6E8}0.212 & \multicolumn{1}{c|}{\cellcolor[HTML]{FBCCCF}0.450} & \multicolumn{1}{c|}{12.7} &  & 10.6 \\
\multicolumn{20}{l}{\cellcolor[HTML]{C0C0C0}\textbf{The Proposed Approach}} \\
\multicolumn{1}{l|}{HGD} & \cellcolor[HTML]{FCFCFF}0.001 & \cellcolor[HTML]{FCF8FB}0.011 & \cellcolor[HTML]{FBC7CA}0.145 & \multicolumn{1}{c|}{\cellcolor[HTML]{F97476}0.371} & \multicolumn{1}{c|}{\textbf{7.0}} &  & \cellcolor[HTML]{FBFBFE}-0.003 & \cellcolor[HTML]{FCFBFE}0.020 & \cellcolor[HTML]{FCE1E4}0.319 & \multicolumn{1}{c|}{\cellcolor[HTML]{F96B6D}1.702} & \multicolumn{1}{c|}{\textbf{6.2}} &  & \cellcolor[HTML]{F8F9FD}-0.005 & \cellcolor[HTML]{FCF9FC}0.031 & \cellcolor[HTML]{FCE2E5}0.247 & \multicolumn{1}{c|}{\cellcolor[HTML]{FBBFC2}0.567} & \multicolumn{1}{c|}{\textbf{8.6}} &  & \textbf{7.3} \\ \hline
\end{tabular}
}
\begin{tablenotes}
    \item{$\cdot$} The results are shown as increments or decrements w.r.t. the BaseLine (OGD). The bold entries stand for the top three highest rankings.
\end{tablenotes}
\label{S:tab:cmp_static_k}
\end{table*}
%===============================================

%===============================================
%===============================================

%===============================================
%===============================================

%===============================================
%===============================================

%===============================================
%===============================================

\clearpage
\newpage

\subsection{Time Efficiency}

\begin{figure}[h]
    \centering
    \subfloat[Perceptron]{
        \includegraphics[width=4cm]{Figs/Exp_Static/Stastic_P_TvsP_AUC.png}
    }
    \subfloat[Linear SVM]{
        \includegraphics[width=4cm]{Figs/Exp_Static/Stastic_V_TvsP_AUC.png}
    }
    \subfloat[Kernel Model]{
        \includegraphics[width=4cm]{Figs/Exp_Static/Stastic_K_TvsP_AUC.png}
    }\\
    \subfloat[Perceptron]{
        \includegraphics[width=4cm]{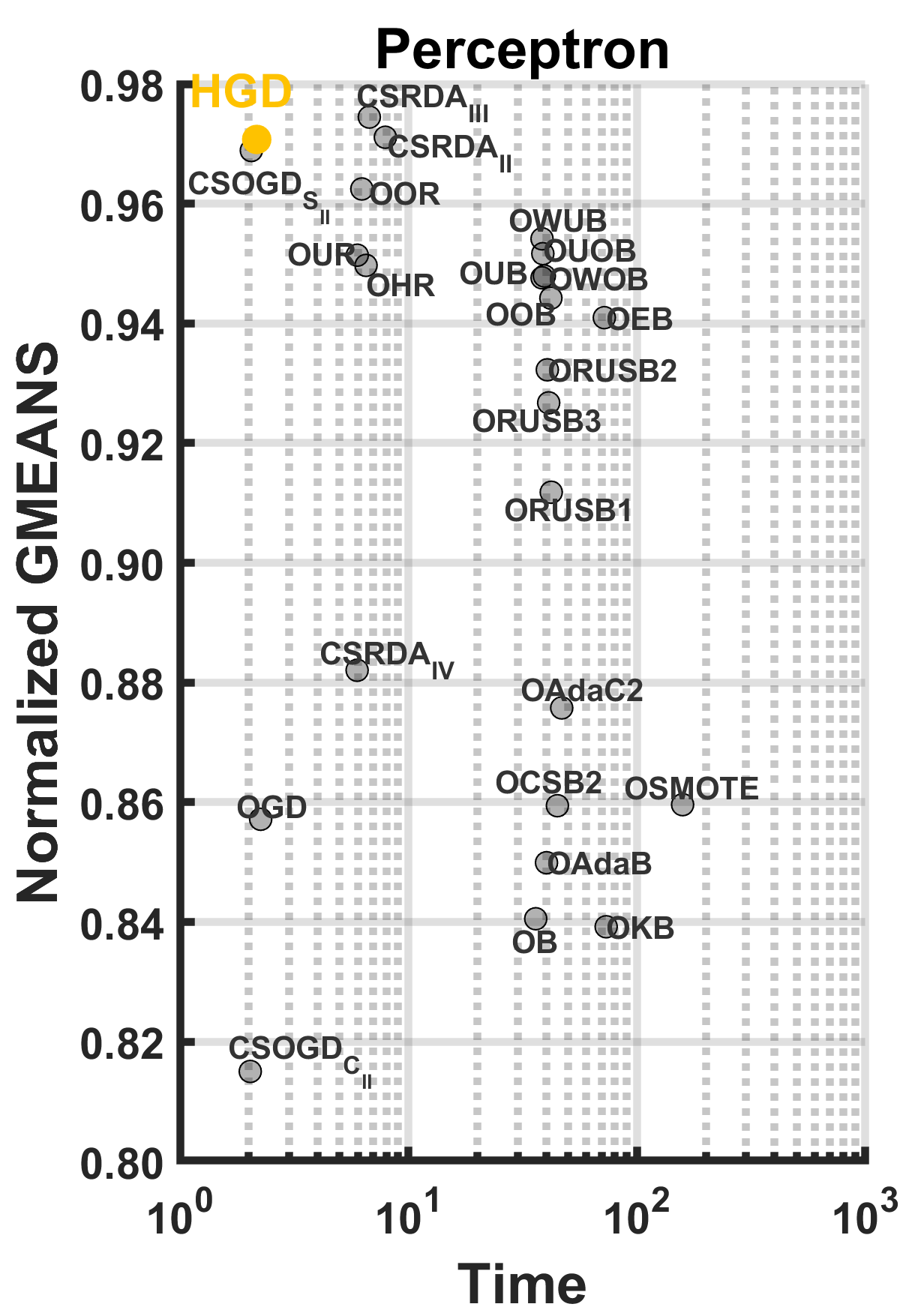}
    }
    \subfloat[Linear SVM]{
        \includegraphics[width=4cm]{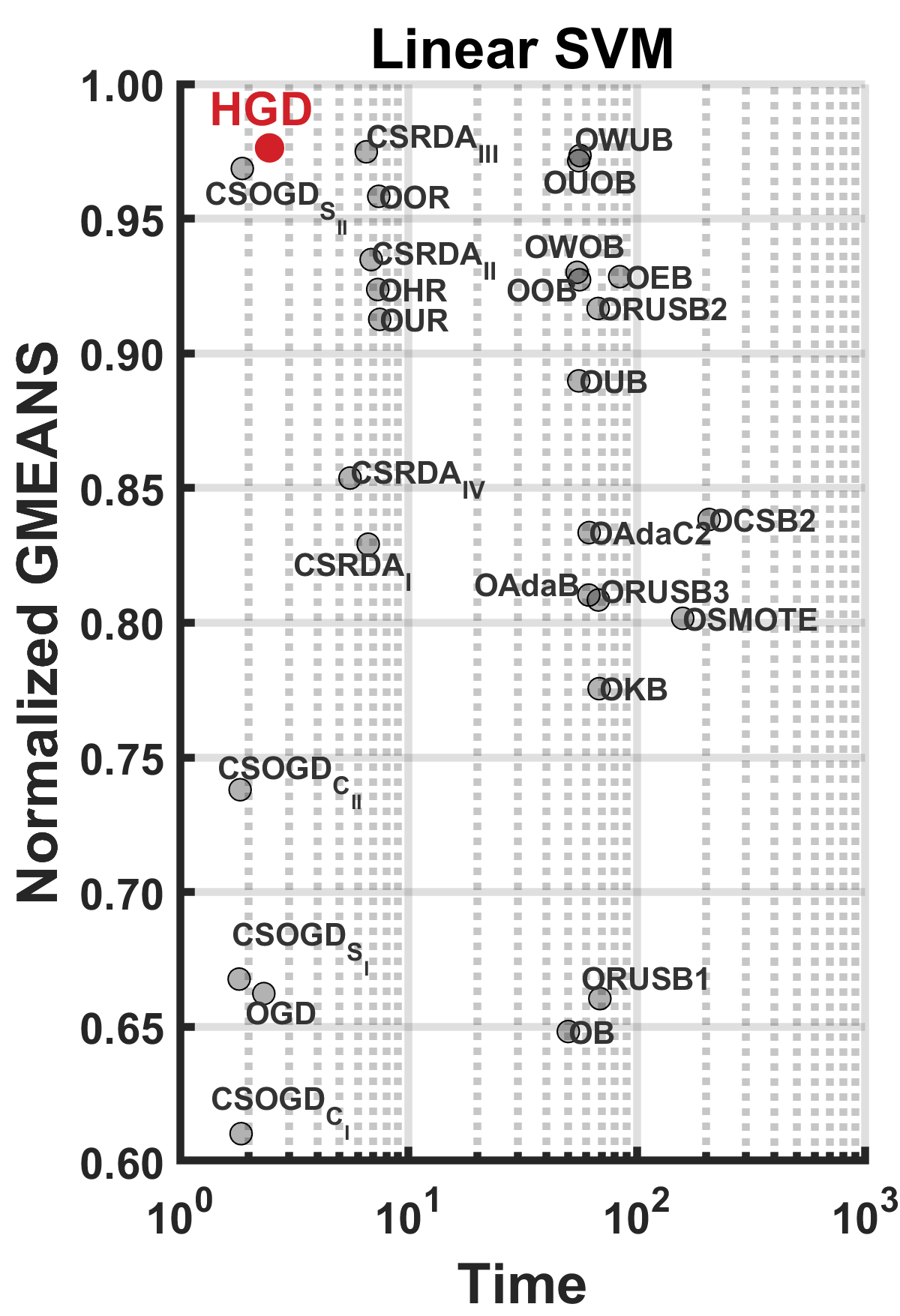}
    }
    \subfloat[Kernel Model]{
        \includegraphics[width=4cm]{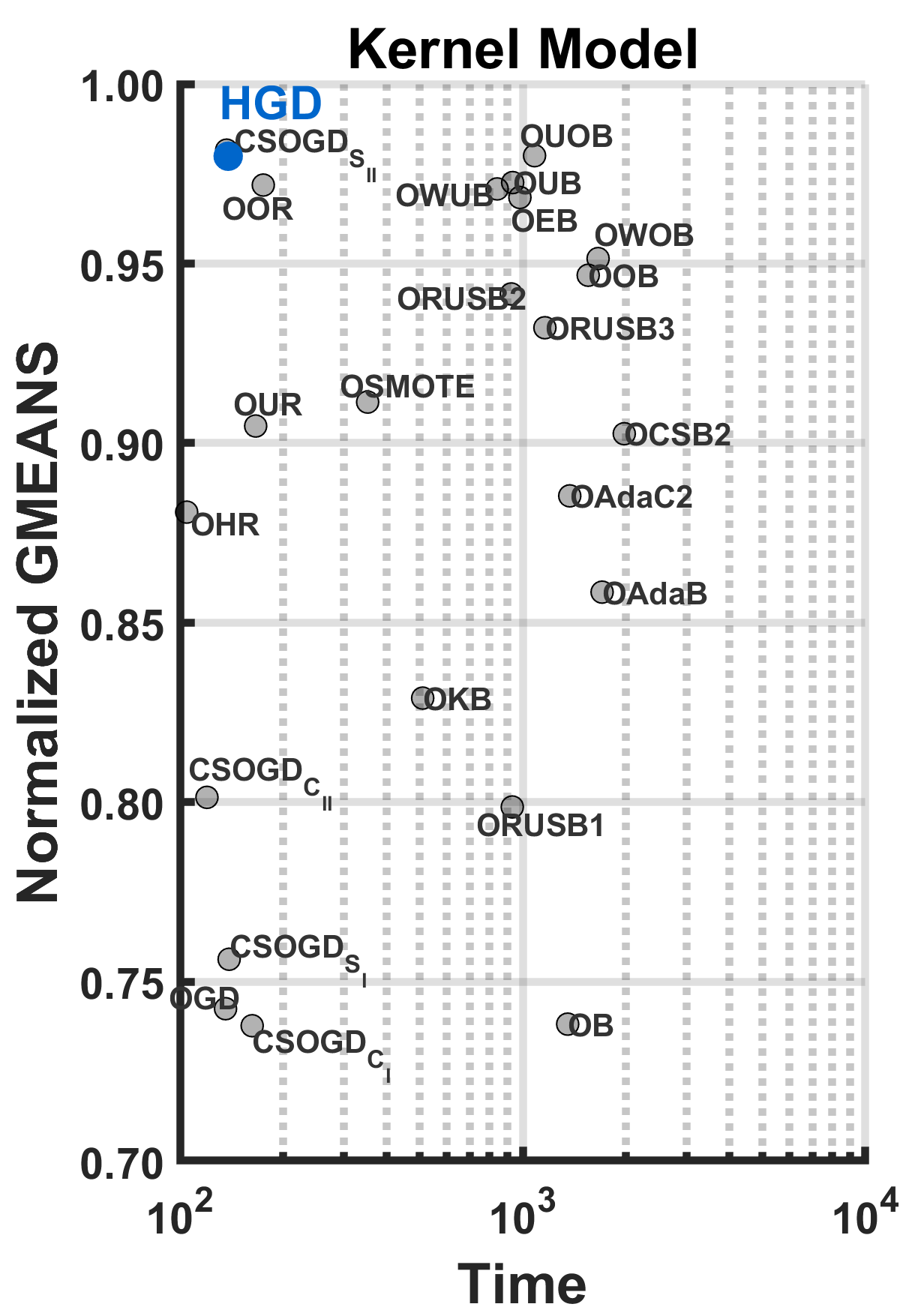}
    }\\
    \subfloat[Perceptron]{
        \includegraphics[width=4cm]{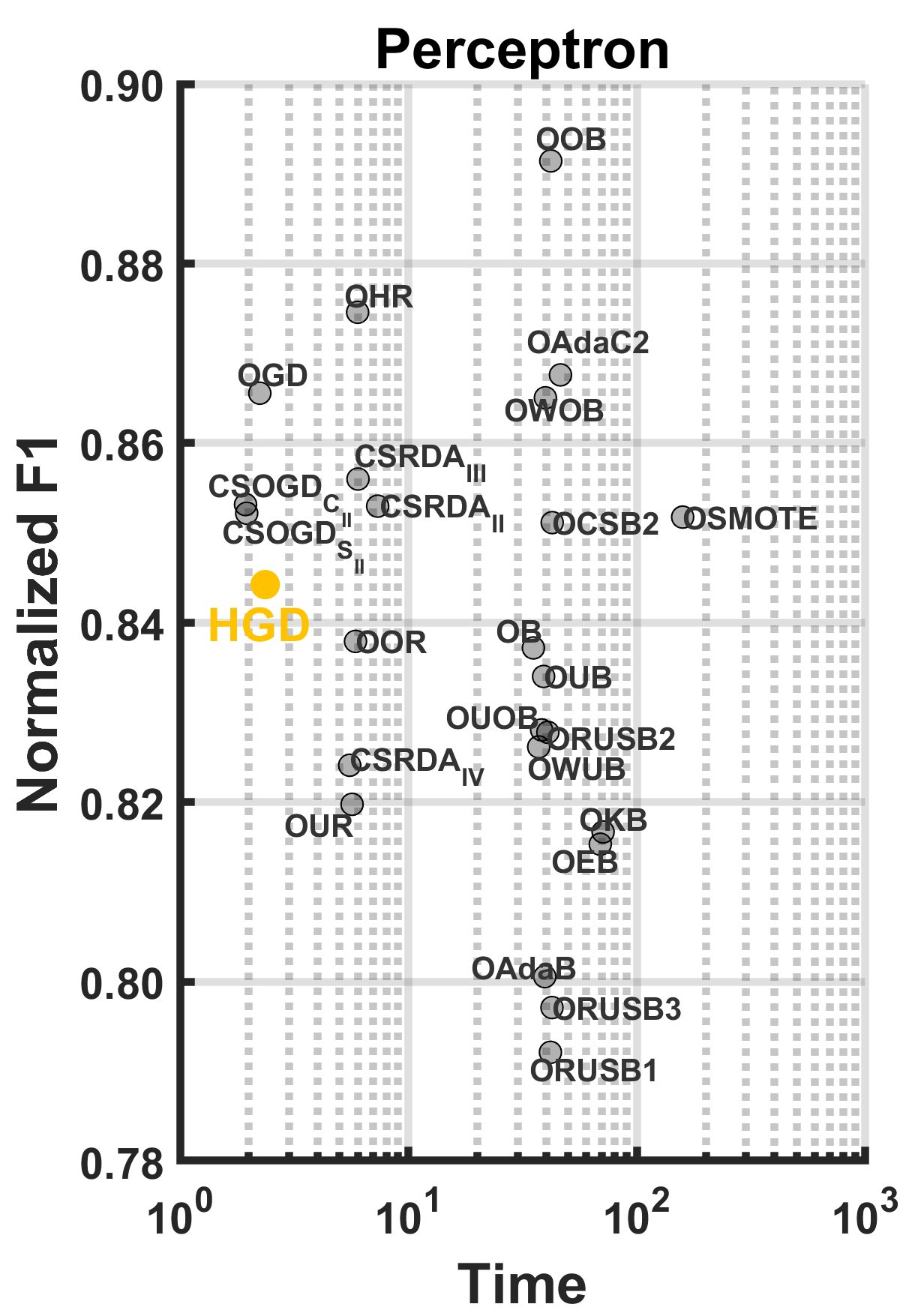}
    } 
    \subfloat[Linear SVM]{
        \includegraphics[width=4cm]{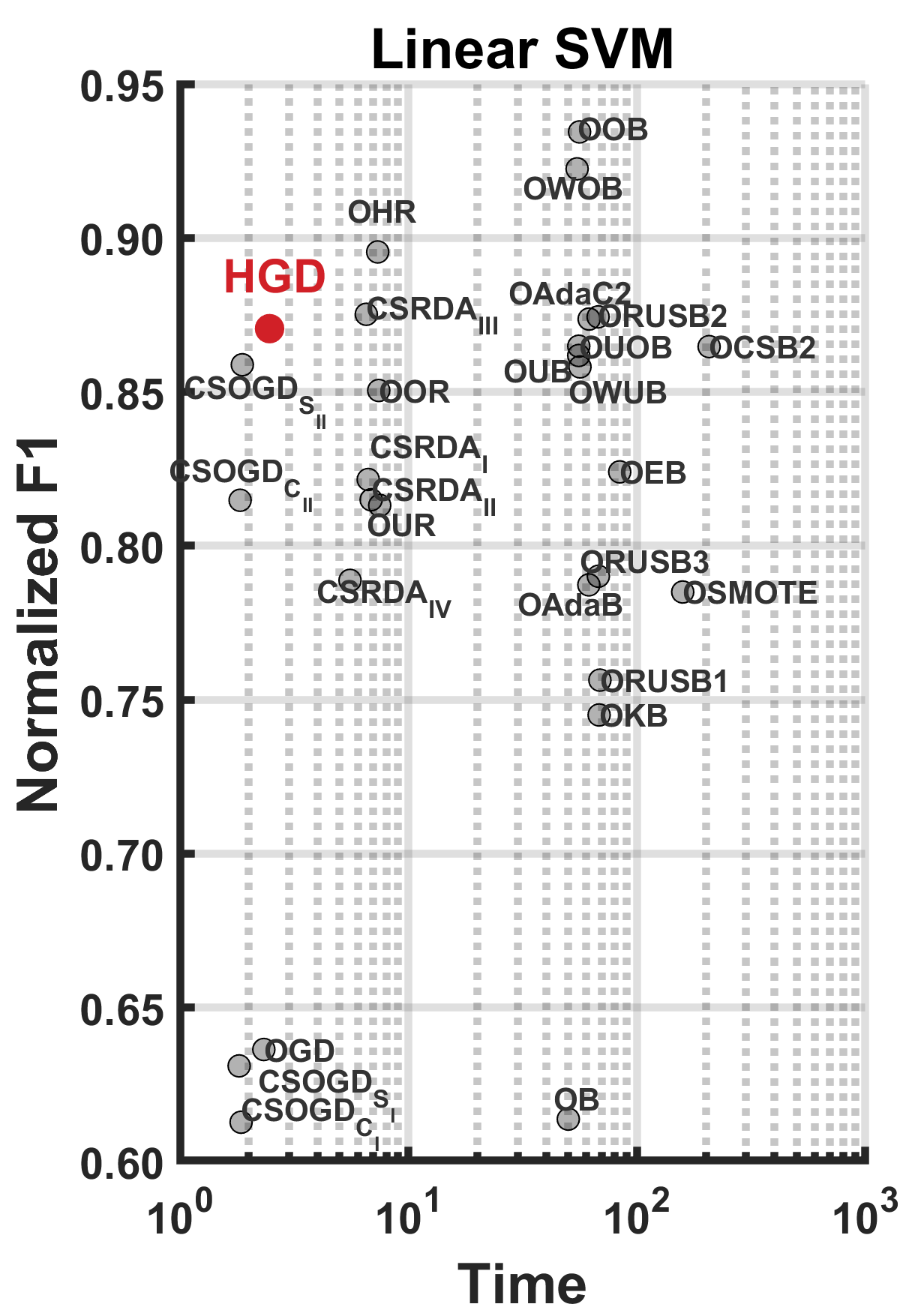}
    }
    \subfloat[Kernel Model]{
        \includegraphics[width=4cm]{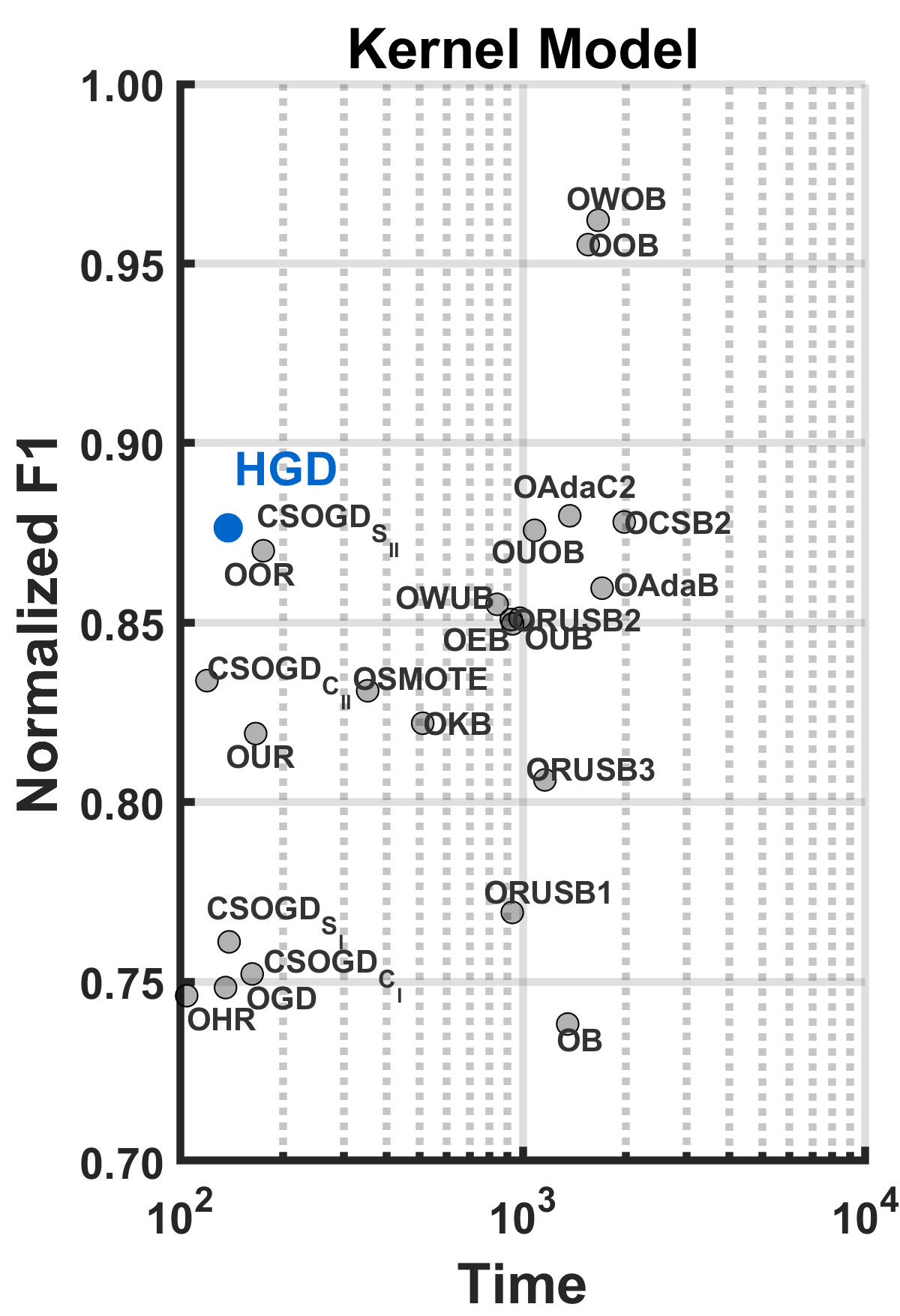}
    }
    \caption{The performance and computational time comparison of all the methods, in terms of (a)-(c) AUC; (d)-(f) GMEANS; (g)-(i) F1 . Up-left denotes higher performance with less computational cost.}
    \label{S:fig:static_TvP}
\end{figure}

To comprehensively evaluate the effectiveness and efficiency of the proposed HGD, we conducted a comparative analysis that examines the performance metrics in relation to computational time across all methods considered. The results are depicted in Figure~\ref{S:fig:static_TvP}. 
Methods positioned in the upper-left part of the figures denote those achieving superior performance while requiring fewer computational resources. Our findings reveal that HGD is consistently positioned favorably in the upper-left part across all the base learner settings, demonstrating both performance efficacy and computational efficiency.

%===============================================
%===============================================

\clearpage
\newpage

\clearpage
\newpage

\section{Results: Learning under Dynamic Imbalanced Data Streams}
\subsection{Datasets}

To further study the performance of HGD under dynamic imbalanced data streams, we selected several datasets from Table~\ref{S:tab:datasets} and simulated data streams featuring dynamic imbalance ratios. A comprehensive details of the dynamic data streams is provided in Table~\ref{S:tab:datasets_dy}. Specifically, our analysis encompassed three types of dynamic imbalance ratios: sudden increase, sudden decrease, and gradual variation, thereby covering distinct characteristics of dynamic imbalance phenomena.

\begin{table*}[h]
\centering
\renewcommand\arraystretch{1.2}
\caption{The Datasets with Simulated Dynamic Imbalance Ratio}
\begin{tabular}{c|c|c}
\hline
\textbf{Datasets} & \textbf{IR Type} & \textbf{Settings} \\ \hline
\multirow{3}{*}{\textbf{a8a}} & \textbf{Sudden Increse} & 1$\xrightarrow{IR:2}$6000$\xrightarrow{IR:2.5}$11000$\xrightarrow{IR:4}$15000 \\
 & \textbf{Sudden Decrease} & 1$\xrightarrow{IR:4}$4000$\xrightarrow{IR:2.5}$9000$\xrightarrow{IR:2}$15000 \\
 & \textbf{Gradual Varying} & 1$\xrightarrow{IR:4\to2}$15000 \\ \hline
\multirow{3}{*}{\textbf{Click}} & \textbf{Sudden Increse} & 1$\xrightarrow{IR:3.5}$11000$\xrightarrow{IR:5}$24000$\xrightarrow{IR:10}$38000 \\
 & \textbf{Sudden Decrease} & 1$\xrightarrow{IR:10}$14000$\xrightarrow{IR:5}$26000$\xrightarrow{IR:3.3}$38000 \\
 & \textbf{Gradual Varying} & 1$\xrightarrow{IR:10\to4}$26000 \\ \hline
\multirow{3}{*}{\textbf{MagicTelescope}} & \textbf{Sudden Increse} & 1$\xrightarrow{IR:1}$4000$\xrightarrow{IR:4}$6500$\xrightarrow{IR:20}$8600 \\
 & \textbf{Sudden Decrease} & 1$\xrightarrow{IR:20}$2100$\xrightarrow{IR:4}$4600$\xrightarrow{IR:1}$8600 \\
 & \textbf{Gradual Varying} & 1$\xrightarrow{IR:2\to20}$5000$\xrightarrow{IR:20\to2}$10000 \\ \hline
\multirow{3}{*}{\textbf{phoneme}} & \textbf{Sudden Increse} & 1$\xrightarrow{IR:1.25}$900$\xrightarrow{IR:1.6}$1700$\xrightarrow{IR:2.5}$2400$\xrightarrow{IR:5}$3000$\xrightarrow{IR:10}$3550 \\
 & \textbf{Sudden Decrease} & 1$\xrightarrow{IR:10}$550$\xrightarrow{IR:5}$1150$\xrightarrow{IR:2.5}$1850$\xrightarrow{IR:1.6}$2650$\xrightarrow{IR:1.25}$3550 \\
 & \textbf{Gradual Varying} & 1$\xrightarrow{IR:10\to1.25}$3500 \\ \hline
\multirow{3}{*}{\textbf{Satellite}} & \textbf{Sudden Increse} & 1$\xrightarrow{IR:33}$1030$\xrightarrow{IR:50}$2050$\xrightarrow{IR:33}$3060 \\
 & \textbf{Sudden Decrease} & 1$\xrightarrow{IR:100}$1100$\xrightarrow{IR:50}$2030$\xrightarrow{IR:33}$3060 \\
 & \textbf{Gradual Varying} & 1$\xrightarrow{IR:100\to50\to100}$3500 \\ \hline
\multirow{3}{*}{\textbf{steel-plates-fault}} & \textbf{Sudden Increse} & 1$\xrightarrow{IR:1}$200$\xrightarrow{IR:2}$800$\xrightarrow{IR:4}$1300 \\
 & \textbf{Sudden Decrease} & 1$\xrightarrow{IR:4}$500$\xrightarrow{IR:2}$1100$\xrightarrow{IR:1}$1300 \\
 & \textbf{Gradual Varying} & 1$\xrightarrow{IR:2.5\to5}$3500 \\ \hline
\multirow{3}{*}{\textbf{pol}} & \textbf{Sudden Increse} & 1$\xrightarrow{IR:1.1}$3800$\xrightarrow{IR:1.3}$7300$\xrightarrow{IR:2}$10300 \\
 & \textbf{Sudden Decrease} & 1$\xrightarrow{IR:2}$3000$\xrightarrow{IR:1.3}$6500$\xrightarrow{IR:1.1}$10300 \\
 & \textbf{Gradual Varying} & 1$\xrightarrow{IR:1.1\to2}$1000 \\ \hline
\multirow{3}{*}{\textbf{mushroom}} & \textbf{Sudden Increse} & 1$\xrightarrow{IR:1}$2000$\xrightarrow{IR:2}$5000 \\
 & \textbf{Sudden Decrease} & 1$\xrightarrow{IR:2}$3000$\xrightarrow{IR:1}$5000 \\
 & \textbf{Gradual Varying} & 1$\xrightarrow{IR:1\to2\to1}$3500 \\ \hline
\end{tabular}
\label{S:tab:datasets_dy}
\end{table*}
%========================================

\clearpage
\newpage

\subsection{Performance Comparison}

\begin{figure}[th]
    \centering
    \subfloat[AUC]{
        \includegraphics[width=2.19cm]{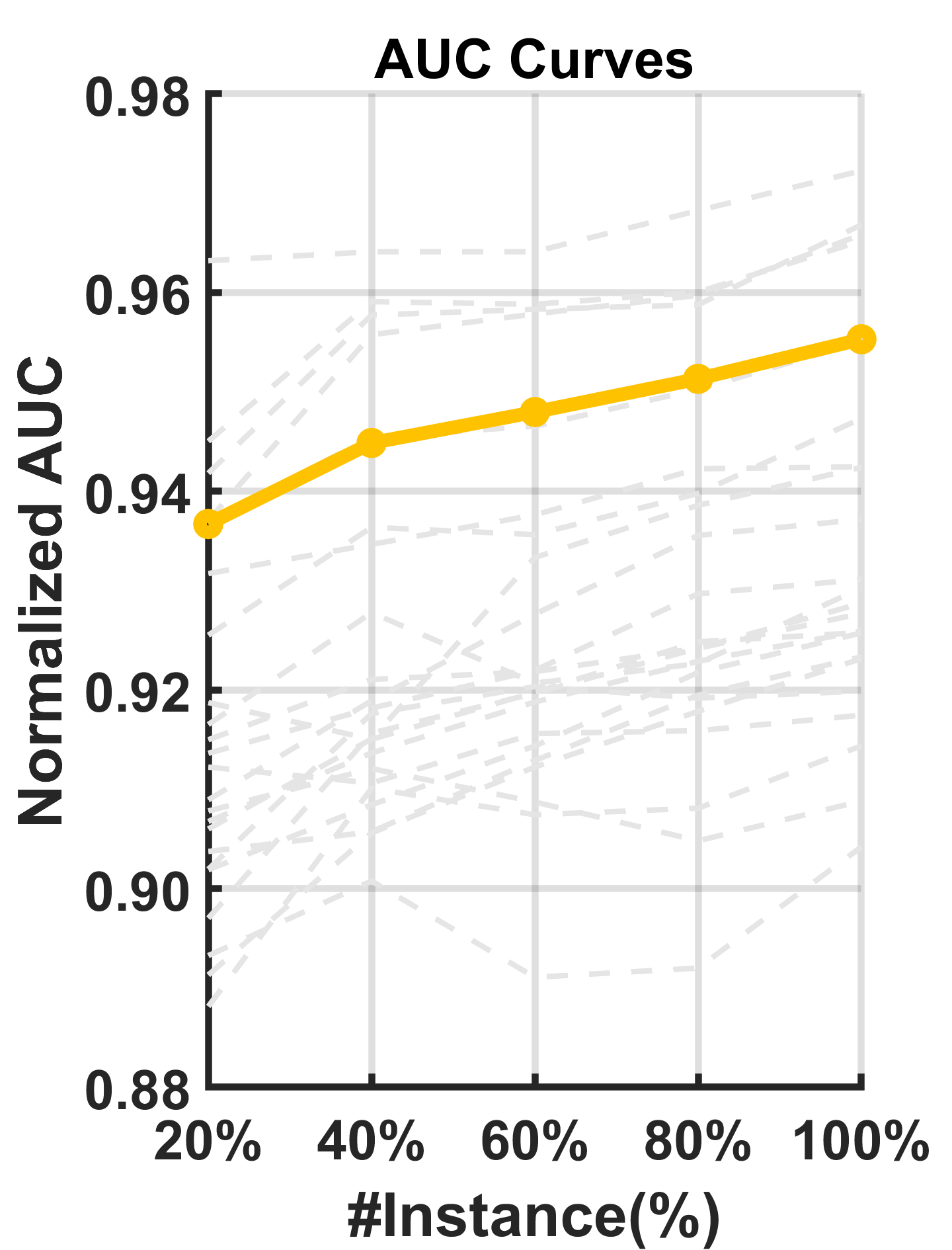}
    }\hspace{-6mm}
    \subfloat[AUC]{
        \includegraphics[width=2.19cm]{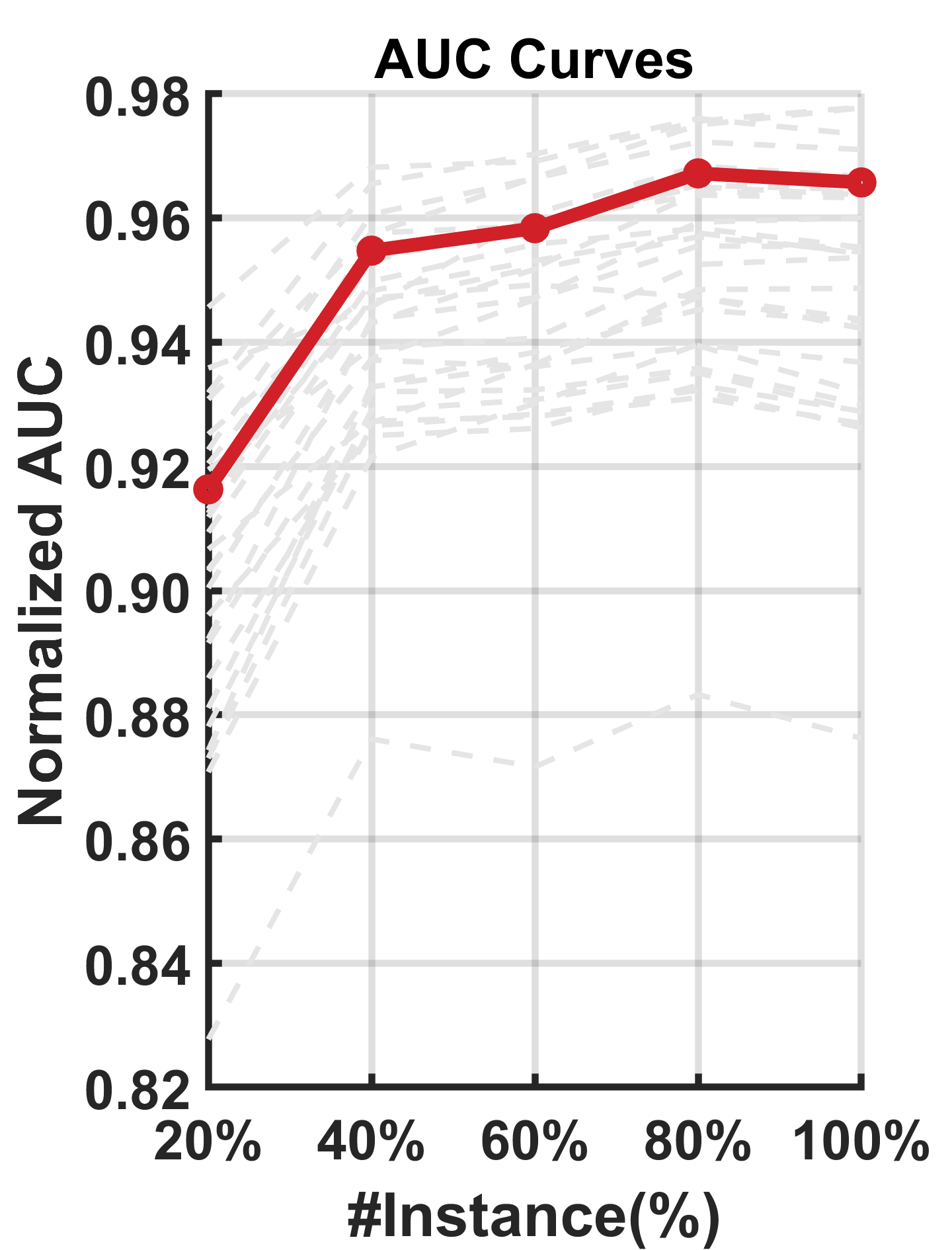}
    }\hspace{-6mm}
    \subfloat[AUC]{
        \includegraphics[width=2.19cm]{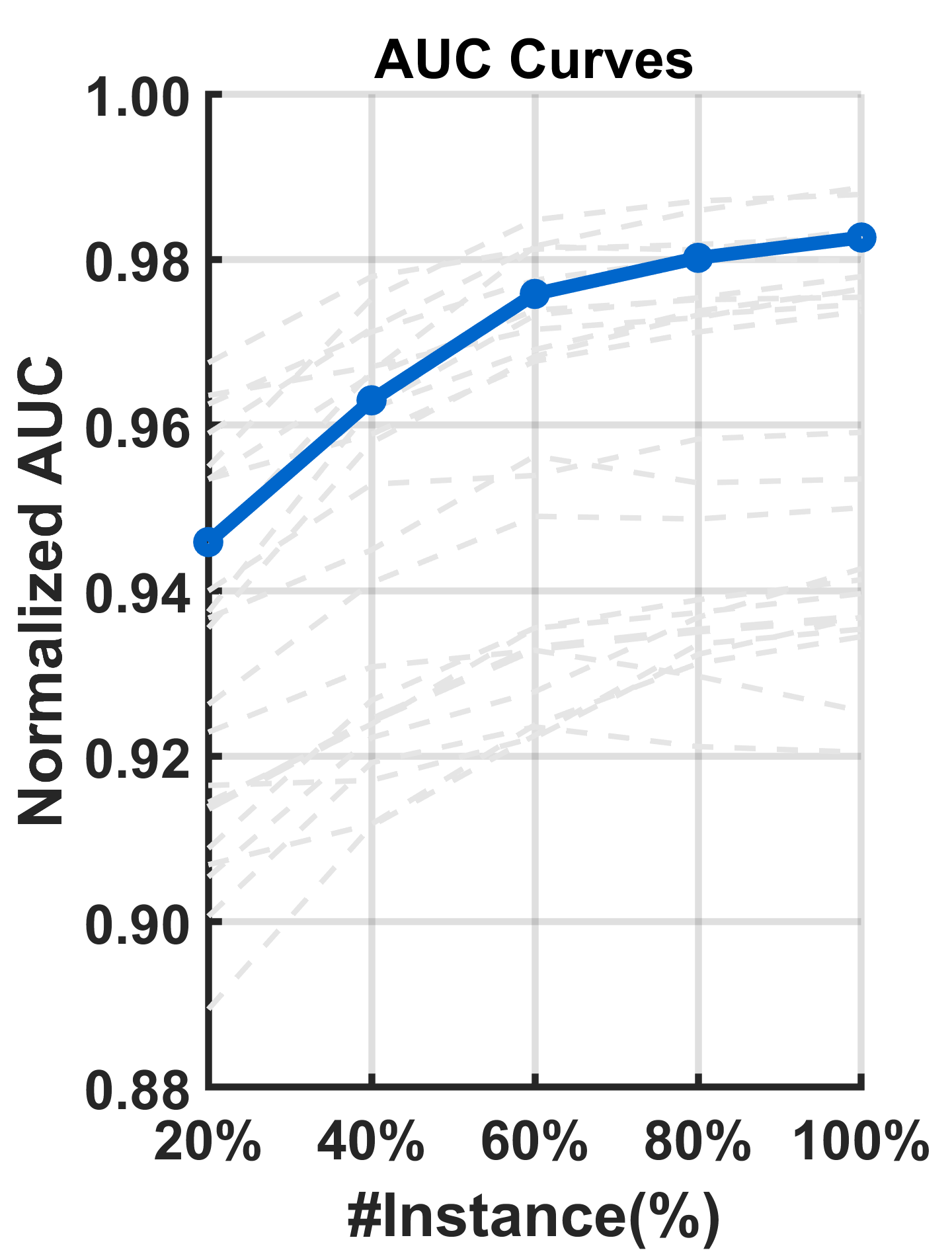}
    }\hspace{-6mm}
    \subfloat[GMEANS]{
        \includegraphics[width=2.19cm]{Figs/Exp_Dy/Dy_P_Curves_GMEANS.png}
    }\hspace{-6mm}
    \subfloat[GMEANS]{
        \includegraphics[width=2.19cm]{Figs/Exp_Dy/Dy_V_Curves_GMEANS.png}
    }\hspace{-6mm}
    \subfloat[GMEANS]{
        \includegraphics[width=2.19cm]{Figs/Exp_Dy/Dy_K_Curves_GMEANS.png}
    }\hspace{-6mm}
    \subfloat[F1]{
        \includegraphics[width=2.19cm]{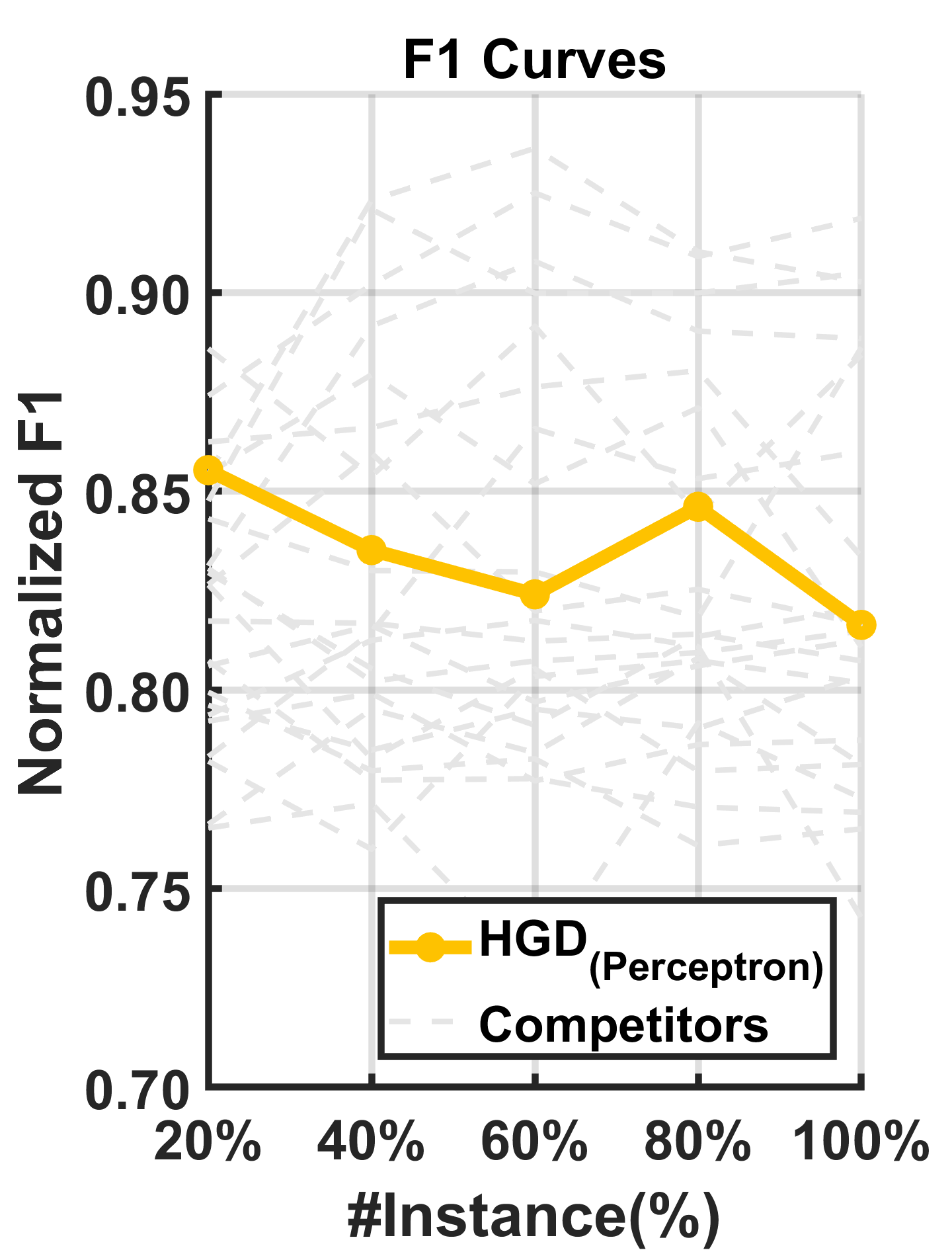}
    }\hspace{-6mm}
    \subfloat[F1]{
        \includegraphics[width=2.19cm]{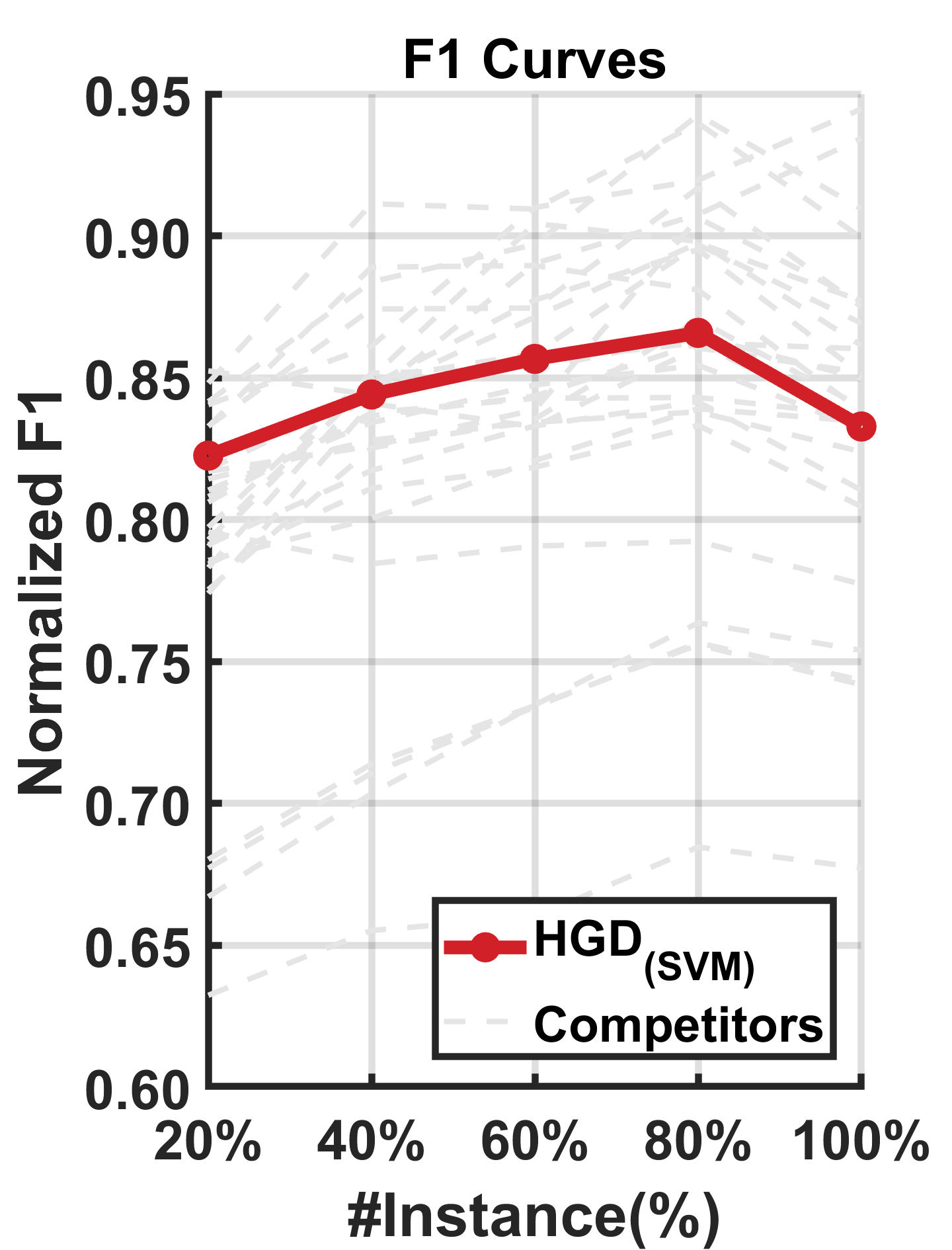}
    }\hspace{-6mm}
    \subfloat[F1]{
        \includegraphics[width=2.19cm]{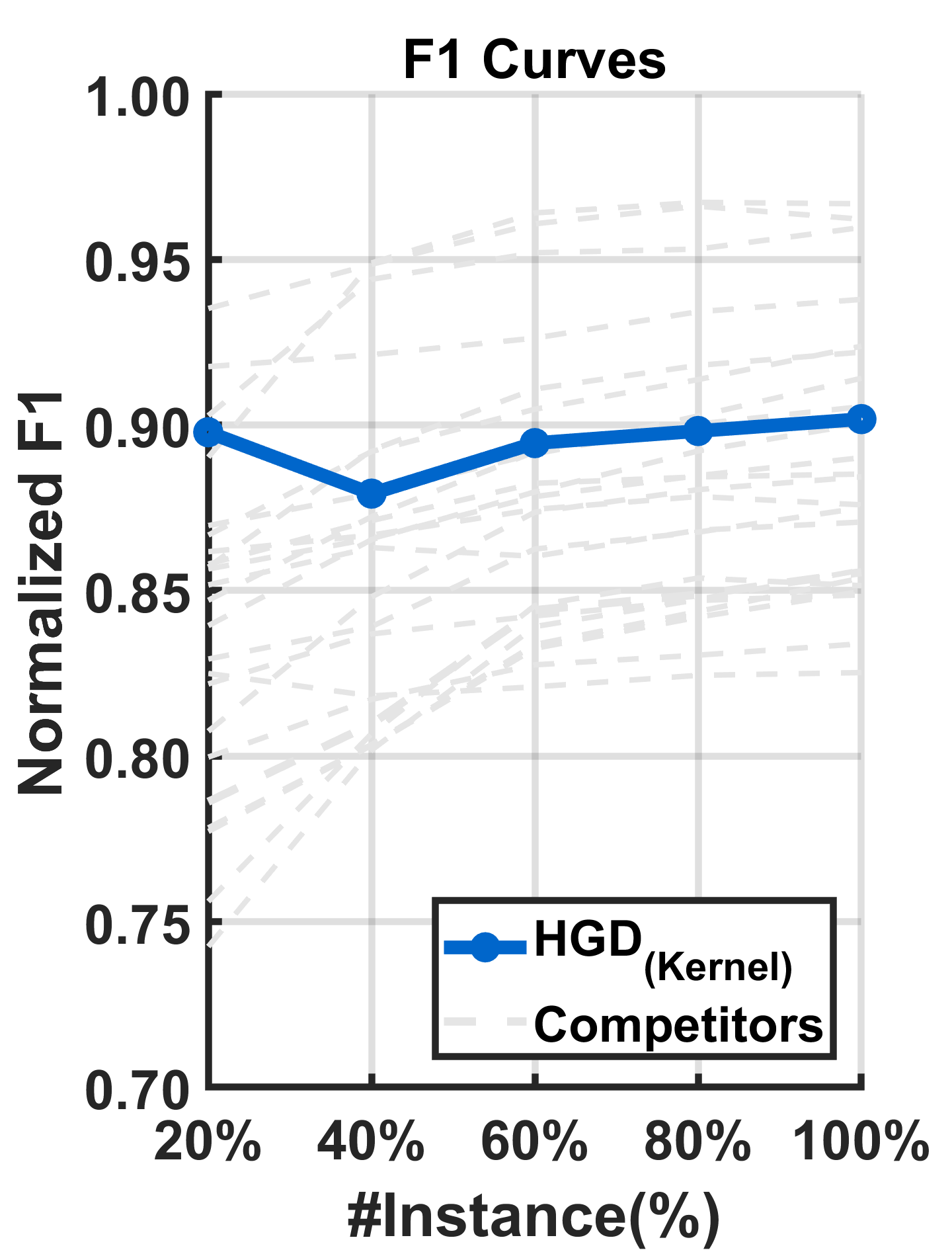}
    }
    \caption{The performance curves of all the methods under dynamic imbalance ratio scenarios. The proposed HGD is highlighted in yellow, red and blue when utilizing perceptron, linear SVM and kernel model as the base learner.}
    \label{S:fig:dy_curves}
\end{figure}

The experimental results in dynamic imbalance ratio scenarios are presented in Table~\ref{S:tab:cmp_dy}. The performance curves throughout the online learning process are depicted in Figure~\ref{S:fig:dy_curves}. As we can see, HGD demonstrates noteworthy improvements over the baseline and exhibits highly competitive performance compared with competitors across all base learners and metrics. Exceptions can be found in the results evaluated by the F1 metric when utilizing the perceptron as the base learner. This performance decrements arises due to the method's attempt towards identifying the minority class, resulting in an inevitable trade-off.

%=========================================
\begin{table*}[h]
\scriptsize
\centering
\renewcommand\arraystretch{1.2}
\caption{The Averaged Performance (Normalized AUC, G-means and F1-Score) Attained by Different Methods Over All Datasets with Dynamic Imbalance Ratio.}
\resizebox{\linewidth}{!}{
\begin{tabular}{lcccccccccccccccccccc}
\hline
\textbf{Base Learner} & \multicolumn{6}{c}{\textbf{Perceptron}} & \textbf{} & \multicolumn{6}{c}{\textbf{Linear SVM}} & \textbf{} & \multicolumn{6}{c}{\textbf{Kernel Model}} \\ \hline
\multicolumn{1}{l|}{\textbf{Metric}} & \textbf{AUC} & \textbf{Rank} & \textbf{GMEANS} & \textbf{Rank} & \textbf{F1} & \textbf{Rank} & \multicolumn{1}{c|}{\textbf{}} & \textbf{AUC} & \textbf{Rank} & \textbf{GMEANS} & \textbf{Rank} & \textbf{F1} & \textbf{Rank} & \multicolumn{1}{c|}{\textbf{}} & \textbf{AUC} & \textbf{Rank} & \textbf{GMEANS} & \textbf{Rank} & \textbf{F1} & \textbf{Rank} \\ \hline
\multicolumn{21}{l}{\cellcolor[HTML]{C0C0C0}\textbf{Baseline}} \\
\multicolumn{1}{l|}{OGD} & \cellcolor[HTML]{FCFCFF}0.967 & 11.1 & \cellcolor[HTML]{FCFCFF}0.935 & 11.0 & \cellcolor[HTML]{FCFCFF}0.934 & \textbf{9.1} & \multicolumn{1}{c|}{} & \cellcolor[HTML]{FCFCFF}0.930 & 14.3 & \cellcolor[HTML]{FCFCFF}0.770 & 13.0 & \cellcolor[HTML]{FCFCFF}0.749 & \textbf{11.7} & \multicolumn{1}{c|}{} & \cellcolor[HTML]{FCFCFF}0.937 & 15.7 & \cellcolor[HTML]{FCFCFF}0.850 & 13.0 & \cellcolor[HTML]{FCFCFF}0.856 & \textbf{9.7} \\
\multicolumn{21}{l}{\cellcolor[HTML]{C0C0C0}\textbf{Data Level Approaches}} \\
\multicolumn{1}{l|}{OSMOTE} & \cellcolor[HTML]{EFF2FA}0.962 & 12.4 & \cellcolor[HTML]{F5F7FC}0.932 & 12.1 & \cellcolor[HTML]{EEF2FA}0.923 & 9.5 & \multicolumn{1}{c|}{} & \cellcolor[HTML]{FAB1B4}0.955 & 14.2 & \cellcolor[HTML]{FA9A9C}0.912 & 12.0 & \cellcolor[HTML]{F9797B}0.899 & \textbf{10.8} & \multicolumn{1}{c|}{} & \cellcolor[HTML]{FBCDD0}0.953 & 15.4 & \cellcolor[HTML]{FA9597}0.949 & 13.5 & \cellcolor[HTML]{F97072}0.962 & 11.6 \\
\multicolumn{1}{l|}{OUR} & \cellcolor[HTML]{FBC2C5}0.970 & \textbf{8.6} & \cellcolor[HTML]{F98082}0.973 & \textbf{8.5} & \cellcolor[HTML]{C5D5EB}0.892 & 9.4 & \multicolumn{1}{c|}{} & \cellcolor[HTML]{FBD2D5}0.944 & 16.7 & \cellcolor[HTML]{F98587}0.941 & 16.6 & \cellcolor[HTML]{FAAEB0}0.838 & 16.4 & \multicolumn{1}{c|}{} & \cellcolor[HTML]{8BACD7}0.925 & 16.8 & \cellcolor[HTML]{FCF8FB}0.854 & 18.7 & \cellcolor[HTML]{5A8AC6}0.825 & 19.4 \\
\multicolumn{1}{l|}{OOR} & \cellcolor[HTML]{ACC3E2}0.937 & 19.6 & \cellcolor[HTML]{C1D3EA}0.906 & 18.8 & \cellcolor[HTML]{BBCEE8}0.885 & 16.0 & \multicolumn{1}{c|}{} & \cellcolor[HTML]{FA9295}0.965 & 13.8 & \cellcolor[HTML]{F97678}0.963 & 13.0 & \cellcolor[HTML]{FAA9AB}0.844 & 14.9 & \multicolumn{1}{c|}{} & \cellcolor[HTML]{FCF4F7}0.940 & 17.8 & \cellcolor[HTML]{F2F5FB}0.846 & 14.5 & \cellcolor[HTML]{E6ECF7}0.851 & 10.8 \\
\multicolumn{1}{l|}{OHR} & \cellcolor[HTML]{B0C6E4}0.939 & 17.8 & \cellcolor[HTML]{C0D1E9}0.905 & 17.9 & \cellcolor[HTML]{AAC2E2}0.872 & 15.2 & \multicolumn{1}{c|}{} & \cellcolor[HTML]{FA9699}0.963 & 14.6 & \cellcolor[HTML]{F97577}0.964 & 13.2 & \cellcolor[HTML]{F97F82}0.891 & 14.3 & \multicolumn{1}{c|}{} & \cellcolor[HTML]{FBD7DA}0.950 & 17.3 & \cellcolor[HTML]{FBC2C5}0.905 & 16.1 & \cellcolor[HTML]{FBD6D9}0.884 & 14.1 \\
\multicolumn{21}{l}{\cellcolor[HTML]{C0C0C0}\textbf{Cost Sensitive Approaches:}} \\
\multicolumn{1}{l|}{CSRDA$_{I}$} & \cellcolor[HTML]{FA9DA0}0.973 & \textbf{7.0} & \cellcolor[HTML]{F97E80}0.974 & \textbf{7.0} & \cellcolor[HTML]{9AB7DC}0.860 & \textbf{9.0} & \multicolumn{1}{c|}{} & \cellcolor[HTML]{FBD7DA}0.942 & 14.3 & \cellcolor[HTML]{FA9B9D}0.911 & 14.2 & \cellcolor[HTML]{F97C7E}0.895 & 13.1 & \multicolumn{1}{c|}{} & - & - & - & - & - & - \\
\multicolumn{1}{l|}{CSRDA$_{II}$} & \cellcolor[HTML]{F8696B}0.976 & \textbf{7.3} & \cellcolor[HTML]{F8696B}0.980 & \textbf{6.1} & \cellcolor[HTML]{A3BDDF}0.867 & \textbf{8.4} & \multicolumn{1}{c|}{} & \cellcolor[HTML]{FAB2B5}0.954 & 14.5 & \cellcolor[HTML]{F97C7E}0.955 & 13.3 & \cellcolor[HTML]{FBB9BC}0.825 & 15.6 & \multicolumn{1}{c|}{} & - & - & - & - & - & -  \\
\multicolumn{1}{l|}{CSRDA$_{III}$} & \cellcolor[HTML]{9EBADE}0.932 & 10.6 & \cellcolor[HTML]{A3BDDF}0.891 & 10.0 & \cellcolor[HTML]{82A6D4}0.841 & 10.8 & \multicolumn{1}{c|}{} & \cellcolor[HTML]{FA9396}0.964 & 14.5 & \cellcolor[HTML]{F97B7D}0.956 & 16.5 & \cellcolor[HTML]{FAA8AB}0.845 & 16.8 & \multicolumn{1}{c|}{} & - & - & - & - & - & - \\
\multicolumn{1}{l|}{CSRDA$_{IV}$} & - & - & - & - & - & - & \multicolumn{1}{c|}{} & \cellcolor[HTML]{FCF7FA}0.932 & 17.2 & \cellcolor[HTML]{FBCDD0}0.838 & 16.6 & \cellcolor[HTML]{FAB2B4}0.834 & 16.3 & \multicolumn{1}{c|}{} & - & - & - & - & - & -  \\
\multicolumn{1}{l|}{CSOGD$_{C_I}$} & - & - & - & - & - & - & \multicolumn{1}{c|}{} & \cellcolor[HTML]{5A8AC6}0.876 & 22.0 & \cellcolor[HTML]{5A8AC6}0.711 & 22.9 & \cellcolor[HTML]{5A8AC6}0.681 & 22.0 & \multicolumn{1}{c|}{} & \cellcolor[HTML]{FCFCFF}0.937 & 17.1 & \cellcolor[HTML]{FAFAFE}0.849 & 14.6 & \cellcolor[HTML]{DCE6F4}0.850 & 11.9 \\
\multicolumn{1}{l|}{CSOGD$_{C_{II}}$} & \cellcolor[HTML]{D2DFF0}0.952 & 12.5 & \cellcolor[HTML]{D0DDEF}0.913 & 15.6 & \cellcolor[HTML]{97B5DB}0.857 & 13.7 & \multicolumn{1}{c|}{} & \cellcolor[HTML]{F8F9FD}0.929 & 20.6 & \cellcolor[HTML]{FAAEB1}0.882 & 21.3 & \cellcolor[HTML]{FBBCBF}0.822 & 22.3 & \multicolumn{1}{c|}{} & \cellcolor[HTML]{FBBDC0}0.959 & 12.8 & \cellcolor[HTML]{FAB0B2}0.923 & 16.4 & \cellcolor[HTML]{FCE8EB}0.871 & 18.6 \\
\multicolumn{1}{l|}{CSOGD$_{S_{I}}$} & - & - & - & - & - & - & \multicolumn{1}{c|}{} & \cellcolor[HTML]{F8F9FD}0.929 & 16.0 & \cellcolor[HTML]{FCFBFE}0.772 & 15.6 & \cellcolor[HTML]{FAFAFE}0.748 & 13.8 & \multicolumn{1}{c|}{} & \cellcolor[HTML]{EDF2FA}0.935 & 18.2 & \cellcolor[HTML]{FCFAFD}0.852 & 13.3 & \cellcolor[HTML]{FCFCFF}0.856 & 10.0 \\
\multicolumn{1}{l|}{CSOGD$_{S_{II}}$} & \cellcolor[HTML]{FBBCBF}0.971 & \textbf{7.5} & \cellcolor[HTML]{F9797B}0.975 & \textbf{7.0} & \cellcolor[HTML]{9CB8DD}0.861 & \textbf{8.9} & \multicolumn{1}{c|}{} & \cellcolor[HTML]{F8696B}0.978 & \textbf{10.4} & \cellcolor[HTML]{F96F71}0.974 & \textbf{10.2} & \cellcolor[HTML]{FA9FA1}0.855 & \textbf{11.7} & \multicolumn{1}{c|}{} & \cellcolor[HTML]{F96C6E}0.988 & \textbf{5.5} & \cellcolor[HTML]{F96A6C}0.990 & \textbf{4.4} & \cellcolor[HTML]{FBD5D8}0.885 & \textbf{7.7} \\
\multicolumn{21}{l}{\cellcolor[HTML]{C0C0C0}\textbf{Ensemble Learning Approaches}} \\
\multicolumn{1}{l|}{OB} & \cellcolor[HTML]{C6D6EC}0.947 & 13.9 & \cellcolor[HTML]{FBC0C3}0.953 & 13.1 & \cellcolor[HTML]{7BA1D1}0.836 & 14.3 & \multicolumn{1}{c|}{} & \cellcolor[HTML]{F0F4FB}0.926 & 15.8 & \cellcolor[HTML]{F8F9FD}0.769 & 13.9 & \cellcolor[HTML]{F5F7FC}0.746 & 12.6 & \multicolumn{1}{c|}{} & \cellcolor[HTML]{FCF0F3}0.941 & 18.1 & \cellcolor[HTML]{FAA7A9}0.931 & 19.0 & \cellcolor[HTML]{FA9092}0.938 & 17.7 \\
\multicolumn{1}{l|}{OAdaB} & \cellcolor[HTML]{F8F9FD}0.965 & 10.1 & \cellcolor[HTML]{F98B8D}0.970 & 9.2 & \cellcolor[HTML]{94B2DA}0.855 & 11.0 & \multicolumn{1}{c|}{} & \cellcolor[HTML]{FAAFB1}0.955 & 15.4 & \cellcolor[HTML]{FAA1A3}0.902 & 15.8 & \cellcolor[HTML]{F9898B}0.881 & 13.3 & \multicolumn{1}{c|}{} & \cellcolor[HTML]{F98285}0.980 & 10.7 & \cellcolor[HTML]{F97375}0.981 & 10.3 & \cellcolor[HTML]{FCE2E4}0.876 & 13.5 \\
\multicolumn{1}{l|}{OAdaC2} & \cellcolor[HTML]{F5F7FC}0.964 & 12.2 & \cellcolor[HTML]{FBCBCE}0.950 & 14.5 & \cellcolor[HTML]{ACC4E3}0.873 & 13.8 & \multicolumn{1}{c|}{} & \cellcolor[HTML]{FAAFB1}0.955 & 16.0 & \cellcolor[HTML]{F98184}0.947 & 17.5 & \cellcolor[HTML]{F9888B}0.881 & 17.8 & \multicolumn{1}{c|}{} & \cellcolor[HTML]{FA9496}0.974 & 10.1 & \cellcolor[HTML]{F98486}0.965 & 15.2 & \cellcolor[HTML]{FBBABD}0.906 & 16.9 \\
\multicolumn{1}{l|}{OCSB2} & \cellcolor[HTML]{CFDCEF}0.950 & 13.8 & \cellcolor[HTML]{FCEBEE}0.940 & 15.3 & \cellcolor[HTML]{94B3DA}0.855 & 13.5 & \multicolumn{1}{c|}{} & \cellcolor[HTML]{FA9597}0.964 & 15.3 & \cellcolor[HTML]{F97F81}0.950 & 17.7 & \cellcolor[HTML]{FA9D9F}0.858 & 19.9 & \multicolumn{1}{c|}{} & \cellcolor[HTML]{F9787A}0.984 & 11.0 & \cellcolor[HTML]{F98385}0.965 & 13.6 & \cellcolor[HTML]{FCE3E6}0.875 & 16.1 \\
\multicolumn{1}{l|}{OKB} & \cellcolor[HTML]{5A8AC6}0.907 & 20.6 & \cellcolor[HTML]{5A8AC6}0.854 & 19.7 & \cellcolor[HTML]{5A8AC6}0.811 & 17.4 & \multicolumn{1}{c|}{} & \cellcolor[HTML]{F2F5FB}0.927 & 14.3 & \cellcolor[HTML]{FBD6D9}0.826 & 12.9 & \cellcolor[HTML]{FBD1D4}0.798 & 12.5 & \multicolumn{1}{c|}{} & \cellcolor[HTML]{E4EBF6}0.934 & 13.6 & \cellcolor[HTML]{FCDCDF}0.880 & 11.9 & \cellcolor[HTML]{F2F5FB}0.854 & 10.1 \\
\multicolumn{1}{l|}{OUOB} & \cellcolor[HTML]{D8E2F2}0.954 & 15.9 & \cellcolor[HTML]{FAB3B5}0.958 & 14.6 & \cellcolor[HTML]{7DA2D2}0.837 & 16.3 & \multicolumn{1}{c|}{} & \cellcolor[HTML]{F96A6C}0.978 & \textbf{11.5} & \cellcolor[HTML]{F8696B}0.981 & \textbf{11.5} & \cellcolor[HTML]{FA979A}0.864 & 13.5 & \multicolumn{1}{c|}{} & \cellcolor[HTML]{F8696B}0.989 & \textbf{8.0} & \cellcolor[HTML]{F8696B}0.990 & \textbf{8.7} & \cellcolor[HTML]{FBCFD1}0.890 & 11.5 \\
\multicolumn{1}{l|}{ORUSB1} & \cellcolor[HTML]{BED0E9}0.944 & 15.3 & \cellcolor[HTML]{BBCEE8}0.903 & 17.0 & \cellcolor[HTML]{759DCF}0.832 & 15.9 & \multicolumn{1}{c|}{} & \cellcolor[HTML]{FCE8EB}0.937 & 14.2 & \cellcolor[HTML]{7AA1D1}0.723 & 17.0 & \cellcolor[HTML]{FBC6C8}0.811 & 15.6 & \multicolumn{1}{c|}{} & \cellcolor[HTML]{5A8AC6}0.920 & 9.7 & \cellcolor[HTML]{5A8AC6}0.786 & 15.7 & \cellcolor[HTML]{88AAD6}0.834 & 16.2 \\
\multicolumn{1}{l|}{ORUSB2} & \cellcolor[HTML]{E1E9F5}0.957 & 14.1 & \cellcolor[HTML]{FAB3B5}0.958 & 14.6 & \cellcolor[HTML]{90B0D9}0.852 & 13.9 & \multicolumn{1}{c|}{} & \cellcolor[HTML]{FBB4B6}0.954 & 13.2 & \cellcolor[HTML]{F9898B}0.936 & 14.5 & \cellcolor[HTML]{FA9193}0.872 & 13.4 & \multicolumn{1}{c|}{} & \cellcolor[HTML]{F98C8F}0.976 & 10.1 & \cellcolor[HTML]{F98184}0.967 & 13.0 & \cellcolor[HTML]{FBC1C4}0.900 & 13.1 \\
\multicolumn{1}{l|}{ORUSB3} & \cellcolor[HTML]{A9C1E1}0.936 & 16.1 & \cellcolor[HTML]{CBD9ED}0.911 & 15.9 & \cellcolor[HTML]{6B96CC}0.824 & 16.9 & \multicolumn{1}{c|}{} & \cellcolor[HTML]{FBD4D6}0.943 & \textbf{11.1} & \cellcolor[HTML]{FBC6C8}0.849 & 13.0 & \cellcolor[HTML]{FAA6A8}0.847 & 12.7 & \multicolumn{1}{c|}{} & \cellcolor[HTML]{FCECEF}0.943 & 12.9 & \cellcolor[HTML]{FBD7DA}0.885 & 14.0 & \cellcolor[HTML]{D7E2F2}0.849 & 13.6 \\
\multicolumn{1}{l|}{OOB} & \cellcolor[HTML]{FCF6F9}0.967 & 10.9 & \cellcolor[HTML]{FA9799}0.966 & 12.0 & \cellcolor[HTML]{B6CAE6}0.881 & 12.2 & \multicolumn{1}{c|}{} & \cellcolor[HTML]{F97779}0.973 & \textbf{10.4} & \cellcolor[HTML]{F96E70}0.974 & \textbf{9.6} & \cellcolor[HTML]{F96A6C}0.916 & \textbf{10.1} & \multicolumn{1}{c|}{} & \cellcolor[HTML]{F9888A}0.978 & 10.5 & \cellcolor[HTML]{F98082}0.968 & \textbf{8.2} & \cellcolor[HTML]{F8696B}0.967 & \textbf{7.3} \\
\multicolumn{1}{l|}{OUB} & \cellcolor[HTML]{D8E3F2}0.954 & 14.1 & \cellcolor[HTML]{FBBABD}0.955 & 14.2 & \cellcolor[HTML]{86A9D5}0.844 & 14.7 & \multicolumn{1}{c|}{} & \cellcolor[HTML]{FBC3C6}0.949 & 16.1 & \cellcolor[HTML]{F9898C}0.935 & 17.6 & \cellcolor[HTML]{FA9294}0.870 & 15.8 & \multicolumn{1}{c|}{} & \cellcolor[HTML]{F98C8E}0.976 & \textbf{8.7} & \cellcolor[HTML]{F9787A}0.976 & 10.6 & \cellcolor[HTML]{FAAFB2}0.914 & 11.9 \\
\multicolumn{1}{l|}{OWOB} & \cellcolor[HTML]{F2F5FB}0.963 & 11.3 & \cellcolor[HTML]{FAA1A3}0.963 & 12.1 & \cellcolor[HTML]{B3C8E5}0.878 & 12.4 & \multicolumn{1}{c|}{} & \cellcolor[HTML]{F97F81}0.971 & \textbf{11.0} & \cellcolor[HTML]{F97678}0.964 & \textbf{10.0} & \cellcolor[HTML]{F8696B}0.916 & \textbf{9.3} & \multicolumn{1}{c|}{} & \cellcolor[HTML]{FA9194}0.975 & \textbf{8.9} & \cellcolor[HTML]{F97F81}0.970 & \textbf{7.0} & \cellcolor[HTML]{F97375}0.960 & \textbf{6.7} \\
\multicolumn{1}{l|}{OWUB} & \cellcolor[HTML]{E0E8F5}0.957 & 16.1 & \cellcolor[HTML]{FAA9AC}0.960 & 15.1 & \cellcolor[HTML]{82A6D4}0.842 & 16.7 & \multicolumn{1}{c|}{} & \cellcolor[HTML]{F98C8E}0.967 & 12.9 & \cellcolor[HTML]{F9787A}0.960 & 14.1 & \cellcolor[HTML]{FA9C9E}0.859 & 14.6 & \multicolumn{1}{c|}{} & \cellcolor[HTML]{F97B7D}0.983 & 11.6 & \cellcolor[HTML]{F97678}0.978 & 11.2 & \cellcolor[HTML]{FAA2A5}0.924 & 12.1 \\
\multicolumn{1}{l|}{OEB} & \cellcolor[HTML]{C2D3EA}0.945 & 17.9 & \cellcolor[HTML]{FBD2D5}0.948 & 16.8 & \cellcolor[HTML]{729BCE}0.830 & 17.8 & \multicolumn{1}{c|}{} & \cellcolor[HTML]{FAA0A3}0.960 & 12.8 & \cellcolor[HTML]{F9797B}0.958 & 12.2 & \cellcolor[HTML]{FA9B9D}0.860 & 12.6 & \multicolumn{1}{c|}{} & \cellcolor[HTML]{FA8F91}0.975 & 13.0 & \cellcolor[HTML]{F9797B}0.975 & 11.9 & \cellcolor[HTML]{FAA5A7}0.922 & 11.9 \\
\multicolumn{21}{l}{\cellcolor[HTML]{C0C0C0}\textbf{The Proposed Approach}} \\
\multicolumn{1}{l|}{HGD} & \cellcolor[HTML]{FCF8FB}0.967 & \textbf{8.4} & \cellcolor[HTML]{F97A7C}0.975 & \textbf{7.0} & \cellcolor[HTML]{A3BDDF}0.866 & \textbf{8.4} & \multicolumn{1}{c|}{} & \cellcolor[HTML]{FA8F91}0.966 & 12.8 & \cellcolor[HTML]{F97072}0.972 & \textbf{10.1} & \cellcolor[HTML]{FA9FA1}0.855 & 12.8 & \multicolumn{1}{c|}{} & \cellcolor[HTML]{F97B7D}0.983 & \textbf{6.4} & \cellcolor[HTML]{F96C6E}0.988 & \textbf{5.1} & \cellcolor[HTML]{FBBFC2}0.902 & \textbf{7.5} \\ \hline
\end{tabular}
}
\begin{tablenotes}
    \item{$\cdot$} The color of cells shows the performance increments (in red) or decrements (in blue) w.r.t. the BaseLine (OGD). 
    \item{$\cdot$} The bold entries stand for the top five highest rankings.
\end{tablenotes}
\label{S:tab:cmp_dy}
\end{table*}
%============================================

%===============================================
%===============================================

\clearpage
\newpage

\subsection{Time Efficiency}

\begin{figure}[h]
    \centering
    \subfloat[Perceptron]{
        \includegraphics[width=4cm]{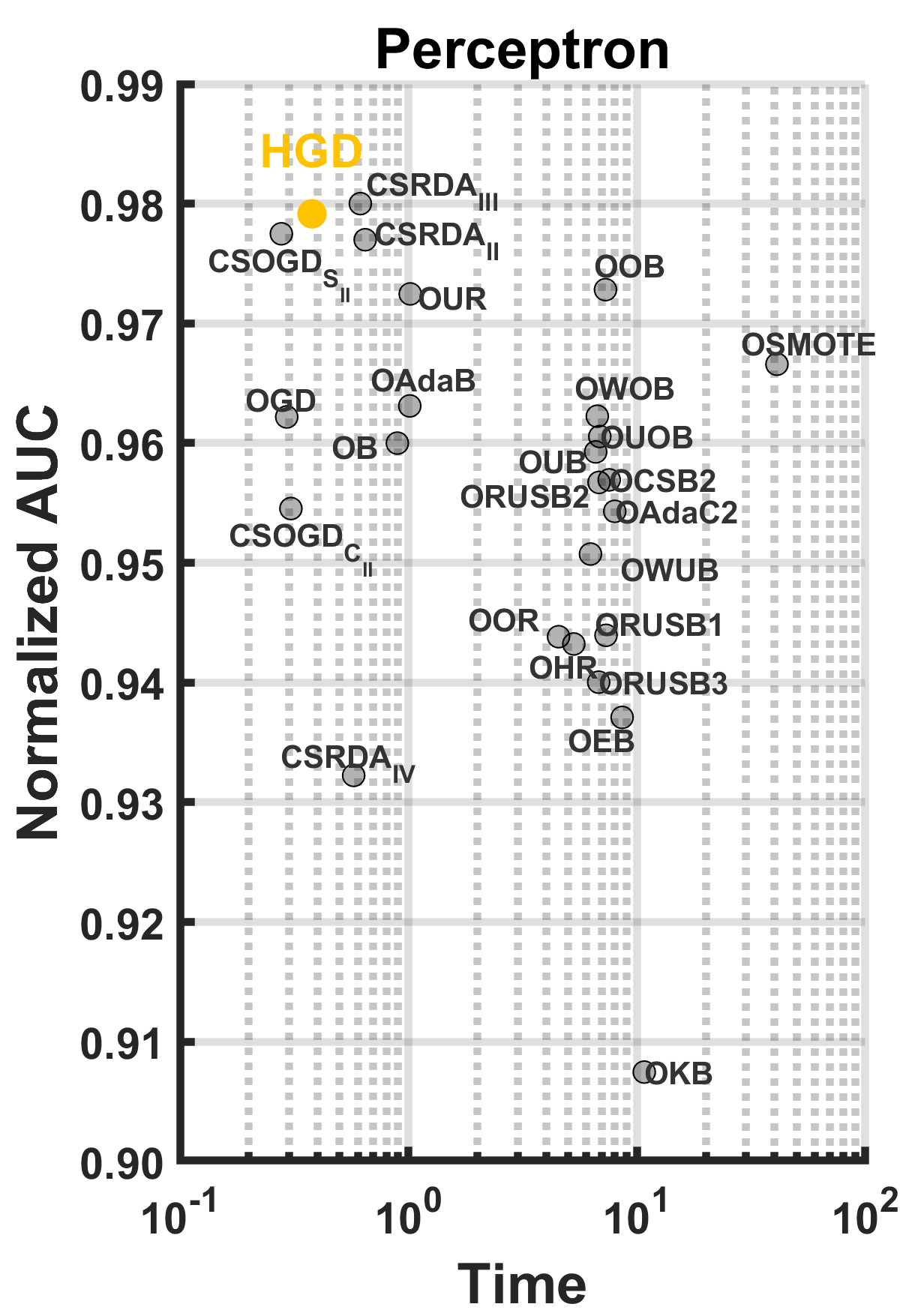}
    }
    \subfloat[Linear SVM]{
        \includegraphics[width=4cm]{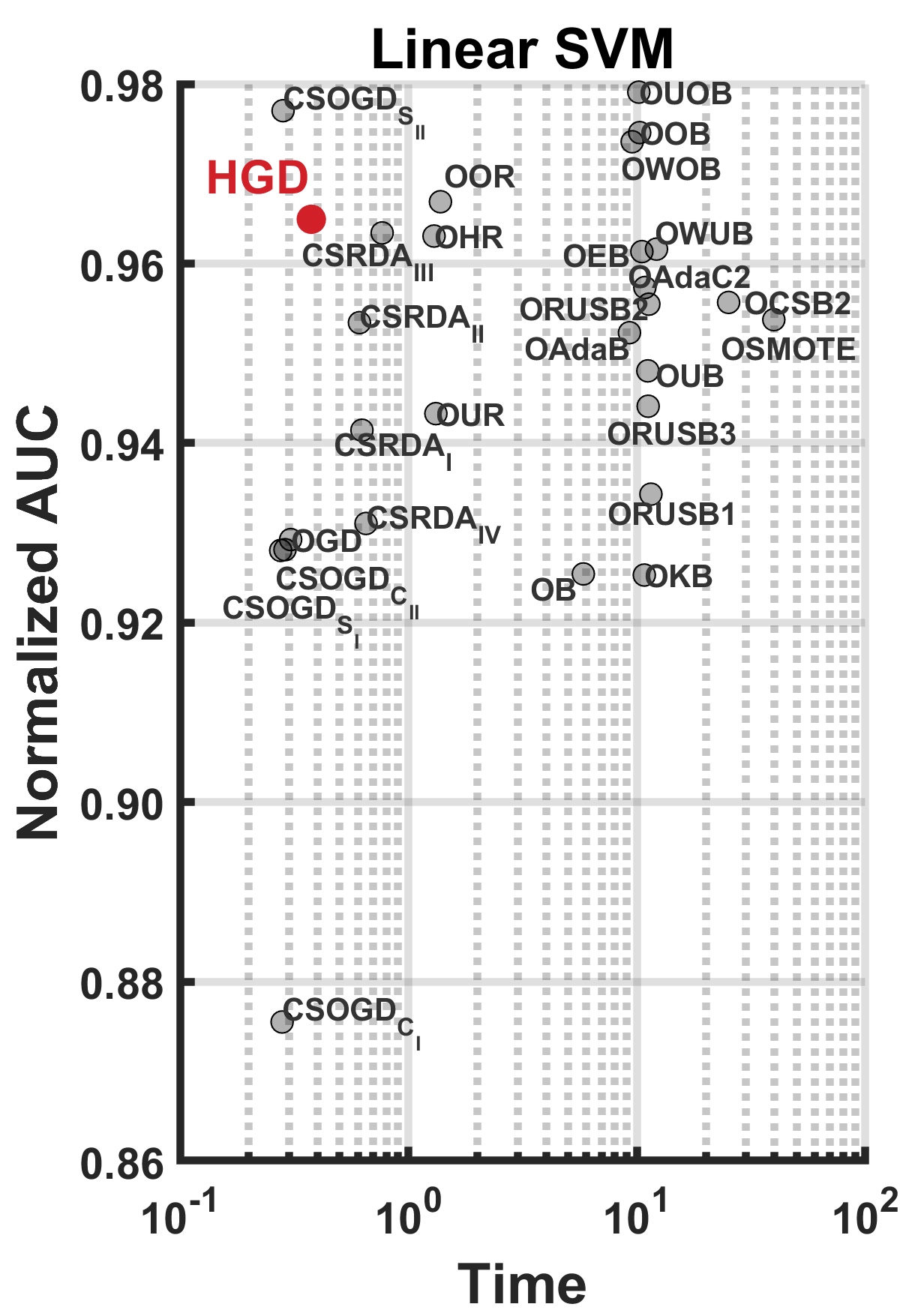}
    }
    \subfloat[Kernel Model]{
        \includegraphics[width=4cm]{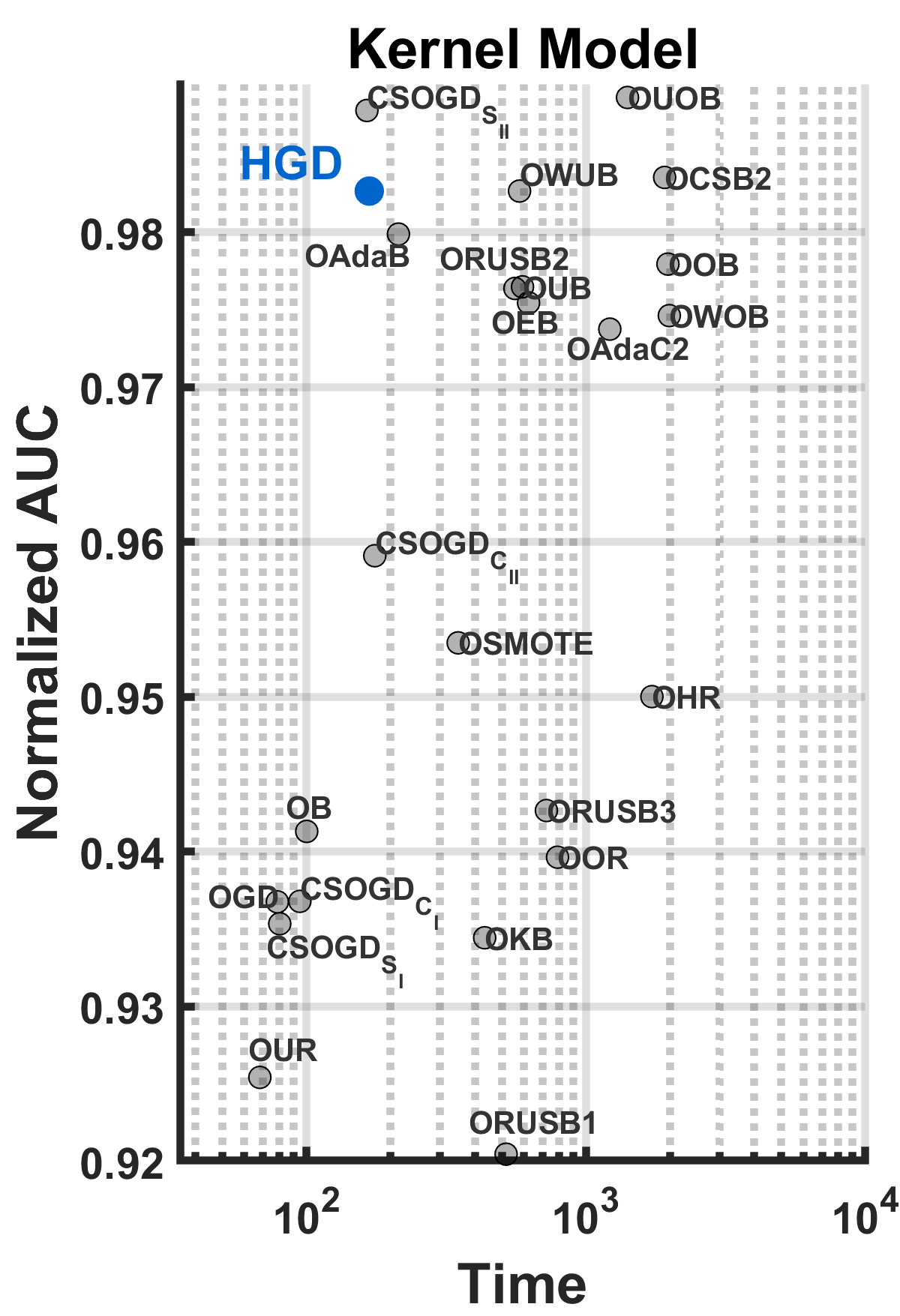}
    }\\
    \subfloat[Perceptron]{
        \includegraphics[width=4cm]{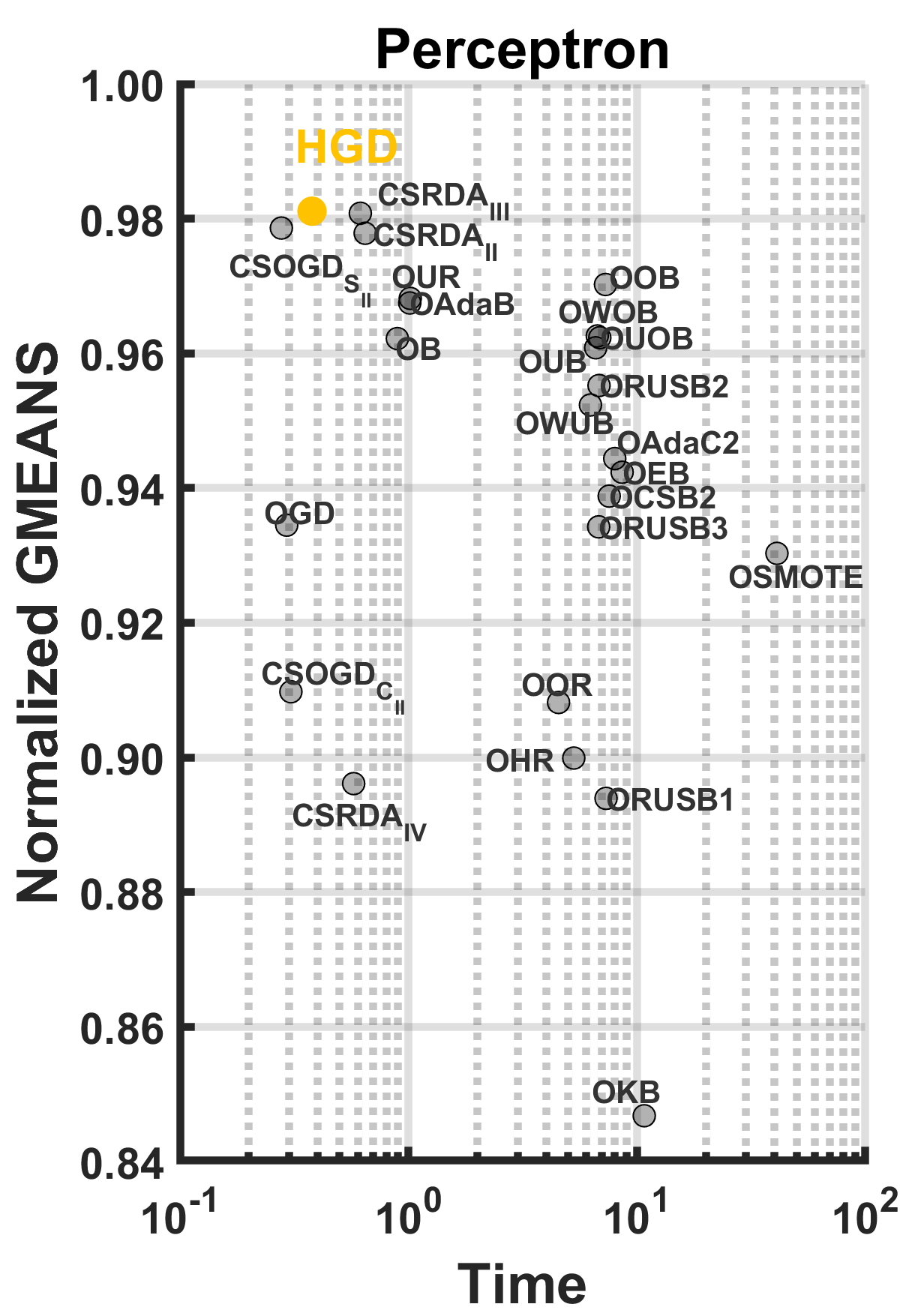}
    }
    \subfloat[Linear SVM]{
        \includegraphics[width=4cm]{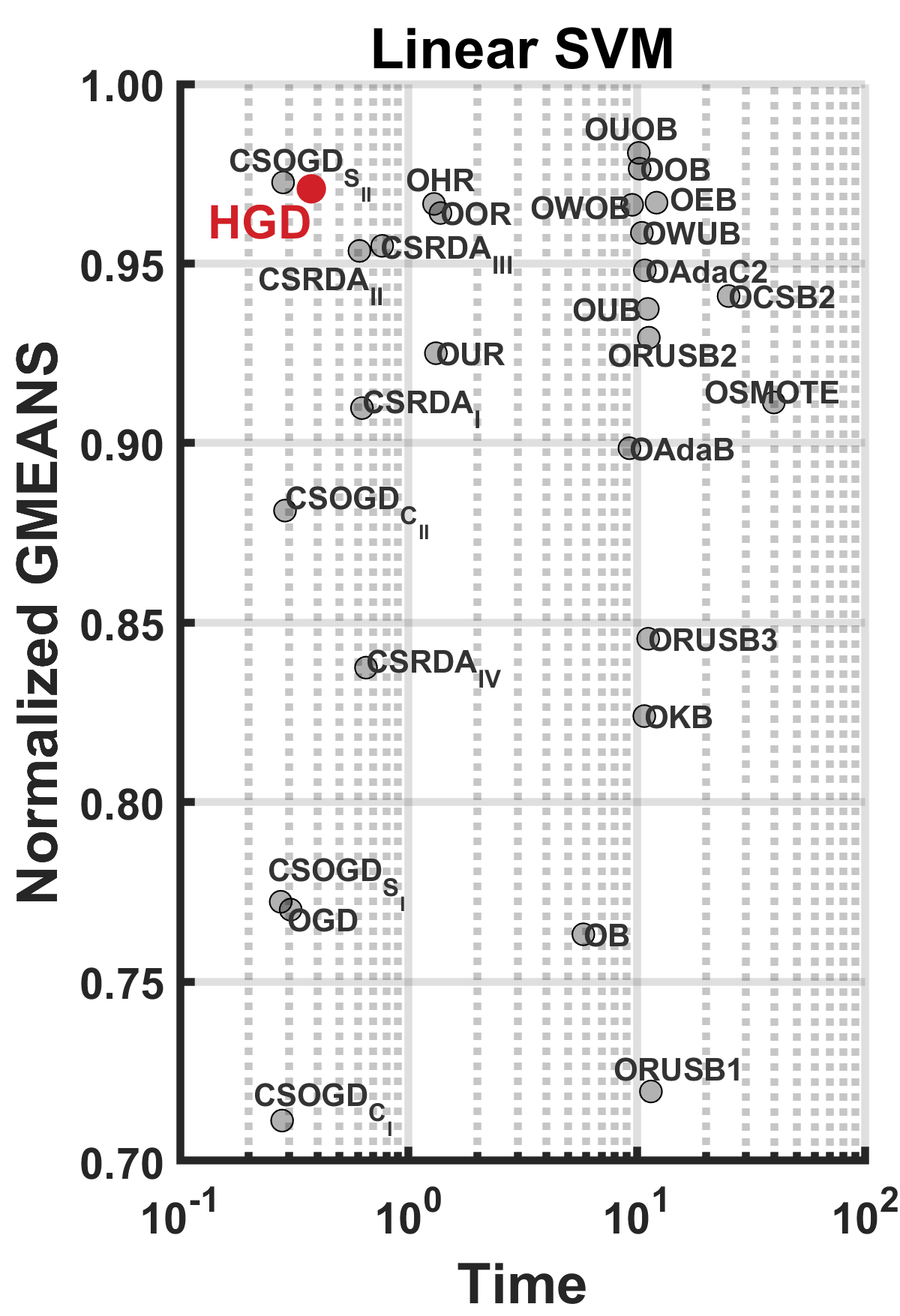}
    }
    \subfloat[Kernel Model]{
        \includegraphics[width=4cm]{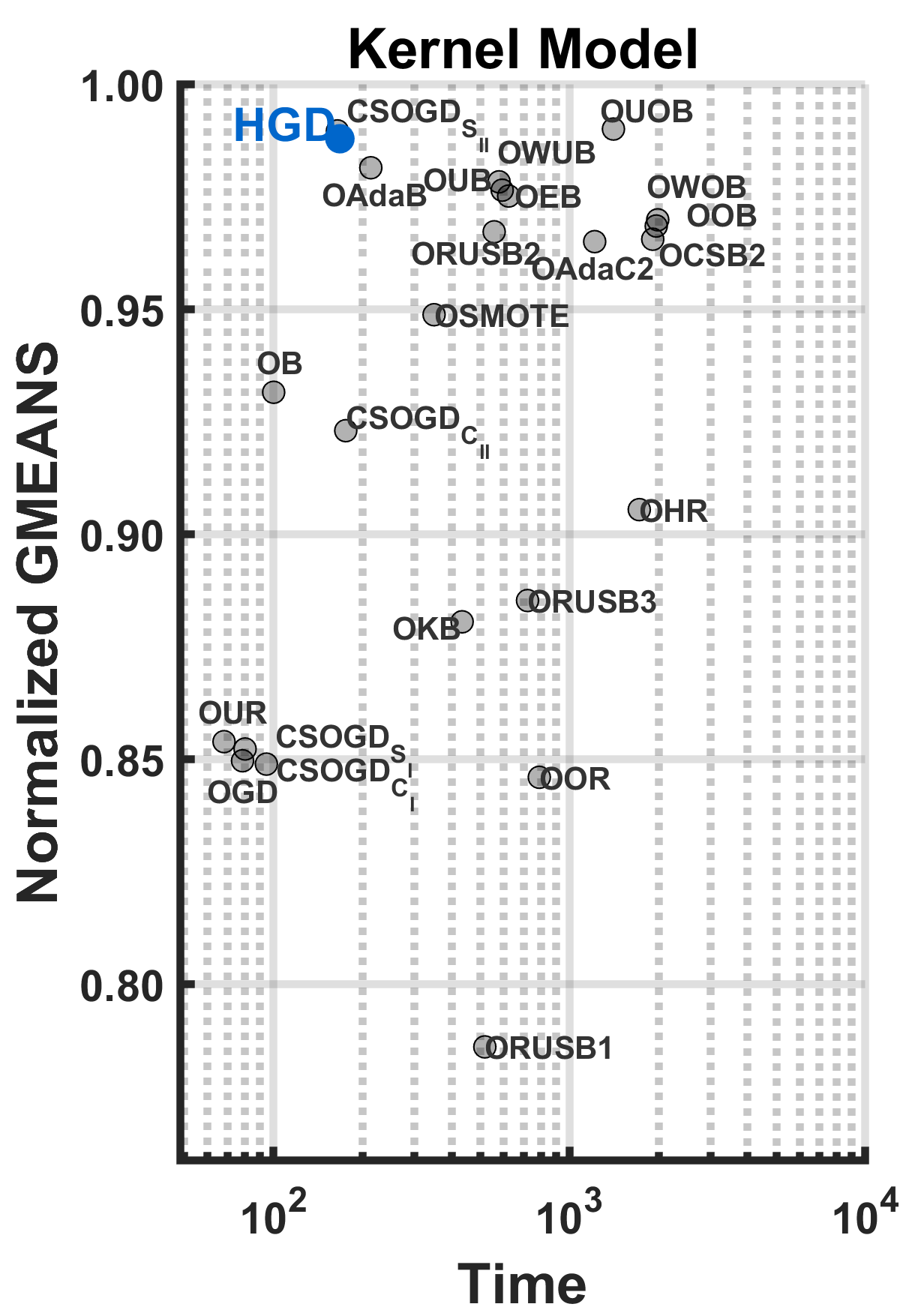}
    }\\
    \subfloat[Perceptron]{
        \includegraphics[width=4cm]{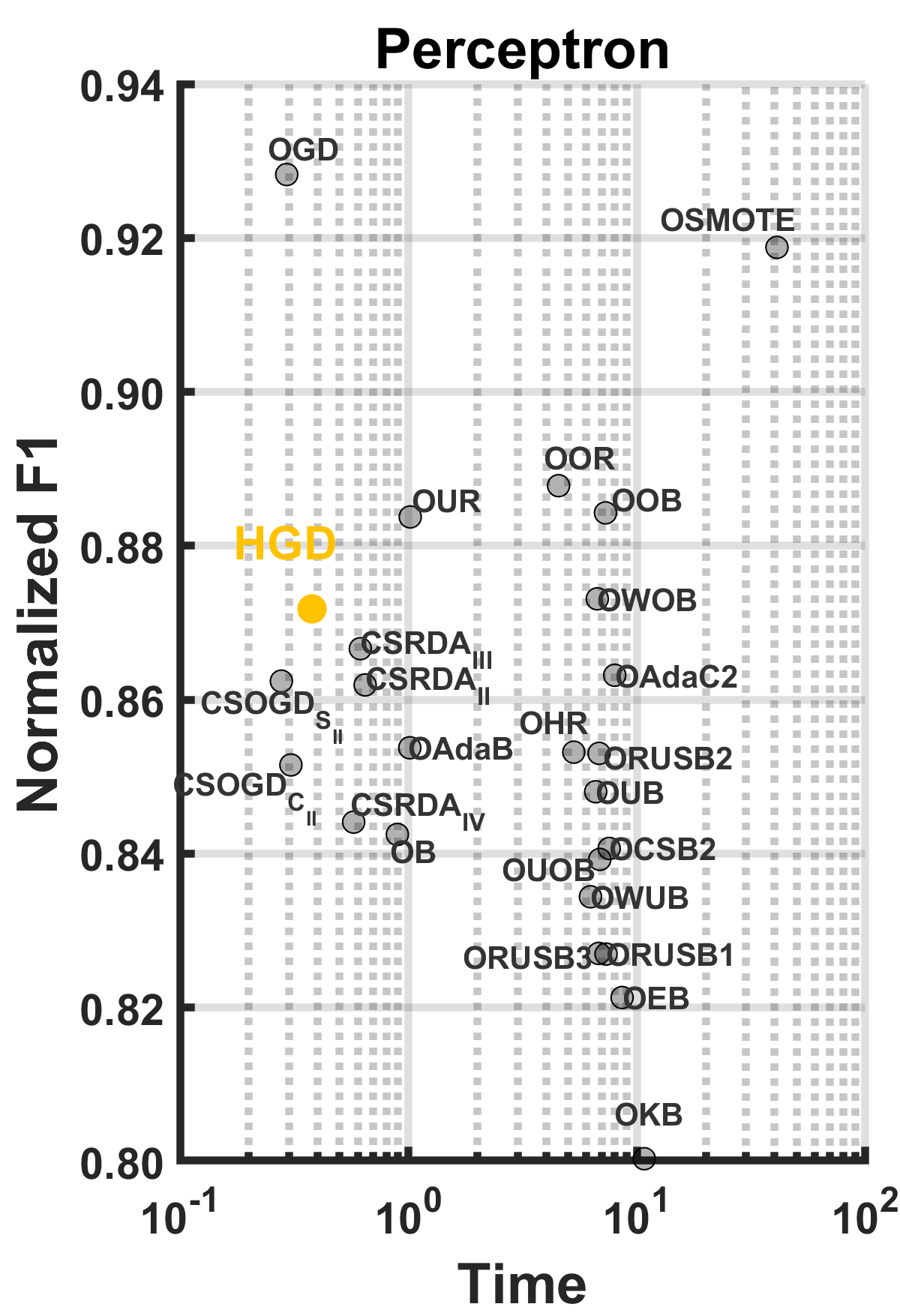}
    } 
    \subfloat[Linear SVM]{
        \includegraphics[width=4cm]{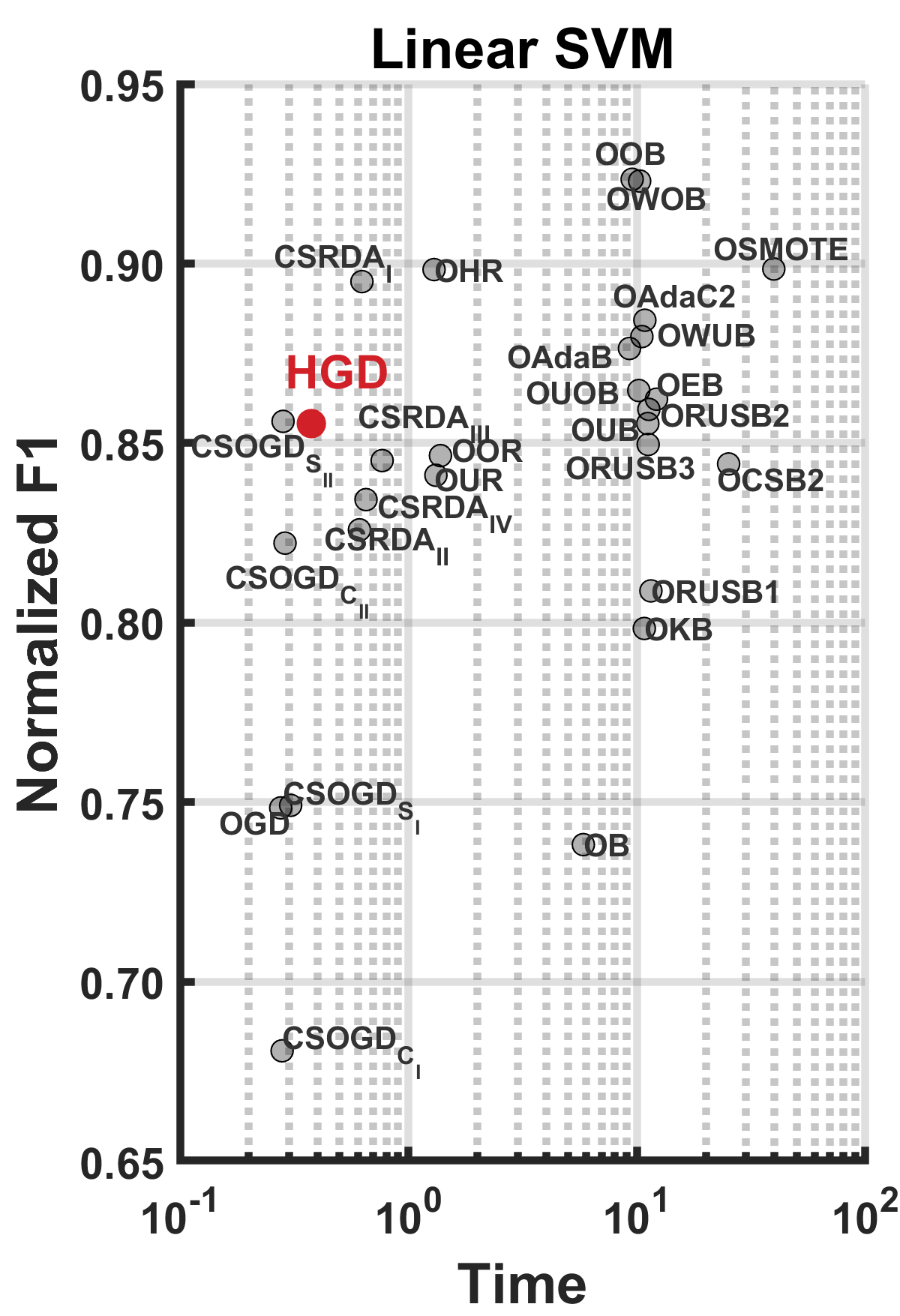}
    }
    \subfloat[Kernel Model]{
        \includegraphics[width=4cm]{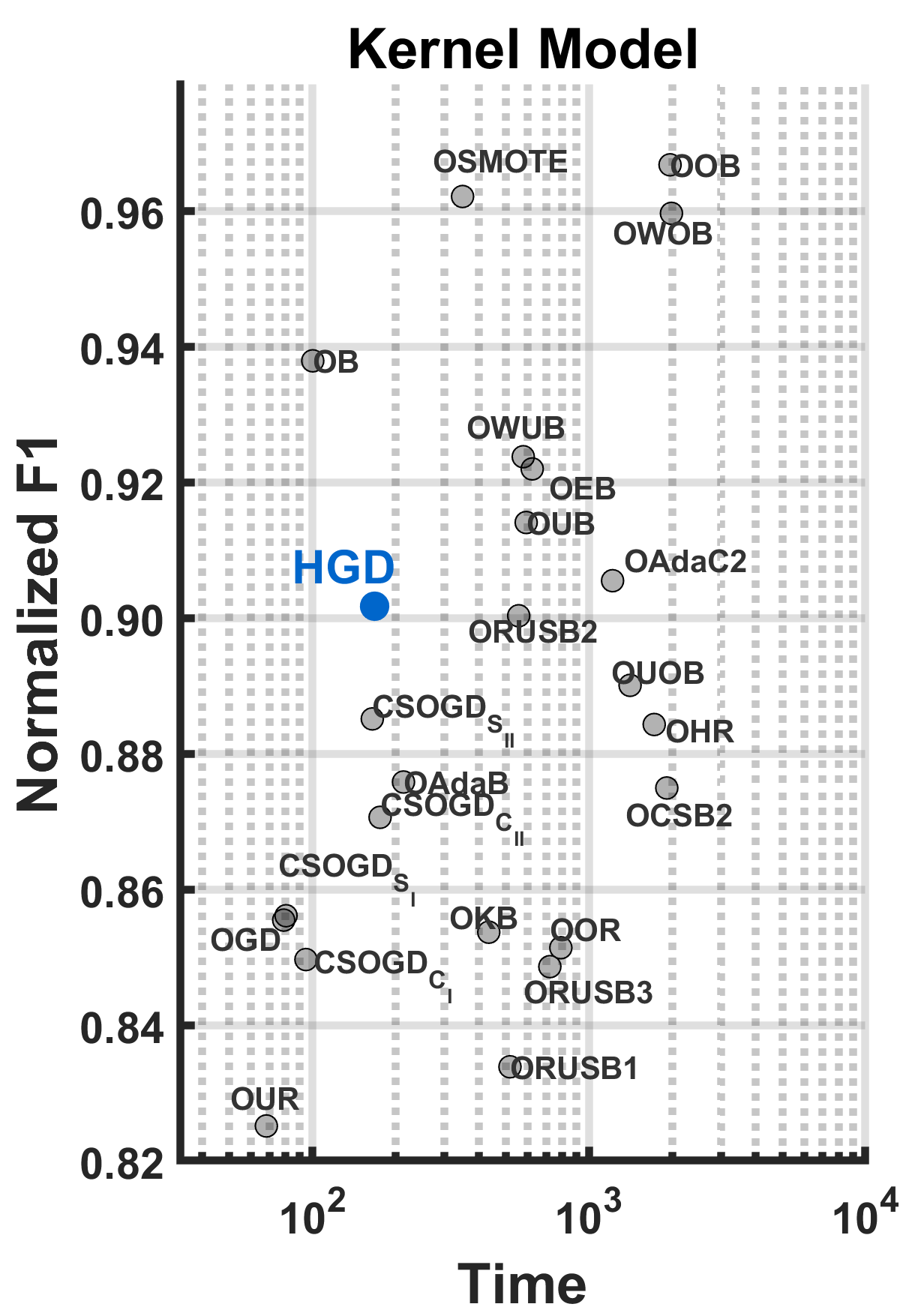}
    }
    \caption{The performance and computational time comparison of all the methods under dynamic imbalance ratio scenarios. Performance is measured by in terms of (a)-(c) AUC; (d)-(f) GMEANS; (g)-(i) F1. Up-left denotes higher performance with less computational cost.}
    \label{S:fig:dy_TvP}
\end{figure}

Comparisons that examines the performance metrics in relation to computational time across all methods are provided in Figure~\ref{S:fig:dy_TvP}. Methods positioned in the upper-left part of the figures denote those achieving superior performance while requiring fewer computational resources. Clearly, HGD holds a favorable position in the upper-left part of figures accross all base learners, thereby highlighting its attributes of performance efficacy and computational efficiency.

%===============================================
%===============================================

\end{document}